\documentclass[sn-mathphys,Numbered]{sn-jnl}

\usepackage{alltt}
\usepackage{amssymb}
\usepackage{caption}
\usepackage{paralist}
\usepackage{nicefrac}
\usepackage{booktabs}
\usepackage{xspace}
\usepackage{amsmath}
\usepackage{marvosym}
\usepackage{wasysym}
\usepackage{verbatim}
\usepackage{float}
\usepackage{hyperref}
\usepackage{multirow}
\usepackage[numbers,sort&compress]{natbib}%
\usepackage[capitalize]{cleveref}
\usepackage[per-mode=symbol,detect-all]{siunitx}
\usepackage{graphics}
\usepackage{amssymb}
\usepackage{enumitem}
\usepackage[table]{xcolor}
\usepackage{colortbl}
\usepackage{ifthen}
\usepackage{mathdots}
\usepackage{arydshln}
\usepackage[outline]{contour}
\usepackage{graphicx}
\usepackage{wrapfig}
\usepackage{algorithm}
\usepackage{algpseudocode}
\usepackage{algorithmicx}
\usepackage{subcaption}
\usepackage{tabularx, stackengine,collcell}

\usepackage{amsfonts}%
\usepackage{amsthm}%
\usepackage{mathrsfs}%
\usepackage{xcolor}%
\usepackage{textcomp}%
\usepackage{manyfoot}%
\usepackage{listings}%

\RequirePackage{multicol} 
\RequirePackage{aliascnt}

\definecolor{navy}{RGB}{0,0,128}
\definecolor{forestGreen}{RGB}{0,110,51}

\newcounter{definitioncounter}

\usepackage{tikz,ifthen,pgfplots}
\usetikzlibrary{arrows,trees,backgrounds,automata,shapes,decorations,plotmarks,fit,calc,positioning,shadows,chains}
\usetikzlibrary{quotes,arrows.meta}
\tikzstyle{every pin edge}=[<-,shorten <=1pt]
\tikzstyle{neuron}=[circle,fill=black!25,minimum size=17pt,inner sep=0pt]
\tikzstyle{input neuron}=[neuron, fill=green!50]
\tikzstyle{output neuron}=[neuron, fill=red!50]
\tikzstyle{hidden neuron}=[neuron, fill=blue!50]
\tikzstyle{small neuron}        =[hidden neuron, draw, minimum size=15pt]
\tikzstyle{small input neuron}  =[input neuron , draw, minimum size=15pt]
\tikzstyle{small output neuron} =[output neuron, draw, minimum size=15pt]
\tikzstyle{annot} = [text width=4em, text centered]
\tikzstyle{nnedge} = [-{stealth},shorten >=0.1cm, shorten <=0.05cm,line 
width=0.8pt,black]
\tikzstyle{edge} = [->,line width = 0.3pt, shorten >=0.2cm]
\tikzstyle{edgeWide} = [->,line width = 2pt, , shorten >=0.2cm]
\usetikzlibrary{calc}

\renewcommand{\arraystretch}{1.2}

\tikzset{every picture/.style={line width=0.75pt}} 
\tikzstyle{BadSquare}=[rectangle,fill=red!30!white,minimum size=25pt,inner 
sep=0pt]
\tikzstyle{InitSquare}=[rectangle,fill=green!30!white,minimum size=25pt,inner 
sep=0pt]

\newcommand{\mysubsection}[1]{\medskip\noindent\textbf{#1}}

\renewcommand{\arraystretch}{1.2}

\newcommand{\relu}{\text{ReLU}\xspace}
\newcommand{\pdt}{\text{PDT}\xspace}

\newcommand{\sat}{\texttt{SAT}\xspace}
\newcommand{\unsat}{\texttt{UNSAT}\xspace}
\newcommand{\init}{\texttt{INIT}\xspace}
\newcommand{\project}{\texttt{PROJECT}\xspace}

\newcommand{\conditionMax}{\texttt{MAX}\xspace}
\newcommand{\conditionPercentile}{\texttt{PERCENTILE}\xspace}
\newcommand{\conditionCombined}{\texttt{COMBINED}\xspace}

\newcommand{\smtsolver}{\texttt{SMT SOLVER }}
\newcommand{\query}{\texttt{query }}
\newcommand{\maxAgree}{\texttt{low}}
\newcommand{\minDisagree}{\texttt{high}}
\newcommand{\disagreementUB}{\texttt{M}}

\newcommand{\marabou}{\textit{Marabou}\xspace}

\newcommand{\modelsSet}{\mathcal{N}}
\newcommand{\modelsSubset}{\mathcal{N'}}
\newcommand{\actionSpace}{\mathcal{A}}
\newcommand{\statesSpace}{\mathcal{S}}
\newcommand{\outputSpace}{\mathcal{O}}
\newcommand{\distanceFn}{d}

\newcommand{\feasibleStatesSpace}{\Psi}
\newcommand{\maxAgg}{\texttt{MAX}\xspace}

\newcommand{\percentileAgg}{\texttt{PERCENTILE}\xspace}
\DeclareMathOperator*{\argmax}{arg\,max}

\usepackage{bussproofs}
\usepackage{gensymb}

\newif\ifcomments
\commentstrue

\newif\ifoutline
\outlinefalse

\newif\iflong
\longtrue

\renewcommand{\paragraph}[1]{\vspace{1mm}\noindent{\bf #1}\ }

\newcounter{experimentCounter}
\newcommand{\experiment}[1]{\noindent%
	\refstepcounter{experimentCounter}\textbf{Experiment 
		(\theexperimentCounter): #1}
}

\usepackage{array}
\newcolumntype{P}[1]{>{\centering\arraybackslash}p{#1}}

\usepackage{placeins}

\newcommand\blfootnote[1]{%
	\begingroup
	\renewcommand\thefootnote{}\footnote{#1}%
	\addtocounter{footnote}{-1}%
	\endgroup
}

\begin{document}
	
	\title{Verifying the Generalization of Deep Learning to Out-of-Distribution 
	Domains}
	
	\author*[1]{\fnm{Guy}
		\sur{Amir}}\email{guyam@cs.huji.ac.il}
	
	\author[1]{\fnm{Osher}
		\sur{Maayan}}\email{osherm@cs.huji.ac.il}
	
	\author[2]{\fnm{Tom}
		\sur{Zelazny}}\email{tzelazny@stanford.edu}
	
	\author[1]{\fnm{Guy}
		\sur{Katz}}\email{guykatz@cs.huji.ac.il}
	
	\author[1]{\fnm{Michael}
		\sur{Schapira}}\email{schapiram@cs.huji.ac.il}
	
	
	
	\affil[1]{\orgname{The Hebrew University of Jerusalem}, 
	\orgaddress{\city{Jerusalem}, \country{Israel}}}

	\affil[1]{\orgname{Stanford University}, \orgaddress{\city{Stanford}, 
	\country{United States}}}





	\abstract{
 \blfootnote{[*] This is an arXiv version of a paper with the same title, appearing in the Journal of Automated Reasoning (JAR), 2024. See \url{https://link.springer.com/journal/10817}.}
 Deep neural networks (DNNs) play a crucial role in the field of 
	machine learning, demonstrating state-of-the-art performance across various 
	application domains. 
		However, despite their success,
		DNN-based models 
		may occasionally exhibit challenges with \emph{generalization}, i.e., 
		may fail to handle inputs that were not encountered during training. 
		This limitation is a significant challenge when it comes to deploying 
		deep learning for safety-critical tasks, as well as in real-world 
		settings characterized by substantial variability. We introduce a novel 
		approach for harnessing DNN
		verification technology to identify DNN-driven decision rules that
		exhibit robust generalization to previously unencountered input 
		domains. 
		Our method assesses generalization within an input domain by measuring 
		the level of agreement between \textit{independently trained} deep 
		neural networks for inputs in this domain. We also efficiently realize 
		our approach by using off-the-shelf DNN verification engines, and 
		extensively evaluate it on both supervised and unsupervised DNN 
		benchmarks, including a deep reinforcement learning (DRL) system for
		Internet congestion control --- demonstrating the applicability of our 
		approach for real-world settings.
		Moreover, our research introduces a fresh objective for formal 
		verification, offering the prospect of mitigating the challenges linked 
		to deploying DNN-driven systems in real-world scenarios.
	}

	\keywords{
		Machine Learning Verification, Deep Learning, Robustness, 
		Generalization}
	
	\maketitle
	\section{Introduction}
	\label{sec:introduction}
	
	In the last decade, deep learning~\cite{GoBeCo16} has demonstrated 
	state-of-the-art performance in natural language processing, image 
	recognition, game playing, computational biology, and numerous other 
	fields~\cite{SiZi14, SiHuMaGuSiVaScAnPaLaDi16, Al19, CoWeBoKaKaKu11, 
	KrSuHi12, BoDeDwFiFlGoJaMoMuZhZhZhZi16, JuLoBrOwKo16}. Despite its 
	remarkable success, deep learning still faces significant challenges that 
	restrict its applicability in domains involving safety-critical tasks or 
	inputs with high variability.

	One critical limitation lies in the well-known challenge faced by deep 
	neural networks (DNNs) when attempting to \emph{generalize} to novel input 
	domains. This refers to their tendency to exhibit suboptimal performance on 
	inputs significantly different from those encountered during training. 
	Throughout the training process, a DNN is exposed to input data sampled 
	from a specific distribution over a designated input domain (referred to as 
	``\textit{in-distribution}'' inputs). The rules derived from this training 
	may falter in generalizing to novel, unencountered inputs, due to several 
	factors: (1) the DNN being invoked in an out-of-distribution (OOD) 
	scenario, where there is a mismatch between the distribution of inputs in 
	the training data and that in the DNN's operational data; (2) certain 
	inputs not being adequately represented in the finite training dataset 
	(such as various, low-probability corner cases); and (3) potential 
	``overfitting'' of the decision rule to the specific training data.

	The importance of establishing the generalizability of (unsupervised) 
	DNN-based decisions is evident in recently proposed applications of deep 
	reinforcement learning (DRL)~\cite{Li17}. 
	Within the framework of DRL, an \textit{agent}, implemented as a DNN, 
	undergoes training through repeated interactions with its environment to 
	acquire a decision-making \textit{policy} achieving high performance 
	concerning a specific objective (``\emph{reward}''). DRL has recently been 
	applied to numerous real-world tasks~\cite{MnKaSi13,ZhLiZhLiXiChXiWaChLi19, 
	MaJaWo20, LiZhChMe18,
		JaRoGoScTa19, VaScShTa17, ChXuWu17, MaAlMeKa16, LeMoSuSa20,
		MaNeAl17}. In many DRL application domains, the learned policy is
	anticipated to perform effectively across a broad spectrum of
	operational environments, with a diversity that cannot possibly be
	captured by finite training data. Furthermore, the consequences of
	inaccurate decisions can be severe. This point is exemplified in our
	examination of DRL-based Internet congestion control (discussed in 
	Sec.~\ref{subsec:aurora}). Good generalization is also crucial for non-DRL 
	tasks, as we shall illustrate through the \emph{supervised-learning} 
	example of Arithmetic DNNs.
	
	
	We introduce a methodology designed to identify DNN-based decision rules 
	that exhibit strong generalization across \emph{a range of distributions} 
	within a specified input domain. Our 
	approach is rooted in the following key observation. 
	The training of a DNN-based model encompasses various stochastic elements, 
	such as the initialization of the DNN's weights and the order in which 
	inputs are encountered during training. As a result, even when DNNs with 
	\emph{the same} architecture undergo training to perform an 
	\emph{identical} task on \emph{the same} training data, the learned 
	decision rules will typically exhibit variations.
	Drawing inspiration from Tolstoy's Anna Karenina~\cite{To77}, we argue
	that ``successful decision rules are all alike; but every
	unsuccessful decision rule is unsuccessful in its own way''. To put it
	differently, we believe that when scrutinizing decisions made by multiple, 
	\textit{independently trained} DNNs on a specific input, consensus is
	more likely to occur when their (similar) decisions are accurate.
	
	Given the above, we suggest the following heuristic for crafting DNN-based 
	decision rules with robust generalization across \textit{an entire} 
	designated input domain: independently train multiple DNNs and identify a 
	subset that exhibits strong consensus across \emph{all} potential inputs 
	within the specified input domain. This implies, according to our 
	hypothesis, that the learned decision rules of these DNNs generalize 
	effectively to all probability distributions over this domain. 
	Our evaluation, as detailed in Sec.~\ref{sec:Evaluation}, underscores the 
	tremendous effectiveness of this methodology in distilling a subset of 
	decision rules that truly excel in generalization across inputs within this 
	domain. As our heuristic aims to identify DNNs whose decisions unanimously 
	align for \emph{every} input in a specified domain, the decision rules 
	derived through this approach consistently achieve high levels of 
	generalization, across all benchmarks.

	Since our methodology entails comparing the outputs of various DNNs across 
	potentially \textit{infinite} input domains, the utilization of formal 
	verification is a natural choice. In this regard, we leverage recent 
	advancements in the formal verification of DNNs~\cite{LuScHe21, AlAvHeLu20,
		AvBlChHeKoPr19, BaShShMeSa19, PrAf20, AnPaDiCh19, SiGePuVe19,
		XiTrJo18, Eh17}.
	Given a verification query comprised of a DNN $N$,
	a precondition $P$, and a postcondition $Q$, a DNN verifier is tasked with
	determining whether there exists an input $x$ to $N$ such that $P(x)$
	and $Q(N(x))$ both hold.
	To date, DNN verification research has primarily concentrated on 
	establishing the local adversarial robustness of DNNs, i.e., identifying 
	small input perturbations that lead to the DNN misclassifying an input of 
	interest~\cite{GeMiDrTsChVe18,
		LyKoKoWoLiDa20, GoKaPaBa18}. Our approach extends the scope of DNN 
		verification by showcasing, for the first time (as far as we are 
		aware), its utility in identifying DNN-based decision rules that 
		exhibit \textit{robust generalization}.
	Specifically, we demonstrate how, within a defined input domain, a DNN 
	verifier can be employed to assign a score to a DNN that indicates its 
	degree of agreement with other DNNs throughout the input domain in 
	question. This, in turn, allows an iterative process for the gradual 
	pruning of the candidate DNN set, retaining only those that exhibit strong 
	agreement and are likely to generalize successfully.

	
	
	

	To assess the effectiveness of our methodology, we concentrate on three 
	widely recognized benchmarks in the field of deep reinforcement learning 
	(DRL):
	\begin{inparaenum}[(i)]
		\item \emph{Cartpole}, where a DRL agent learns to control a cart while 
		balancing a pendulum;
		\item \emph{Mountain Car}, which requires controlling a car to escape 
		from a valley; and
		\item \emph{Aurora}, designed as an Internet congestion controller.
	\end{inparaenum}
	Aurora stands out as a compelling case for our approach. While Aurora is 
	designed to manage network congestion in a diverse range of real-world 
	Internet environments, its training relies solely on synthetically 
	generated data. Therefore, for the deployment of Aurora in real-world 
	scenarios, it is crucial to ensure the soundness of its policy across 
	numerous situations not explicitly covered by its training inputs.
	
	Additionally, we consider a benchmark from the realm of supervised 
	learning, namely, DNN-based arithmetic learning, in which the goal is to 
	train a DNN to correctly perform arithmetic operations. Arithmetic DNNs are 
	a natural use-case for demonstrating the applicability of our approach to a 
	supervised
	learning (and so, non-DRL) setting, and since generalization to OOD domains 
	is a primary focus in this context and is perceived to be especially 
	challenging~\cite{TrHiReRaDyBl18, MaJo20}. We demonstrate how our approach 
	can be employed to assess the capability of Arithmetic DNNs to execute 
	learned operations on ranges of real numbers not encountered in training. 
	
	The results of our evaluation indicate that, across all benchmarks, 
	our verification-driven approach effectively ranks DNN-based decision
	rules based on their capacity to generalize successfully to inputs
	beyond their training distribution.
	In addition, we present compelling evidence that our
	formal verification method is superior to competing methods, namely 
	gradient-based optimization methods and predictive 
	uncertainty methods. 
	These findings highlight the efficacy of our approach. 
	Our code and benchmarks are publicly available as an artifact accompanying 
	this work~\cite{ArtifactRepository}.
	
	The rest of the paper is organized in the following manner.
	Sec.~\ref{sec:Background} provides background on DNNs and their 
	verification procedure. In Sec.~\ref{sec:Approach} we present
	our verification-driven approach for identifying DNN-driven decision rules 
	that generalize
	successfully to OOD input domains. Our evaluation
	is presented in Sec.~\ref{sec:Evaluation}, and a comparison to competing 
	optimization methods is presented in Sec.~\ref{sec:competingMethods}.
	Related work is covered in Sec.~\ref{sec:RelatedWork}, 
	limitations are covered in Sec.~\ref{sec:Limitations},
	and our conclusions are provided in Sec.~\ref{sec:Conclusion}.
	We include appendices with
	additional information regarding our
	evaluation.
	
	
	\medskip
	\noindent{\bf Note.} This is an extended version of our paper, titled 
	``\emph{Verifying Generalization in Deep Learning}''~\cite{AmMaZeKaSc23}, 
	which appeared at the Computer Aided Verification (CAV) 2023 conference. 
	In the original paper, we presented a brief
	description of our method, and
	evaluated it on two DRL
	benchmarks, while giving a high-level description of its applicability to 
	additional benchmarks. In this extended version, we significantly enhance 
	our
	original paper along multiple axes, as explained next. In terms of our 
	approach, we
	elaborate on how to strategically design a DNN verification query
	for the purpose of executing our methods, and we also elaborate on various 
	distance 
	functions
	leveraged in this context.
	We also incorporate a section on competing optimization methods, and
	showcase the advantages of our approach compared to gradient-based
	optimization techniques.
	We significantly enhance our evaluation in the following manner:
	\begin{inparaenum}[(i)]
		\item
		we demonstrate the applicability of our approach to supervised learning,
		and specifically to \emph{Arithmetic} DNNs (in fact, to the best of 
		our knowledge, we are the first to verify
		Arithmetic DNNs); and
		\item
		we enhance the previously presented DRL case study to include additional
		results and benchmarks.
	\end{inparaenum}
	We believe these additions merit an extended paper, which complements our 
	original, shorter one~\cite{AmMaZeKaSc23}.

	\section{Background}
	\label{sec:Background}
	
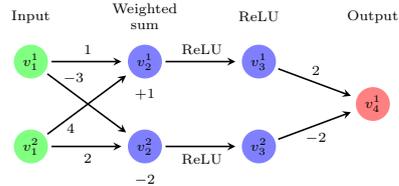
\begin{wrapfigure}[7]{r}{0.45\textwidth}
		\vspace{-1.1cm}
		\begin{center}
			\scalebox{0.75} {
				\def\layersep{2.0cm}
				\begin{tikzpicture}[shorten >=1pt,->,draw=black!50, node 
					distance=\layersep,font=\footnotesize]
					
					\node[input neuron] (I-1) at (0,-1) {$v^1_1$};
					\node[input neuron] (I-2) at (0,-2.5) {$v^2_1$};
					
					\node[hidden neuron] (H-1) at (\layersep,-1) {$v^1_2$};
					\node[hidden neuron] (H-2) at (\layersep,-2.5) {$v^2_2$};
					
					\node[hidden neuron] (H-3) at (2*\layersep,-1) {$v^1_3$};
					\node[hidden neuron] (H-4) at (2*\layersep,-2.5) {$v^2_3$};
					
					\node[output neuron] at (3*\layersep, -1.75) (O-1) 
					{$v^1_4$};
					
					\draw[nnedge] (I-1) --node[above] {$1$} (H-1);
					\draw[nnedge] (I-1) --node[above, pos=0.3] {$\ -3$} (H-2);
					\draw[nnedge] (I-2) --node[below, pos=0.3] {$4$} (H-1);
					\draw[nnedge] (I-2) --node[below] {$2$} (H-2);
					
					\draw[nnedge] (H-1) --node[above] {$\relu$} (H-3);
					\draw[nnedge] (H-2) --node[below] {$\relu$} 
					(H-4);https://www.overleaf.com/project/63a03b9d4341f61c76dcdcf0
					
					\draw[nnedge] (H-3) --node[above] {$2$} (O-1);
					\draw[nnedge] (H-4) --node[below] {$-2$} (O-1);

					\node[below=0.05cm of H-1] (b1) {$+1$};
					\node[below=0.05cm of H-2] (b2) {$-2$};
					
					\node[annot,above of=H-1, node distance=0.8cm] (hl1) 
					{Weighted 
						sum};
					\node[annot,above of=H-3, node distance=0.8cm] (hl2) {ReLU 
					};
					\node[annot,left of=hl1] {Input };
					\node[annot,right of=hl2] {Output };
				\end{tikzpicture}
			}	
		\end{center}
		\caption{A toy DNN.}
		\label{fig:toyDnn}
	\end{wrapfigure}

	\noindent{\bf Deep Neural Networks (DNNs)}~\cite{GoBeCo16} are directed 
	graphs 
	comprising several layers, that subsequently compute various mathematical 
	operations. Upon receiving an input, i.e., assignment values to the 
	nodes of the DNN's
	first (input) layer, the DNN propagates these values, layer after layer, 
	until 
	eventually
	reaching the  final (output) layer, which computes the assignment of the 
	received input.  Each node computes the
	value based on the \emph{type} of operations to which it is associated.
	For example, nodes in weighted-sum layers, compute 
	affine combinations of the values of the nodes in the preceding layer to 
	which 
	they are connected. Another popular layer type is the \emph{rectified 
	linear unit} (\emph{ReLU}) layer, in which each node $y$
	computes the value $y=\relu{}(x)=\max(x,0)$, in which $x$ is the output 
	value of a single
	node from the preceding layer. 
	For more details on DNNs and
	their training procedure, see~\cite{GoBeCo16}. 
	Fig.~\ref{fig:toyDnn} depicts an example of a toy DNN. Given
	input $V_1=[2, 1]^T$, the second layer of this toy DNN computes the 
	(weighted sum)
	$V_2=[7,-6]^T$. Subsequently, the \relu{} functions are applied
	in the third layer, resulting in $V_3=[7,0]^T$. Finally, the
	DNN's single output is accordingly calculated as $V_4=[14]$.

	\medskip
	\noindent{\bf Deep Reinforcement Learning (DRL)}~\cite{Li17} is a popular 
	paradigm in machine 
	learning, in which
	a reinforcement learning (RL) agent, realized as a DNN, interacts with an
	\textit{environment} across multiple time-steps $t\in\{0,1,2,\ldots\}$. At 
	each discrete
	time-step, the DRL 
	agent observes the environment's \textit{state}
	$s_{t} \in \statesSpace$, and selects an \textit{action}
	$N(s_t)=a_{t} \in \actionSpace$ accordingly. As a result of this action, 
	the environment may change and transition to its next
	state $s_{t+1}$, and so on. During training, at each time-step, the 
	environment also presents the agent with a \textit{reward} $r_t$ based on 
	its 
	previously chosen action. The agent is trained by repeatedly interacting 
	with the 
	environment, with the goal of maximizing its \textit{expected cumulative
		discounted reward}
	$R_t=\mathbb{E}\big[\sum_{t}\gamma^{t}\cdot r_t\big]$, where
	$\gamma \in \big[0,1\big]$ is a \textit{discount 
		factor}, i.e., a hyperparameter that controls the accumulative effect 
		of past decisions on the reward. For additional details, 
		see~\cite{SuBa18, ZhJoBr20, ShWoDh17, vHaGuSi16, SuMcSi99,
		HaZhAbLe18}.
	
	\medskip
	\noindent{\bf Supervised Learning (SL)} is another popular machine learning 
	(ML) paradigm. In SL, the input is a dataset of training data comprising 
	pairs of inputs and their \emph{ground-truth} labels $(x_i, y_i)$, drawn 
	from some (possibly unknown) distribution $\mathcal{D}$. The dataset is 
	used to train a model to predict the correct output label for new inputs 
	drawn from the same distribution. 
	
	\medskip
	\noindent{\bf Arithmetic DNNs.} Despite the success of DNNs in many
	SL tasks, they (surprisingly) fail to generalize for the simple
	SL task of attempting to learn arithmetic
	operations~\cite{TrHiReRaDyBl18}. When trained to perform such tasks,
	they often succeed for inputs sampled from the distribution on which
	they were trained, but their performance \emph{significantly}
	deteriorates when tested on inputs drawn OOD, e.g., input values from
	another domain. This behavior is indicative of Arithmetic DNNs tending to 
	overfit their training data rather than systematically learning from it. 
	This is observed even in the context of simple arithmetic tasks, such as 
	approximating the identity function, or learning to sum up inputs. A common 
	belief is that the limitations of the classic learning processes, combined 
	with DNNs' over-parameterized nature, prevent them from learning to 
	generalize arithmetic operations successfully~\cite{TrHiReRaDyBl18, MaJo20}.
	
	\mysubsection{DNN Verification.}  A DNN
	verifier~\cite{KaBaDiJuKo17} receives the following inputs: 
	\begin{inparaenum}[(i)]
		\item a (\textit{trained}) DNN $N$;
		\item a \textit{precondition} $P$ on the inputs of the DNN, effectively 
		limiting the possible assignments 
		to be part of a domain of interest; and	
		\item a \textit{postcondition} $Q$ on the outputs of the DNN.
	\end{inparaenum}
	A sound DNN verifier can then respond in one of the following two ways:
	\begin{inparaenum}[(i)]
		\item \sat, along with a concrete input $x'$ for which the query
		$P(x') \wedge Q(N(x'))$ is satisfied; or
		\item \unsat, indicating no such input $x'$ exists.
	\end{inparaenum}
	Typically, the postcondition $Q$ encodes the \textit{negation} of the DNN's 
	desirable
	behavior for all inputs satisfying $P$. Hence, a \sat result indicates
	that the DNN may err, and that $x'$ is an example of an input in our domain 
	of interest, that triggers a bug;
	whereas an \unsat result indicates that the DNN always performs correctly.
	
	
	For example, let us revisit the DNN in Fig.~\ref{fig:toyDnn}. Suppose that 
	we wish to verify 
	that for all non-negative
	inputs the toy DNN outputs a
	value strictly smaller than $25$, i.e., for all inputs
	$x=\langle v_1^1,v_1^2\rangle \in \mathbb{R}^2_{\geq 0}$, it holds
	that $N(x)=v_4^1 < 25$. This is encoded as a verification query by
	choosing a precondition restricting the inputs to be non-negative, i.e., 
	$P= ( 
	v^1_1\geq 0 \wedge v_1^2\geq 0)$, and
	by setting $Q=(v_4^1\geq 25)$, which is the \textit{negation} of our
	desired property. For this specific verification query, a
	sound verifier will return \sat, alongside a feasible counterexample
	such as $x=\langle 1, 3\rangle$, which produces $v_4^1=26 \geq
	25$. Hence, this property does not hold for the DNN described in
	Fig.~\ref{fig:toyDnn}. 
	To date, a plethora of DNN verification engines have been
	put forth~\cite{Al21,KaBaDiJuKo17,
		GeMiDrTsChVe18, WaPeWhYaJa18, LyKoKoWoLiDa20, HuKwWaWu17}, mostly used 
		in the context of validating the robustness of a general DNN to local 
		adversarial perturbations.

	\section{Quantifying Generalizability via Verification}
	\label{sec:Approach}

	Our strategy for evaluating a DNN's potential for generalization on 
	out-of-distribution inputs is rooted in the ``Karenina hypothesis'': while 
	there might be numerous (potentially infinite) ways to generate 
	\emph{incorrect results}, correct outputs are likely to be quite 
	similar~\footnote{Not to be confused with the ``Anna Karenina Principle'' 
	in statistics, for describing significance tests.}. 
	Therefore, to pinpoint DNN-based decision rules that excel at generalizing 
	to new input domains, we propose the training of multiple DNNs and 
	assessing the learned decision models based on the alignment of their 
	outputs with those of other models in the domain. As we elaborate next, 
	this scoring procedure can be conducted using a backend DNN verifier.
	We show how to effectively distill DNNs that successfully generalize OOD, 
	by iteratively filtering out models that tend to disagree with their peers. 
	
	\subsection{Our Iterative Procedure}
	
	To facilitate our reasoning about the agreement between two DNN-based 
	decision rules over an input domain, we introduce the following definitions.

	\vspace{0.5em} 
	\begin{tikzpicture}
		\node[rectangle, draw, inner sep=10pt, fill=white] (box) {
			\begin{minipage}{0.9\textwidth}
				\begin{definition}[\textbf{Distance Function}]
					\newline
					Let $\outputSpace$ be the space of possible outputs for a 
					DNN. A \emph{distance function} for $\outputSpace$ is a 
					function $\distanceFn: \outputSpace \times \outputSpace 
					\mapsto \mathbb{R^+}$.
				\end{definition}
			\end{minipage}
		};
	\end{tikzpicture}
	
	\noindent
	Intuitively, a distance function allows to quantify 
	the (dis)agreement level between the decisions of two DNNs, when fed the 
	same input.
	We elaborate later on examples of various distance functions that were used.


	
	
	
	\vspace{0.5em} 
	\begin{tikzpicture}
		\node[rectangle, draw, inner sep=10pt, fill=white] (box) {
			\begin{minipage}{0.9\textwidth}
				\begin{definition}[\textbf{Pairwise Disagreement 
				Threshold}]\label{def:disagreementThreshold}
					\newline
					Let $N_{1}, N_{2}$ be a pair of DNNs mapping inputs from 
					the same input domain $\feasibleStatesSpace$ to the same 
					output space $\outputSpace$, and let 
					$\distanceFn$ be a distance function. We
					define the \emph{pairwise disagreement threshold} (PDT) of 
					the DNNs $N_1$ and
					$N_2$ as:
					\[
					\alpha =
					\pdt{}_{\distanceFn, \feasibleStatesSpace}(N_1, N_2) 
					\triangleq \min 
					\left\{\alpha' \in \mathbb{R}^{+} \mid \forall x \in 
					\feasibleStatesSpace 
					\colon d(N_1(x),N_2(x)) \leq \alpha' \right\}
					\]
				\end{definition}
			\end{minipage}
		};
	\end{tikzpicture}

	\noindent
	This definition captures the notion that for \textit{every} possible input 
	in our domain
	$\feasibleStatesSpace$, DNNs $N_{1}$ and $N_{2}$ produce
	outputs that are (at most) $\alpha$-distance apart from each other. Small 
	$\alpha$ values 
	indicate
	that the $N_1$ and $N_2$ produce ``close'' values for all inputs in the 
	domain
	$\feasibleStatesSpace$, whereas a large $\alpha$ values indicate that there
	exists an input in $\feasibleStatesSpace$ for which there is a notable 
	divergence between both decision models.
	
	
	To calculate $\pdt$ values, our method utilizes verification to perform a 
	binary search aiming to find the maximum distance between the outputs of a 
	pair of DNNs; see 
	Alg.~\ref{alg:algorithmPairDisagreementScores}. 
	

	
	\begin{algorithm}[ht]\caption{Pairwise Disagreement Threshold}
		\textbf{Input:} 
		DNNs ($N_{i}$, $N_{j}$),
		input domain $\feasibleStatesSpace$, distance function $\distanceFn$,  
		max. disagreement $\disagreementUB > 0$ \\
		%
		\textbf{Output:} $\pdt(N_{i}, N_{j})$
		
		\begin{algorithmic}[1]  
			\State $\maxAgree \gets 0$, $\minDisagree \gets \disagreementUB$
			
			\While {$\left( \maxAgree < \minDisagree  \right)$} 
			
			\State $\alpha \gets \frac{1}{2} \cdot (\maxAgree+\minDisagree)$
			\State \query $\gets$ \smtsolver $\langle P \gets 
			\feasibleStatesSpace, 
			[N_i;N_j], Q \gets d(N_i,N_j)\geq\alpha \rangle$
			\label{line:SMTsolverForPdt}
			\State \textbf{if} \query is \sat \textbf{then}:  $\maxAgree \gets 
			\alpha$
			
			\State \textbf{else if} \query is \unsat \textbf{then}:   
			$\minDisagree 
			\gets \alpha$ 
			
			
			\EndWhile 
			
			
			
			\State \Return $\alpha$
		\end{algorithmic}
		\label{alg:algorithmPairDisagreementScores}
	\end{algorithm}

	After being calculated, the Pairwise disagreement thresholds can 
	subsequently be aggregated to
	measure the overall disagreement between a decision model and a 
	\textit{set} of other 
	decision models, as defined next.

	\vspace{0.5em} 
	\begin{tikzpicture}
		\node[rectangle, draw, inner sep=10pt, fill=white] (box) {
			\begin{minipage}{0.9\textwidth}
				\begin{definition}[\textbf{Disagreement 
				Score}]\label{def:disagreementScores}\\
					Let 
					$\modelsSet=\{N_{1}, N_{2},\ldots,N_{k}\}$
					be a set of $k$ DNN-based decision models over an input 
					domain $\feasibleStatesSpace$, and
					let $\distanceFn$ be a distance function over the DNNs' 
					output domain.
					We define a model's \emph{disagreement score} (DS) with 
					respect to $\modelsSet$, as:
					\[
					DS_{\modelsSet,\distanceFn, \feasibleStatesSpace}(N_i) = 
					\frac{1}{|\modelsSet|-1}\sum_{j \in [k], j\neq 
					i}\pdt_{\distanceFn, 
						\feasibleStatesSpace}(N_{i}, N_{j})
					\]
					
				\end{definition}
			\end{minipage}
		};
	\end{tikzpicture}

	\noindent
	Intuitively, a disagreement score of a single DNN decision model measures 
	the degree to which it tends to disagree, on average, with the remaining 
	models. %
	
	
	\medskip
	\noindent{\textbf{Iterative Scheme.}}
	Leveraging disagreement scores, our heuristic employs an iterative process 
	(see Alg.~\ref{alg:modelSelection}) to choose a subset of models that 
	exhibit generalization to out-of-distribution scenarios --- as encoded by 
	inputs in 
	$\feasibleStatesSpace$.
	At first, $k$
	DNNs $\{N_1, N_2,\ldots,N_k\}$ are trained \emph{independently} on the 
	training 
	data. Next, a backend verifier is invoked in order to calculate, 
	per each of the ${k \choose 2} $ DNN pairs, their
	respective pairwise-disagreement threshold (up to some accuracy, 
	$\epsilon$).
	Next, our algorithm iteratively: \begin{inparaenum}[(i)]
		\item calculates the disagreement score of each model in the remaining 
		model subset;
		\item identifies models with (relatively) high $DS$ scores; and 
		\item removes them from the model set (Line~\ref{lst:line:removeModels} 
		in 
		Alg.~\ref{alg:modelSelection}).
	\end{inparaenum}
	We also note that the algorithm is given an 
	upper bound (\disagreementUB) on the maximum difference, as informed by the 
	user's 
	domain-specific knowledge. 
	
	\medskip
	\noindent{\textbf{Termination.}}
	The procedure terminates after it exceeds a predefined number of
	iterations (Line~\ref{lst:line:mainLoopStart} in 
	Alg.~\ref{alg:modelSelection}), or alternatively, when all remaining
	models ``agree'' across the input domain $\feasibleStatesSpace$, as 
	indicated by nearly identical 
	disagreement scores
	(Line~\ref{lst:line:checkDifferenceBetweenModels} in 
	Alg.~\ref{alg:modelSelection}). 
	
	

	
	\begin{algorithm}[ht]\caption{Model Selection 
	Procedure}\label{alg:modelSelection}
		\textbf{Input:} Set of models $\modelsSet=\{N_{1},\ldots,N_{k}\}$,
		max disagreement $\disagreementUB$,  number of \texttt{ITERATIONS}  \\
		\textbf{Output:} $\modelsSubset \subseteq \modelsSet$
		\begin{algorithmic}[1]
			\State \texttt{PDT} $\gets $ \Call{Pairwise Disagreement 
				Thresholds}{$\modelsSet, d, \feasibleStatesSpace, 
				\disagreementUB$} 
			\Comment{table with all PDTs} 
			\label{def:disagreementThreshold}
			\State $\modelsSubset \gets \modelsSet$
			\For {$l=1 \ldots $\texttt{ITERATIONS} } 
			\label{lst:line:mainLoopStart} 
			\For{$N_{i} \in \modelsSubset$}
			
			\State \texttt{currentDS}[$N_{i}$] $\gets DS_{\modelsSubset}(N_{i}, 
			\texttt{PDT})$ 
			\Comment{based on Definition~\ref{def:disagreementScores}} 
			\label{line:getDS}
			\EndFor
			\State \textbf{if}  
			\texttt{modelScoresAreSimilar(\texttt{currentDS})} 
			\textbf{then}: \texttt{break}
			\label{lst:line:checkDifferenceBetweenModels}
			\State $\texttt{modelsToRemove} \gets 
			\texttt{findModelsWithHighestDS(\texttt{currentDS})} $
			\label{lst:line:modelsToRemove}
			\State $\modelsSubset \gets \modelsSubset
			\setminus \texttt{modelsToRemove}$
			\Comment{remove models that may disagree} 
			\label{lst:line:removeModels}
			\EndFor
			\State \Return $\modelsSubset$
		\end{algorithmic}
	\end{algorithm}
	
	\medskip
	\noindent{\textbf{DS Removal Threshold.}}
	There are various possible criteria for determining the DS threshold above 
	for 
	which models are removed, as well as the number of models to remove in each 
	iteration 
	(Line~\ref{lst:line:modelsToRemove} in Alg.~\ref{alg:modelSelection}). In 
	our evaluation, we used a 
	simple and natural approach, of iteratively removing the $p\%$ 
	models with the \emph{highest} disagreement scores, for some choice of $p$ 
	($p= 25\%$ in our case). A thorough discussion of 
	additional filtering criteria (all of which proved successful, on all 
	benchmarks) is relegated to
	Appendix~\ref{sec:appendix:AlgorithmAdditionalInformation}.
	

	\subsection{Verification  Queries}    
	\label{subsec:verification-queries}
	Next, we elaborate on how we encoded the queries, which we later fed to our 
	backend verification engine (Line~\ref{line:SMTsolverForPdt} in 
	Alg.~\ref{alg:algorithmPairDisagreementScores}), in order to compute the 
	PDT scores for a DNN pair. 
	
\begin{figure}[ht]
		\centering
		\captionsetup{justification=centering}
		\includegraphics[width=0.6\textwidth]{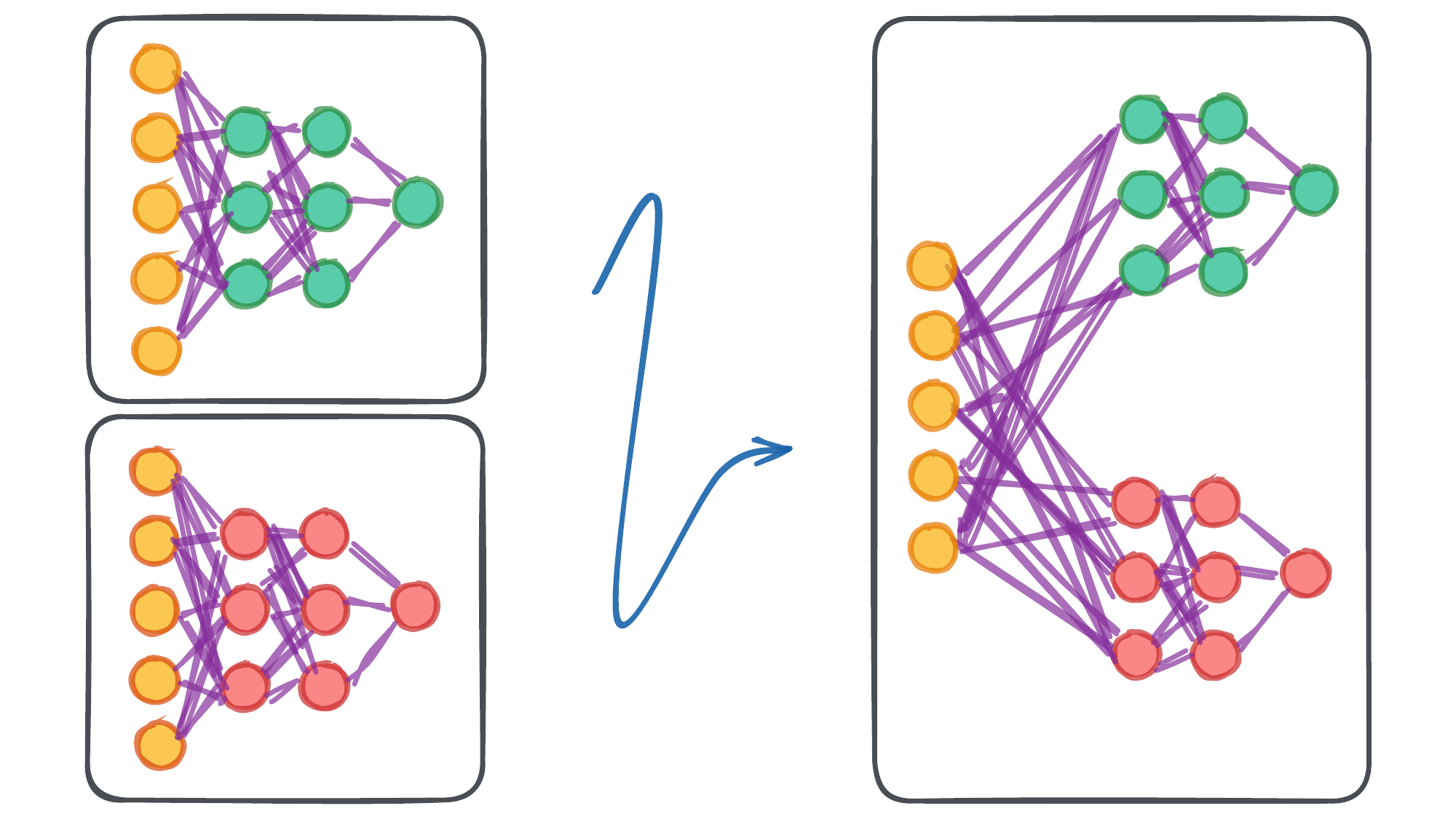}
		\caption{To calculate the PDT scores, we generated a new DNN that
			is the concatenation of each pair of DNNs (sharing the same 
			input).} 
		\label{fig:concatenatedDnns}
	\end{figure}

	\medskip
	\noindent
	Given a DNN pair, $N_1$ and $N_2$, we execute the following stages:
	
	\begin{enumerate}
		
		\item {\bf Concatenate $N_1$ and $N_2$ to a new DNN $N_3=[N_1; N_2]$}, 
		which is roughly twice the size of each of the original DNNs (as both 
		$N_1$ and $N_2$ have the same architecture). The input of $N_3$ is of 
		the same original size as each single DNN and is connected to the 
		second layer of each DNN, consequently allowing the same input to flow 
		throughout the network to the output layers of $N_1$ and 
		$N_2$. Thus, the output layer of $N_3$ is a concatenation of the 
		outputs of both $N_1$ and $N_2$. A scheme depicting the construction of 
		a concatenated DNN appears in Fig.~\ref{fig:concatenatedDnns}.

		\item \textbf{Encode a \emph{precondition} P} which represents the 
		ranges of 
		value assignments to the input variables. As we mentioned before, the 
		value-range bounds are supplied by the system designer, 
		based on prior knowledge of the input domain. In some cases, these 
		values can be predefined 
		to match a specific OOD setting evaluated. In others, these values can 
		be extracted based on empirical simulations of the models post-training.
		For additional details, we refer the reader to 
		Appendix~\ref{sec:appendix:VerificationQueries}.

		\item \textbf{Encode a \emph{postcondition} Q} which encapsulates (for 
		a fixed slack 
		$\alpha$) and a given distance function $\distanceFn: 
		\outputSpace\times\outputSpace \mapsto \mathbb{R^+}$, that for an input 
		$x'\in 
		\feasibleStatesSpace$ the following holds: 
		
		\[     \distanceFn(N_{1}(x'),N_{2}(x')) 
		\geq \alpha
		\]

		\medskip
		\noindent
		Examples of distance functions include:
		
		\begin{enumerate}
			
			\item \textbf{$L_{1}$ norm:}
			
			\[\distanceFn(N_{1}, N_{2}) = \argmax_{x\in 
				\feasibleStatesSpace}(|N_{1}(x) - N_{2}(x)|)\]
			
			This distance function is used in our evaluation of the Aurora and 
			Arithmetic DNNs benchmarks.

			\item $\mathbf{\textbf{condition-distance} 
			(``\text{c-distance}'')}$\textbf{:} 
			
			This function returns the maximal $L_{1}$ norm of two 
			DNNs, for all inputs $x \in \feasibleStatesSpace$ such that both 
			outputs $N_{1}(x)$, $N_{2}(x)$ comply to constraint $\mathbf{c}$.
			
			\[\text{c-distance}(N_{1}, N_{2}) \triangleq \max_{x\in 
				\feasibleStatesSpace \text{ s.t. } N_{1}(x),N_{2}(x) \vDash 
				c}(|N_{1}(x) - N_{2}(x)|)\]

			This distance function is used in our evaluation of the Cartpole 
			and Mountain Car 
			benchmarks. In these cases, we defined the distance 
			function to be:
			\[ \distanceFn (N_{1}, N_{2}) =\min_{c, c'} (\text{c-distance} 
			(N_{1}, N_{2}), 
			\text{c'-distance}(N_{1}, N_{2})) \] 
			
		\end{enumerate}
		
	\end{enumerate}
	
	\section{Evaluation}
	\label{sec:Evaluation}

	\medskip
	\noindent{\textbf{Benchmarks.}} 
	We extensively evaluated our method using four benchmarks: 
	\begin{inparaenum}[(i)]
		\item Cartpole;
		\item Mountain Car;
		\item Aurora; and
		\item Arithmetic DNNs.
	\end{inparaenum}
	The first three are DRL benchmarks, whereas the fourth is a challenging 
	supervised learning benchmark.  Our evaluation of DRL systems spans two 
	classic DRL settings, 
	Cartpole~\cite{BaSuAn83} and Mountain Car~\cite{Mo90}, as well as the 
	recently 
	proposed Aurora congestion controller for Internet 
	traffic~\cite{JaRoGoScTa19}. We also extensively evaluate our approach on
	\emph{Arithmetic DNNs}, i.e., DNNs trained to approximate mathematical
	operations (such as addition, multiplication, etc.).

	\medskip
	\noindent{\textbf{Setup.}} 
	For each of the four benchmarks, we initially trained multiple DNNs with 
	identical architectures, varying only the random seed employed in the 
	training process. 
	Subsequently, we
	removed from this set all the DNNs but the ones that achieved high reward 
	values (in the DRL benchmarks) or high precision (in the 
	supervised-learning benchmark)
	in-distribution, in order to rule out the chance that a decision model 
	exhibits poor generalization solely because of inadequate training.
	Next, we specified out-of-distribution input domains of interest for each 
	specific benchmark and employed Alg.~\ref{alg:modelSelection} to choose the 
	models deemed most likely to exhibit good generalization on those domains 
	according to our framework.
	To determine the ground truth regarding the actual generalization 
	performance of different models in practice, we applied the models to 
	inputs drawn from the considered OOD domain, and ranked them based on 
	empirical performance (average reward/maximal error, depending on the 
	benchmark).
	To assess the robustness of our results, we performed the last step with 
	different choices of probability distributions over the inputs in the 
	domain. 
	
	\medskip
	\noindent
	\textbf{Verification.} 
	All queries were dispatched using
	\marabou~\cite{KaHuIbJuLaLiShThWuZeDiKoBa19,WuIsZeTaDaKoReAmJuBaHuLaWuZhKoKaBa24}
	 --- a sound and complete DNN verification engine, which is capable of 
	addressing queries regarding a DNN's characteristics by converting them 
	into SMT-based constraint satisfaction problems.
	The Cartpole benchmark included $48,000$ queries ($24,000$ queries per each 
	of the two platform
	sides), all of which terminated within $12$ hours. 
	The Mountain Car benchmark included $10,080$ queries, all of which 
	terminated within one hour.
	The Aurora benchmark included $24,000$ verification queries, out of which 
	all but $12$ queries terminated within $12$ hours; and the remaining ones 
	hit the time-out threshold. 
	Finally, the Arithmetic DNNs benchmark included $2,295$ queries, running 
	with a time-out value of $24$ hours; all queries terminated, with over 
	$96\%$ running in less than an hour, and the longest non-DRL query taking 
	slightly less than $13.8$ hours. 
	All benchmarks ran on a single CPU, and with a memory limit of either $1$ 
	GB (for Arithmetic DNNs) or $2$ GB (for the DRL benchmarks). 
	We note that in the case of the Arithmetic DNNs benchmark --- Marabou 
	internally used the \emph{Guorobi} LP 
	solver\footnote{\url{https://www.gurobi.com}} as a backend engine when 
	dealing with these queries.
	
	\medskip
	\noindent{\textbf{Results.}} 
	The findings support our claim that models chosen using our approach are 
	expected to \emph{significantly outperform} other models for inputs drawn 
	from the OOD domain considered. This is the case for all evaluated settings 
	and benchmarks, regardless of the chosen hyperparameters and filtering 
	criteria. 
	We note that although our approach can potentially also remove some of the 
	successful models, in all benchmarks, and across all evaluations, it 
	managed to remove \emph{all} unsuccessful models.
	Next, we provide an overview of our evaluation. A comprehensive exposition 
	and additional details can be found in the appendices. 
	Our code and benchmarks are publicly available 
	online~\cite{ArtifactRepository}.

	\subsection{Cartpole}\label{subsec:cartpole}
	\begin{wrapfigure}{r}{0.5\textwidth}
		\vspace{-1.5cm}
		\centering
		\begin{center}
			\includegraphics[width=0.45\textwidth]{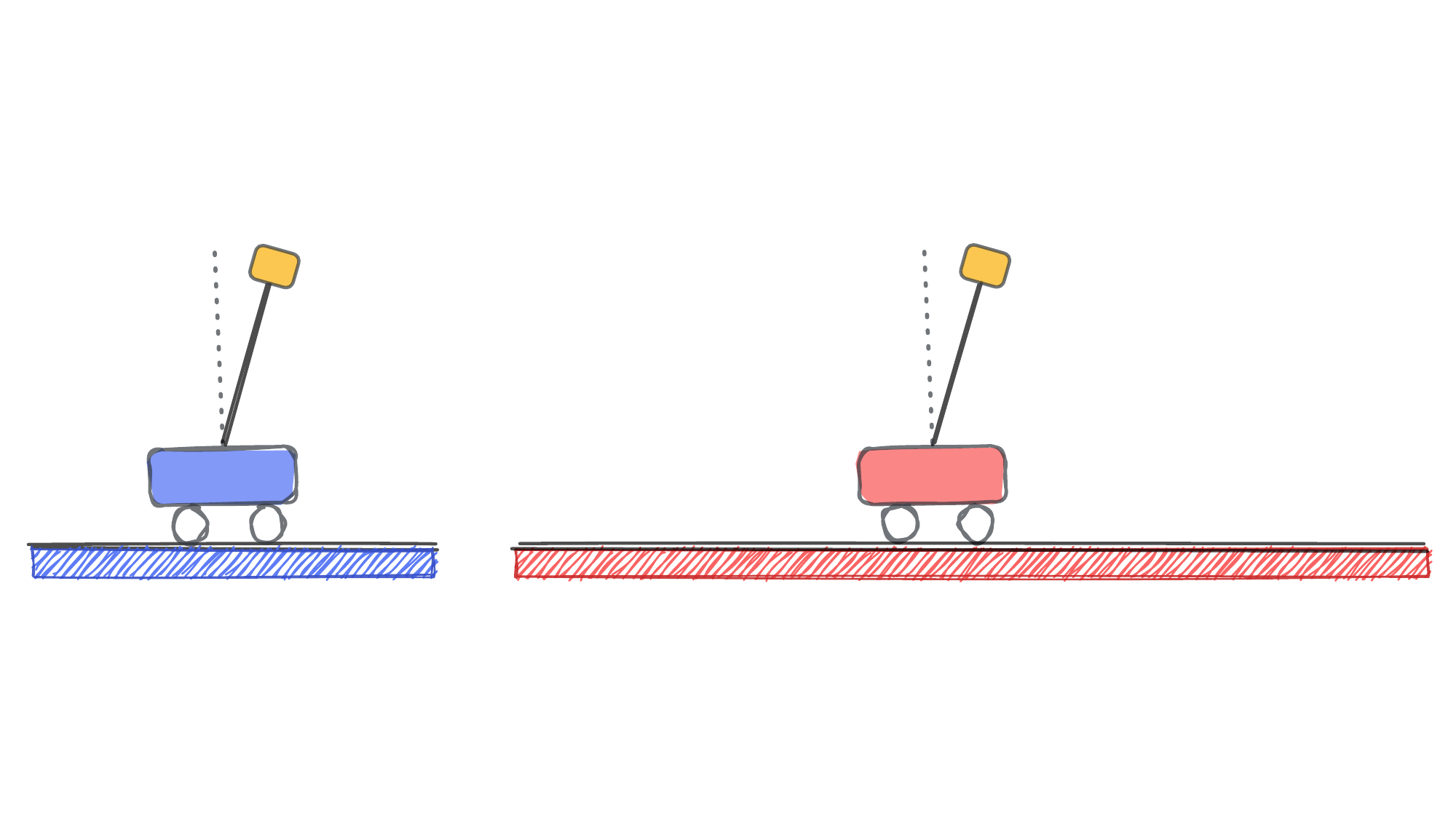}
		\end{center}
		\vspace{-1.0cm}
		\caption{Cartpole: in-distribution setting (blue) and OOD setting 
		(red).}
		\vspace{-0.5cm}
		\label{fig:CartpoleInDistributionAndOod}
	\end{wrapfigure}
	Cartpole~\cite{GeSi93} is a widely known RL benchmark where an agent
	controls the motion of a cart with an inverted pendulum (``pole'') affixed 
	to its top. The cart traverses a platform, and the objective of the agent 
	is to maintain balance for the pole for as long as possible (see 
	Fig.~\ref{fig:CartpoleInDistributionAndOod}).

	\medskip
	\noindent{\textbf{Agent and Environment.}}
	The agent is provided with inputs, denoted as $s=(x, v_{x}, \theta, 
	v_{\theta})$, where $x$ represents the cart's position on the platform, 
	$\theta$ represents the angle of the pole (with $|\theta|$ approximately 
	$0$ for a balanced pole and $|\theta|$ approximately $90\degree$ for an 
	unbalanced pole), $v_{x}$ indicates the cart's horizontal velocity, and 
	$v_{\theta}$ denotes the pole's angular velocity.
	
	\medskip
	\noindent{\textbf{In-Distribution Inputs.}} 
	During the training process, the agent is encouraged to balance the pole 
	while remaining within the boundaries of the platform. In each iteration, 
	the agent produces a single output representing the cart's acceleration 
	(both sign and magnitude) for the subsequent step. Throughout the training, 
	we 
	defined the platform's limits as $[-2.4, 2.4]$, and the initial position of 
	the cart as nearly static and close to the center of the platform (as 
	depicted on the left-hand side of 
	Fig.~\ref{fig:CartpoleInDistributionAndOod}). This was accomplished by 
	uniformly sampling the initial state vector values of the cart from the 
	range $[-0.05, 0.05]$.



	\medskip
	\noindent{\textbf{(OOD) Input Domain.}}
	We examine an input domain with larger platforms compared to those utilized 
	during training. Specifically, we extend the range of the $x$ coordinate in 
	the 
	input vectors to cover [-10, 10]. The bounds for the other inputs remain 
	the same as during training. For additional details, see 
	Appendices~\ref{sec:appendix:trainingAndEvaluation} 
	and~\ref{sec:appendix:VerificationQueries}.
	
	\medskip
	\noindent{\textbf{Evaluation.}}
	We trained a total of $k=16$ models, all of which demonstrated high rewards 
	during training on the short platform. Subsequently, we applied 
	Alg.~\ref{alg:modelSelection} until convergence (requiring $7$ iterations 
	in our experiments) on the
	aforementioned input domain. 
	This resulted in a collection of 3 models. We then subjected all $16$ 
	original models to inputs that were drawn from the new, OOD domain. The 
	generated distribution was crafted to represent a novel scenario: the cart 
	is now positioned at the center of a considerably longer, shifted platform 
	(see the red-colored cart depicted in 
	Fig.~\ref{fig:CartpoleInDistributionAndOod}).

\begin{figure}[h]
		\centering
		\subfloat[In-distribution 
		\label{subfig:cartpoleRewards:inDist}]{\includegraphics[width=0.49\textwidth]{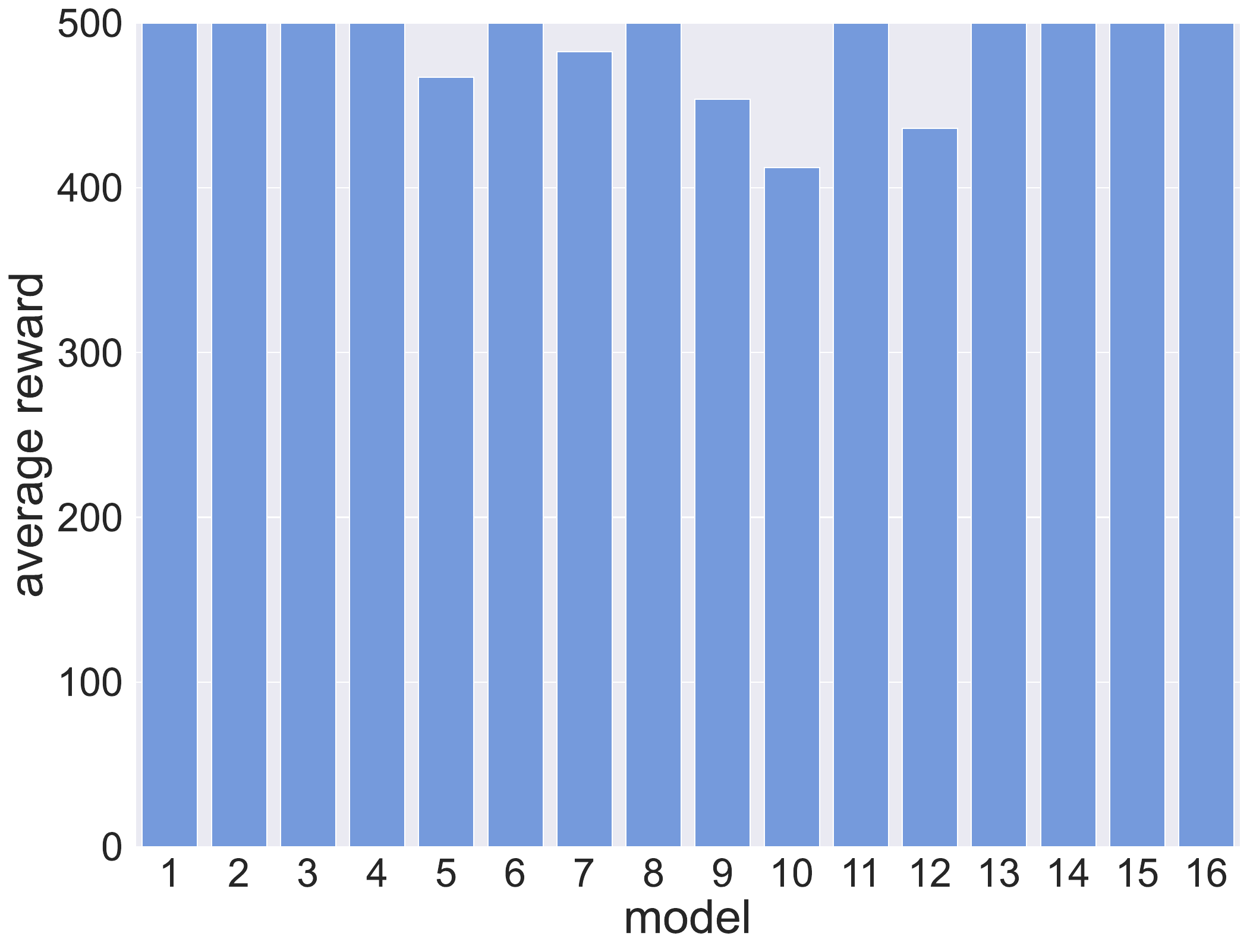}}
		\hfill
		\subfloat[OOD\label{subfig:cartpoleRewards:OOD}]{\includegraphics[width=0.49\textwidth]{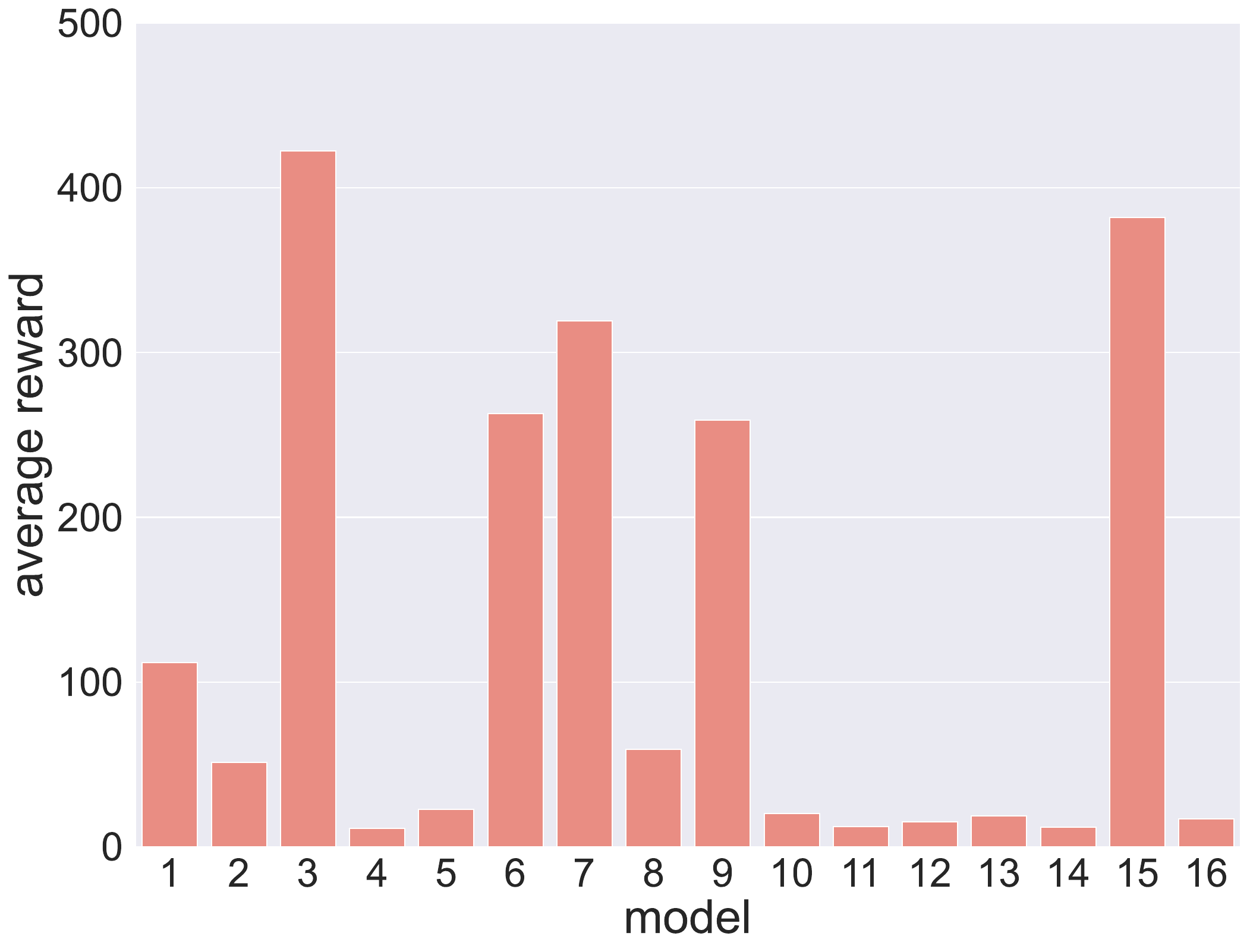}}\\
		
		\caption{Cartpole: models' average rewards in different distributions.}
		\label{fig:cartpoleRewards}
	\end{figure}

\begin{figure}[h] 
		\centering
		\begin{center}
			\includegraphics[width=0.55\textwidth]{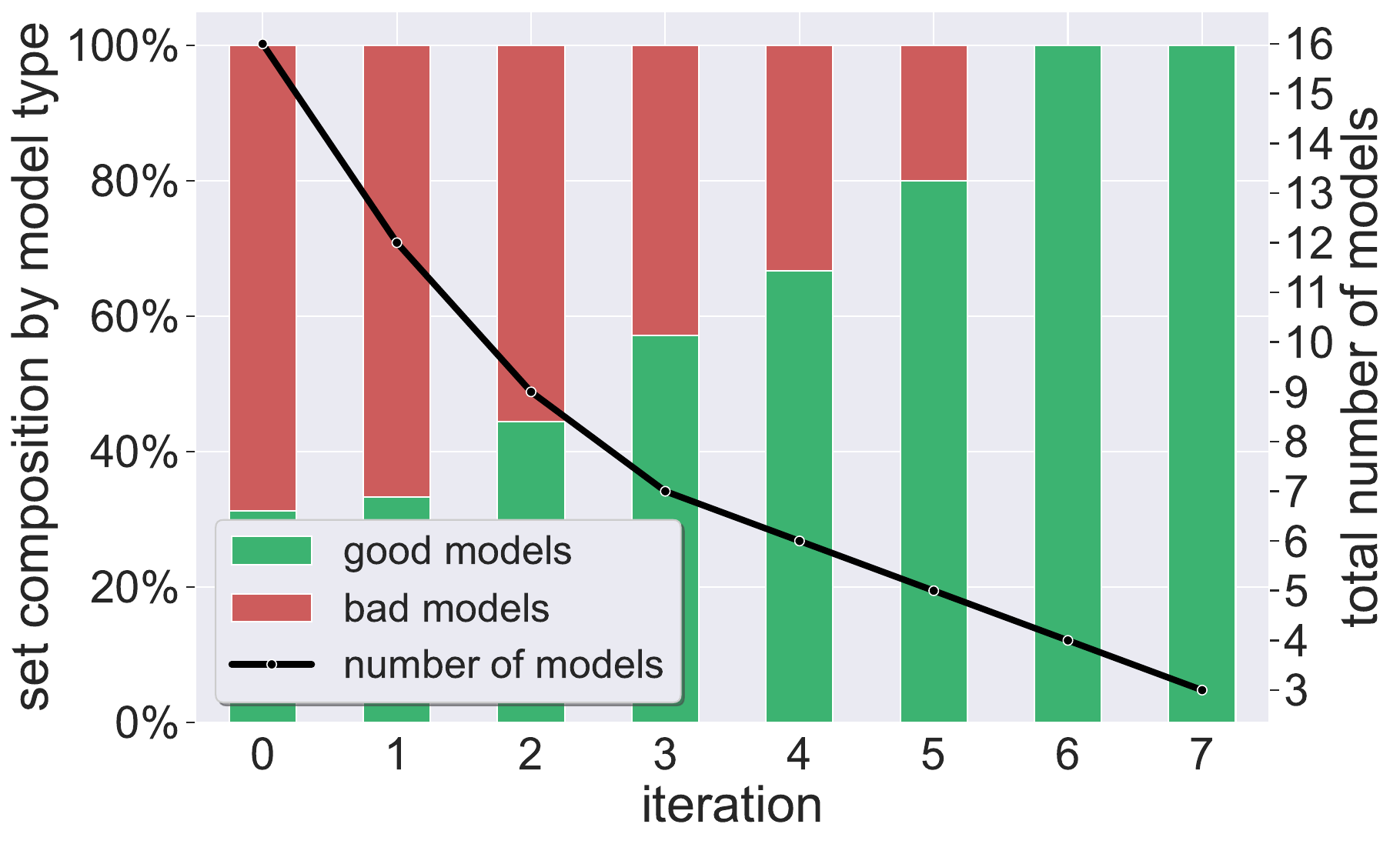}
		\end{center}
		\caption{Cartpole: Alg.~\ref{alg:modelSelection}'s results, per 
		iteration: the bars represent the ratio of good and bad models in the 
		surviving set (left y-axis), while the curve indicates the number of 
		surviving models (right y-axis). 
			Our technique  selected models \{6,7,9\}.}
		
		\label{fig:cartpolePercentileGoodBadResults}
	\end{figure}

	All remaining parameters in the OOD environment matched those used for the 
	original training. Figure~\ref{fig:cartpoleRewards} presents the outcomes 
	of evaluating the models on $20,000$ OOD instances. Out of the initial $16$ 
	models, $11$ achieved low to mediocre average rewards, demonstrating their 
	limited capacity to generalize to this new distribution. Only $5$ models 
	attained high reward values on the OOD domain, including the $3$ models 
	identified by our approach;
	thus indicating that our method successfully eliminated all $11$ models 
	that would have otherwise exhibited poor performance in this OOD setting
	(see Fig.~\ref{fig:cartpolePercentileGoodBadResults}).  For more
	information, we refer the reader to 
	Appendix~\ref{sec:appendix:CartPoleSupplementaryResults}.

	\subsection{Mountain Car}
	\label{subsec:mountaincar} 
	
\begin{figure}
		\centering
		\captionsetup[subfigure]{justification=centering}
		\captionsetup{justification=centering}
		\begin{subfigure}[t]{0.49\linewidth}
			\includegraphics[width=\textwidth]{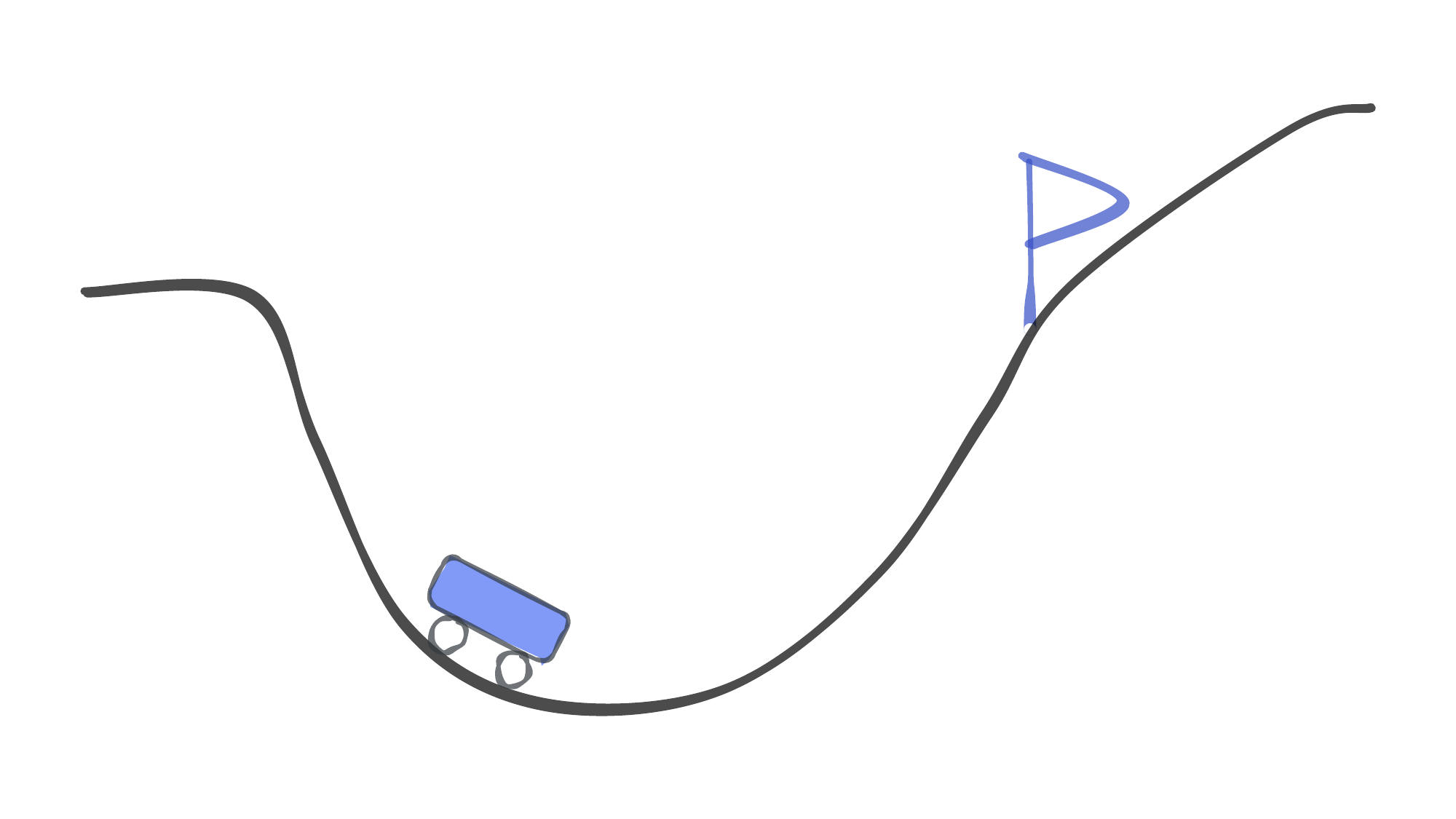}
			\caption{Training (in-distribution) setting: the agent's initial 
				point is on the left side, and the goal is at a nearby point}
			\label{fig:MountainCarInDistribution:a}
		\end{subfigure}
		\hfill
		\begin{subfigure}[t]{0.49\linewidth}
			\includegraphics[width=\textwidth]{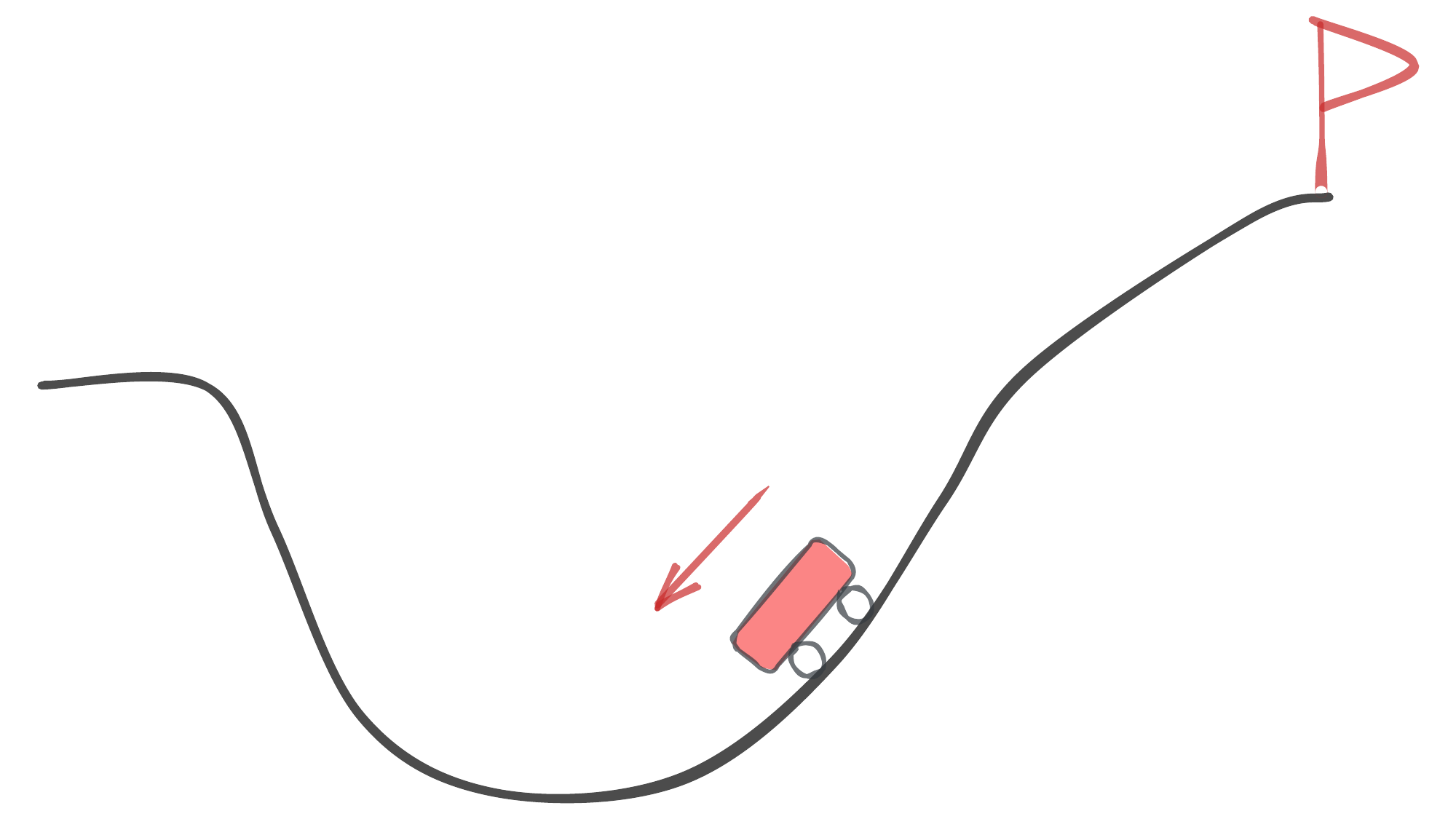}
			
			\caption{OOD setting: the agent's initial point is farther away, 
			with 
				initial negative velocity; the goal is significantly farther up 
				than 
				during training}
			\label{fig:MountainCarOod:b}
		\end{subfigure}
		
		\caption{Mountain Car: figure (a) depicts the setting in which the 
		agents 
			were trained, and figure (b) depicts the harder, OOD setting.}
		\label{fig:MountainCarInDistributionAndOod}
	\end{figure}

	For our second experiment, we evaluated our method on the
	Mountain Car~\cite{Ri05} benchmark, in which an agent controls a car
	that needs to learn how to escape a valley and reach a target (see 
	Fig.~\ref{fig:MountainCarInDistributionAndOod}).

	\medskip
	\noindent{\textbf{Agent and Environment.}}
	The car (agent) is placed in a valley between two hills (at $x\in[-1.2, 
	0.6]$), and 
	needs to reach a flag on top of one of the hills. The state, $s=(x, v_{x})$ 
	represents the car's location (along the x-axis) and velocity.
	The agent's action (output) is the \emph{applied force}: a continuous
	value indicating the magnitude and direction in which the agent wishes
	to move. During training, the agent is incentivized to reach the flag 
	(placed 
	at the top of a valley, originally at $x=0.45$). For each time-step until 
	the 
	flag is reached, the agent receives a small, negative reward; if it reaches 
	the 
	flag, the agent is rewarded with a large positive reward. An episode 
	terminates 
	when the flag is reached, or when the
	number of steps exceeds some predefined value ($300$ in our
	experiments). Good and bad models are distinguished by an average reward 
	threshold of \emph{90}.

	\medskip
	\noindent{\textbf{In-Distribution Inputs.}}
	During training (in-distribution), the car is initially placed on the 
	\emph{left} side
	of the valley's bottom, with a low, random velocity (see
	Fig.~\ref{fig:MountainCarInDistribution:a}). We trained $k=16$ agents 
	(denoted 
	as
	$\{1, 2, \ldots 16\}$), which all perform well, i.e., achieve an average 
	reward 
	higher than our
	threshold, in-distribution. This evaluation was conducted  over $10,000$ 
	episodes.

	
	\medskip
	\noindent{\textbf{(OOD) Input Domain.}}
	According to the scenarios used by the training environment, we specified 
	the 
	(OOD) input domain by:
	\begin{inparaenum}[(i)]
		\item extending the x-axis, from $[-1.2, 0.6]$ to $[-2.4,
		0.9]$;
		\item moving the flag further to the right, from $x=0.45$ to
		$x=0.9$; and 
		\item setting the car's initial location further to the right of
		the valley's bottom, and with a large initial \textit{negative}
		velocity (to the left).
	\end{inparaenum}
	An illustration appears in Fig.~\ref{fig:MountainCarOod:b}.  These new
	settings represent a novel state distribution, which causes the agents
	to respond to states that they had not observed during training:
	different locations, greater velocity, and different combinations of
	location and velocity directions.
	
\begin{figure}
		\centering
		\subfloat[In-distribution 
		\label{subfig:mountaincarRewards:inDist}]{\includegraphics[width=0.49\textwidth]{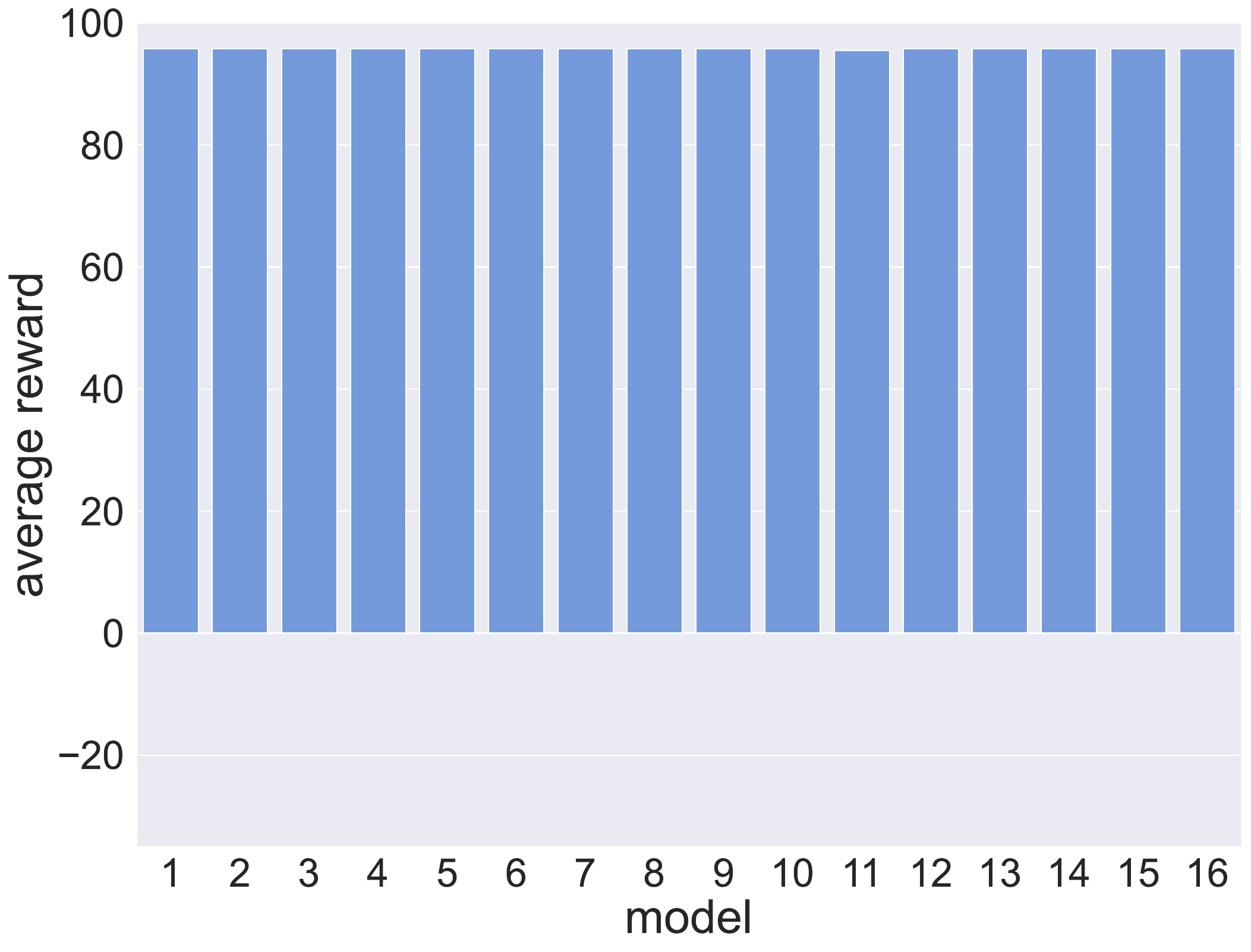}}
		\hfill
		\subfloat[OOD\label{subfig:mountaincarRewards:OOD}]{\includegraphics[width=0.49\textwidth]{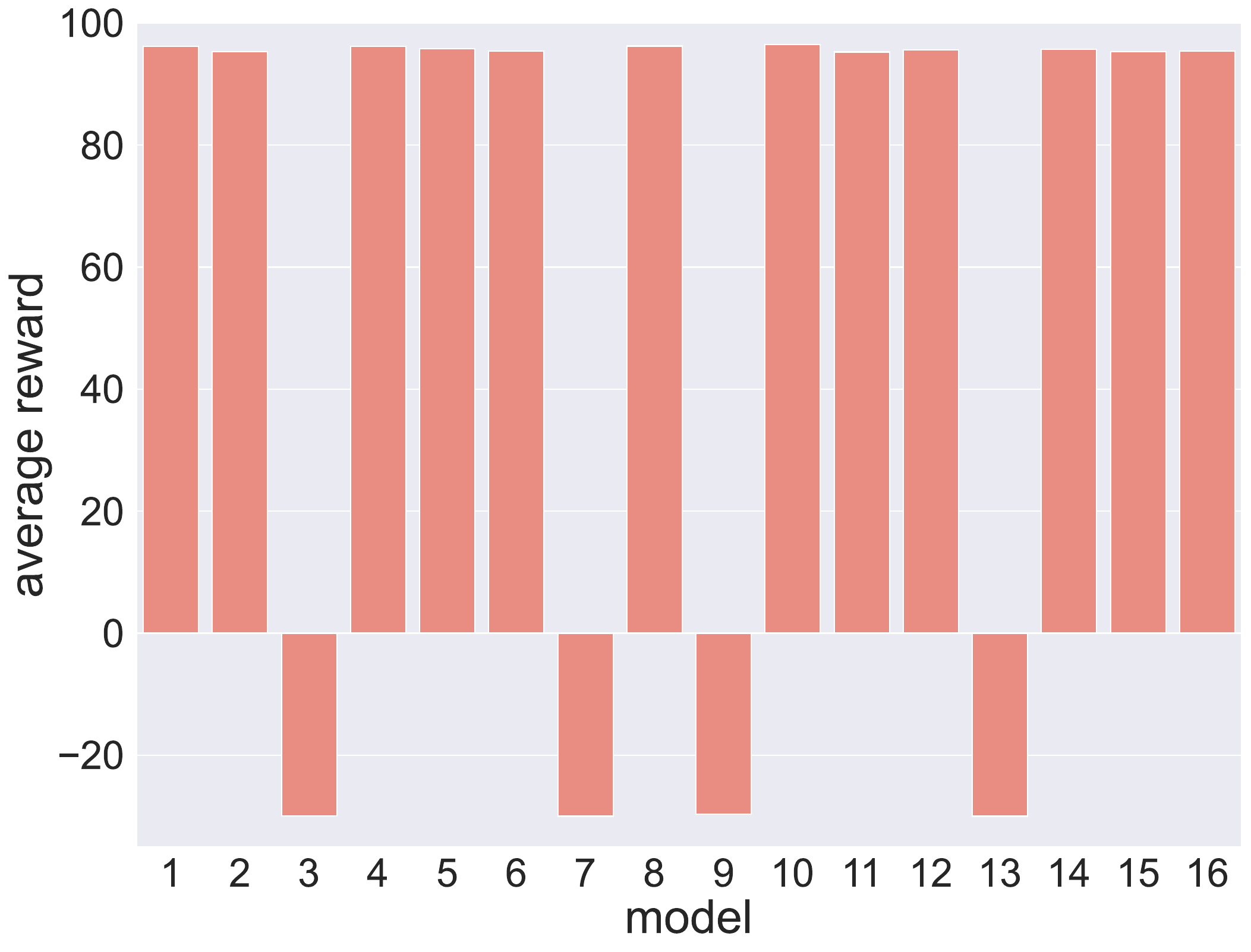}}\\
		
		\caption{Mountain Car: the models' average rewards in different 
			distributions.}\label{fig:mountaincarRewards}
	\end{figure}

	\medskip
	\noindent{\textbf{Evaluation.}}
	Out of the $k=16$ models that performed well in-distribution, $4$ models
	failed (i.e., did not reach the flag, ending their episodes with a negative
	average reward) in the OOD scenario, while the remaining $12$
	succeeded, i.e., reached a high average reward when simulated on the OOD 
	data 
	(see Fig.~\ref{fig:mountaincarRewards}).  
	The large ratio
	of successful models is not surprising, as Mountain Car is a relatively 
	easy 
	benchmark. 
	
	To evaluate our algorithm, we ran it on these models, and the 
	aforementioned 
	(OOD) input domain, and checked whether it removed the
	models that (although successful in-distribution) fail in the new, harder,
	setting. Indeed, our method was able to filter out all unsuccessful
	models, leaving only a subset of $5$ models ($\{2,4,8,10,15\}$), all of 
	which 
	perform well in the OOD scenario.
	For additional information, see 
	Appendix~\ref{sec:appendix:MountainCarSupplementaryResults}.

	
	
	\subsection{The Aurora Congestion Controller}\label{subsec:aurora}
	
	In the third benchmark, we applied our methodology to an intricate 
	system that enforces a policy for the real-world task of Internet 
	congestion control. 
	Congestion control aims to determine, for each traffic source in a 
	communication network, the appropriate rate at which data packets should be 
	dispatched into the network.
	Managing congestion is a notably challenging and fundamental issue in 
	computer networking~\cite{LoPaDo02, Na84}; transmitting packets too quickly 
	can result in network congestion, causing data loss and delays. Conversely, 
	employing low sending rates may result in the underutilization of available 
	network bandwidth.
	Developed by~\cite{JaRoGoScTa19}, Aurora is a DNN-based congestion 
	controller trained to optimize network performance. Recent research has 
	delved into formally verifying the reliability of DNN-based systems, with 
	Aurora serving as a key example~\cite{ElKaKaSc21,  AmScKa21}. Within each 
	time-step, an Aurora agent collects network statistics and determines the 
	packet transmission rate for the next time-step.
	For example, if the agent observes poor network
	conditions (e.g., high packet loss), we expect it to decrease the
	packet sending rate to better utilize the bandwidth.
	We note that Aurora handles a much harder task than the previous RL 
	benchmarks 
	(Cartpole and Mountain Car): congestion controllers must gracefully respond 
	to diverse potential events, interpreting nuanced signals presented by 
	Aurora's inputs. Unlike in prior benchmarks, determining the optimal policy 
	in this scenario is not a straightforward endeavor.

	\medskip
	\noindent{\textbf{Agent and Environment.}}
	Aurora receives as input an ordered set of $t$ vectors $v_{1}, 
	\ldots,v_{t}$, that collectively represent
	observations from the previous $t$ time-steps (each of the vectors 
	$v_{i}\in\mathbb{R}^3$ includes three
	distinct values that represent statistics on the network's condition, as 
	detailed in Appendix~\ref{sec:appendix:AuroraSupplementaryResults}).
	The agent has a single
	output indicating the change in the packet sending rate over the
	following time-step. In 
	line with~\cite{ElKaKaSc21,JaRoGoScTa19, AmScKa21}, we set $t=10$
	time-steps, hence making Aurora's inputs of dimension $3t=30$.
	During training, Aurora's reward function is a linear combination of
	the data sender's packet loss, latency, and throughput, as observed by
	the agent (see~\cite{JaRoGoScTa19} for more details).
	
\begin{figure}[h]
		\centering
		\captionsetup[subfigure]{justification=centering}
		\captionsetup{justification=centering}
		\begin{subfigure}[t]{0.48\linewidth}
			\includegraphics[width=\textwidth]{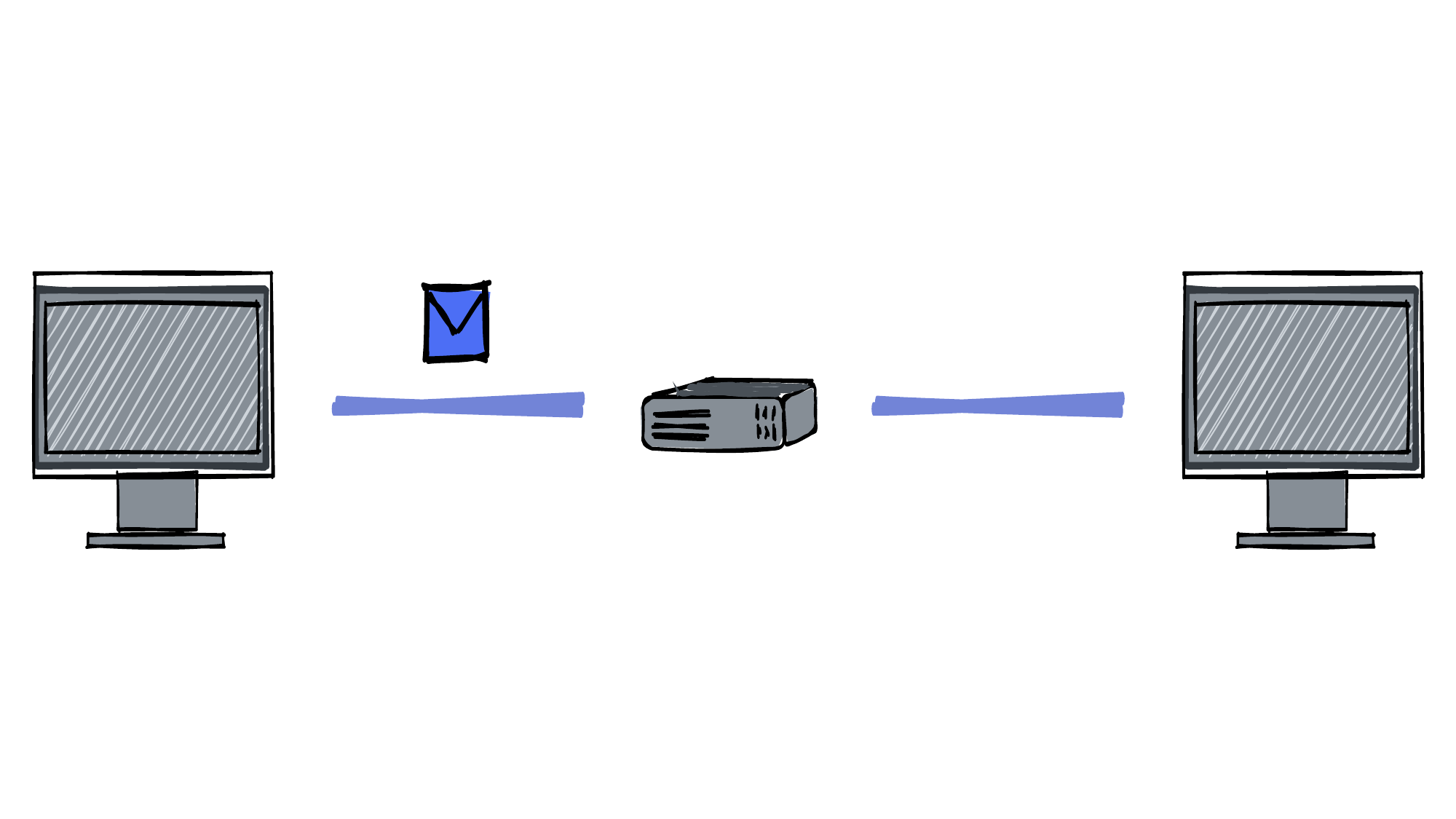}
			\caption{In-distribution setting, in which the agent was trained, 
			with 
				low packet congestion}
			\label{fig:AuroraInDistribution:a}
		\end{subfigure}
		\hfill
		\begin{subfigure}[t]{0.48\linewidth}
			\includegraphics[width=\textwidth]{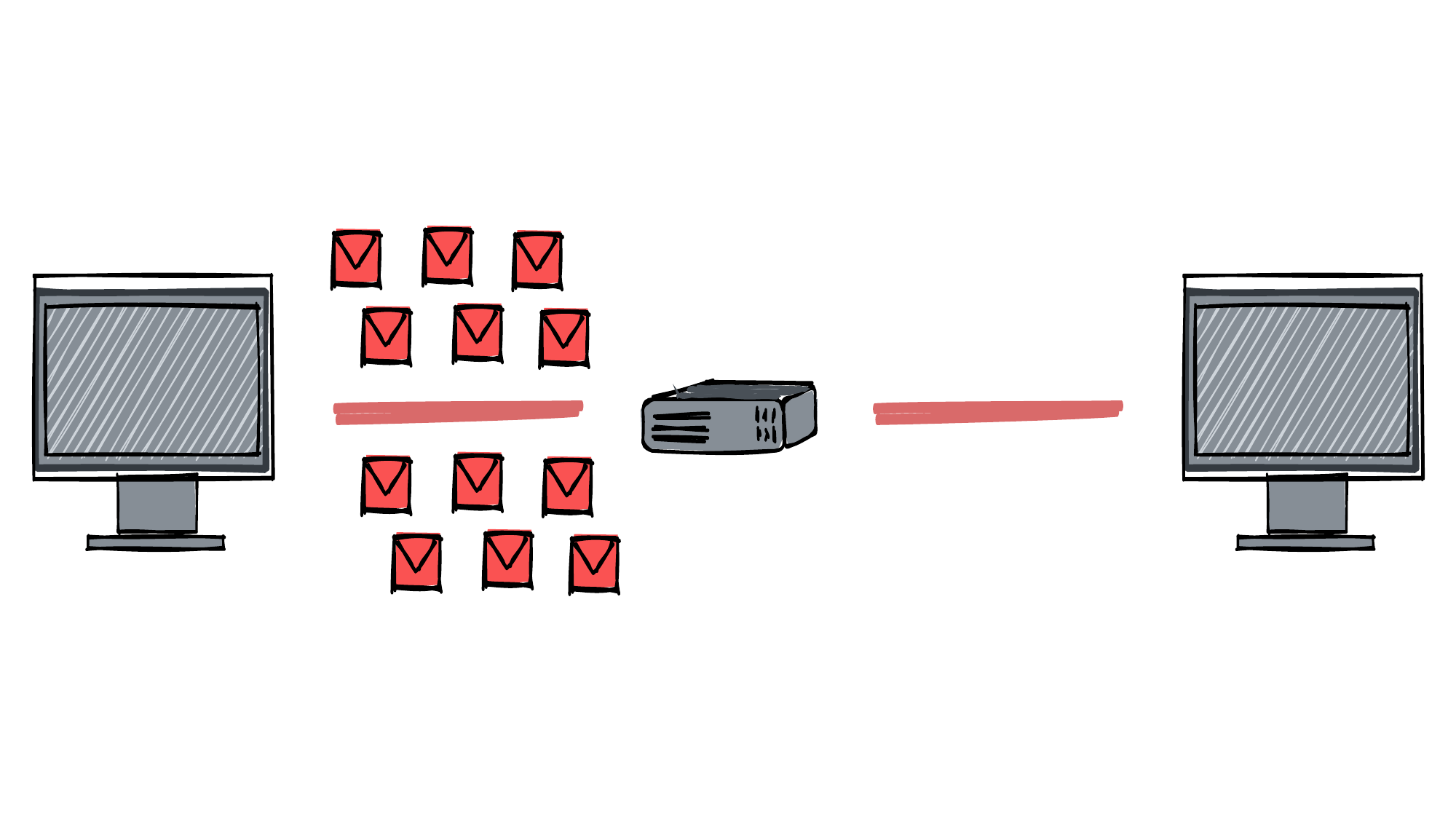}
			\caption{OOD settings, of a much higher congestion rate, with a 
				significantly greater packet loss}
			\label{fig:AuroraOod:b}
		\end{subfigure}
		
		\caption{Aurora: illustration of in-distribution and OOD settings.}
		\label{fig:AuroraInDistributionAndOod}
	\end{figure}

	\medskip
	\noindent{\textbf{In-Distribution Inputs.}} 
	During training, Aurora executes congestion control on basic network 
	scenarios --- a \emph{single} sender node 
	sends traffic to a \emph{single} receiver node across a \emph{single} 
	network 
	link. 
	Aurora undergoes training across a range of options for the initial sending 
	rate, link bandwidth, link packet-loss rate, link latency, and the size of 
	the link's packet buffer. During the training phase, data packets are 
	initially sent by Aurora at a rate that 
	corresponds to $0.3-1.5$ times the link's bandwidth, leading mostly to low 
	congestion, as depicted in Fig.~\ref{fig:AuroraInDistribution:a}.

	\medskip
	\noindent{\textbf{(OOD) Input Domain.}}
	In our experiments, the input domain represented a link with a 
	\emph{limited packet buffer}, indicating that the network can only store a 
	small number of packets (with most surplus traffic being discarded), 
	resulting in the link displaying erratic behavior. This is reflected in the 
	initial sending rate being set to up to $8$ times (!) the link's bandwidth, 
	simulating the potential for a significant reduction in available bandwidth 
	(for example, due to competition, traffic shifts, etc.).
	For additional details, see 
	Appendix~\ref{sec:appendix:AuroraSupplementaryResults}.

	\FloatBarrier
	
	\medskip
	\noindent{\textbf{Evaluation.}} 
	We executed our algorithm and evaluated the models by assessing their 
	disagreement upon this extensive domain, encompassing inputs that were not 
	encountered during training, and representing the aforementioned conditions 
	(depicted in Fig.~\ref{fig:AuroraOod:b}).
	
	\medskip
	\noindent

\begin{figure}[ht]
		\centering
		\captionsetup{justification=centering}
		\subfloat[In-distribution 
		\label{subfig:auroraRewards:inDist}]{\includegraphics[width=0.49\textwidth]{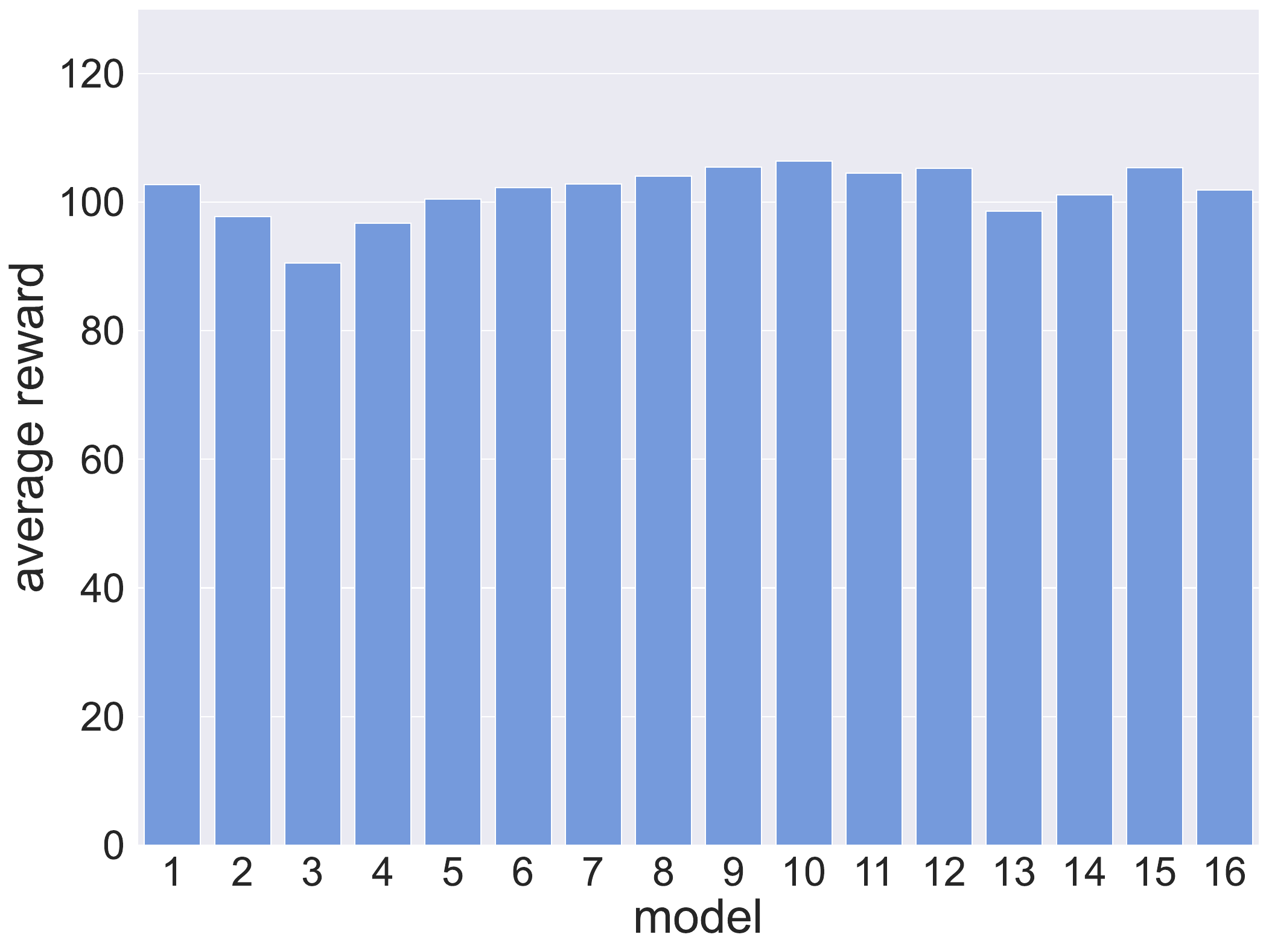}}
		\hfill
		\subfloat[OOD\label{subfig:auroraRewards:OOD}]{\includegraphics[width=0.49\textwidth]{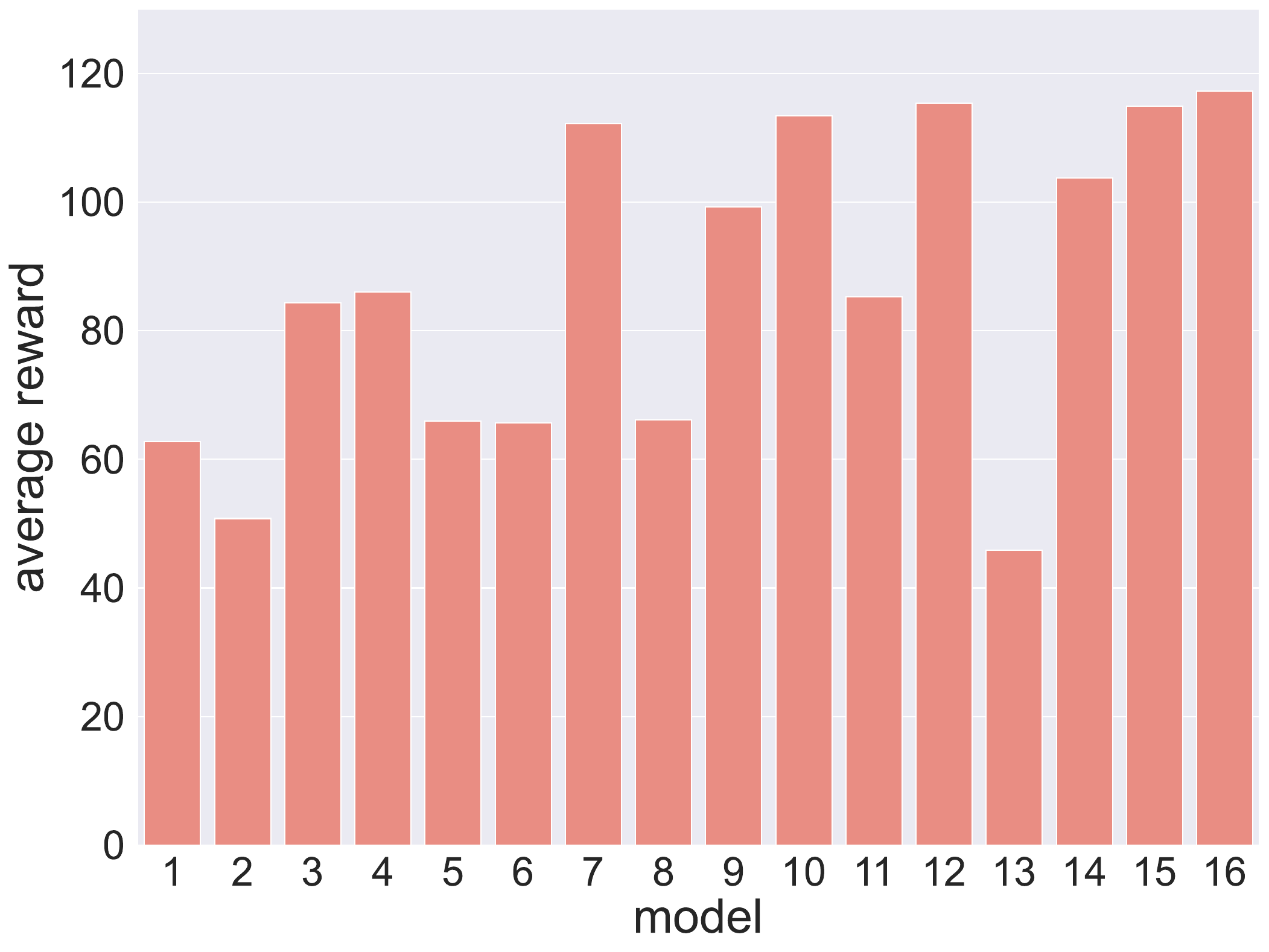}}\\
		
		\caption{Aurora Experiment~\ref{exp:auroraShort}: the models' average 
			rewards when simulated on different 
			distributions.}\label{fig:auroraRewards}
	\end{figure}

	\experiment{High Packet Loss.}\label{exp:auroraShort} 
	In this experiment, we trained more than $100$ Aurora agents in the 
	original (in-distribution) environment. From this pool, we chose $k=16$ 
	agents that attained a high average reward in the in-distribution setting 
	(see Fig.~\ref{subfig:auroraRewards:inDist}),
	as evaluated over
	$40,000$ episodes from the same distribution on which the models
	were trained. 
	Subsequently, we assessed these agents using out-of-distribution inputs 
	within the previously outlined domain. The primary distinction between the 
	training distribution and the new (OOD) inputs lies in the potential 
	occurrence of exceptionally high packet loss rates during initialization.
	
\begin{figure}[htbp]
		\centering
		\captionsetup[subfigure]{justification=centering}
		\captionsetup{justification=centering}
		\begin{subfigure}[t]{0.49\linewidth}
			\includegraphics[width=\textwidth, 
			height=0.67\textwidth]{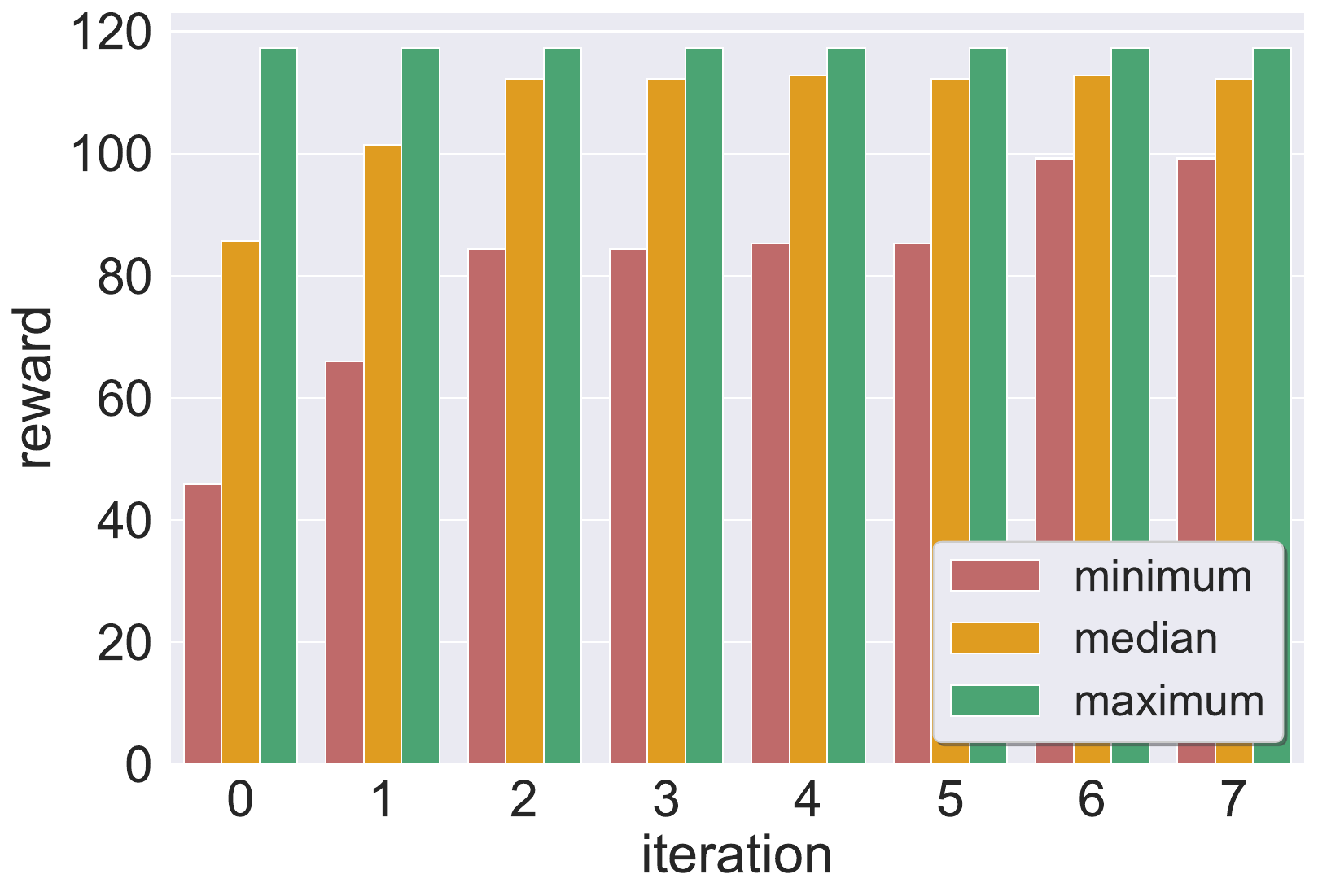}
			
			\caption{Reward statistics of remaining models}
			\label{fig:auroraShortMinMaxRewardsAlongIterations}
		\end{subfigure}
		\hfill
		\begin{subfigure}[t]{0.49\linewidth}
			\includegraphics[width=\textwidth, 
			height=0.67\textwidth]{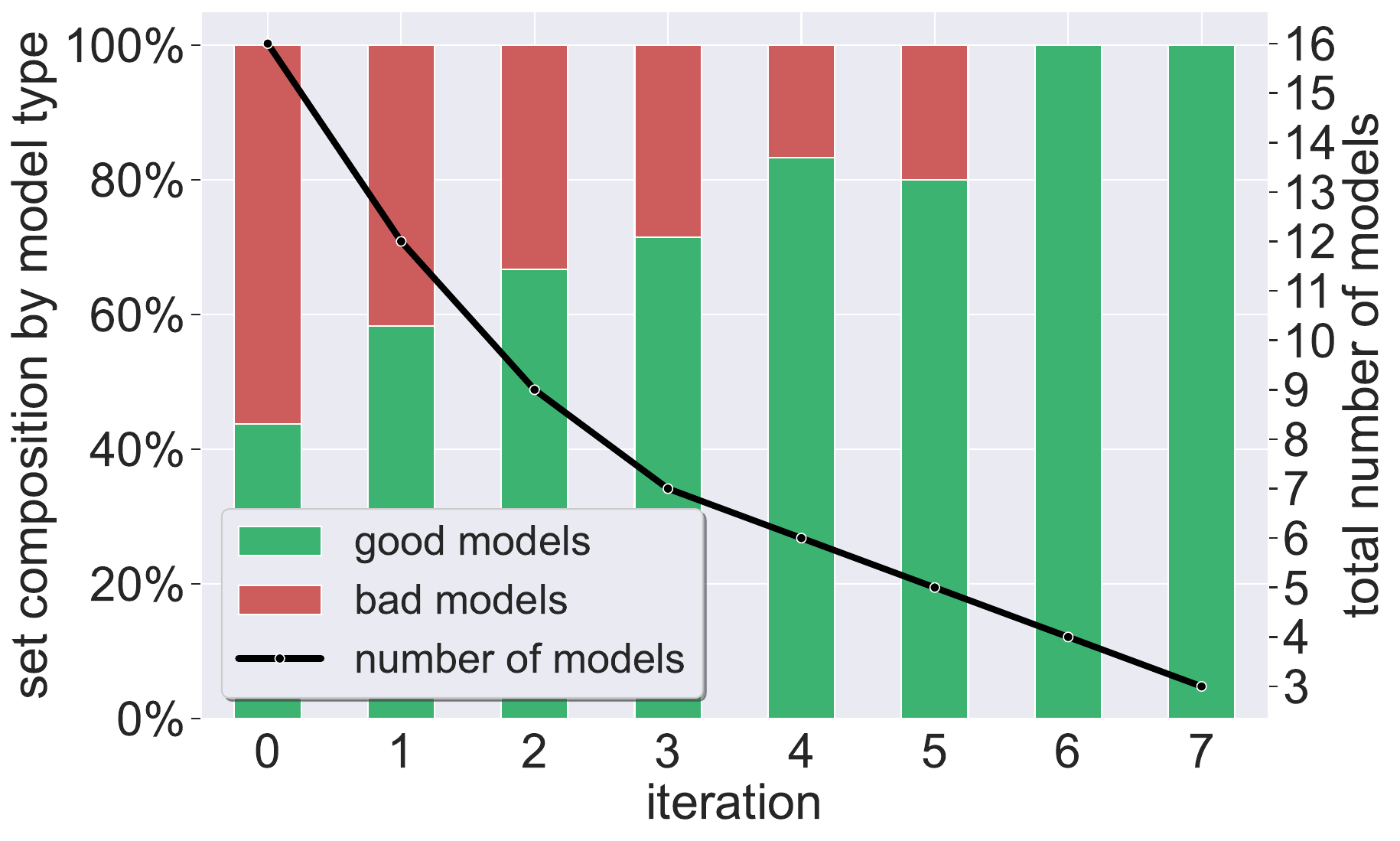}
			
			\caption{Ratio between good/bad models}
			\label{fig:auroraShortGoodBadModelsPercentages}
		\end{subfigure}
		
		\caption{Aurora: Alg.~\ref{alg:modelSelection}'s results, per iteration.
			Our technique selected models \{7,9,16\}.}
		\label{fig:AuroraShortTrainingMinMaxRewardAndGoodBadRatioPerIteration}
		
	\end{figure}

	\medskip
	\noindent
	Our assessment of out-of-distribution inputs within the domain reveals that 
	while all $16$ models excelled in the in-distribution setting, only $7$ 
	agents demonstrated the ability to effectively handle such OOD inputs (see 
	Fig.~\ref{subfig:auroraRewards:OOD}).
	When Algorithm~\ref{alg:modelSelection} was applied to the $16$ models, it 
	successfully identified and removed \emph{all} $9$ models that exhibited 
	poor generalization on the out-of-distribution inputs (see
	Fig.~\ref{fig:AuroraShortTrainingMinMaxRewardAndGoodBadRatioPerIteration}).
	Additionally, it is worth mentioning that during the initial iterations, 
	the four models chosen for exclusion were $\{1, 2, 6, 13\}$ --- which 
	constitute the poorest-performing models on the OOD inputs   (see 
	Appendix~\ref{sec:appendix:AuroraSupplementaryResults}).
	

	\medskip 
	\experiment{Additional Distributions over OOD Inputs.}
	To further demonstrate that our method is apt to retain superior-performing 
	models and eliminate inferior ones within the given input domain, we 
	conducted additional Aurora experiments by varying the distributions 
	(probability density functions) over the OOD inputs. Our assessment 
	indicates that all models filtered out by 
	Algorithm~\ref{alg:modelSelection} consistently exhibited low reward values 
	also for these alternative distributions
	(see
	Fig.~\ref{fig:auroraShortDifferentPdfGoodBadModelsPercentages} and
	Fig.~\ref{fig:auroraShortDifferentPdfGoodRewardsStats} in
	Appendix~\ref{sec:appendix:AuroraSupplementaryResults}). 
	These
	results highlight an important advantage of our approach: it applies
	to all inputs within the considered domain, and so it applies to \emph{all
		distributions over these inputs}. 
	We note again that 
	our model filtering process is based on verification queries in which
	the imposed bounds can represent \emph{infinitely} many distribution
	functions, on these bounds. In other words, our method, if correct, should 
	also apply to 
	additional OOD
	settings, beyond the ones we had originally considered, which share
	the specified input range but may include a different probability
	density function (PDF) over this range.
	

	
	\FloatBarrier

	\medskip
	\noindent{\textbf{Additional Experiments.}}
	We additionally created a fresh set of Aurora models by modifying the 
	training process to incorporate substantially longer interactions 
	(increasing from $50$ to $400$ steps). Subsequently, we replicated the 
	aforementioned experiments. The outcomes, detailed in 
	Appendix~\ref{sec:appendix:AuroraSupplementaryResults}, affirm that our 
	approach once again effectively identified a subset of models capable of 
	generalizing well to distributions across the OOD input domain.

	\subsection{Arithmetic DNNs}
	\label{subsec:arithmetic-dnns}

	In our last benchmark, we applied our approach to supervised-learning
	models, as opposed to models trained via DRL. In supervised learning,
	the agents are trained using inputs that have accompanying
	``ground truth'' results, per data point. Specifically, 
	we focused here on an Arithmetic DNNs benchmark, in which the DL
	models are trained to receive an input vector, and to approximate a
	simple arithmetic operation on some (or all) of the vector's entries.
	We note that this supervised-learning benchmark is considered quite
	challenging~\cite{TrHiReRaDyBl18, MaJo20}.

\begin{figure}[!h]
		\centering
		\begin{center}
			\includegraphics[width=0.65\textwidth]{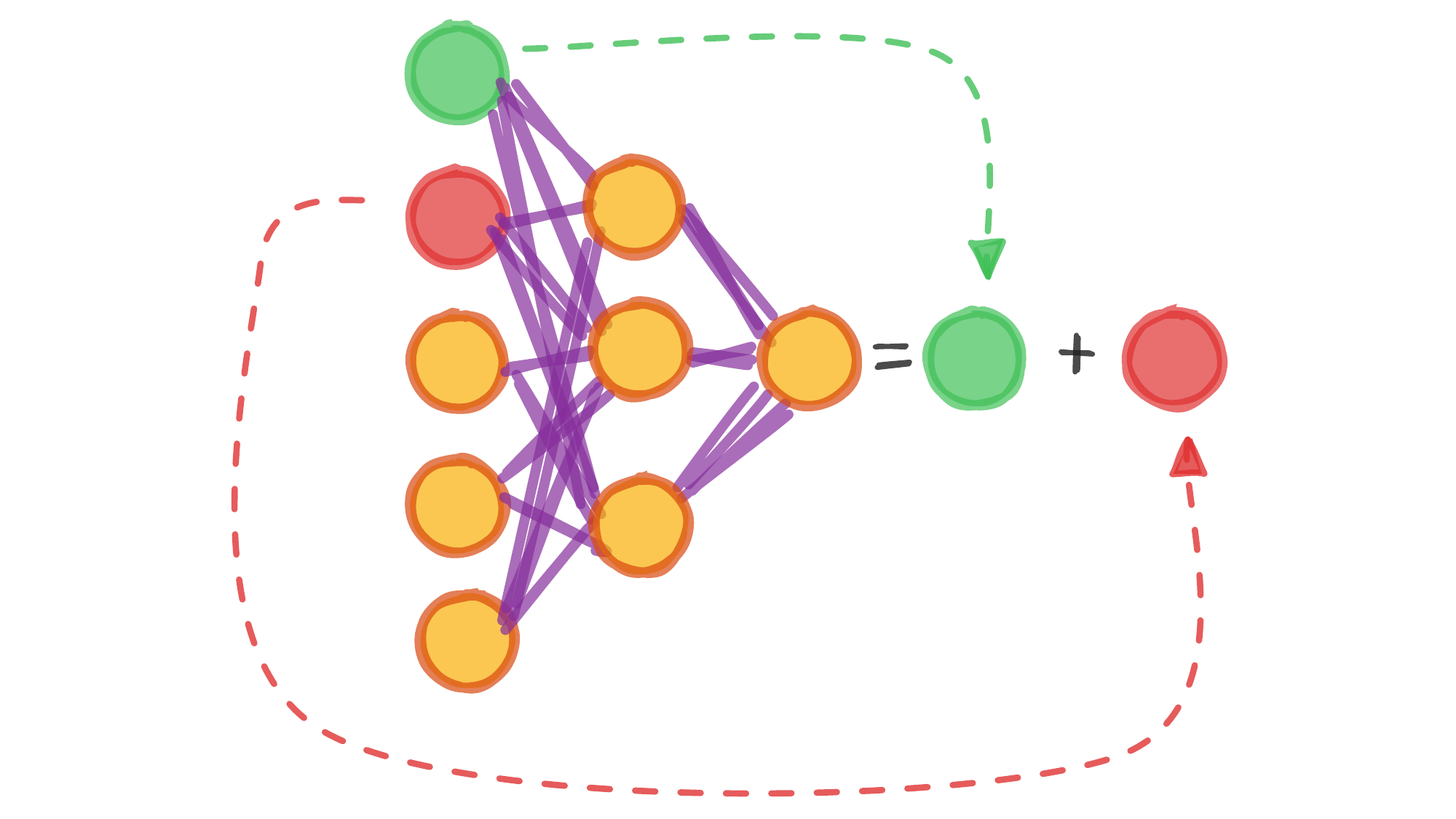}
		\end{center}
		
		\caption{A toy example of a DNN that performs simple arithmetic. The 
		DNN receives a d-dimensional input and learns to output a single value 
		which constitutes the sum of the first two inputs of the vector, while 
		ignoring the remaining (d-2) inputs.}
		\label{fig:arithmetic-dnns}
	\end{figure}

	\medskip
	\noindent{\textbf{Agent and Environment.}}
	We trained a DNN for the following supervised task. The input is a
	vector of size $10$ of real numbers, drawn uniformly at random from some 
	range $[l, u]$. The output is a single scalar, representing the \emph{sum} 
	of two hidden (yet consistent across the task) indices of the input vector; 
	in our case, the first $2$ input indices,
	as depicted in Fig.~\ref{fig:arithmetic-dnns}. 
	Differently put, the agent needs to learn to model the sum of the
	relevant (initially unknown) indices, while learning to ignore the rest of 
	the inputs. 
	We trained our networks for 10 epochs over a dataset consisting of
	$10,000$ input vectors drawn uniformly at random from the range $[l=-10, 
	u=10]$, using the Adam optimization algorithm~\cite{KiBa15} with a learning 
	rate of $\gamma = 0.001$ and using the mean squared error (MSE) loss 
	function.
	For additional details, see 
	Appendix~\ref{sec:appendix:trainingAndEvaluationArithmeticDnns}.

	\medskip
	\noindent{\textbf{In-Distribution Inputs.}} 
	During training, we presented the models with input values sampled from a 
	multi-modal uniform distribution [-10,10]$^{10}$, resulting in a single 
	output in the range [-20,20]. 
	As expected, the models performed well over this distribution, as depicted 
	in Fig.~\ref{subfig:arithmeticDnns:inDist} of the Appendix.
	
	\medskip
	\noindent{\textbf{(OOD) Input Domain.}} 
	A natural OOD distribution includes any $d$-dimensional multi-modal
	distribution, in which each input is drawn from a range different than
	$[l=-10, u=10]$ --- and hence, can necessarily be assigned values on
	which the model was not trained initially. In our case, we chose the 
	multi-modal distribution of $[l=-1,000, u=1,000]$$^{10}$. Unlike the case 
	for the in-distribution inputs, there was a high variance among the 
	performance of the models in this novel, unseen OOD setting, as depicted in 
	Fig.~\ref{subfig:arithmeticDnns:OOD} of the Appendix.

\begin{figure}[htbp]
		\centering
		\captionsetup[subfigure]{justification=centering}
		\captionsetup{justification=centering}
		\begin{subfigure}[t]{0.49\linewidth}
			\includegraphics[width=\textwidth, 
			height=0.67\textwidth]{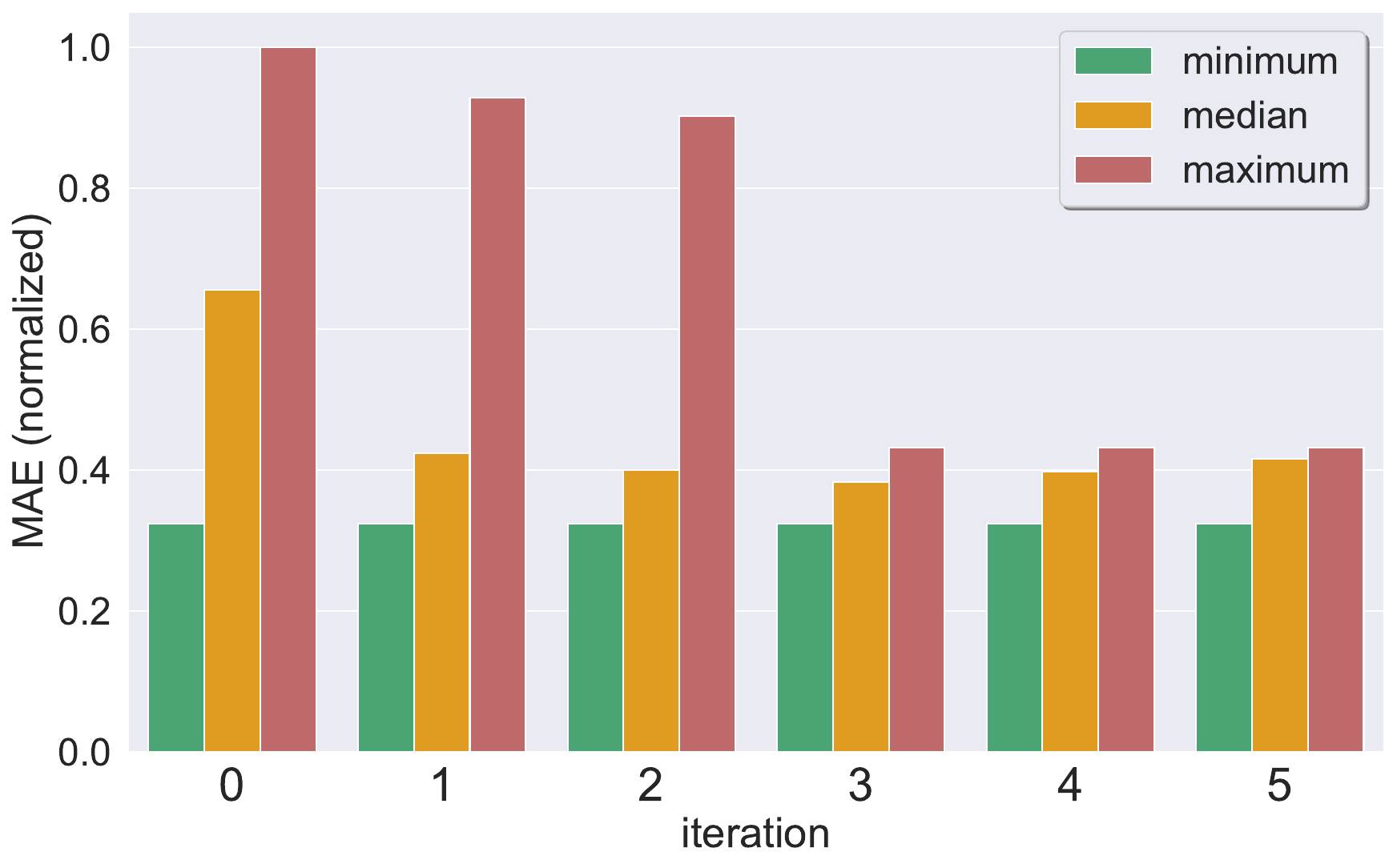}
			
			\caption{MAE statistics of remaining models}
			\label{fig:ArithmeticDnnsErrorAlongIterations}
		\end{subfigure}
		\hfill
		\begin{subfigure}[t]{0.49\linewidth}
			\includegraphics[width=\textwidth, 
			height=0.67\textwidth]{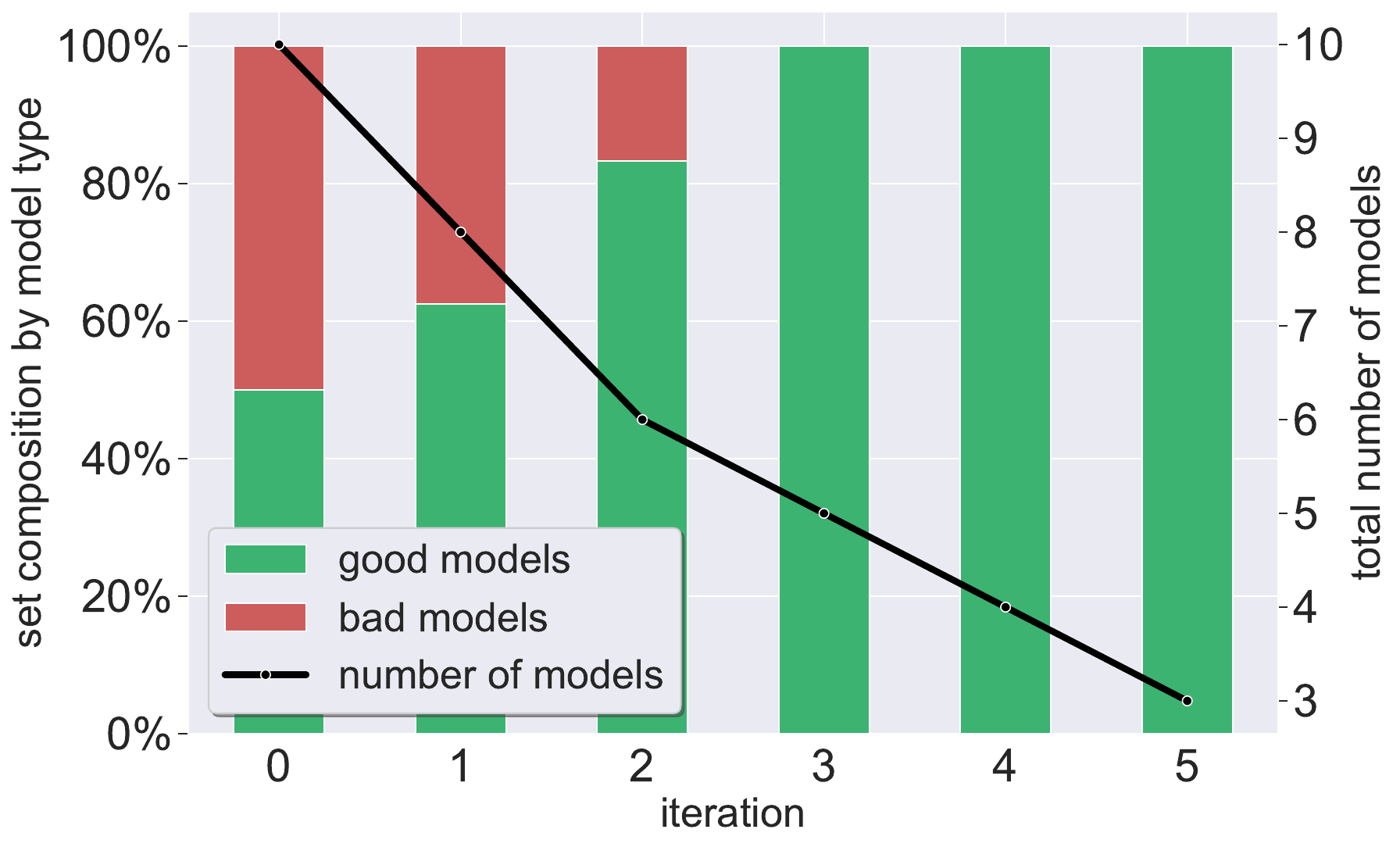}
			
			\caption{Ratio between good/bad models}
			\label{fig:ArithmeticDnnsShortGoodBadModelsPercentages}
		\end{subfigure}
		
		\caption{Arithmetic DNNs: Alg.~\ref{alg:modelSelection}'s results, per 
		iteration.
			Our technique selected models \{5,8,10\}.} 
		\label{fig:ArithmeticDnnsSetCompositionAndError}
		
	\end{figure}

	\medskip
	\noindent{\textbf{Evaluation.}} 
	We originally trained $n=50$ models. After validating that all models 
	succeed in-distribution, we generated a pool of $k=10$ models. This pool 
	was generated by collecting the five best and five worst models OOD (based 
	on 
	their maximal normalized error, over the same $100,000$ points sampled 
	OOD). We then executed our algorithm and checked whether it was able to 
	identify and remove all unsuccessful models, which consisted of half of the 
	original model pool. Indeed, as can be seen in 
	Fig.~\ref{fig:ArithmeticDnnsSetCompositionAndError}, all bad models were 
	filtered out within three iterations. After convergence, three models 
	remained 
	in the model pool, including model \{8\} --- which constitutes the best 
	model, OOD.
	This experiment was successfully repeated with additional filtering 
	criteria (see Fig.~\ref{fig:arithmeticDnns:MaxCritResults} 
	in Appendix~\ref{sec:appendix:arithmeticDNNs}).

	\subsection{Averaging the Selected Models}
	\label{subsec:generatingGoodEnsembles}
	
\begin{figure}[htbp]
		\centering
		\captionsetup[subfigure]{justification=centering}
		\captionsetup{justification=centering}
		\begin{subfigure}[t]{0.49\linewidth}
			\includegraphics[width=\textwidth, 
			height=0.67\textwidth]{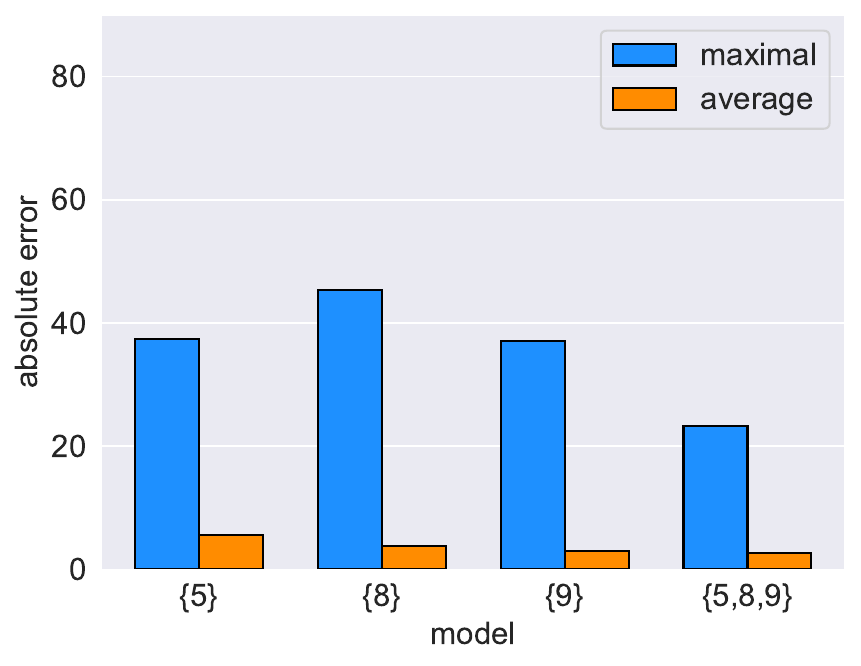}
			
			\caption{In-distribution, from left to right: the maximal absolute 
			errors are $37.44$, $45.33$, $37.01$, $23.28$, and the average 
			absolute errors are: $5.6$, $3.84$, $2.98$, $2.65$.}
			\label{fig:ArithmeticDnnsGoodEnsembleInDist}
		\end{subfigure}
		\hfill
		\begin{subfigure}[t]{0.49\linewidth}
			\includegraphics[width=\textwidth, 
			height=0.67\textwidth]{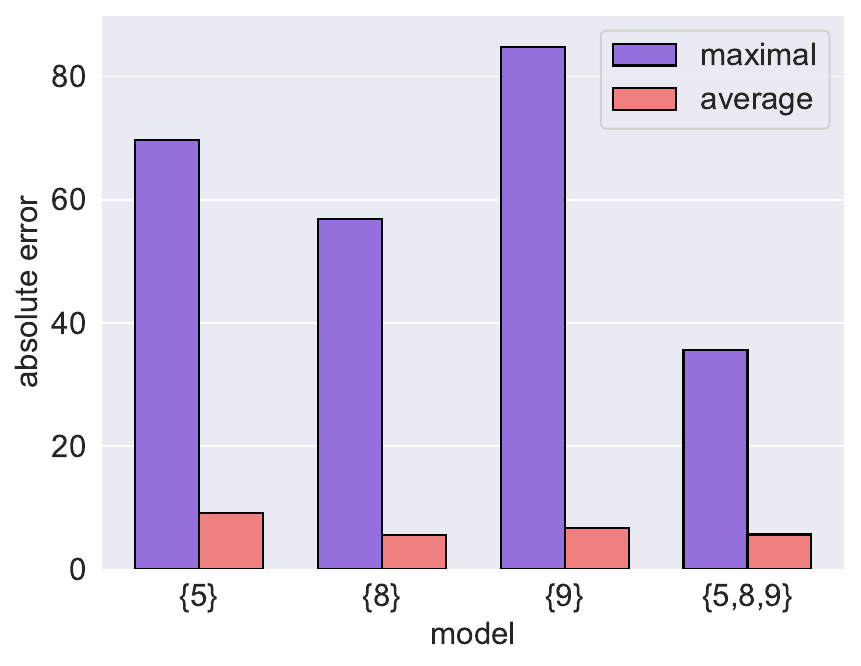}
			\caption{OOD, from left to right: the maximal absolute errors are 
			$69.7$, $56.9$, $84.81$, $35.59$, and the average absolute errors 
			are: $9.23$, $5.61$, $6.77$, $5.67$.}
			\label{fig:ArithmeticDnnsGoodEnsembleOOD}
		\end{subfigure}
		
		\caption{ 
			Arithmetic DNNs:
			Ensemble results. After Alg.~\ref{alg:modelSelection} selected 
			models \{$5$\}, \{$8$\}, and \{$9$\}, we generated an ensemble 
			\{$5$,$8$,$9$\} and sampled $5,000$ inputs in-distribution (a) and 
			OOD (b). 
			We note that we multiplied the errors attained in the 
			in-distribution experiment by $100$ in order to normalize the 
			selected range.} 
		\label{fig:ArithmeticDnnsEnsemblesOfGoodModels}
		
	\end{figure}

	To improve performance even further, it is possible to create (in 
	polynomial time) an ensemble of the surviving ``good'' models, instead of 
	selecting a single model. 
	As DNN robustness is linked to uncertainty, and due to the use of ensembles 
	as a prominent approach for uncertainty prediction, it has been shown that 
	averaging ensembles may improve performance~\cite{LaPrBl17}.
	For example, in the  Arithmetic DNNs benchmark, our approach eventually 
	selected three models  (\{$5$\}, \{$8$\}, and \{$9$\}, as depicted in 
	Fig.~\ref{fig:ArithmeticDnnsSetCompositionAndError}). 
	Subsequently,  we generated an ensemble comprised of these three DNN 
	models. Now, when the ensemble evaluates a given input, that 
	input is first independently passed to each of the ensemble members; and 
	the final ensemble prediction is the average of each of the members' 
	original outputs.
	We then sampled $5,000$ inputs drawn in-distribution (see 
	Fig.~\ref{fig:ArithmeticDnnsGoodEnsembleInDist}) and $5,000$ inputs drawn 
	OOD (see Fig.~\ref{fig:ArithmeticDnnsGoodEnsembleOOD}), and compared the 
	average and maximal errors of the ensemble on these sampled inputs to that 
	of its constituents. 
	In both cases, the ensemble had a maximal absolute error that was 
	significantly lower than each of its three constituent DNNs, as well as a 
	lower average error (with the sole exception of the average error OOD, 
	which was the second-smallest error, by a margin of only $0.06$). 
	Although the use of ensembles is not directly related to our approach, it 
	demonstrates how our technique can be extended and built upon additional 
	robustness techniques, for improving performance even further.


	
	\subsection{Analyzing the Eliminated Models}
	
	We conducted an additional analysis of the eliminated models, in order to 
	compare the average PDT scores of eliminated ``good'' models to those of 
	eliminated ``bad'' ones.
	For each of the five benchmarks, we divided the eliminated models into two 
	separate clusters, of either ``good'' or ``bad'' models (note that the 
	latter necessarily includes \emph{all} bad models, as in all our benchmarks 
	we return strictly ``good'' models).
	For each cluster, we calculated the average PDT score for all the DNN pairs.
	The results, summarized in 
	Table~\ref{tab:GoodAndBadEliminatedModelClusters}, demonstrate a clear 
	decrease in the average PDT score among the cluster of DNN pairs comprising 
	successful models, compared to their peers. This trend is observed across 
	all benchmarks, resulting in an average PDT score difference of between 
	$21.2\%$ to $63.2\%$, between the clusters, per benchmark.
	We believe that these results further support our hypothesis that good 
	models tend to make similar decisions.

	\begin{table}[htbp]
		\captionsetup{justification=centering}
		\centering
		\renewcommand{\arraystretch}{1.5} 
		\begin{tabular}{c|*{5}{c}}
			\hline\hline
			& \texttt{\textbf{cartpole}} & \texttt{\textbf{mountain car}} & 
			\texttt{\textbf{aurora (short)}} & \texttt{\textbf{aurora (long)}} 
			& \texttt{\textbf{arithmetic}} \\
			\hline
			\textbf{\textcolor{green}{good} models PDT}  & \textcolor{green}{4} 
			& \textcolor{green}{1.39} & \textcolor{green}{20.67} & 
			\textcolor{green}{9.32} & \textcolor{green}{180} \\
			\hdashline
			\textbf{\textcolor{red}{bad} models PDT}  & \textcolor{red}{8.44} & 
			\textcolor{red}{2.08} & \textcolor{red}{26.22} & 
			\textcolor{red}{25.3} & \textcolor{red}{299} \\
			\hline
			\textbf{ratio} & \textbf{47.4\%} & \textbf{66.6\%} & 
			\textbf{78.8\%} & \textbf{36.8\%} & \textbf{60.2\%} \\
			\hline
		\end{tabular}
		\caption{Evaluating the eliminated models throughout all benchmarks. 
		Columns (from left to right) represent the benchmarks: Cartpole, 
		Mountain Car, Aurora (short training), Aurora (long training), and 
		Arithmetic DNNs. The first row (in \textcolor{green}{green}) represents 
		the average PDT score for pairs of ``good'' models, and the second row 
		(in \textcolor{red}{red}) represents the average PDT score for pairs of 
		``bad'' models. The third row represents the ratio between the average 
		PDT scores of both clusters, per benchmark.}
		\label{tab:GoodAndBadEliminatedModelClusters}
	\end{table}

	\section{Comparison to Gradient-Based Methods \& Additional Techniques}
	\label{sec:competingMethods}

	

	The methods presented in this paper build upon DNN verification 
	(e.g., Line~\ref{line:SMTsolverForPdt} in 
	Alg.~\ref{alg:algorithmPairDisagreementScores}) in order to solve the 
	following 
	optimization problem: \emph{given a pair of DNNs, an input domain, 
		and a distance function, what is the maximal distance between their 
		outputs?} 
	In other words, verification is rendered to find an input that maximizes 
	the 
	difference between the outputs of two neural networks, under certain 
	constraints. Although DNN verification requires significant computational 
	resources~\cite{KaBaDiJuKo21}, nonetheless, we 
	demonstrate that it is crucial in our setting.  To support this claim, we 
	show the 
	results of our method when verification is replaced with other, more 
	scalable, techniques, such as gradient-based 
	algorithms (``attacks'')~\cite{Ru16,MaDiMe20, KuGoBe16}.
	In recent years, these optimization techniques have become popular due to 
	their simplicity and scalability, albeit the trade-off of inherent 
	incompleteness and reduced precision~\cite{WuZeKaBa22, AmZeKaSc22}. 
	As we demonstrate next, when using gradient-based methods (instead of 
	verification), at times, suboptimal PDT values were computed. This, in 
	turn, resulted in retaining unsuccessful models, which were successfully 
	removed when using DNN verification.

	\subsection{Comparison to Gradient-Based Methods}
	
	For our comparison, we 
	generated three gradient attacks:
	\begin{itemize}
		\item \emph{Gradient attack \# 1}: a \emph{Non-Iterative Fast Gradient 
			Sign Method (FGSM)}~\cite{HuPaGoDuAb17} attack, used when 
			optimizing linear 
		constraints, e.g., \emph{$L_{1}$ norm}, as in the case of 
		Aurora and Arithmetic DNNs;
		
		\item \emph{Gradient attack \# 2}: an \emph{Iterative 
			PGD}~\cite{MaMaScTsVl17} attack, also used when optimizing linear 
		constraints. 
		We note that we used this attack in cases where the previous attack 
		failed.
		
		\item \emph{Gradient attack \# 3}: a \emph{Constrained Iterative 
			PGD}~\cite{MaMaScTsVl17} attack, used in the case of encoding 
			non-linear 
		constraints (e.g., \emph{c-distance} functions; see 
		Sec.~\ref{sec:Approach}), as in the case of 
		Cartpole and Mountain Car. 
		This attack is a modified version of popular gradient attacks, that 
		were altered in order for them to succeed in our setting. 

	\end{itemize}
	
	\medskip
	\noindent
	Next, we formalize these attacks as constrained optimization problems. 
	
	\subsection{Formulation}
	Given an input domain $\mathcal{D}$, an output space 
	$\outputSpace=\mathbb{R}$, and  a pair of neural networks $N_1: \mathcal{D} 
	\to \mathbb{R}$ and $N_2: \mathcal{D} \to \mathbb{R}$, we wish to 
	find an input $\boldsymbol{x}\in\mathcal{D}$ that maximizes the difference 
	between the outputs of these neural networks. 
	
	Formally, in the case of the $L_{1}$ norm, we wish to solve the following 
	optimization problem:
	\begin{alignat*}{2}
		&\underset{x}{\textbf{max}} \quad &&|N_1(\boldsymbol{x}) - 
		N_2(\boldsymbol{x})| 
		\\
		&\textbf{s.t.} &&\boldsymbol{x}\in\mathcal{D}
	\end{alignat*}
	
	\subsubsection{Gradient Attack \# $1$}
	In cases where only input constraints are present, a local maximum can be 
	obtained via 
	conventional gradient attacks, that maximize the following objective 
	function:
	\[L(\boldsymbol{x}) = |N_1(\boldsymbol{x}) - N_2(\boldsymbol{x})|\]
	by taking steps in the direction of its gradient, and projecting them into 
	the 
	domain $\mathcal{D}$, that is:
	\begin{align*}
		&\boldsymbol{x_0} \in \mathcal{D} \\
		&\boldsymbol{x}_{t+1} = [\boldsymbol{x}_t + \epsilon \cdot 
		\nabla_{\boldsymbol{x}} L(\boldsymbol{x}_t)]_{\mathcal{D}}
	\end{align*}
	Where $[\cdot]_\mathcal{D}: \mathbb{R}^n \to \mathcal{D}$  projects the 
	result 
	onto $\mathcal{D}$, and $\epsilon$ being the step size.
	We note that $[\cdot]_\mathcal{D}$ may be non-trivial to implement, however 
	for our cases, in which each input of the DNN is encoded as some range, 
	i.e., 
	$\mathcal{D} \equiv \{\boldsymbol{x}\ |\ x\in\mathbb{R}^n \,\, \forall 
	i\in[n]: 
	l_i \leq x_i \leq u_i \}$, this can be implemented by clipping every 
	coordinate to 
	its appropriate range, and $\boldsymbol{x_0}$ can be obtained by taking 
	$\boldsymbol{x_0} = \frac{\boldsymbol{l} + \boldsymbol{u}}{2}$.
	
	\medskip
	\noindent
	In our context, the gradient attacks maximize a loss function for a pair of 
	DNNs, relative to their input. The popular FGSM attack  (\emph{gradient 
	attack \# 1}) achieves this by moving 
	in a single step toward the direction of the gradient. This simple attack 
	has been shown to be quite efficient in causing 
	misclassification~\cite{HuPaGoDuAb17}. In our setting, we can formalize 
	this (projected) FGSM as follows:
	\begin{algorithm}[H]
		\caption{FGSM} 
		\hspace*{\algorithmicindent} \textbf{Input:} objective $L$, variables 
		$\boldsymbol{x}$, input domain $\mathcal{D}$:(\init, \project), step 
		size 
		$\epsilon$ \\
		\hspace*{\algorithmicindent} \textbf{Output:} adversarial input 
		$\boldsymbol{x}$
		\begin{algorithmic}[1]
			\State $\boldsymbol{x_0} \gets \Call{Init}{\mathcal{D}}$
			\State $\boldsymbol{x}_{\textbf{adv}} \gets 
			\Call{Project}{\boldsymbol{x}_0 + \epsilon \cdot 
				\textbf{sign}(\nabla_{\boldsymbol{x}} L(\boldsymbol{x}_0))}$
			\State \Return{$\boldsymbol{x}_{\textbf{adv}}$}
		\end{algorithmic}
	\end{algorithm}
	
	\noindent
	In the context of our algorithms, we define $\mathcal{D}$ by two 
	functions:  
	\init, which returns an initial value from $\mathcal{D}$; and \project, 
	which 
	implements $[\cdot]_\mathcal{D}$. 
	
	
	\subsubsection{Gradient Attack \# $2$}
	
	A more powerful extension of this attack is the PGD algorithm, which we 
	refer 
	to as \emph{gradient attack \# 2}. This attack \emph{iteratively} moves in 
	the 
	direction of the gradient, often yielding superior results when compared to 
	its 
	single-step (FGSM) counterpart. The attack can be formalized as follows:
	\begin{algorithm} [H]
		\caption{PGD (maximize)}
		\hspace*{\algorithmicindent} \textbf{Input:} objective $L$, variables 
		$\boldsymbol{x}$, input domain $\mathcal{D}$:(\init, \project), 
		iterations 
		$T$, step size $\epsilon$ \\
		\hspace*{\algorithmicindent} \textbf{Output:} adversarial input 
		$\boldsymbol{x}$
		\begin{algorithmic}[1]
			\State $\boldsymbol{x_0} \gets \init({\mathcal{D}})$
			\For{$t=0 \ldots T-1$}
			\State $\boldsymbol{x}_{t+1} \gets \project({\boldsymbol{x}_t + 
				\epsilon \cdot \nabla_{\boldsymbol{x}} L(\boldsymbol{x}_t)})$ 
			\Comment{$\textbf{sign}(\nabla_{\boldsymbol{x}} 
				L(\boldsymbol{x}_t))$ may also be used}
			\EndFor
			\State \Return{$\boldsymbol{x}_{T}$}
		\end{algorithmic}
	\end{algorithm}
	We note that the case for using PGD in order to \emph{minimize} the 
	objective  
	function is symmetric.
	
	\subsubsection{Gradient Attack \# $3$}
	In some cases, the gradient attack needs to optimize a loss function 
	that represents constraints on the \emph{outputs} of the DNN pairs as well. 
	For example, in the case of the Cartpole and Mountain Car benchmarks, we 
	used the c-distance function. 
	In this scenario, we may need to 
	encode constraints of the form:
	
	\begin{align*}
		&N_1(\boldsymbol{x}) \leq 0\\
		&N_2(\boldsymbol{x}) \leq 0
	\end{align*}
	resulting in the following \emph{constrained} optimization problem:
	\begin{alignat*}{2}
		& \underset{x}{\textbf{max}} \quad &&|N_1(\boldsymbol{x}) - 
		N_2(\boldsymbol{x})| \\
		& \textbf{s.t.} &&\boldsymbol{x}\in\mathcal{D} \\
		& &&N_1(\boldsymbol{x}) \leq 0 \\
		& &&N_2(\boldsymbol{x}) \leq 0
	\end{alignat*}
	
	However, conventional gradient attacks are typically not geared for solving 
	such 
	optimizations. Hence, we tailored an additional gradient attack 
	(\emph{gradient 
		attack \# 3}) that can efficiently bridge this gap, and optimize the 
		aforementioned constraints by combining our Iterative PGD attack 
	with  \emph{Lagrange Multipliers}~\cite{Ro93} $\boldsymbol{\lambda} \equiv 
	(\lambda^{(1)}, \lambda^{(2)})$, hence allowing to penalize solutions for 
	which the constraints 
	do not hold. 
	To this end, we introduce a novel objective function:
	\[
	L_{-}(\boldsymbol{x},\boldsymbol{\lambda}) = |N_1(\boldsymbol{x}) - 
	N_2(\boldsymbol{x})| - \lambda^{(1)} \cdot \text{ReLU}(N_1(\boldsymbol{x})) 
	- 
	\lambda^{(2)} \cdot \text{ReLU}(N_2(\boldsymbol{x}))
	\]
	resulting in the following optimization problem:
	\begin{alignat*}{2}
		& \underset{x}{\textbf{max}}  \, \underset{\lambda}{\textbf{min}} \quad 
		&&L_{-}(\boldsymbol{x},\boldsymbol{\lambda}) \\
		& \textbf{s.t.} && \boldsymbol{x}\in\mathcal{D} \\
		& &&\lambda^{(1)} \geq 0 \\
		& &&\lambda^{(2)} \geq 0
	\end{alignat*}
	Next, we implemented a Constrained Iterative PGD algorithm that 
	approximates a 
	solution to this optimization problem:
	
	\begin{algorithm} [H]
		\caption{Constrained Iterative PGD}
		\hspace*{\algorithmicindent} \textbf{Input:} objective $L$, input 
		domain 
		$\mathcal{D}$, constraints: $C_i(\boldsymbol{x})$, iterations: $T, T_x, 
		T_\lambda$, step sizes: $\epsilon_x, \epsilon_\lambda$ \\
		\hspace*{\algorithmicindent} \textbf{Output:} adversarial input 
		$\boldsymbol{x}$
		\begin{algorithmic}[H]
			\State $\boldsymbol{x_0} \gets \Call{Init}{\mathcal{D}}$
			\State $L_{C}(\boldsymbol{x},\boldsymbol{\lambda}) \equiv 
			L(\boldsymbol{x}) - \sum^k_{i=0}\lambda^{(i)} \cdot 
			\text{ReLU}(C_i(\boldsymbol{x}))$ \Comment{the new objective}
			\For{$t=0 \ldots T-1$}
			\State $\boldsymbol{\lambda}_{t+1} \gets \Call{PGD\_min}{L_C, 
				\boldsymbol{\lambda}, ( \boldsymbol{\lambda} \gets 0, 
				\boldsymbol{\lambda} \geq 0), T_{\lambda}, \epsilon_{\lambda}}$ 
			\Comment{minimize $L_C(\boldsymbol{x}_t, \boldsymbol{\lambda})$ 
				with $\boldsymbol{x}_t$ as constant}
			\State $\boldsymbol{x}_{t+1} \gets \Call{PGD\_max}{L_C, 
				\boldsymbol{x}, \mathcal{D}, T_{x}, \epsilon_{x}}$ 
			\Comment{maximize $L_C(\boldsymbol{x}, \boldsymbol{\lambda}_t)$ 
				with $\boldsymbol{\lambda}_t$ as constant}
			\EndFor
			\State \Return{$\boldsymbol{x}_{T}$}
		\end{algorithmic}
	\end{algorithm}
	
	\subsection{Results}
\begin{figure}[!ht]
		\centering
		\captionsetup[subfigure]{justification=centering}
		\captionsetup{justification=centering} 
		\begin{subfigure}[t]{0.49\linewidth}
			\centering
			\includegraphics[width=\textwidth, 
			height=0.67\textwidth]{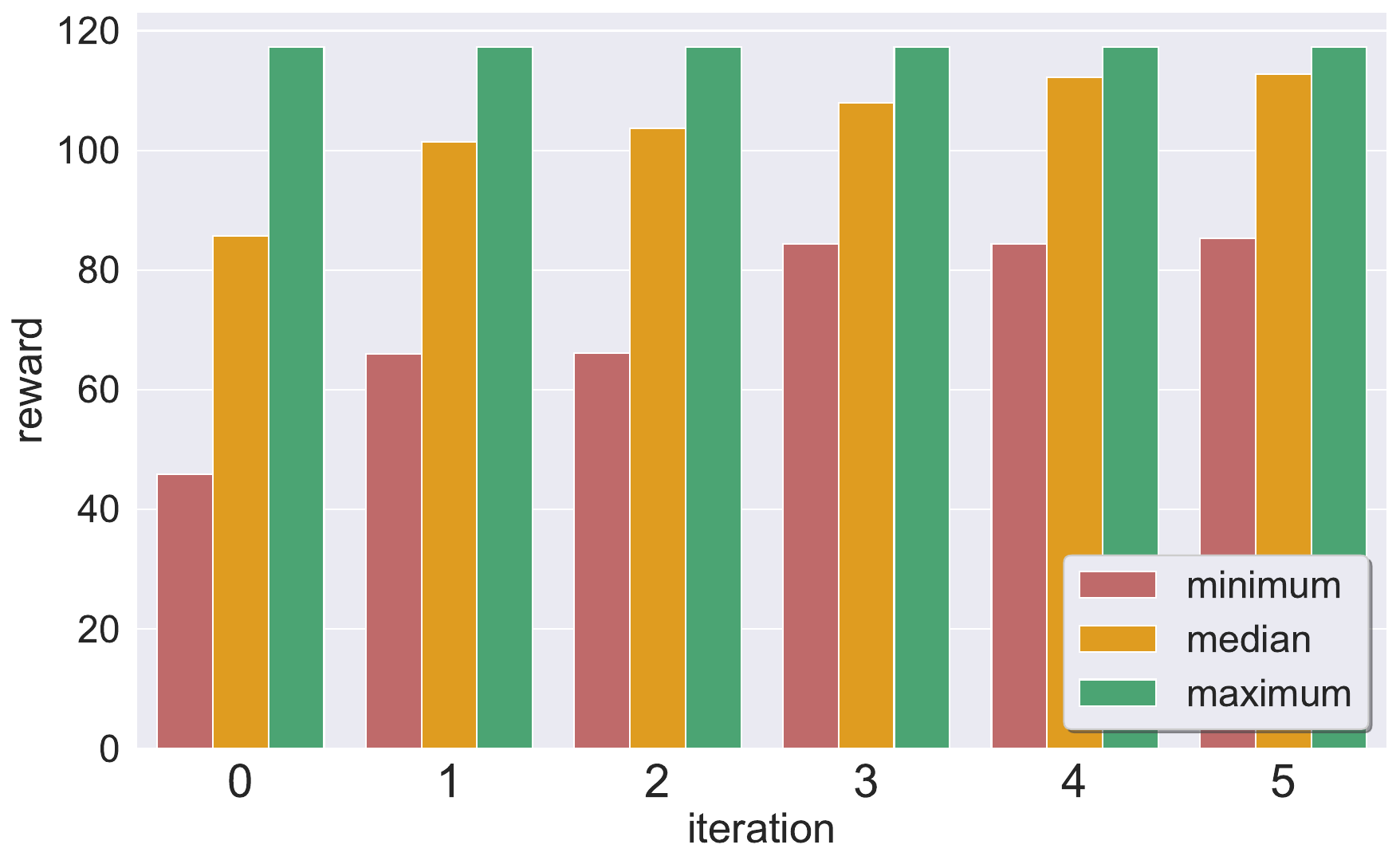}
			\caption{Reward statistics of remaining models}
			\label{fig:gradientAttacksSingleStep:aurora:rewardsStats}
		\end{subfigure}
		\hfill
		\begin{subfigure}[t]{0.49\linewidth}
			\includegraphics[width=\textwidth, 
			height=0.67\textwidth]{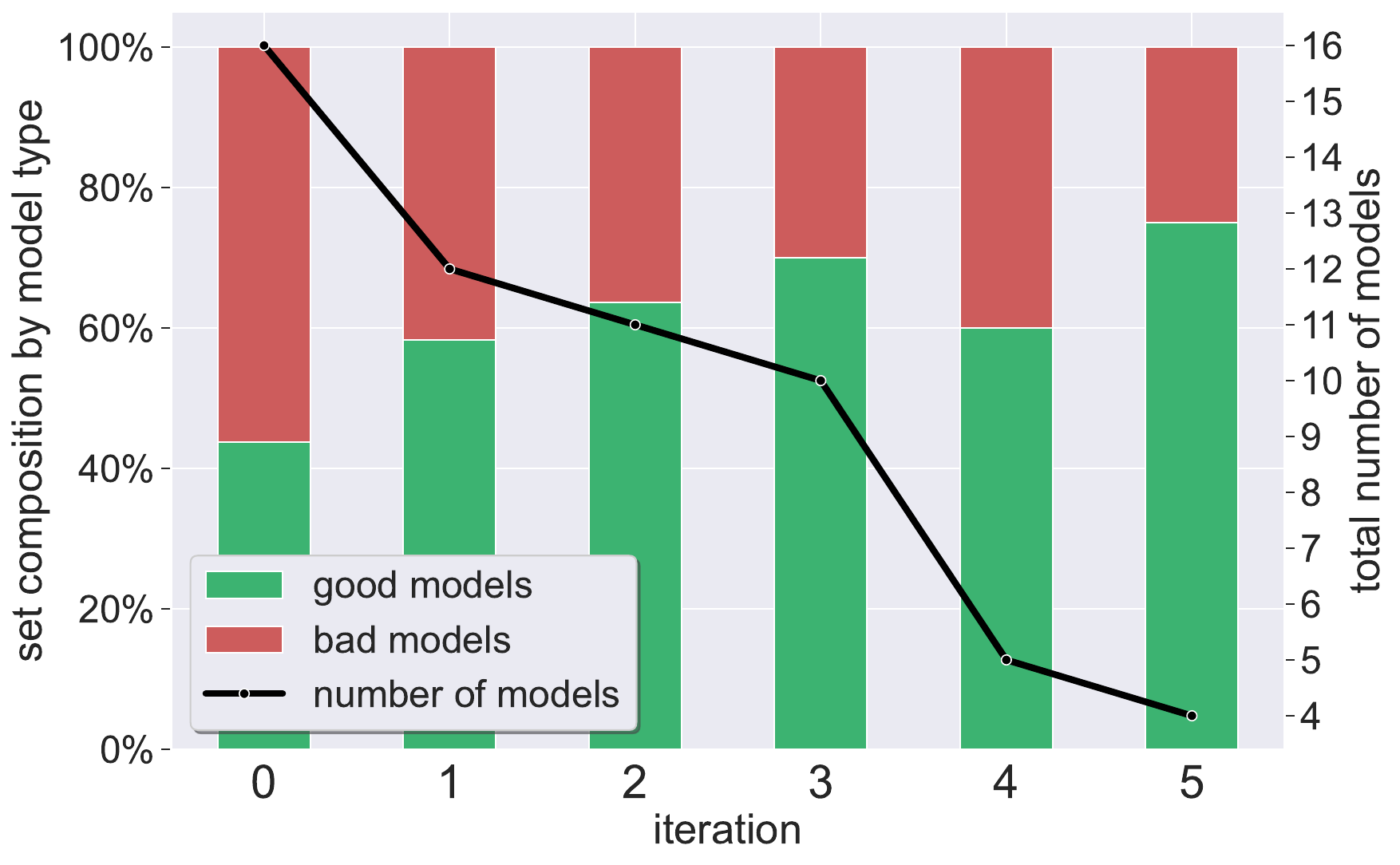}
			\caption{Ratio between good/bad models}
			\label{fig:gradientAttacksSingleStep:aurora:goodBadRatio}
		\end{subfigure}
		\caption{Aurora: Gradient attack \# 1 (single-step FGSM): Results of 
		models 
		filtered 
			using \emph{PDT} scores approximated by \emph{gradient attacks} 
			(instead of 
			a verification engine) on short-trained Aurora models, using the 
			\maxAgg 
			criterion (and terminating in advance if the disagreement scores 
			are no 
			larger than $2$). In contrast to our verification-driven approach, 
			the 
			final result contains a bad model. Compare to 
			Fig.~\ref{fig:auroraShortMaxFiltering}.}
		\label{fig:auroraGradienAttack1}
	\end{figure}

\begin{figure}[!ht]
		\centering
		\captionsetup[subfigure]{justification=centering}
		\captionsetup{justification=centering} 
		\begin{subfigure}[t]{0.49\linewidth}
			\centering
			\includegraphics[width=\textwidth, 
			height=0.67\textwidth]{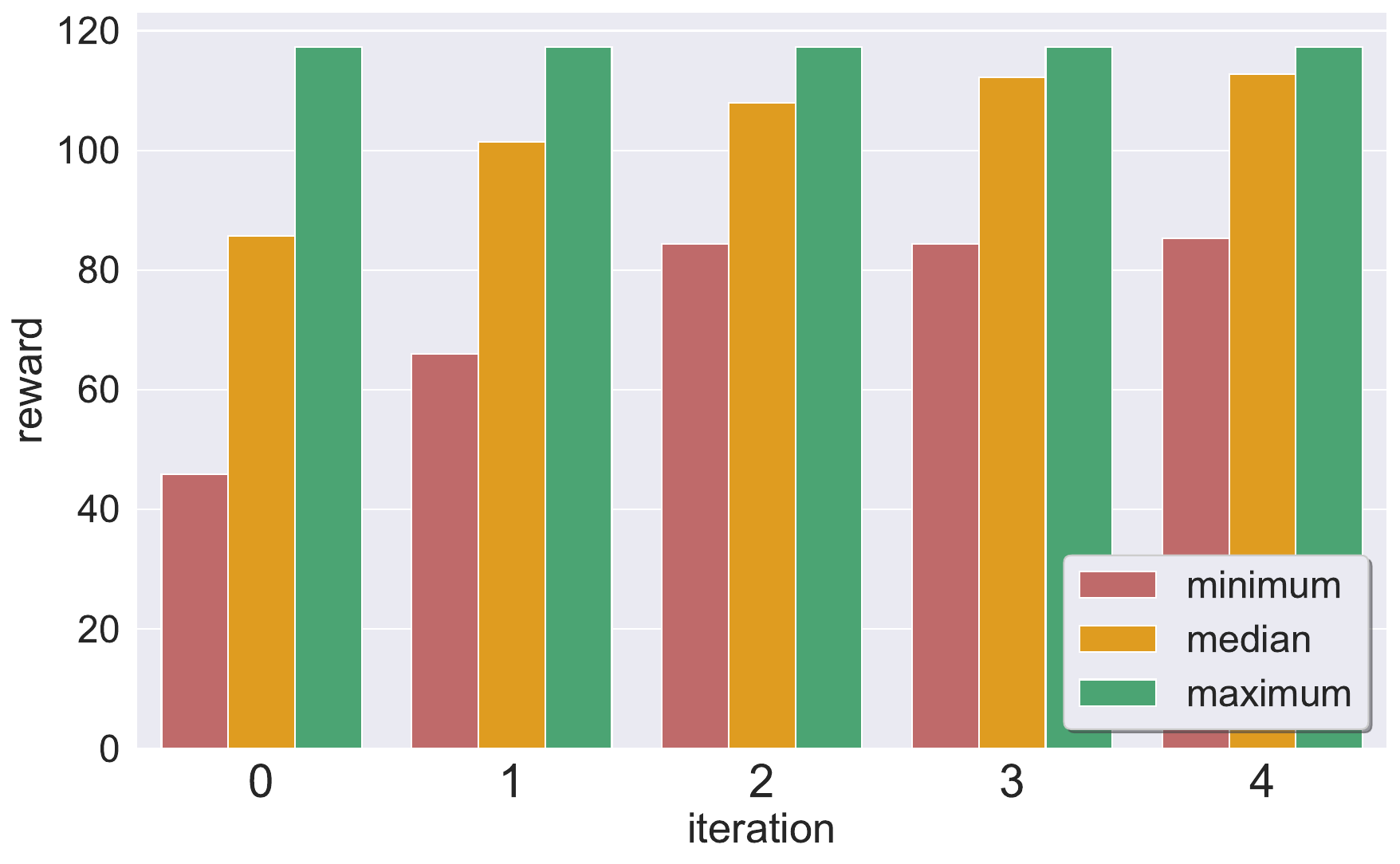}
			\caption{Reward statistics of remaining models}
			\label{fig:gradientAttacksMultiStep:aurora:rewardsStats}
		\end{subfigure}
		\hfill
		\begin{subfigure}[t]{0.49\linewidth}
			\includegraphics[width=\textwidth, 
			height=0.67\textwidth]{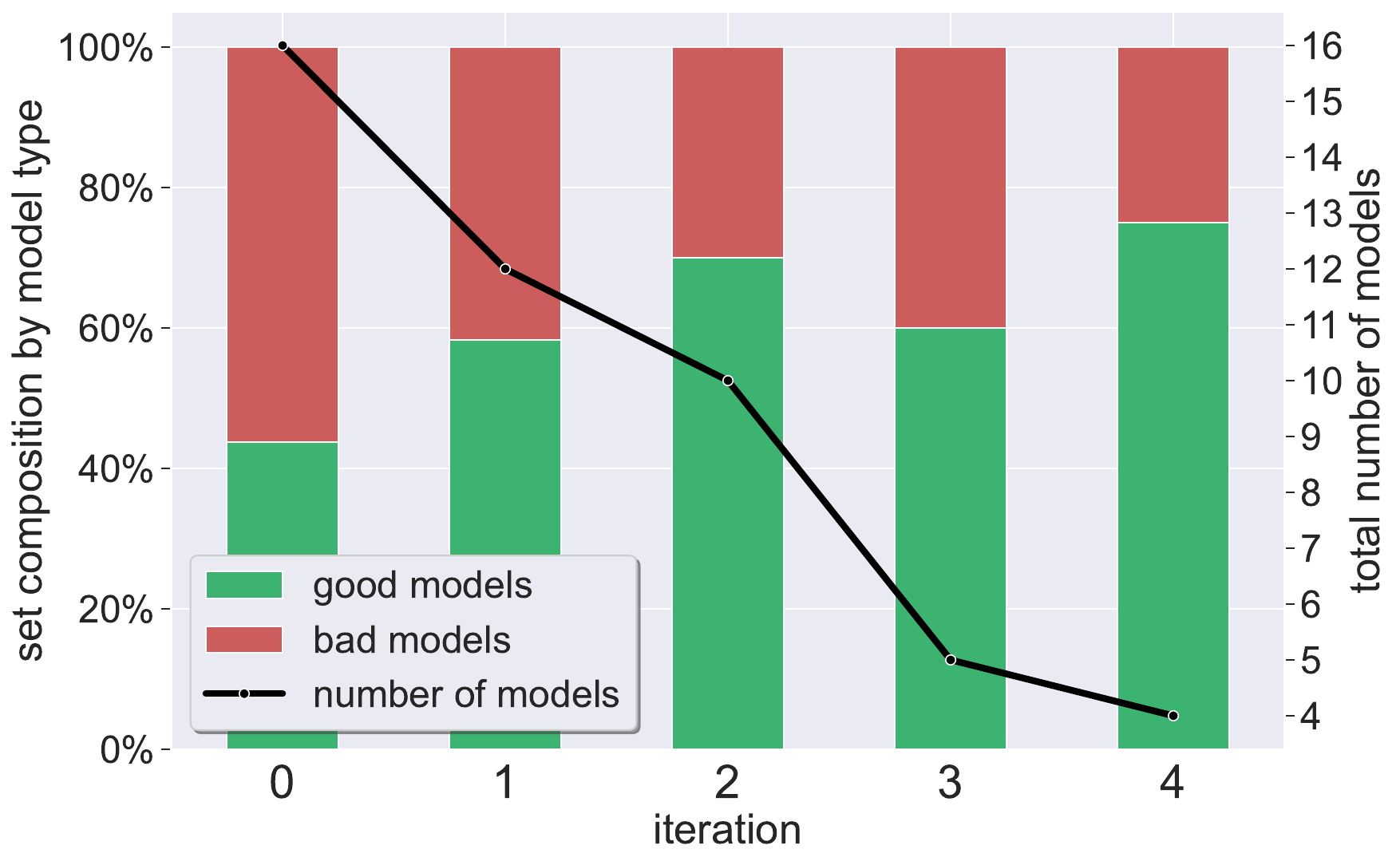}
			\caption{Ratio between good/bad models}
			\label{fig:gradientAttacksMultiStep:aurora:goodBadRatio}
		\end{subfigure}
		\caption{Aurora: Gradient attack \# 2 (Iterative PGD): Results of 
		models 
		filtered 
			using \emph{PDT} scores approximated by \emph{gradient attacks} 
			(instead of 
			a verification engine) on short-trained Aurora models, using the 
			\maxAgg 
			criterion (and terminating in advance if the disagreement scores 
			are no 
			larger than $2$). In contrast to our verification-driven approach, 
			the 
			final result contains a bad model. Compare to 
			Fig.~\ref{fig:auroraShortMaxFiltering}.}
		\label{fig:auroraGradienAttack2}
	\end{figure}

\begin{figure}[!ht]
		\centering
		\captionsetup[subfigure]{justification=centering}
		\captionsetup{justification=centering} 
		\begin{subfigure}[t]{0.49\linewidth}
			\centering
			\includegraphics[width=\textwidth]{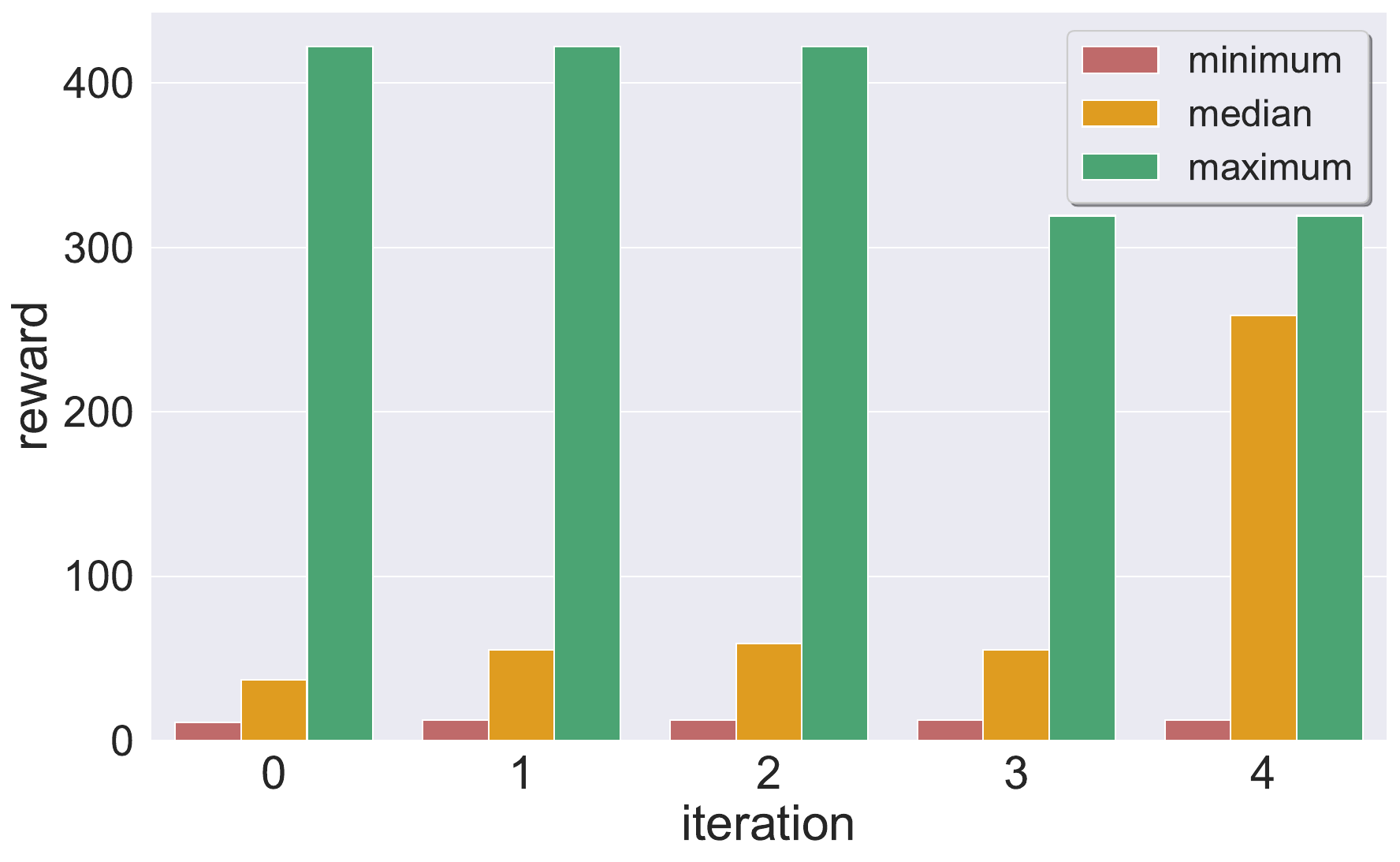}
		\end{subfigure}
		\hfill
		\begin{subfigure}[t]{0.49\linewidth}
			\includegraphics[width=\textwidth]{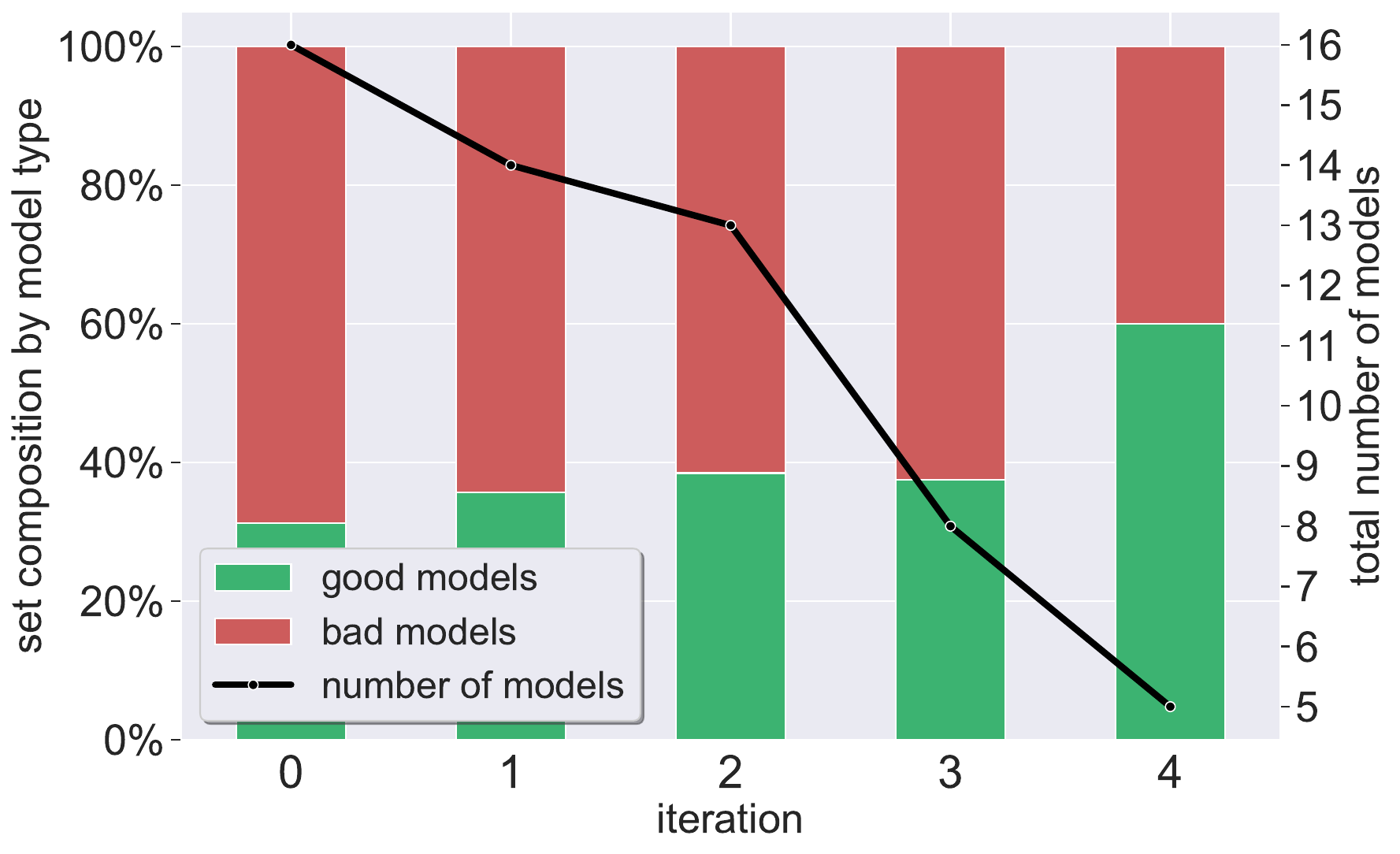}
		\end{subfigure}
		
		\begin{subfigure}[t]{0.49\linewidth}
			\centering
			\includegraphics[width=\textwidth]{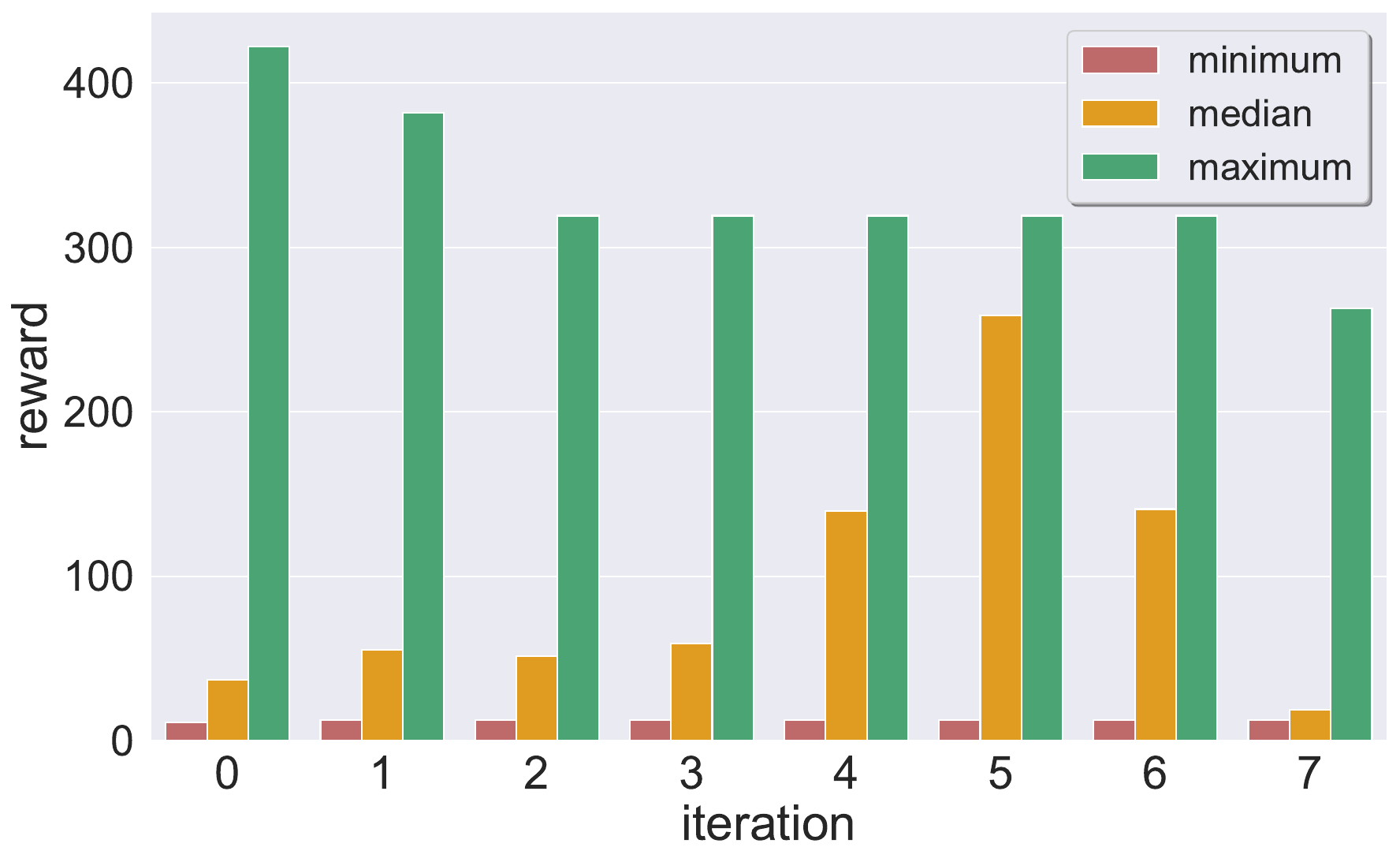}
			
		\end{subfigure}
		\hfill
		\begin{subfigure}[t]{0.49\linewidth}
			\includegraphics[width=\textwidth]{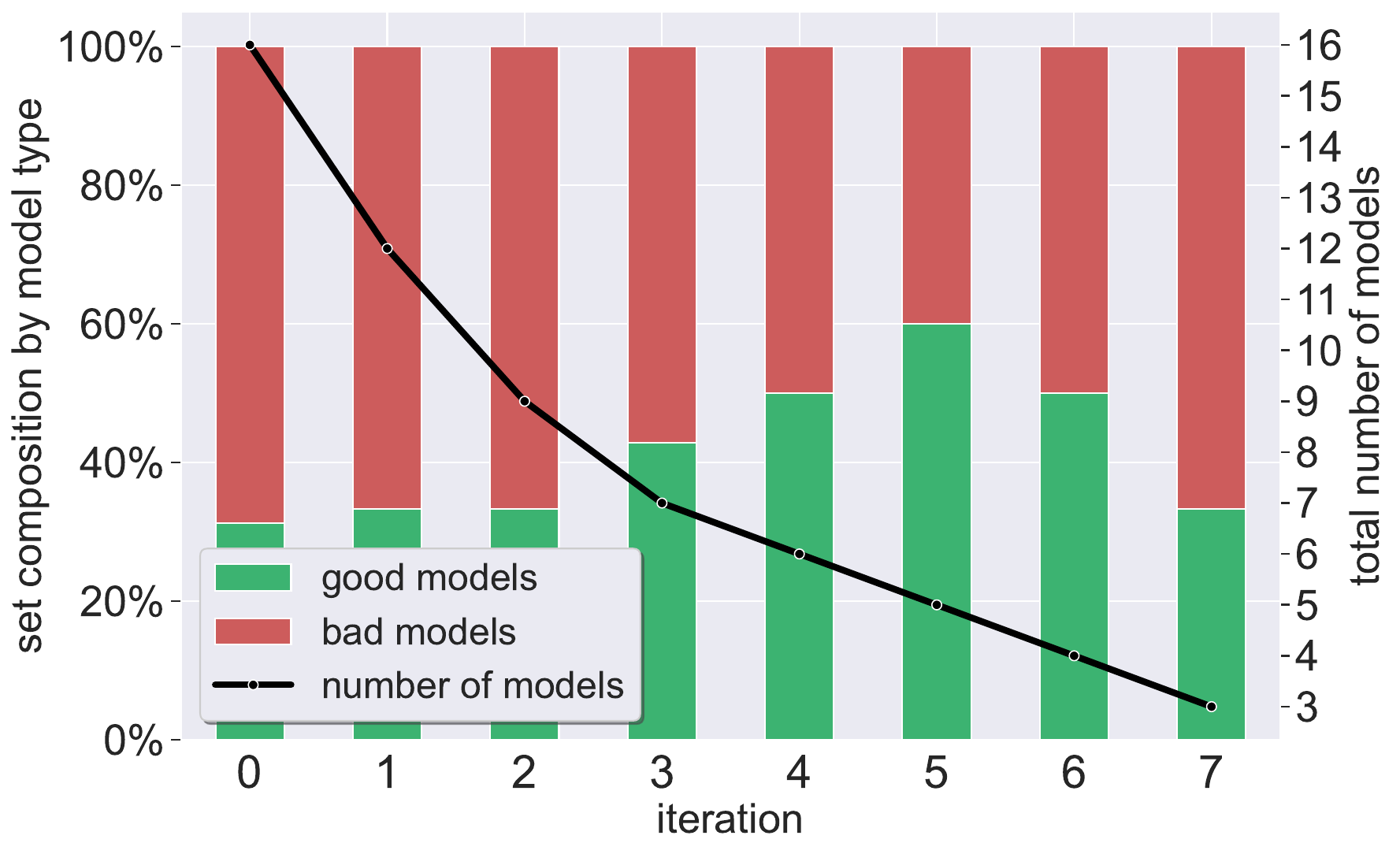}
		\end{subfigure}
		
		\begin{subfigure}[t]{0.49\linewidth}
			\centering
			\includegraphics[width=\textwidth]{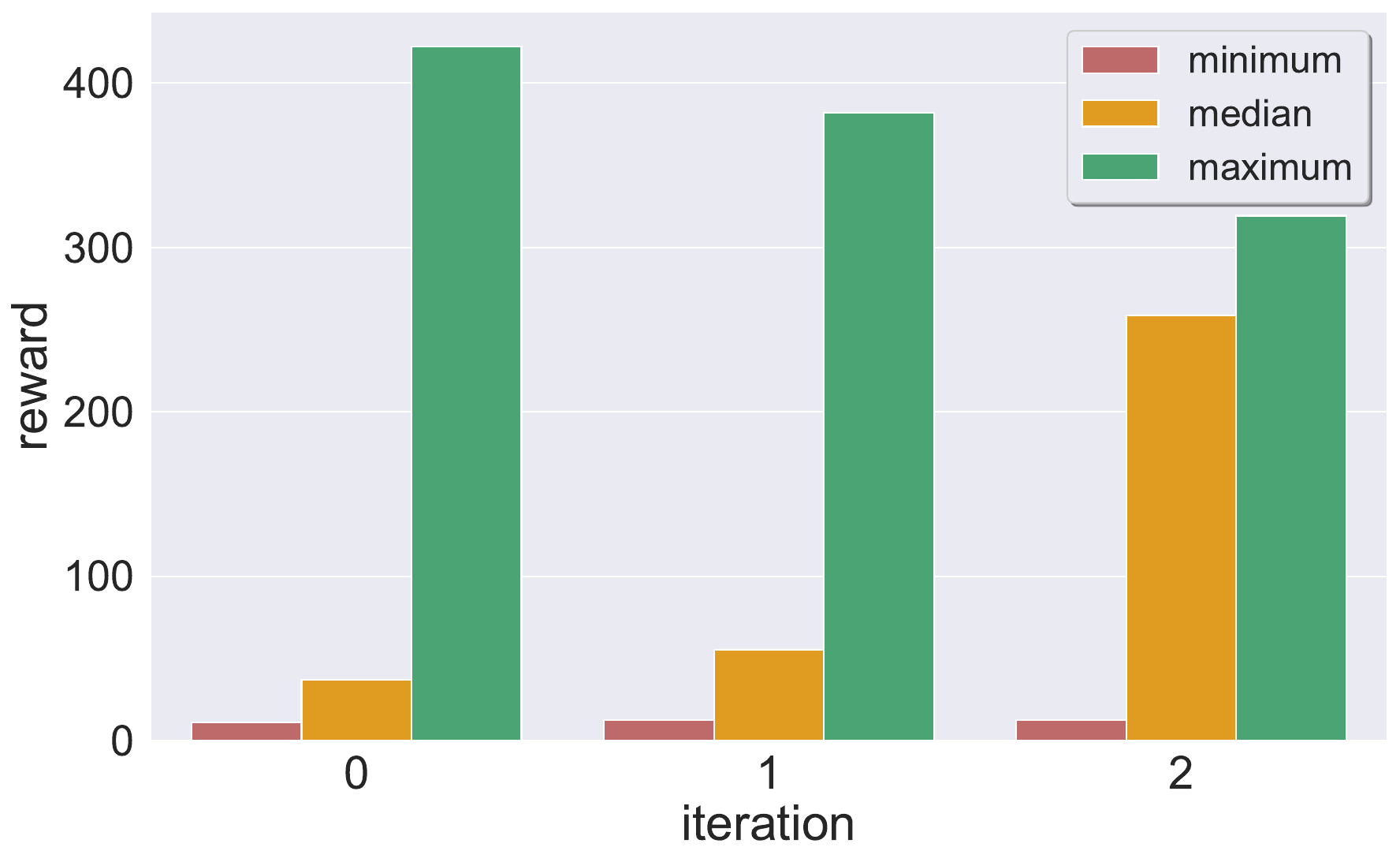}
			\caption{Reward statistics of remaining models}
			
		\end{subfigure}
		\hfill
		\begin{subfigure}[t]{0.49\linewidth}
			\includegraphics[width=\textwidth]{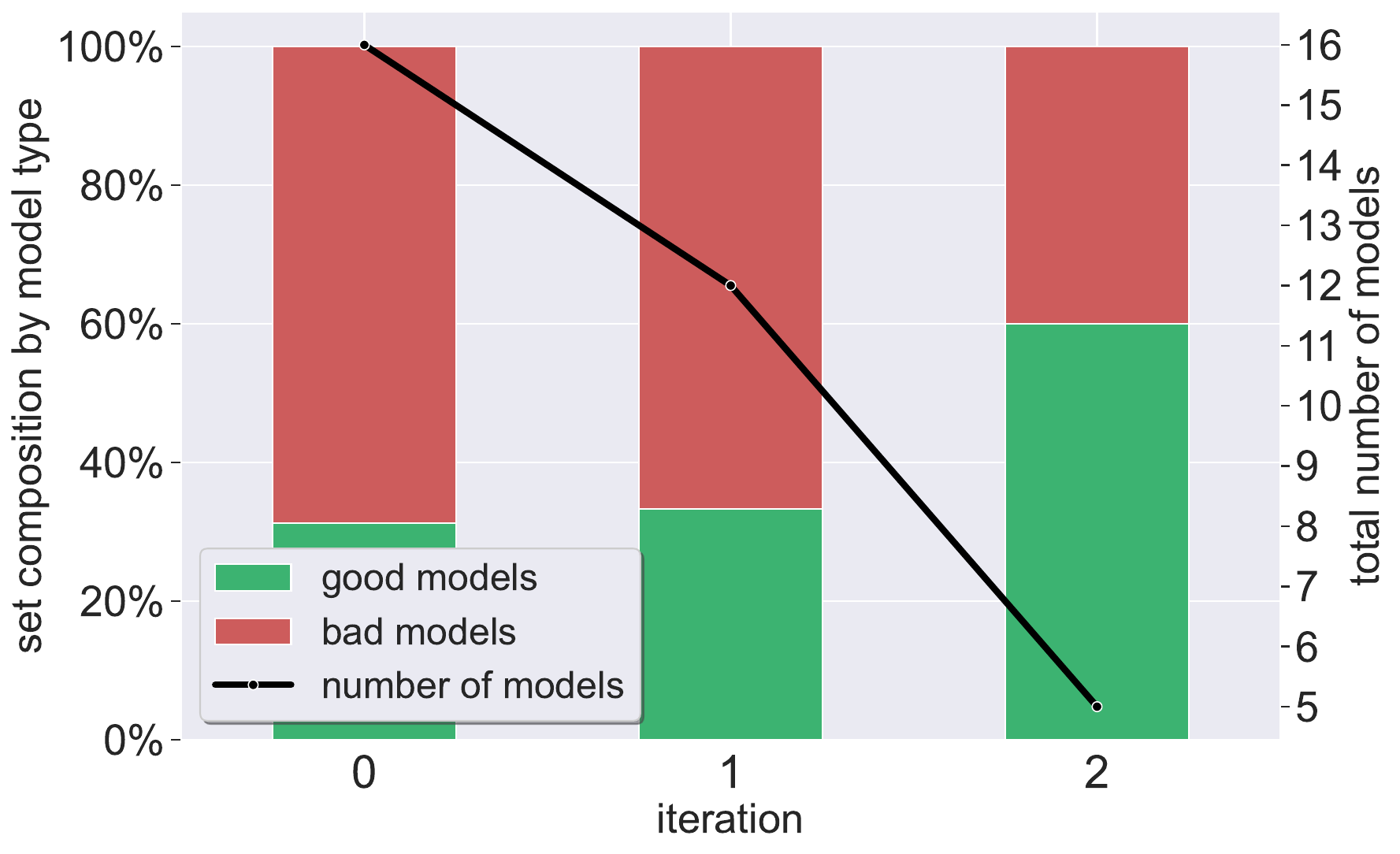}
			\caption{Ratio between good/bad models}
			\label{fig:gradientAttacks:cartPole:goodBadRatio}
		\end{subfigure}
		
		\caption{Cartpole: Gradient attack \# 3 (Constrained Iterative PGD): 
		Results of 
			models filtered using \emph{PDT} scores approximated by 
			\emph{gradient 
				attacks} (instead of a verification engine) on the Cartpole 
				models.\\Each 
			row, from top to bottom, contains results using a different 
			filtering 
			criterion (and terminating in advance if the disagreement scores 
			are no 
			larger than $2$): \percentileAgg (compare 
			to Fig.~\ref{fig:cartpolePercentileGoodBadResults} and 
			Fig.~\ref{fig:cartpolePercentileMinMaxRewards}), \maxAgg (compare 
			to 
			Fig.~\ref{fig:cartpole:maxResults}), and \conditionCombined 
			(compare to 
			Fig.~\ref{fig:cartpole:CombinedResults}). \\ In all cases, the 
			algorithm 
			returned at least one bad model (and usually more than one), 
			resulting in 
			models with lower average rewards than the models returned with our 
			verification-based approach.}
		\label{fig:cartPoleGradienAttack3}
	\end{figure}

	We ran our 
	algorithm on all original DRL benchmarks, with the sole difference being 
	the 
	replacement of the backend verification engine 
	(Line~\ref{line:SMTsolverForPdt} 
	in Alg.~\ref{alg:algorithmPairDisagreementScores}) with the described 
	gradient attacks.
	The first two attacks (i.e., FGSM and Iterative PGD) were used for both 
	Aurora batches (``short'' and ``long'' training), and the third attack 
	(Constrained 
	Iterative PGD) was used in the case of Cartpole and Mountain Car, as for 
	these benchmarks we required the encoding of a distance function with 
	constraints on the DNNs' outputs as well. 
	We note that 
	in the case of Aurora, we ran the Iterative PGD attack only when the 
	weaker attack failed (hence, only on the models from 
	Experiment~\ref{exp:auroraShort}).
	Our results,  summarized in Table~\ref{table:gradientAttackResultsSummary}, 
	demonstrate the advantages of using formal verification, compared to 
	competing, gradient attacks. These attacks, although scalable, resulted in 
	various cases to suboptimal PDT values, and in turn, retained unsuccessful 
	models that were successfully removed when using verification. For 
	additional results,
	we also refer the reader to Fig.~\ref{fig:auroraGradienAttack1}, 
	Fig.~\ref{fig:auroraGradienAttack2},  
	and Fig.~\ref{fig:cartPoleGradienAttack3}.


	\begin{table}[h]
		
		\centering
		\captionsetup{justification=centering}
		\centerline{
			\begin{tabular}{|c|c|c|c|c|c|c|c|c|}
				%
				\hline 
				\texttt{\textbf{ATTACK}} & \texttt{\textbf{BENCHMARK}} & 
				\texttt{\textbf{FEASIBLE}} & \texttt{\textbf{$\#$ PAIRS}} & 
				\texttt{\textbf{$\#$ ALIGNED}} & \texttt{\textbf{$\#$ 
				UNTIGHTENED}} & 
				\texttt{\textbf{$\#$ FAILED}} & 
				\texttt{\textbf{CRITERION}} & 
				\texttt{\textbf{SUCCESSFUL}} \\ \hline
				\multirow{3}{*}{1} & \multirow{3}{*}{Aurora (short)} & 
				\multirow{3}{*}{yes} & \multirow{3}{*}{120} & 
				\multirow{3}{*}{70} & 
				\multirow{3}{*}{50} & \multirow{3}{*}{0} & \conditionMax & 
				\color{red}{no} \\
				& & & & & & & \conditionCombined & \color{forestGreen}{yes} \\ 
				& & & & & & & \conditionPercentile & \color{forestGreen}{yes} 
				\\ \hline
				
				\multirow{3}{*}{1} & 
				\multirow{3}{*}{Aurora (long)} & \multirow{3}{*}{yes} & 
				\multirow{3}{*}{120} & \multirow{3}{*}{111} & 
				\multirow{3}{*}{9} & 
				\multirow{3}{*}{0} & \conditionMax & \color{forestGreen}{yes} 
				\\ 
				& & & & & & & \conditionCombined & \color{forestGreen}{yes} \\ 
				& & & & & & & \conditionPercentile & \color{forestGreen}{yes} 
				\\ 
				\hline
				
				\multirow{3}{*}{2} & \multirow{3}{*}{Aurora (short)} & 
				\multirow{3}{*}{yes} & \multirow{3}{*}{120} & 
				\multirow{3}{*}{104} & 
				\multirow{3}{*}{16} & \multirow{3}{*}{0} & \conditionMax & 
				\color{red}{no} \\
				& & & & & & & \conditionCombined & \color{forestGreen}{yes} \\ 
				& & & & & & & \conditionPercentile & \color{forestGreen}{yes} 
				\\ \hline

				\multirow{3}{*}{3} & 
				\multirow{3}{*}{Mountain Car} & \multirow{3}{*}{no} & 
				\multirow{3}{*}{120} & \multirow{3}{*}{38} & 
				\multirow{3}{*}{35} & 
				\multirow{3}{*}{47} & 
				\conditionMax &  \color{red}{no} \\
				& & & & & & & \conditionCombined &  \color{red}{no} \\ 
				& & & & & & & \conditionPercentile &  \color{red}{no} \\ \hline
				
				\multirow{3}{*}{3} & 
				\multirow{3}{*}{Cartpole} & \multirow{3}{*}{partially} & 
				\multirow{3}{*}{120} & \multirow{3}{*}{56} & 
				\multirow{3}{*}{61} & 
				\multirow{3}{*}{3} & \conditionMax &  \color{red}{no} \\
				& & & & & & & \conditionCombined &  \color{red}{no} \\ 
				& & & & & & & \percentileAgg  &  \color{red}{no}  \\ \hline
				
			\end{tabular}
		}
		
		\vspace{3mm}
		\caption{Summary of the gradient attack comparison. The first two 
		columns 
			describe the attack chosen and the benchmark on which it was 
			evaluated; the 
			third column states if the incomplete attack allowed a 
			gradient-based 
			approximation of all PDT scores; the next four columns respectively 
			represent the total number of DNN pairs, the number of pairs in 
			which the 
			attack returned a PDT score identical to the original one received 
			by our 
			verification engine, the number of pairs in which the attack 
			returned a 
			score that is less precise than the one returned by our 
			verification 
			engine, and the number of cases in which that attack failed to 
			generate outputs with both signs 
			(however, this does not mean that it cannot succeed by relaxing the 
			constraint and observing partial outputs). The 
			second-to-last column states the filtering criterion, and the last 
			column 
			indicates whether using the attack-based scores resulted in solely 
			successful models 
			(as was the case when verification was used).}
		\label{table:gradientAttackResultsSummary}
	\end{table}

	\subsection{Comparison to Sampling-Based Methods}
	
	In yet another line of experiments, we again replaced the verification 
	sub-procedure of our technique, and calculated the PDT scores 
	(Line~\ref{line:SMTsolverForPdt} in 
	Alg.~\ref{alg:algorithmPairDisagreementScores}) with sampling heuristics 
	instead.
	We note that, as any sampling technique is inherently incomplete, this can 
	be used solely for \emph{approximating} the PDT scores. 
	In our experiment, we sampled $1,000$ inputs from the OOD domain, and fed 
	them to all DNN pairs, per each benchmark. 
	Based on the outputs of the DNN pairs, we approximated the PDT scores, and 
	ran our algorithm in order to assess if scalable sampling techniques can 
	replace our verification-driven procedure. Our experiment raised two main 
	concerns regarding the use of sampling techniques instead of verification.

	First, in many cases, sampling could not result in constrained outputs. For 
	instance, in the Mountain Car benchmark, we use the \emph{c-distance} 
	function (see subsec.~\ref{subsec:verification-queries}), which requires 
	outputs with multiple signs. However, even extensive sampling cannot 
	guarantee this --- over a third (!) of all Mountain Car DNN pairs had 
	non-negative outputs, for \emph{all} $1,000$ OOD samples, hence requiring 
	approximation of the PDT scores even further, based only on partial 
	outputs. On the other hand, encoding the c-distance conditions in SMT is 
	straightforward in our case, and guarantees the required constraints.

	The second setback of this approach is that, as in the case of gradient 
	attacks, sampling may result in suboptimal PDT scores, that skew the 
	filtering process to retain unwanted models. 
	For example, in our results (summarized in 
	Table~\ref{table:sampling1000ResultsSummary}), in both the Mountain Car and 
	Aurora (short-training) benchmarks the algorithm returned unsuccessful 
	(``bad'') models in some cases, while these models are
	effectively removed when using verification. 
	We believe that these results further motivate the use of verification, 
	instead of applying more scalable and simpler methods.

	\begin{table}[htbp]
		\centering
		\captionsetup{justification=centering}
		\begin{tabular}{|c|c|c|}
			\hline
			\texttt{\textbf{BENCHMARK}} & \texttt{\textbf{CRITERION}} & 
			\texttt{\textbf{SURVIVING MODELS}} \\
			\hline
			\multirow{3}{*}{Cartpole} & \conditionMax & \{7\} \\
			& \conditionPercentile & \{6,7,9\} \\
			& \conditionCombined & \{7,9\} \\
			\hline
			\multirow{3}{*}{Mountain Car} & \conditionMax & \{1, 2, 
			\textcolor{red}{3}, 5, 6, \textcolor{red}{7}, 8, 
			\textcolor{red}{9}, 10, 11, \textcolor{red}{13}, 16\} \\
			& \conditionPercentile & \{1, 5, 8\} \\
			& \conditionCombined & \{1, 5, 6, 8, \textcolor{red}{9}, 10, 
			\textcolor{red}{13}\} \\
			\hline
			\multirow{3}{*}{Aurora (short)} & \conditionMax & \{7, 9, 
			\textcolor{red}{11}, 16\} \\
			& \conditionPercentile & \{7, 15, 16\} \\
			& \conditionCombined & \{7, 16\} \\
			\hline
			\multirow{3}{*}{Aurora (long)} & \conditionMax & \{20, 22, 27, 28\} 
			\\
			& \conditionPercentile & \{20, 27, 28\} \\
			& \conditionCombined & \{20, 22, 27, 28\} \\
			\hline
			\multirow{3}{*}{Arithmetic DNNs} & \conditionMax & \{1,5,6,8,10\} \\
			& \conditionPercentile & \{6, 8, 10\} \\
			& \conditionCombined & \{1,5,6,8,10\} \\
			\hline
		\end{tabular}
		\vspace{3mm}
		\caption{A summary of Alg.~\ref{alg:modelSelection}'s results, per each 
		of the five benchmarks: Cartpole, Mountain Car, Aurora (short \& long 
		training), and Arithmetic DNNs. For each benchmark, we sampled $1,000$ 
		OOD inputs, and approximated the PDT scores, based on which we ran 
		Alg.~\ref{alg:modelSelection}. The columns, from left to right, 
		indicate the benchmark, the filtering criterion, and the surviving 
		models, 
		with the unsuccessful models colored in \textcolor{red}{red}.
			We note that for cases in which the c-condition was used and could 
			not be approximated on both signs, we approximated it based on the 
			partial results, as afforded by the sampling technique.}
		\label{table:sampling1000ResultsSummary}
	\end{table}

	\subsection{Comparison to Predictive Uncertainty Methods}
	\label{subsec:predictiveUncertaintyComparison}
	In yet another experiment, we evaluated whether our verification-driven 
	approach can be replaced with \emph{predictive uncertainty 
	methods}~\cite{OvFeReNaScNoDiLaSn19, AbPoHuReLiGhFiCaKhAcMaNa21}.
	These methods are \emph{online} techniques, that assess uncertainty, i.e., 
	discern whether an encountered input aligns with the training distribution.
	Among these techniques, ensembles~\cite{KrVe94, Di00, GaHuMaTaSu22} are a 
	popular approach for predicting the uncertainty of a given input, by 
	comparing the variance among the ensemble members; intuitively, the higher 
	the variance is for a given input, the more ``uncertain'' the models are 
	with regard to the desired output.
	We note that in subsec.~\ref{subsec:generatingGoodEnsembles} we demonstrate 
	that \emph{after} using our verification-driven approach, ensembling the 
	resulting models may improve the overall performance relative to each 
	individual member. However, now we set to explore whether ensembles can not 
	only extend our verification-driven approach, but also \emph{replace} it 
	completely.
	As we demonstrate next, ensembles, like gradient attacks and sampling 
	techniques, are not a reliable replacement for verification in our setting.
	For example, in the case of Cartpole, we generated all possible $k$-sized 
	ensembles (we chose $k=3$ as this was the number of selected models via our 
	verification-driven approach, see 
	Fig.~\ref{fig:cartpolePercentileGoodBadResults}), resulting in 
	$ {n \choose k}={16 \choose 3}=560$ ensemble combinations. 
	Next, we randomly sampled $10,000$ OOD inputs (based on the specification 
	in Appendix~\ref{sec:appendix:VerificationQueries}) and utilized a 
	variance-based metric (inspired by~\cite{LoSeSc20}) to identify ensemble 
	subsets exhibiting low output variance on these OOD-sampled inputs.
	However, even the subset represented by the ensemble with the lowest 
	variance, included the ``bad'' model $\{8\}$ (see 
	Fig.~\ref{fig:cartpoleRewards}), which was successfully removed in our 
	equivalent verification-driven technique.
	We believe that this too demonstrates the merits of our verification-driven 
	approach.

	\section{Related Work}
	\label{sec:RelatedWork}
	Due to 
	its widespread occurrence, the phenomenon of adversarial inputs has 
	gained considerable attention~\cite{SzZaSuBrErGoFe13,
		GoShSz14, PaMcJhFrCeSw16, PaMcGoJhCeSw17, MoSeFaFr16, FaFoWeIdMu19, 
		ZuAkGu18}. Specifically, The machine learning community has dedicated 
		substantial effort to measure and enhance the robustness of 
		DNNs~\cite{CiBoGrDa17, MaMaScTsVl17, CoRoKo19,
		QiMaGoKrDvFaDeStKo19, WoRiKo20, ShNaGhXuDiStDaTaGo19,
		GaUsAjGeLaLaMaLe16, HaYaYuNiXuHuTsSu18, YuHaYaNiTsSu19, LuLoWaJo19,
		ShSaZhGhStJaGo19}. 
	The formal methods community has also 
	been looking into the problem, by devising methods for DNN verification, 
	i.e., techniques that can automatically and formally 
	guarantee the correctness of DNNs~\cite{KaBaDiJuKo17,  
		GoKaPaBa18, SiGePuVe19, TjXiTe17, 
		KoLoJaBl20, WuOzZeIrJuGoFoKaPaBa20, SeDeDrFrGhKiShVaYu18, PoAbKr20,
		OkWaSeHa20, GoPaPuRuSa21, LyKoKoWoLiDa20, XiTrJo18, 
		VaPeWaNiSiKh22, BaGiPa21, DuChSa19, DuJhSaTi18b, SuKhSh19, 
		FuPl18,CoMaFa21,
		GeLeXuWaGuSi22,RuHuKw18, IsBaZhKa22, UrChWuZh20, SoTh19,
		YaYaTrHoJoPo21, GoAdKeKa20, DoSuWaWaDa20, UsGoSuNoPa21, ZhShGuGuLeNa20, 
		JaBaKa20,AlAvHeLu20,CoAmKaFa24, MaAmWuDaNeRaMeDuGaShKaBa24}. 
	These techniques include SMT-based approaches
	(e.g.,~\cite{HuKwWaWu17, KaBaDiJuKo21,
		KaHuIbJuLaLiShThWuZeDiKoBa19, KuKaGoJuBaKo18}) as used in this work, 
		methods based on MILP and LP solvers (e.g.,~\cite{LoMa17,Eh17,TjXiTe19,
		BuTuToKoMu18}), methods based on abstract interpretation or symbolic 
		interval propagation (e.g.,~\cite{WaPeWhYaJa18, GeMiDrTsChVe18,
		WeZhChSoHsBoDhDa18, TrBkJo20}), as well as abstraction-refinement
	(e.g.,~\cite{ElGoKa20, PrAf20, AnPaDiCh19, 
		SiGePuVe19, ZeWuBaKa22, OsBaKa22,
		AsHaKrMo20}),
	size reduction~\cite{Pr22}, 
	quantitative
	verification~\cite{BaShShMeSa19}, synthesis~\cite{AlAvHeLu20}, 
	monitoring~\cite{LuScHe21},
	optimization~\cite{AvBlChHeKoPr19, StWuZeJuKaBaKo21}, 
	and also tools for verifying recurrent 
	neural networks (RNNs)~\cite{ZhShGuGuLeNa20, JaBaKa20}. 
	In addition, efforts have been undertaken to offer verification with 
	provable guarantees~\cite{RuHuKw18, IsBaZhKa22},  verification of DNN 
	fairness~\cite{UrChWuZh20}, and DNN repair and modification after 
	deployment~\cite{SoTh19, YaYaTrHoJoPo21, GoAdKeKa20, DoSuWaWaDa20, 
	UsGoSuNoPa21}.
	We also note that some sound and incomplete techniques~\cite{BePaWiLaKw22, 
	TjXiTe17} have put forth an alternative strategy for DNN verification, via 
	convex relaxations. These techniques are relatively fast, and can also be 
	applied by our approach, which is generally agnostic to the underlying DNN 
	verifier. 
	In the specific case of DRL-based systems, various non-verification 
	approaches have been put forth to increase the reliability of such 
	systems~\cite{RaAcAm19, 
		ZhZhWu21, WaSu20,GaFe15, AcHeTa17}. 
	These techniques rely mostly on \emph{Lagrangian 
	Multipliers}~\cite{StAcAb20, RoGiRo21, 
		LiDiLi20}.

	In addition to DNN verification techniques, another approach that 
	guarantees safe behavior is \emph{shielding}~\cite{BlKoKoWa15,AlBloEh18}, 
	i.e., 
	incorporating an external component (a ``shield'') 
	that \emph{enforces} the safe behavior of the agent, according to a given 
	specification on the input/output relation of the DNN in question. 
	Classic shielding approaches~\cite{WuMaDeWa19, BlKoKoWa15,
		AlBloEh18,PraKoPoBlo21,PraKoTa21} focus on simple properties that can 
		be expressed in Boolean LTL formulas.
	However, proposals for reactive synthesis methods within infinite theories 
	have also emerged recently~\cite{ChoFinkPisSant22, MadBlo22,FinkHeiPass22}. 
	Yet another relevant approach is \emph{Runtime 
	Enforcement}~\cite{Sc00,LiBauWa09,FalFerMou12}, which is akin to shielding 
	but incompatible with reactive systems~\cite{BlKoKoWa15}.
	In a broader sense, these aforementioned techniques can be viewed as part 
	of ongoing research on improving the safety of \emph{Cyber-Physical 
		Systems} 
		(CPS)~\cite{SaCrAbNo16,TrCaDiMuJoKo19,PeTh20,LiXuLiYu19,GuEa19}.
	
	Variability among machine learning models has been widely employed to 
	enhance performance, often through the use of 
	\emph{ensembles}~\cite{KrVe94, 
	Di00, GaHuMaTaSu22}. 
	However, only a limited number of methodologies utilize ensembles to tackle 
	generalization concerns~\cite{YaZeZhWu13, OsAsCa18, 
		RoScTa20, OrCaMa22}. 
	In this context, we note that our approach can also be used for additional 
	tasks, such as \emph{ensemble selection}~\cite{AmZeKaSc22}, as it can 
	identify subsets of models that have a high variance in their outputs.
	Furthermore, alternative techniques beyond verification for assessing 
	generalization involve evaluating models across predefined new 
	distributions~\cite{PaGaKoKrKoSo18}.


	In the context of learning, 
	there is ample research on identifying and mitigating
	\emph{data drifts}, i.e., changes in the distribution of inputs that are 
	fed to the 
	ML model, during deployment~\cite{MaArJo22,SaChSaPe23,FiHsCh19, 
	KhAdKaMaPaCh23,
		BeCaFiBiGaMo06,GeCoGiDo20}.
	In addition, 
	certain studies employ verification for novelty detection with respect 
	to DNNs concerning a \emph{single} distribution~\cite{HaKrRiSc22}. 
	Other work focused on applying verification to evaluate the performance of 
	a model relative to \emph{fixed} distributions~\cite{BaSt19, 
	WuTaRoYaMaOaHaPaBa22}, while non-verification approaches, such as 
	ensembles~\cite{YaZeZhWu13, OsAsCa18, RoScTa20, OrCaMa22}, runtime 
	monitoring~\cite{HaKrRiSc22}, and other techniques~\cite{PaGaKoKrKoSo18}, 
	have been applied for OOD input detection.
	Unlike the aforementioned approaches, our objective is to establish 
	verification-guided \textit{generalization} \textit{scores} that encompass 
	an \textit{input domain}, spanning \emph{multiple} distributions within 
	this domain.
	Furthermore, as far as we are aware, our approach represents the first 
	endeavor to harness the diversity among models to distill a subset with 
	enhanced \emph{generalization} capabilities. 
	Particularly, it is also the 
	first endeavor to apply formal verification for this goal.

	\section{Limitations}
	\label{sec:Limitations}

	Although our evaluation results indicate that our approach is applicable to 
	varied settings and problem domains, it may suffer from multiple 
	limitations. First, by design, our approach assumes a single solution to a 
	given generalization problem. This does not allow selecting DNNs with 
	different generalization strategies to the same problem. 
	We also note that although our approach builds upon 
	verification techniques, it cannot \emph{assure} correctness or 
	generalization guarantees of the selected models (although, in practice, 
	this can happen in various scenarios --- as our evaluation demonstrates).

	In addition, our approach relies on the underlying assumption that the 
	range of inputs is known apriori. 
	In some situations, this assumption may turn out to be  highly non-trivial 
	--- for example, in cases where 
	the DNN's inputs are themselves produced by another DNN, or some other 
	embedding mechanism.  
	Furthermore, even when the range of inputs is known, bounding their exact 
	values may require domain-specific knowledge for encoding various distance 
	functions, and the metrics that build upon them (e.g., PDT scores). 
	For example, in the case of Aurora, routing expertise is required in order 
	to translate various Internet congestion levels to actual bounds on 
	Aurora's input variables. We note that such knowledge may be highly 
	non-trivial in various domains.

	Finally, we note that other limitations stem from the use of the 
	underlying 
	DNN verification technology, which may serve as a computational bottleneck. 
	Specifically, while our approach requires dispatching a polynomial number 
	of DNN verification queries, solving each of these queries is 
	NP-complete~\cite{KaBaDiJuKo17}. In addition, the underlying DNN verifier
	itself may limit the \emph{type} of encodings it affords, which, in turn, 
	restricts various use-cases to which our approach can be applied. For 
	example, sound and complete DNN verification engines are currently suitable 
	solely for DNNs encompassing piecewise-linear activations. However, as DNN 
	verification technology improves, so will our approach.

	\section{Conclusion}
	\label{sec:Conclusion}
	
	This case study presents a novel, verification-driven approach to identify 
	DNN models that effectively generalize to an input domain of interest.
	We introduced an iterative scheme that utilizes a backend DNN verifier, 
	enabling us to assess models by scoring their capacity to generate similar 
	outputs for multiple distributions over a specified domain.
	We extensively evaluated our approach on multiple benchmarks of both 
	supervised, and unsupervised learning, and demonstrated that, indeed, it is 
	able to effectively distill models capable of successful generalization 
	capabilities. 
	As DNN verification technology advances, our approach will gain scalability 
	and broaden its applicability to a more diverse range of DNNs.

	
	\section*{Acknowledgments}
	Amir, Zelazny, and Katz received partial support for their work from the 
	Israel Science Foundation (ISF grant 683/18). Amir received additional 
	support through a scholarship from the Clore Israel Foundation. The work of 
	Maayan and Schapira received partial funding from Huawei. 
	We thank Aviv Tamar for his contributions to this project.

	
	\section*{Declarations}
	We made use of Large Language Models 
	(LLMs) for assistance in rephrasing certain parts of the text. We do not 
	have further disclosures or declarations.

	\bibliography{references}

	\newpage
	\appendix
	\renewcommand{\thesection}{\Alph{section}}
	
	\section*{\huge Appendices}
	
	\section{DRL Benchmarks: Training and Evaluation}
	\label{sec:appendix:trainingAndEvaluation}
	
	In this appendix, we elaborate on the hyperparameters and the training 
	procedure, for reproducing all models and environments of all three DRL 
	benchmarks. We also provide a thorough overview of various implementation 
	details. The code is based on the
	\textit{Stable-Baselines 3}~\cite{RaHiGlKaErDo21} and \textit{OpenAI 
		Gym}~\cite{BrChPeScScTaZa16} packages. Unless stated otherwise, the 
		values 
	of the various parameters used during training and evaluation are the 
	default values (per training algorithm, environment, etc.).
	
	\subsection{Training Algorithm}
	We trained our models with \emph{Actor-Critic} algorithms. These are 
	state-of-the-art RL training algorithms that iteratively optimize two
	neural networks: 
	
	\begin{itemize}
		\item a \textit{critic} network, that learns a value
		function~\cite{MnKaSi13} (also known as a \emph{Q-function}), that 
		assigns a
		value to each $\langle$state,action$\rangle$ pair; and 
		\item an \textit{actor} network, which is the DRL-based agent trained 
		by 
		the algorithm. This network iteratively maximizes the value function
		learned by the critic, thus improving the learned policy.
	\end{itemize}
	
	Specifically, we used two implementations of Actor-Critic algorithms: 
	\textit{Proximal Policy Optimization} (PPO)~\cite{ShWoDh17} and
	\emph{Soft Actor-Critic} (SAC)~\cite{HaZhAbLe18}. 
	
	Actor-Critic algorithms are considered very advantageous, due to their 
	typical requirement of relatively few samples to learn from, and also due 
	to their ability to allow the agent to learn policies for continuous spaces 
	of
	$\langle$state,action$\rangle$ pairs. 
	
	In each training process, all models were trained using the same 
	hyperparameters, with the exception of the \textit{Pseudo Random Number 
		Generator's (PRNG) seed}. Each training phase consisted of $10$ 
	checkpoints, while each checkpoint included a constant number of 
	environment steps, as described below. For model evaluation, we used the 
	last checkpoint of each training process (per benchmark).
	
	\subsection{Architecture}
	In all benchmarks, we used DNNs with a feed-forward architecture. We 
	refer the reader to Table~\ref{table:benchmarksTrainingInfo} for a summary 
	of the chosen architecture per each benchmark.

	\begin{table}[ht]
		\centering
		\begin{tabular}{| P{0.2\linewidth} | P{0.15\linewidth} | 
		P{0.2\linewidth} | 
				P{0.3\linewidth} | P{0.15\linewidth} |}
			\hline
			\textbf{benchmark} & \textbf{hidden layers} & \textbf{layer size} & 
			\textbf{activation function} & \textbf{training algorithm} \\ \hline
			Cartpole	& 2	& [32, 16]	& \relu	& PPO \\ \hline
			Mountain Car & 2 & [64, 16] & \relu & SAC \\ \hline
			Aurora & 2 & [32, 16] & \relu & PPO \\ \hline
		\end{tabular}
		
		\caption{DNN architectures and training algorithms, per benchmark.}
		\label{table:benchmarksTrainingInfo}
	\end{table}
	
	\newpage
	
	\subsection{Cartpole Parameters}
	\label{subsec:appendix:trainingAndEvaluation:Cartpole}
	
	\hfill
	
	\subsubsection{Architecture and Training}
	
	\begin{enumerate}
		
		\item
		\textbf{Architecture}
		\begin{itemize}
			\item \textit{hidden layers}: 2
			\item \textit{size of hidden layers}: 32 and 16, respectively
			\item \textit{activation function}: \relu
		\end{itemize}
		
		\item
		\textbf{Training}
		\begin{itemize}
			\item \textit{algorithm}: Proximal Policy Optimization (PPO)
			\item \textit{gamma ($\gamma$)}: 0.95
			\item \textit{batch size}: 128
			
			\item \textit{number of checkpoints}: $10$
			\item \textit{total time-steps} (number of training steps for each 
			checkpoint): $50,000$
			\item \textit{PRNG seeds} (each one used to train a different 
			model):\\
			$\{1, 2, 3, 4, 5, 6, 7, 8, 9, 10, 11, 12, 13, 14, 15, 16\}$
		\end{itemize}
		
	\end{enumerate}
	
	\begin{figure}[ht]
		\centering
		\captionsetup{justification=centering}
		\includegraphics[width=0.6\textwidth]{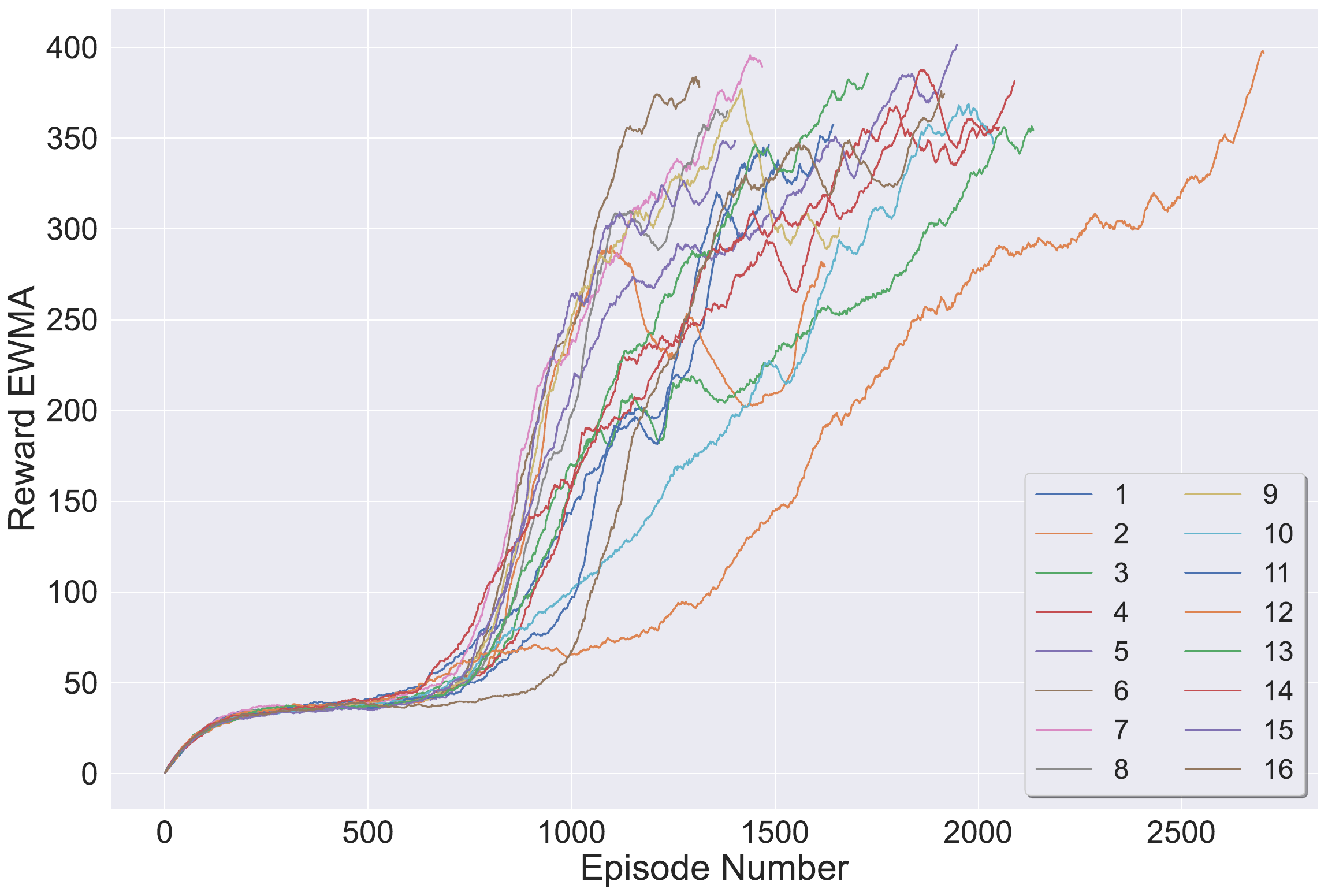}
		\caption{Cartpole: models' exponential weighted moving average 
			(EWMA) reward during training. All models achieved a high reward 
			(at the end of 
			their 
			training phase).}
		\label{fig:cartpole:trainingRewards}
	\end{figure}

	\subsubsection{Environment}
	\hfill
	
	\noindent We used the configurable \textit{CartPoleContinuous-v0} 
	environment.
	Given lower and upper bounds for the x-axis location, denoted as 
	$[low,high]$, and $mid=\frac{high+low}{2}$, the initial x position is 
	randomly, uniformly drawn from the interval $[mid-0.05, mid+0.05]$.
	
	An \emph{episode} is a sequence of interactions between the agent and the
	environment, such that the episode ends when a terminal state is reached. 
	In the Cartpole 
	environment, an episode terminates after the first of the following occurs: 
	\begin{enumerate}
		\item The cart's location exceeds the platform's boundaries (as 
		expressed via the $x$-axis location); or
		
		\item The cart was unable to balance the pole, which fell (as 
		expressed via the $\theta$-value); or
		
		\item $500$ time-steps have passed.
		
	\end{enumerate}
	
	\hfill
	
	\subsubsection{Domains}
	\begin{enumerate}
		
		\item \textbf{(Training) In-Distribution}
		
		\begin{itemize}
			\item \textit{action min magnitude}: True
			\item \textit{x-axis lower bound} (x\_threshold\_low): $-2.4$
			\item \textit{x-axis upper bound} (x\_threshold\_high): $2.4$
		\end{itemize}
		
		\item  \textbf{(OOD) Input Domain}
		
		Two symmetric OOD scenarios were evaluated: the cart's $x$ position 
		represented significantly extended platforms in a single direction, 
		hence, 
		including areas previously unseen during training. Specifically, we
		generated a domain of input points characterized by $x$-axis
		boundaries that were selected, with an equal probability, either from 
		$[-10, -2.4]$ or from $[2.4, 10]$ (instead of the in-distribution range 
		of 
		$[-2.4,2.4])$. The cart's initial location was uniformly drawn from the 
		range's \emph{center} $\pm0.05$: $[-6.4-0.05, -6.4+0.05]$ and 
		$[6.4-0.05, 
		6.4+0.05]$, respectively. 
		All other parameters were the same as the ones used 
		in-distribution.
		
		\ \\
		
		\noindent \underline{OOD scenario 1}
		\begin{itemize}
			\item \textit{x-axis lower bound} (x\_threshold\_low): $-10.0$
			\item \textit{x-axis upper bound} (x\_threshold\_high): $-2.4$
		\end{itemize}
		\hfill
		
		\noindent \underline{OOD scenario 2}
		\begin{itemize}
			\item \textit{x-axis lower bound} (x\_threshold\_low): $2.4$
			\item \textit{x-axis upper bound} (x\_threshold\_high): $10.0$
		\end{itemize}
		
	\end{enumerate}

	\subsection{Mountain Car Parameters}
	\label{subsec:appendix:trainingAndEvaluation:mountaincar}
	
	\subsubsection{Architecture and Training}
	
	\hfill
	
	\begin{enumerate}
		
		\item
		\textbf{Architecture}
		
		\begin{itemize}
			\item \textit{hidden layers}: 2
			\item \textit{size of hidden layers}: 64 and 16, respectively
			\item \textit{activation function}: \relu
			\item \textit{clip mean} parameter: 5.0
			\item \textit{log stdinit} parameter: -3.6
		\end{itemize}
		
		\item
		\textbf{Training}

		\begin{itemize}
			\item \textit{algorithm:} Soft Actor-Critic (SAC)
			\item \textit{gamma ($\gamma$)}: 0.9999	
			\item \textit{batch size}: 512
			\item \textit{buffer size}: 50,000
			
			\item \textit{gradient steps}: 32
			\item \textit{learning rate}: $\num{3e-4}$
			\item \textit{learning starts}: 0
			\item \textit{tau ($\tau$)}: 0.01
			\item \textit{train freq}: 32
			\item \textit{use sde}: True
			
			\item \textit{number of checkpoints}: $10$
			\item \textit{total time-steps} (number of training steps for each 
			checkpoint): $5,000$
			\item \textit{PRNG seeds} (each one used to train a different 
			model):\\
			$\{1, 2, 3, 4, 5, 6, 7, 8, 9, 10, 11, 12, 13, 14, 15, 16\}$
		\end{itemize}
		
	\end{enumerate}

	\begin{figure}[ht]
		\centering
		\captionsetup{justification=centering}
		\includegraphics[width=0.6\textwidth]{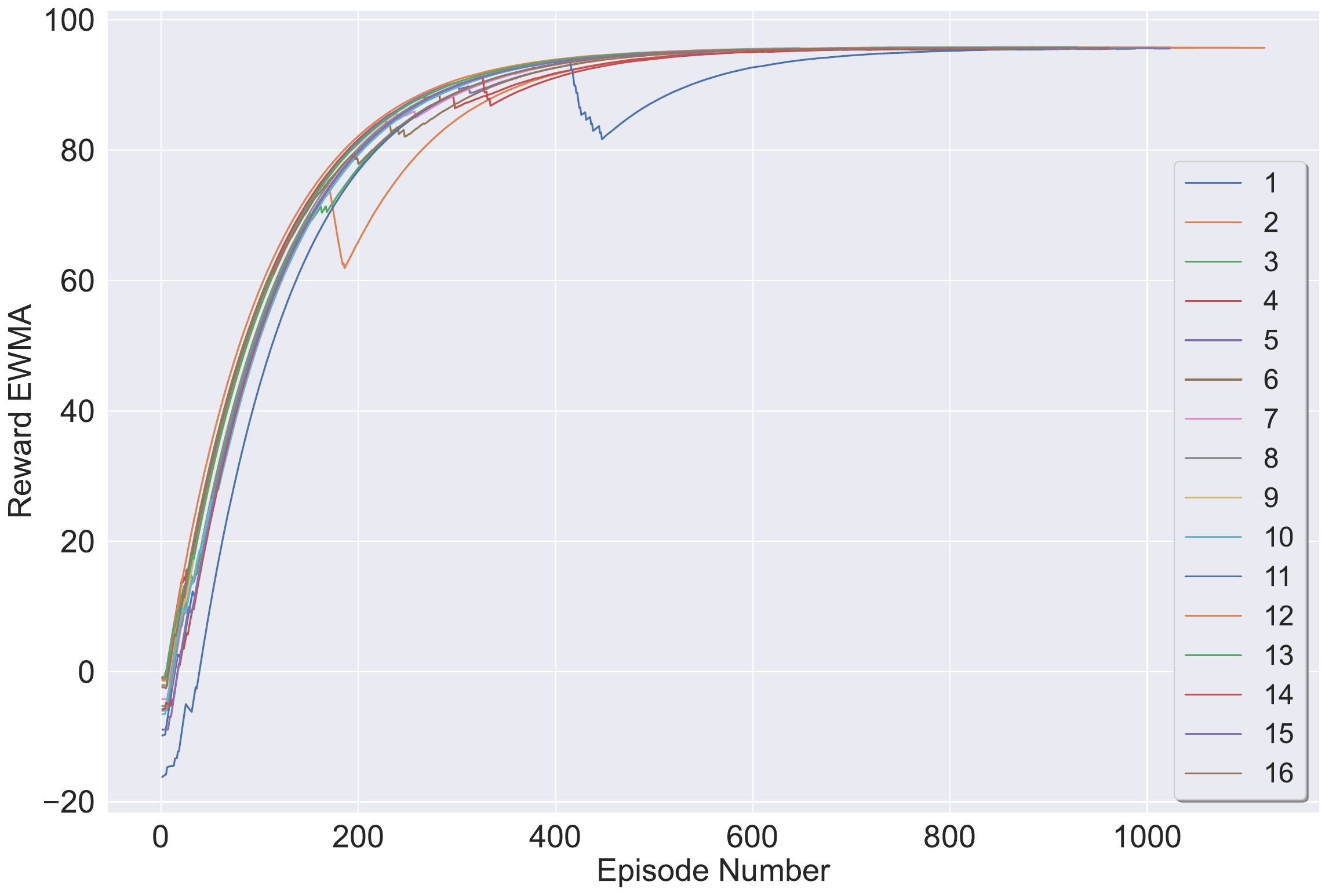}
		\caption{Mountain Car: models' exponential weighted moving 
		average 
			(EWMA) reward during training. All models achieved a high reward 
			(at the end of 
			their 
			training phase).}
		\label{fig:mountaincar:trainingRewards}
	\end{figure}
	
	\subsubsection{Environment}
	\hfill
	
	\noindent We used the \textit{MountainCarContinuous-v1} environment.
	
	\hfill
	
	\subsubsection{Domains}
	\begin{enumerate}
		
		\item \textbf{(Training) In-Distribution}
		
		\begin{itemize}
			\item \textit{min position}: $-1.2$
			\item \textit{max position}: $-0.6$
			\item \textit{goal position}: $0.45$
			\item \textit{min action} (if the agent's action is negative and 
			under 
			this value, this value is used): $-2$
			\item \textit{max action} (if the agent's action is positive and 
			above 
			this value, this value is used): $2$
			\item \textit{max speed}: $0.4$
			\item \textit{initial location range} (from which the initial 
			location 
			is uniformly drawn): $[-0.9, -0.6]$
			\item \textit{initial velocity range} (from which the initial 
			velocity 
			is uniformly drawn): $[0, 0]$ (i.e., the initial velocity in this 
			scenario is always $0$) 
			\item \textit{x scale factor} (used for scaling the x-axis): $1.5$
		\end{itemize}
		
		\item  \textbf{(OOD) Input Domain}
		
		The inputs are the same as the ones used in-distribution, except for 
		the 
		following:
		\begin{itemize}
			\item \textit{min position}: $-2.4$
			\item \textit{max position}: $1.2$
			\item \textit{goal position}: $0.9$
			\item \textit{initial location range}: $[0.4, 0.5]$
			\item \textit{initial location velocity}: $[-0.4, -0.3]$
		\end{itemize}
		
	\end{enumerate}
	
	\subsection{Aurora Parameters}
	\label{subsec:appendix:trainingAndEvaluation:aurora}
	
	
	\hfill
	
	\subsubsection{Architecture and Training}
	
	\begin{enumerate}
		
		\item
		\textbf{Architecture}
		
		\begin{itemize}
			\item \textit{hidden layers}: 2
			\item \textit{size of hidden layers}: 32 and 16, respectively
			\item \textit{activation function}: \relu
		\end{itemize}
		
		\item
		\textbf{Training}
		\begin{itemize}
			\item \textit{algorithm}: Proximal Policy Optimization (PPO)
			
			\item \textit{gamma ($\gamma$)}: 0.99
			\item \textit{number of steps to run for each environment, per 
			update} 
			(n\_steps): $8,192$
			\item \textit{number of epochs when optimizing the surrogate loss} 
			(n\_epochs): $4$
			\item \textit{learning rate}: $\num{1e-3}$
			\item \textit{value function loss coefficient} (vf\_coef): $1$
			\item \textit{entropy function loss coefficient} (ent\_coef): 
			$\num{1e-2}$
			
			\item \textit{number of checkpoints}: $6$
			\item \textit{total time-steps} (number of training steps for each 
			checkpoint): $656,000$ (as used in the original 
			paper~\cite{JaRoGoScTa19})
			\item \textit{PRNG seeds} (each one used to train a different 
			model): 
			\\$\{4, 52, 105, 666, 850, 854, 857, 858, 885, 897, 901, 906, 907, 
			929, 
			944, 945\}$ \\
			We note that for simplicity, these were mapped to indices $\{1 
			\ldots 
			16\}$, accordingly (e.g., $\{4\} \rightarrow \{1\}$, $\{52\} 
			\rightarrow \{2\}$, etc.).
			
		\end{itemize}
		
	\end{enumerate}
	
	\begin{figure}[ht]
		\centering
		\captionsetup{justification=centering}
		\includegraphics[width=0.6\textwidth]{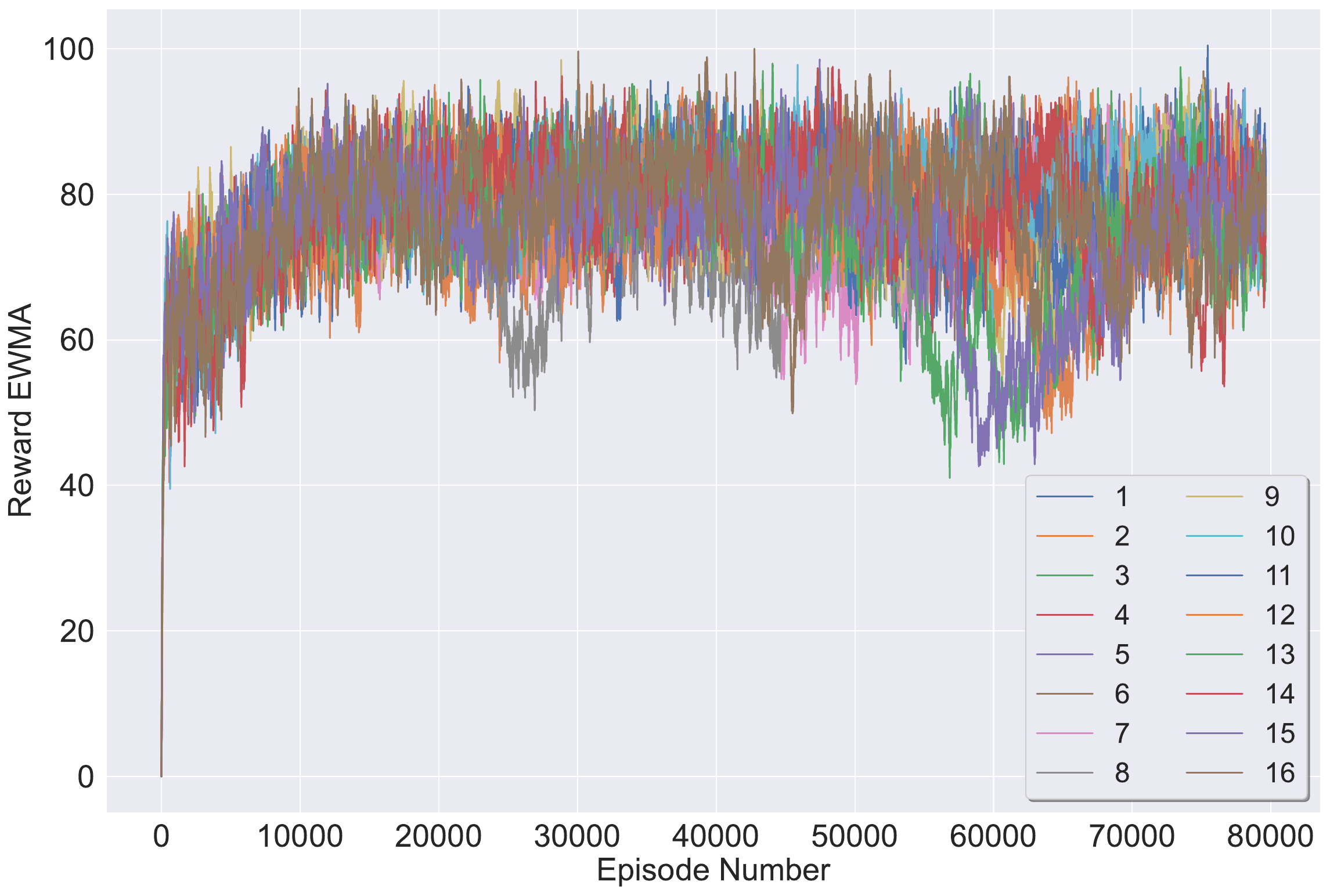}
		\caption{Aurora (short training): models' exponential weighted 
			moving average (EWMA) reward during training. All models achieved a 
			high 
			reward 
			(at the end of their training phase).}
		\label{fig:auroraShort:trainingRewards}
	\end{figure}
	
	\noindent \subsubsection{Environment}
	\hfill
	
	\noindent We used a configurable version of the \textit{PccNs-v0} 
	environment. 
	For models in Exp.~\ref{exp:auroraShort} (with the \emph{short} training), 
	each episode consisted of $50$ steps. For models in 
	Exp.~\ref{exp:auroraLong} (with the \emph{long} training), each episode 
	consisted of $400$ steps.

	\hfill
	
	\subsubsection{Domains}
	\begin{enumerate}
		
		\item \textbf{(Training) In-Distribution}
		
		\begin{itemize}
			\item \textit{minimal initial sending rate ratio (to the link's 
				bandwidth)} (min\_initial\_send\_rate\_bw\_ratio): $0.3$
			\item \textit{maximal initial sending rate ratio (to the link's 
				bandwidth)} (max\_initial\_send\_rate\_bw\_ratio): $1.5$
		\end{itemize}
		
		\item  \textbf{(OOD) Input Domain}
		
		To bound the \textit{latency gradient} and \textit{latency ratio} 
		elements 
		of the input, we used a shallow buffer setup, with a bounding parameter 
		$\delta>0$ such that \textit{latency gradient} $\in [-\delta, \delta]$ 
		and 
		\textit{latency ratio} $\in [1.0, 1.0 +\delta]$.

		\begin{itemize}
			\item \textit{minimal initial sending rate ratio (to the link's 
				bandwidth)} (min\_initial\_send\_rate\_bw\_ratio): $2.0$
			\item \textit{maximal initial sending rate ratio (to the link's 
				bandwidth)} (max\_initial\_send\_rate\_bw\_ratio): $8.0$
			\item \textit{use shallow buffer}: True
			\item \textit{shallow buffer $\delta$ bound parameter}: $\num{1e-2}$
			
		\end{itemize}
		
	\end{enumerate}

	\newpage
	
	\section{Arithmetic DNNs: Training and Evaluation}
	\label{sec:appendix:trainingAndEvaluationArithmeticDnns}

	

	In this appendix, we elaborate on the hyperparameters and the training 
	procedure, for reproducing all models and environments of the supervised 
	learning Arithmetic DNNs 
	benchmark. We also provide a thorough overview of various implementation 
	details. 
	
	\medskip
	\noindent
	To train our neural networks we used the
	\textit{pyTorch} package, version $2.0.1$. Unless stated otherwise, the 
	values 
	of the various parameters used during training and evaluation are the 
	default values (per training algorithm, environment, etc.).
	\subsection{Training Algorithm}
	We trained our models with the Adam optimizer~\cite{KiBa15}, for $10$ 
	epochs, and with a batch size of $32$. 
	All models were trained using the same 
	hyperparameters, with the exception of the \textit{Pseudo Random Number 
		Generator's (PRNG) seed}.
	
	\subsection{Architecture}
	In all benchmarks, we used DNNs with a fully connected feed-forward 
	architecture with ReLU activations.
	
	\begin{table}[ht]
		\centering
		\begin{tabular}{| P{0.2\linewidth} | P{0.15\linewidth} | 
		P{0.2\linewidth} | 
				P{0.3\linewidth} | P{0.15\linewidth} |}
			\hline
			\textbf{benchmark} & \textbf{hidden layers} & \textbf{layer size} & 
			\textbf{activation function} & \textbf{training algorithm} \\ \hline
			Arithmetic DNNs	& 3	& [10, 10, 10]	& \relu	& Adam \\ \hline
		\end{tabular}
		
		\caption{Arithmetic DNNs: benchmark parameters.}
		\label{table:ArithmeticDNNTrainingInfo}
	\end{table}
	
	\subsection{Arithmetic DNNs Parameters}
	\label{subsec:appendix:trainingAndEvaluation:Arithmetic}
	
	\hfill
	
	\subsubsection{Architecture and Training}
	
	\begin{enumerate}
		
		\item
		\textbf{Architecture}
		\begin{itemize}
			\item \textit{hidden layers}: 3
			\item \textit{size of (each) hidden layer}: 10
			\item \textit{activation function}: \relu
		\end{itemize}
		
		\item
		\textbf{Training}
		\begin{itemize}
			\item \textit{algorithm}: Adam~\cite{KiBa15}
			\item \textit{learning rate}: $\gamma = 0.001$
			\item \textit{batch size}: 32
			\item \textit{PRNG seeds} (each one used to train a different 
			model): $[0, 49]$. 
			The $5$ models with the \emph{best} seeds OOD are (from best to 
			worse): $\{37, 4, 22, 20, 47\}$, and the $5$ models with the 
			\emph{worst} seeds OOD are (from best to worse): $\{15, 12, 11, 44, 
			30\}$. 
			We note that for simplicity, these were mapped to indices $\{1 
			\ldots 
			10\}$, based on their order (e.g., $\{4\} \rightarrow \{1\}$, 
			$\{11\} 
			\rightarrow \{2\}$, etc.).
			
			\item \textit{loss function}: mean squared error (MSE)
		\end{itemize}
		
	\end{enumerate}
	
	\subsubsection{Domains}
	\begin{enumerate}
		
		\item \textbf{(Training) In-Distribution}
		We have generated a dataset of $10,000$ vectors of dimension $d=10$, in 
		which every entry is sampled from the multi-modal uniform distribution 
		$[l=-10,u=10]^{10}$. Hence, $x_1, x_2, ..., x_{10000} \sim [-10, 
		10]^{10}$. and the output label is $y_i = x_i[0] + x_i[1]$.
		The random seed used for generating the dataset is $0$.
		
		\item  \textbf{(OOD) Input Domain}
		We evaluated our networks on $100,000$ input vectors of dimension 
		$d=10$, where every entry is uniformly distributed between $[l=-1,000, 
		u=1,000]$.
		All other parameters were identical to the ones used 
		in-distribution.
		
	\end{enumerate}

	\clearpage
	
	\section{Verification Queries: Additional Details}
	\label{sec:appendix:VerificationQueries}
	

	\subsection{Precondition}    
	
	In our experiments, we used the following bounds for the (OOD) input domain:
	
	\begin{enumerate}
		\item \textbf{Cartpole:}
		\begin{itemize}
			
			\item x position: $x \in [-10,-2.4]$ or $x \in [2.4,10]$
			The PDT was set to be the maximum PDT score of each of these 
			two scenarios.
			
			\item $x$ velocity: $v_{x} \in [-2.18,2.66]$
			
			\item angle: $\theta \in [-0.23, 0.23]$ 
			
			\item angular velocity: $v_{\theta} \in [-1.3, 1.22]$
			
		\end{itemize}
		
		\item \textbf{Mountain Car:}
		\begin{itemize}
			\item x position: $x \in [-2.4,0.9]$

			\item $x$ velocity: $v_{x} \in [-0.4,0.134]$
		\end{itemize}
		
		\item \textbf{Aurora:}
		\begin{itemize}
			\item \emph{latency gradient}:
			$x_{t} \in [-0.007, 0.007]$, for all $t$ s.t.
			$(t \mod 3) = 0$
			
			\item \emph{latency ratio}:
			$x_{t} \in [1, 1.04]$, for all $t$ s.t.
			$(t \mod 3) = 1$
			
			\item \emph{sending ratio}:
			$x_{t} \in [0.7, 8]$, for all $t$ s.t.
			$(t \mod 3) = 2$
			
		\end{itemize}
		
		\item \textbf{Arithmetic DNNs:}
		
		\begin{itemize}
			\item 
			for all $0\leq i \leq 9$: 
			$x_{i} \in [-1000,1000]$
		\end{itemize}
		
	\end{enumerate}

	\subsection{Postcondition}
	As elaborated in subsection~\ref{subsec:verification-queries}, we encode an 
	appropriate \emph{distance function} on the DNNs' outputs. 
	
	\medskip
	\noindent
	\textbf{Note.} In the case of the \emph{c-distance} function, we chose, for 
	Cartpole and Mountain Car, $c: = N_{1}(x)\geq 0  \wedge N_{2}(x) \geq 0$ 
	and $c':= N_{1}(x)\leq 0 
	\wedge N_{2}(x) \leq 0$.
	This distance function is tailored to find the maximal difference between 
	the 
	outputs (actions) of two models, in a given category of inputs 
	(non-positive or 
	non-negative, in our case). The intuition behind this function is that in 
	some 
	benchmarks, good and bad models may differ in the \emph{sign} (rather than 
	only 
	the magnitude) of their actions. For example, consider a scenario of the 
	Cartpole benchmark where the cart is located on the ``edge'' of the 
	platform: 
	an 
	action in one direction (off the platform) will cause the episode to end, 
	while 
	an action in the other direction will allow the agent to increase its 
	reward 
	by continuing the episode, and possibly reaching the goal.

	
	\subsection{Verification Engine}
	
	All queries were dispatched to 
	\marabou~\cite{KaHuIbJuLaLiShThWuZeDiKoBa19,WuIsZeTaDaKoReAmJuBaHuLaWuZhKoKaBa24}
	 --- a sound and complete 
	verification engine, previously used in other DNN-verification-related 
	work~\cite{AmZeKaSc22, AmWuBaKa21, AmCoYeMaHaFaKa23, CaKoDaKoKaAmRe22, 
		JaBaKa20, 
		WuZeKaBa22, WaPeWhYaJa18, SiGePuVe19, ElGoKa20, OsBaKa22, 
		CoYeAmFaHaKa22, 
		BaKa22, AmScKa21, ElCoKa22,AmFrKaMaRe23, CoElBaKa22, BaAmCoReKa23}.

	\clearpage
	
	\section{Algorithm Variations and Hyperparameters}
	\label{sec:appendix:AlgorithmAdditionalInformation}
	
	In this appendix, we elaborate on our algorithms' additional 
	hyperparameters and filtering criteria, used throughout our evaluation. As 
	the results demonstrate, our method is highly robust in a myriad of 
	settings.
	
	\subsection{Precision}
	For each benchmark and each experiment, we arbitrarily selected $k$ 
	models which reached our reward threshold for the in-distribution data. 
	Then, we used these models for our empirical evaluation. The PDT scores 
	were calculated up to a finite precision of $0.5\leq \epsilon\leq 20$, 
	depending on the
	benchmark ($0.5$ for Mountain Car, $1$ for Cartpole and Aurora, and $20$ 
	for Arithmetic DNNs).

	\subsection{Filtering Criteria}
	As elaborated in Sec.~\ref{sec:Approach}, our algorithm iteratively 
	filters 
	out (Line~\ref{lst:line:removeModels} in Alg.~\ref{alg:modelSelection}) 
	models with a relatively high disagreement score, i.e., models that may 
	disagree with their peers in the input domain. We present three different 
	criteria based on which we may select the models to remove in a given 
	iteration, 
	after sorting the models based on their DS score:
	
	\begin{enumerate}
		\item \conditionPercentile: in which we remove the \emph{top $p \% $} 
		of 
		models with the highest disagreement scores, for a predefined value 
		$p$. 
		In our experiments, we chose $p=25\%$.
		
		\item \conditionMax: in which we: \begin{enumerate}
			\item sort the DS scores of all models in a descending order;
			\item calculate the difference between every two adjacent scores;
			\item search for the \emph{greatest difference} of any two 
			subsequent 
			DS scores;
			\item for this difference, use the larger DS as a threshold; and
			\item remove all models with a DS that is greater than or equal to 
			this 
			threshold.
		\end{enumerate}
		
		\item \conditionCombined: in which we remove models based either on 
		\conditionMax or  
		\conditionPercentile, depending on which criterion eliminates more 
		models in 
		a specific iteration.  
	\end{enumerate}

	\clearpage
	
	\section{Cartpole: Supplementary Results}
	\label{sec:appendix:CartPoleSupplementaryResults}
	
	Throughout our evaluation of this benchmark, we use a threshold of 
	\textbf{250} to distinguish between \emph{good} and \emph{bad} models --- 
	this threshold value induces a large margin from rewards pertaining to 
	poorly-performing models (which usually reached rewards lower than $100$).
	
	Note that as seen in Fig.~\ref{fig:cartpolePercentileGoodBadResults}, our 
	algorithm eventually also removes \textit{some} of the more successful 
	models. 
	However, the final result contains \textit{only} well-performing models, as 
	in 
	the other benchmarks.

	\begin{figure}[ht]
		\centering
		\captionsetup{justification=centering}
		\includegraphics[width=0.5\textwidth]{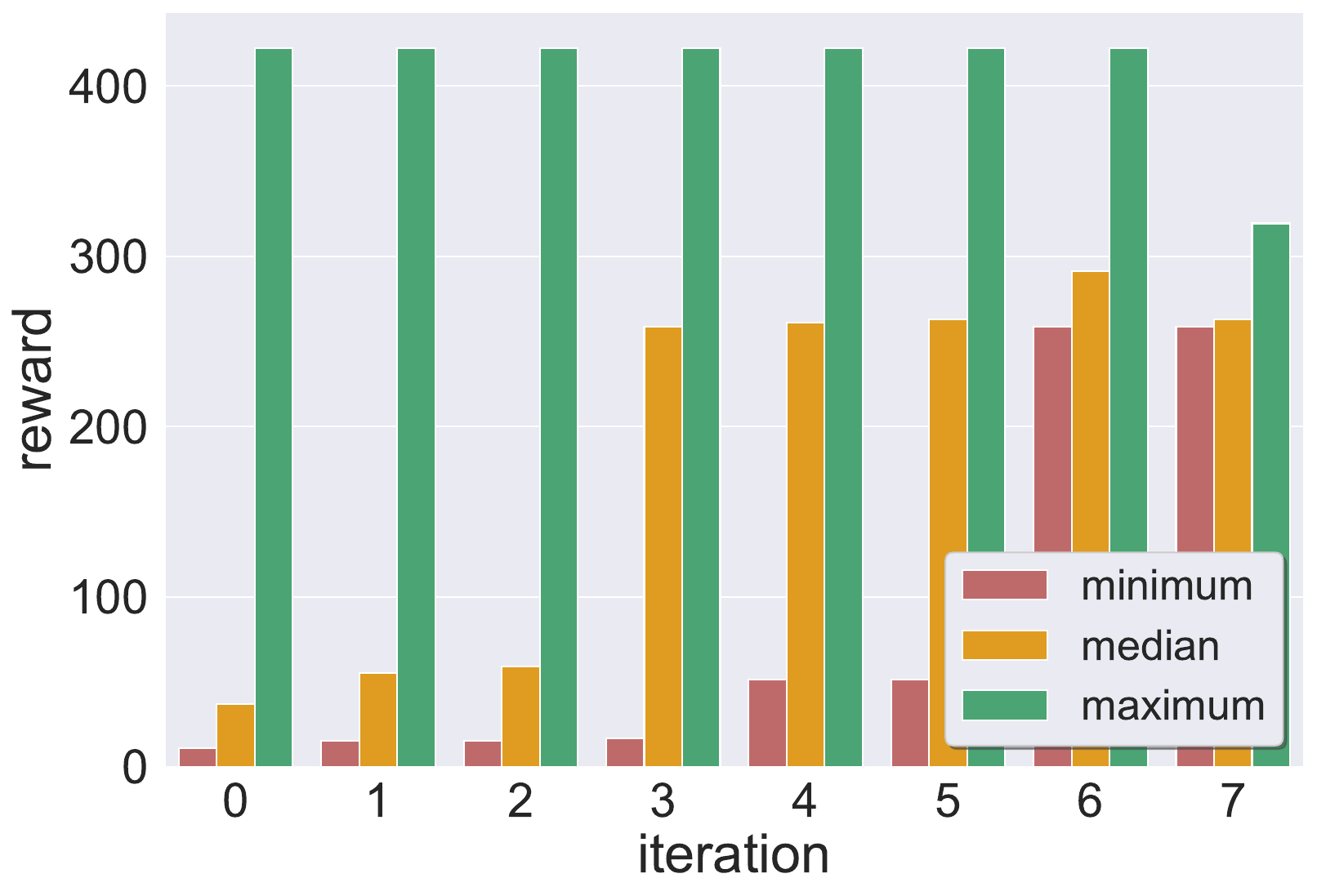}
		\caption{Cartpole: model selection results: minimum, median, and 
		maximum 
			rewards of the models selected after each iteration. Our technique  
			selected models \{6,7,9\}.
		}
		\label{fig:cartpolePercentileMinMaxRewards}
	\end{figure}
	\FloatBarrier
	
	\clearpage
	
	\subsection{Result per Filtering Criteria}
	
	\begin{figure}[!ht]
		\centering
		\captionsetup[subfigure]{justification=centering}
		\captionsetup{justification=centering} 
		\begin{subfigure}[t]{0.49\linewidth}
			\centering
			\includegraphics[width=\textwidth]{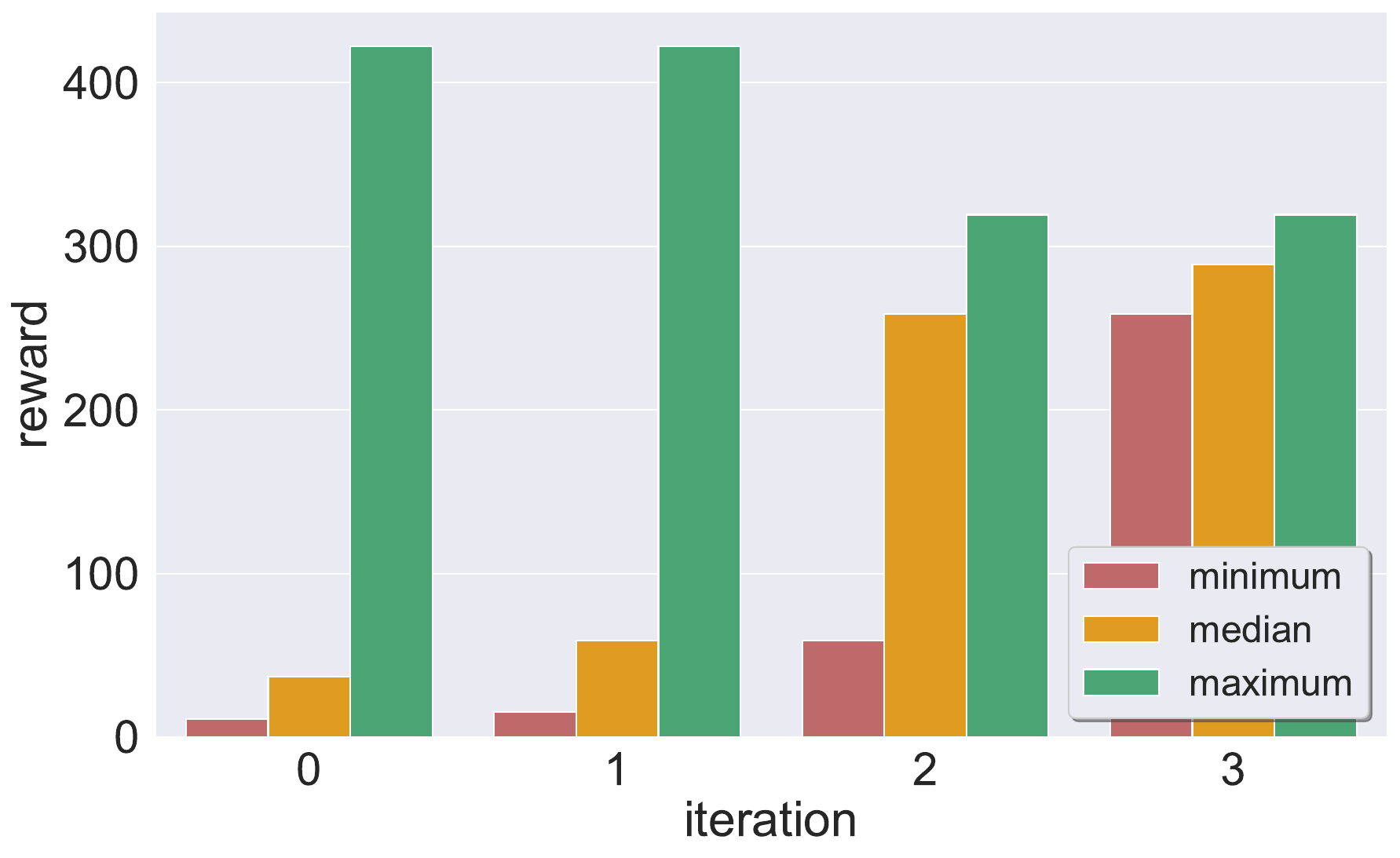}
			\caption{Reward statistics of remaining models}
			\label{}
		\end{subfigure}
		\hfill
		\begin{subfigure}[t]{0.49\linewidth}
			\includegraphics[width=\textwidth, 
			height=0.61\textwidth]{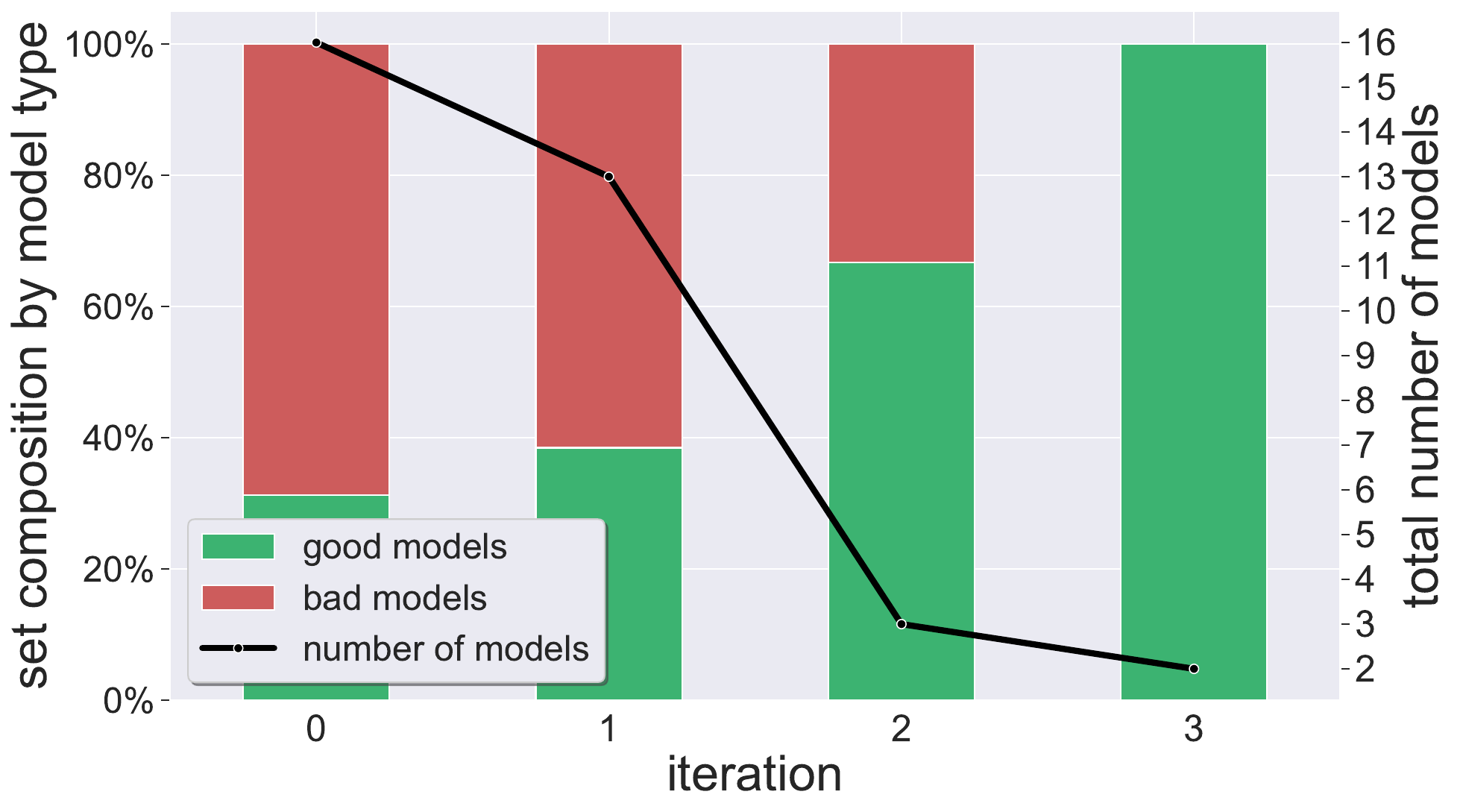}
			\caption{Ratio between good/bad models}
			\label{}
		\end{subfigure}
		
		\caption{Cartpole: results using the \maxAgg filtering criterion.
			Our technique  selected models \{7, 9\}.}
		\label{fig:cartpole:maxResults}
	\end{figure}
	
	\begin{figure}[!h]
		\centering
		\captionsetup[subfigure]{justification=centering}
		\captionsetup{justification=centering} 
		\begin{subfigure}[t]{0.49\linewidth}
			\centering
			\includegraphics[width=\textwidth]{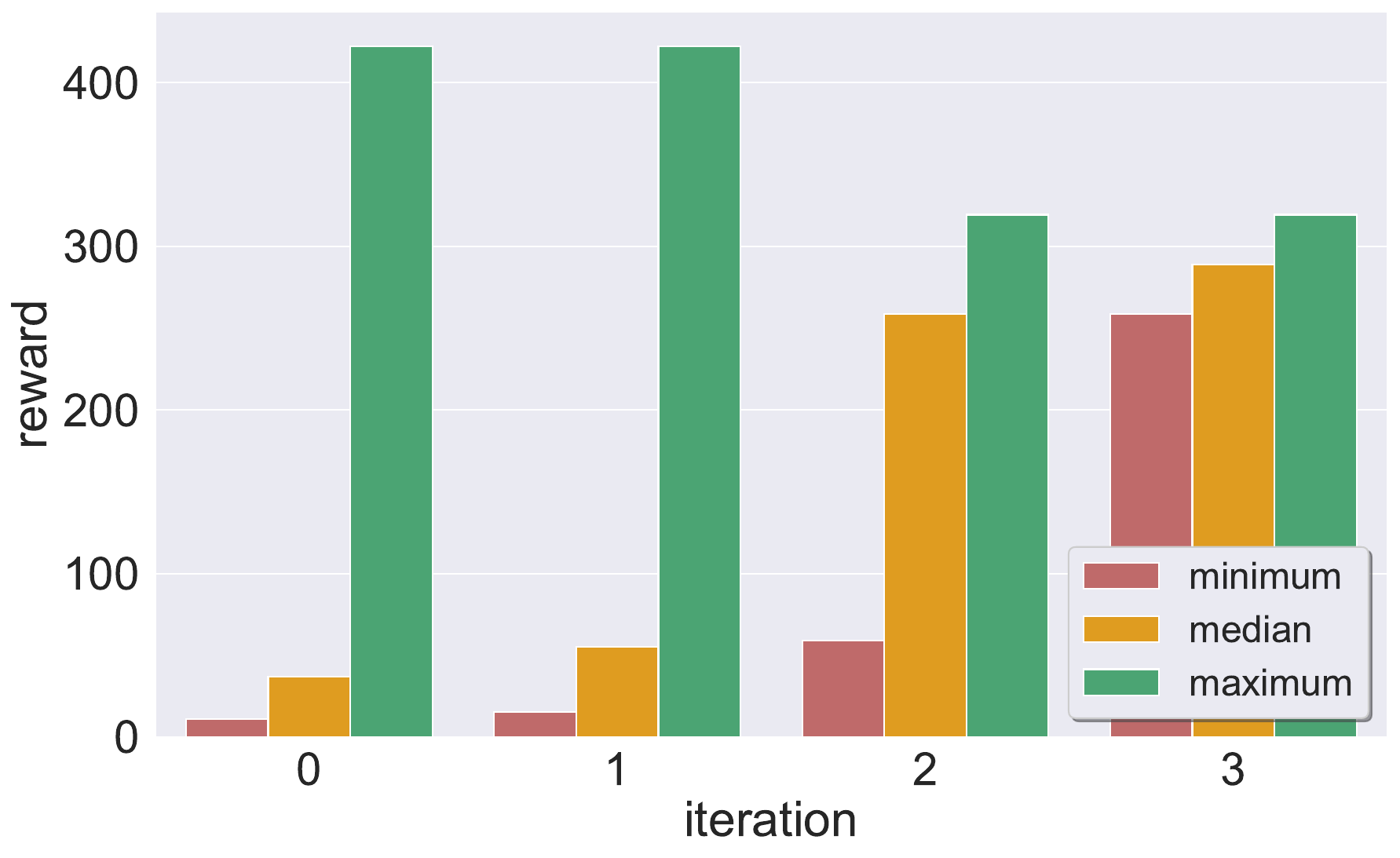}
			\caption{Reward statistics of remaining models}
			\label{}
		\end{subfigure}
		\hfill
		\begin{subfigure}[t]{0.49\linewidth}
			\includegraphics[width=\textwidth]{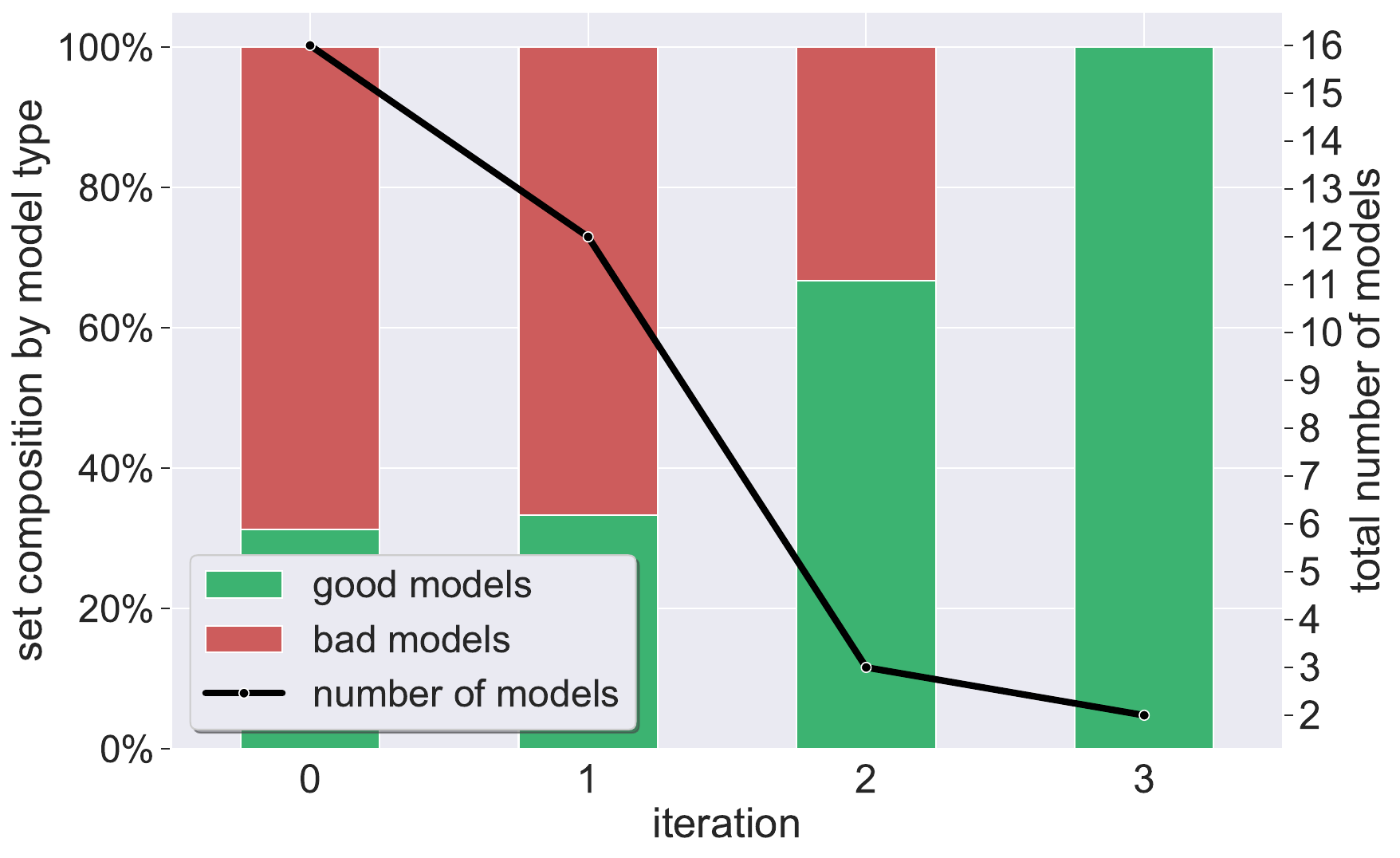}
			\caption{Ratio between good/bad models}
			\label{}
		\end{subfigure}
		\caption{Cartpole: results using the \conditionCombined filtering 
			criterion. Our technique  selected models \{7, 9\}.}
		\label{fig:cartpole:CombinedResults}
	\end{figure}
	\FloatBarrier
	
	\newpage

	\section{Mountain Car: Supplementary Results}
	\label{sec:appendix:MountainCarSupplementaryResults}
	
	\subsection{The Mountain Car Benchmark}

\begin{figure}[ht]
		\centering
		\captionsetup[subfigure]{justification=centering}
		\captionsetup{justification=centering} 
		\begin{subfigure}[t]{0.49\linewidth}
			\includegraphics[width=\textwidth]{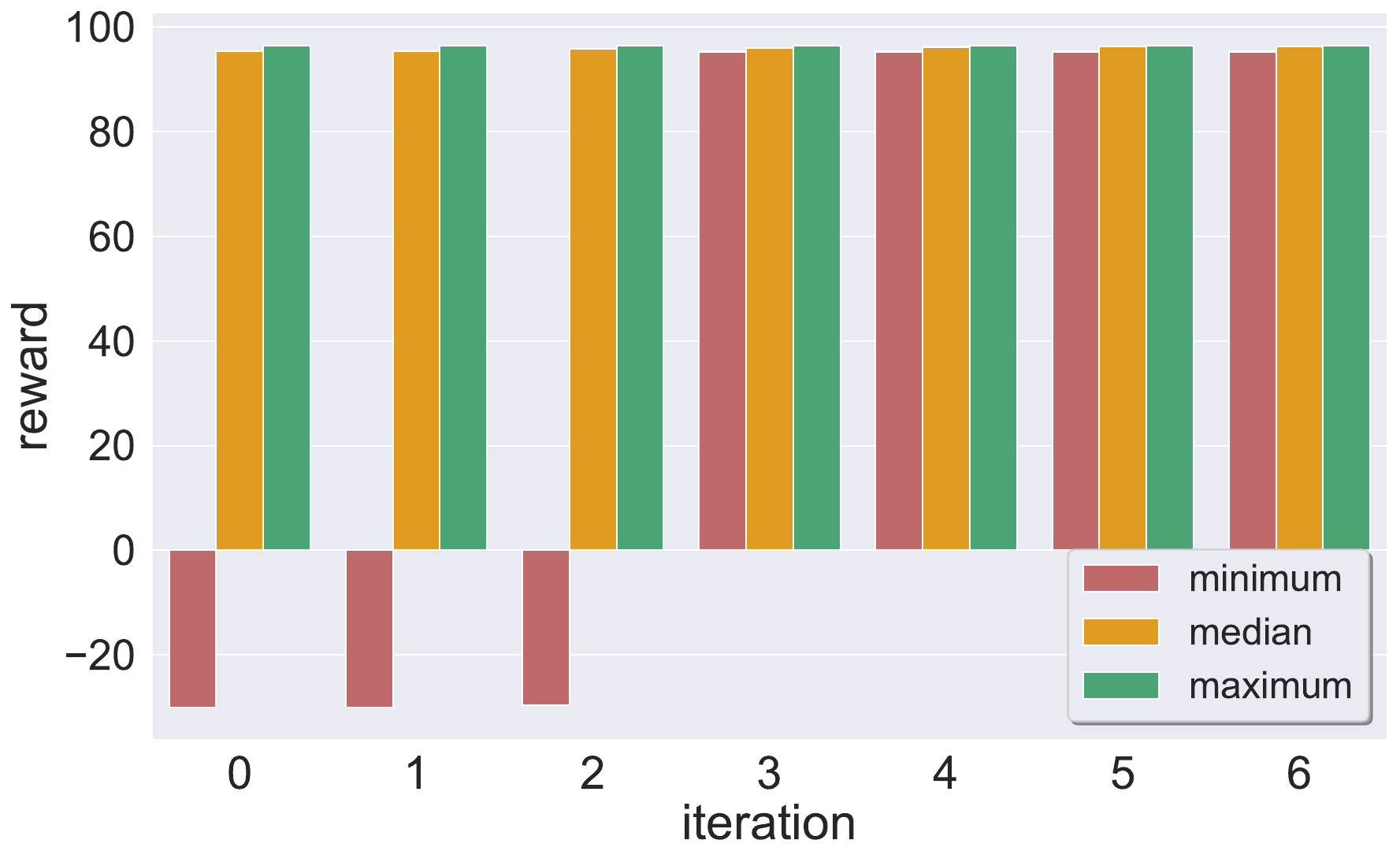}
			\caption{Reward statistics of remaining models}
			\label{fig:mountaincar:PercentileMinMaxRewards}
		\end{subfigure}
		\hfill
		\begin{subfigure}[t]{0.49\linewidth}
			\centering
			\includegraphics[width=\textwidth]{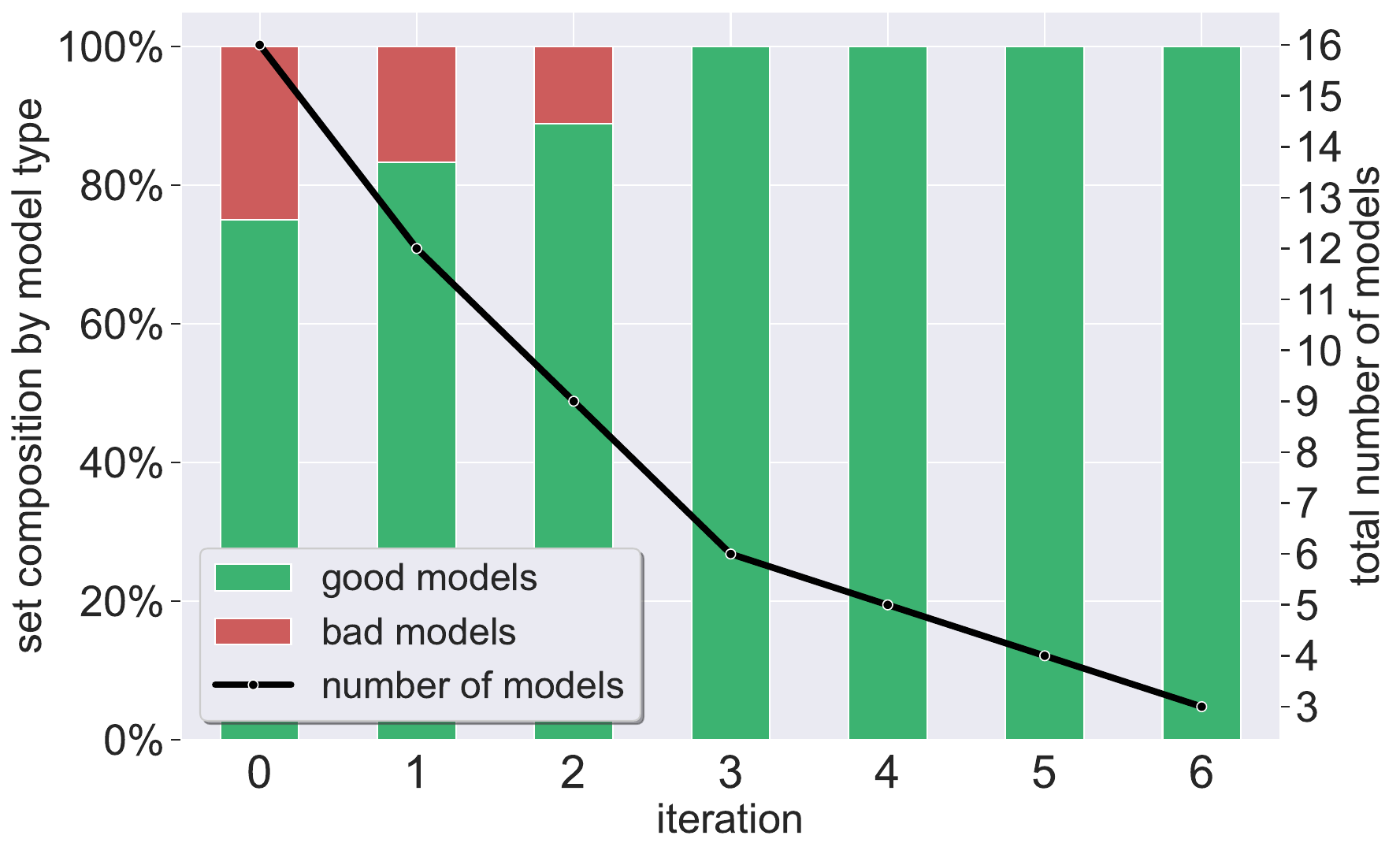}
			\caption{Ratio between good/bad models}
			\label{fig:mountaincar:goodBadModelsPercentages}
		\end{subfigure}
		\caption{Mountain Car: results using the \conditionPercentile filtering 
			criterion. Our technique  selected models \{8, 10, 15\}.}
		\label{fig:mountaincar:PrecentileCritResults}
	\end{figure}

\begin{figure}[!h]
		\centering
		\captionsetup[subfigure]{justification=centering}
		\captionsetup{justification=centering} 
		\begin{subfigure}[t]{0.49\linewidth}
			\centering
			\includegraphics[width=\textwidth]{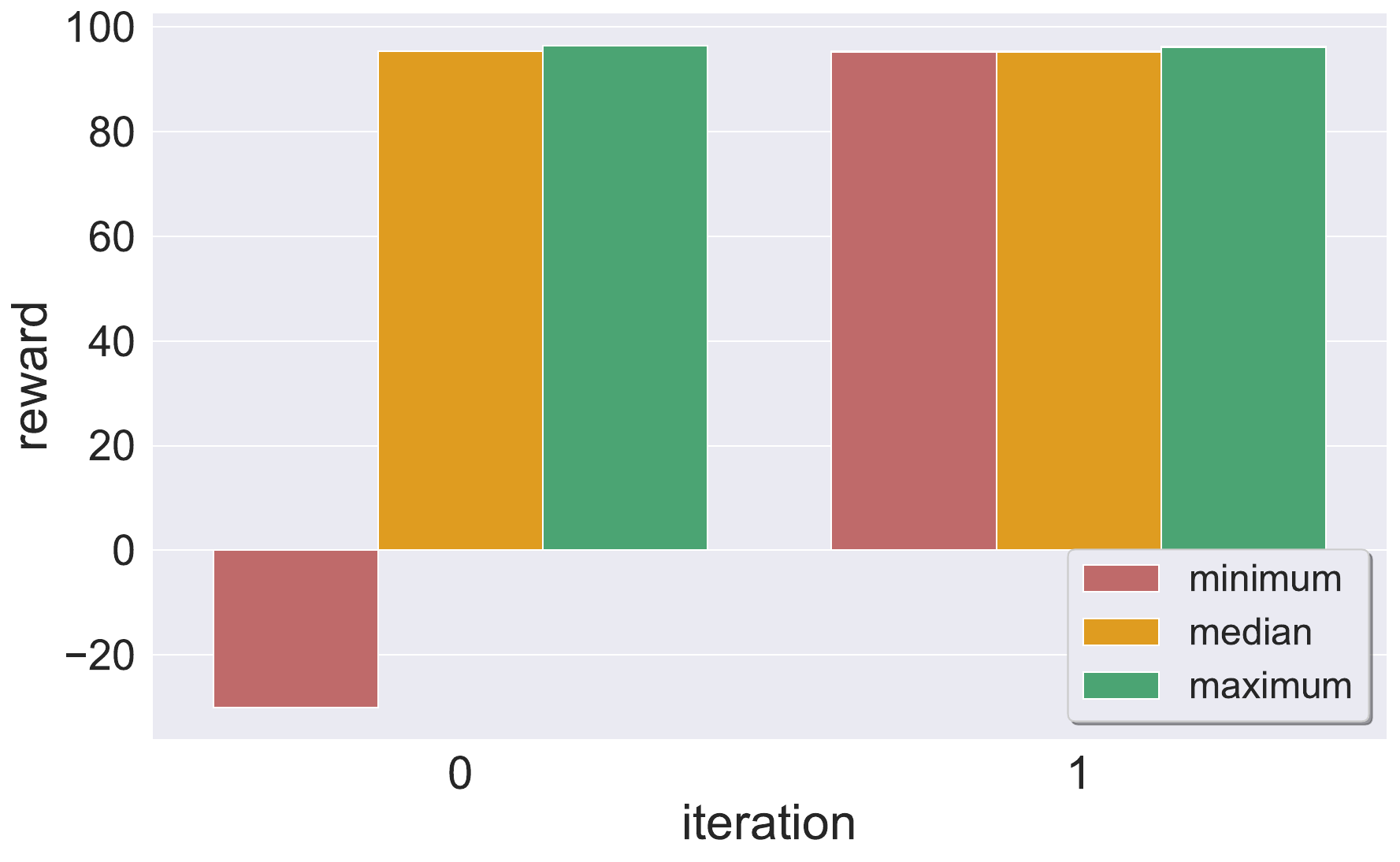}
			\caption{Reward statistics of remaining models}
			\label{}
		\end{subfigure}
		\hfill
		\begin{subfigure}[t]{0.49\linewidth}
			\includegraphics[width=\textwidth]{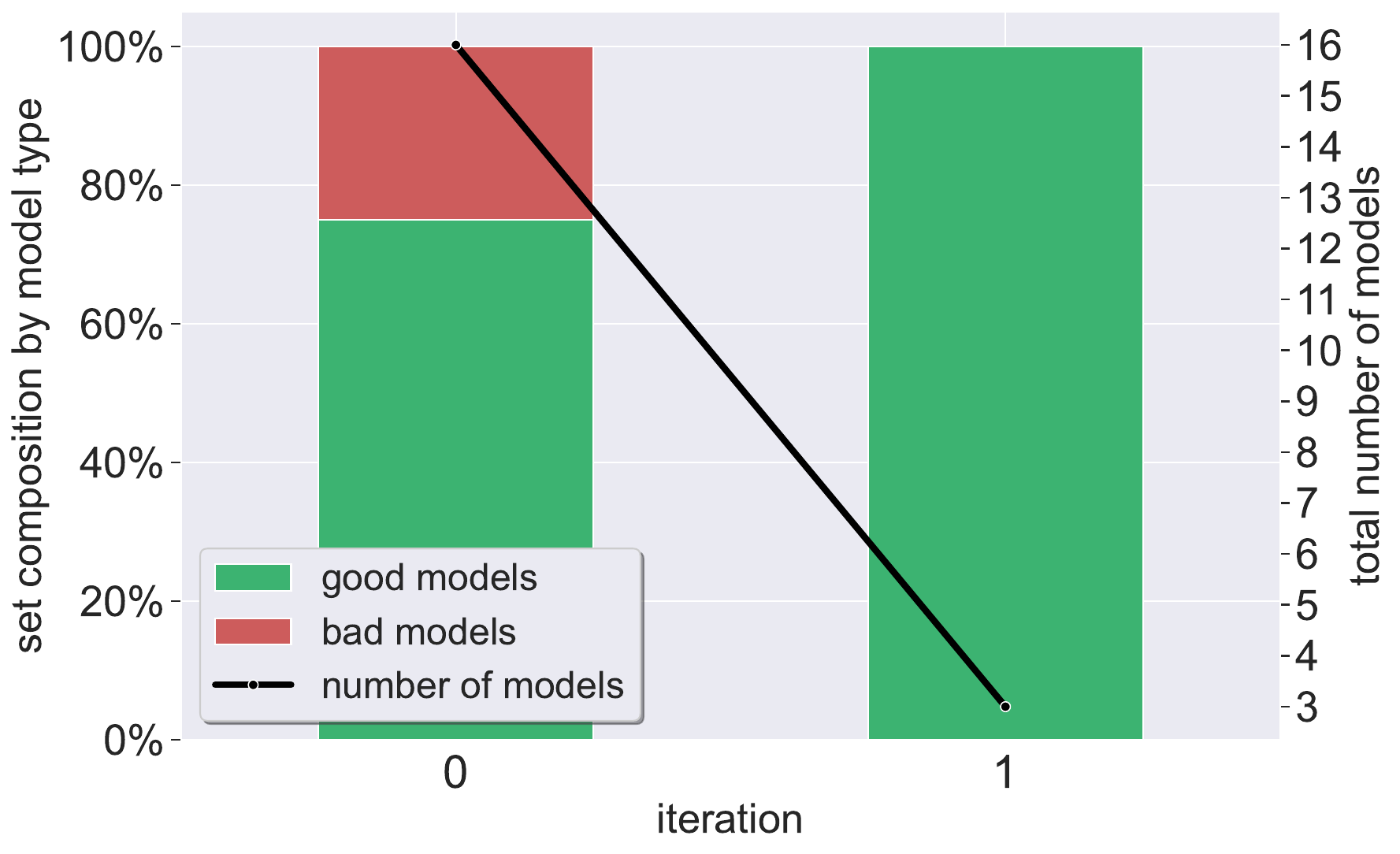}
			\caption{Ratio between good/bad models}
			\label{}
		\end{subfigure}
		\caption{Mountain Car: results using the \maxAgg filtering criterion.
			Our technique selected models \{2, 4, 15\}.}
		\label{fig:mountaincar:MaxCritResults}
	\end{figure}

\begin{figure}[!h]
		\centering
		\captionsetup[subfigure]{justification=centering}
		\captionsetup{justification=centering} 
		\begin{subfigure}[t]{0.49\linewidth}
			\centering
			\includegraphics[width=\textwidth]{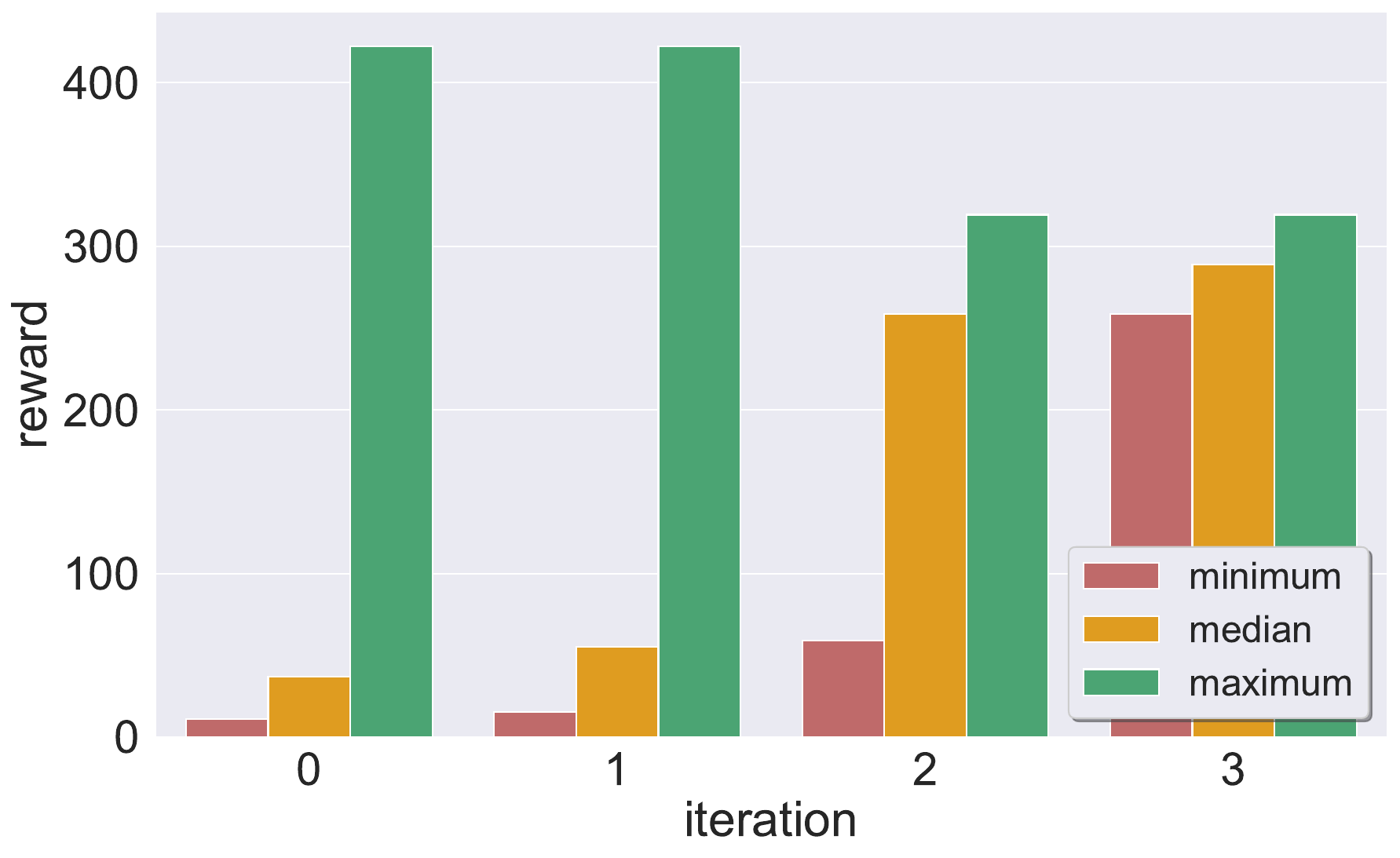}
			\caption{Reward statistics of remaining models}
			\label{}
		\end{subfigure}
		\hfill
		\begin{subfigure}[t]{0.49\linewidth}
			\includegraphics[width=\textwidth]{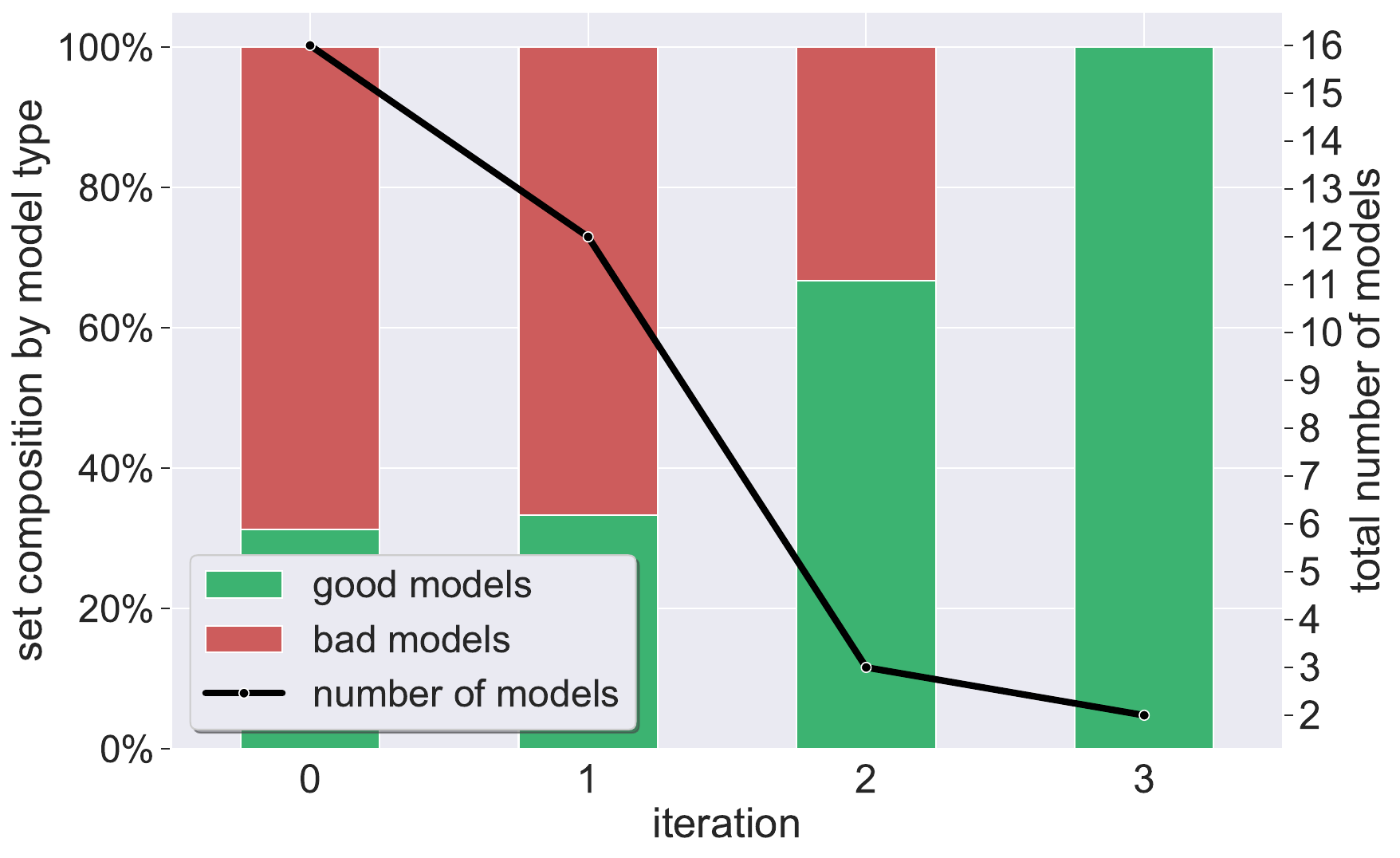}
			\caption{Ratio between good/bad models}
			\label{}
		\end{subfigure}
		\caption{Mountain Car: results using the \conditionCombined filtering 
			criterion.
			Our technique  selected models \{2, 4, 15\}.}
		\label{fig:mountaincar:CombinedCritResults}
	\end{figure}

	We note that our algorithm is robust to various hyperparameter choices, as 
	demonstrated in Fig.~\ref{fig:mountaincar:PrecentileCritResults}, 
	Fig.~\ref{fig:mountaincar:MaxCritResults} and 
	Fig.~\ref{fig:mountaincar:CombinedCritResults} which depict the results of 
	each 
	iteration of our algorithm, when applied with different filtering criteria 
	(elaborated in Appendix~\ref{sec:appendix:AlgorithmAdditionalInformation}).

	\FloatBarrier


	\newpage
	\subsection{Additional Filtering Criteria}
	
	\FloatBarrier

	
	\subsection{Combinatorial Experiments}
	
\begin{figure}[!h]
		\centering
		\captionsetup{justification=centering}
		\includegraphics[width=0.6\textwidth]{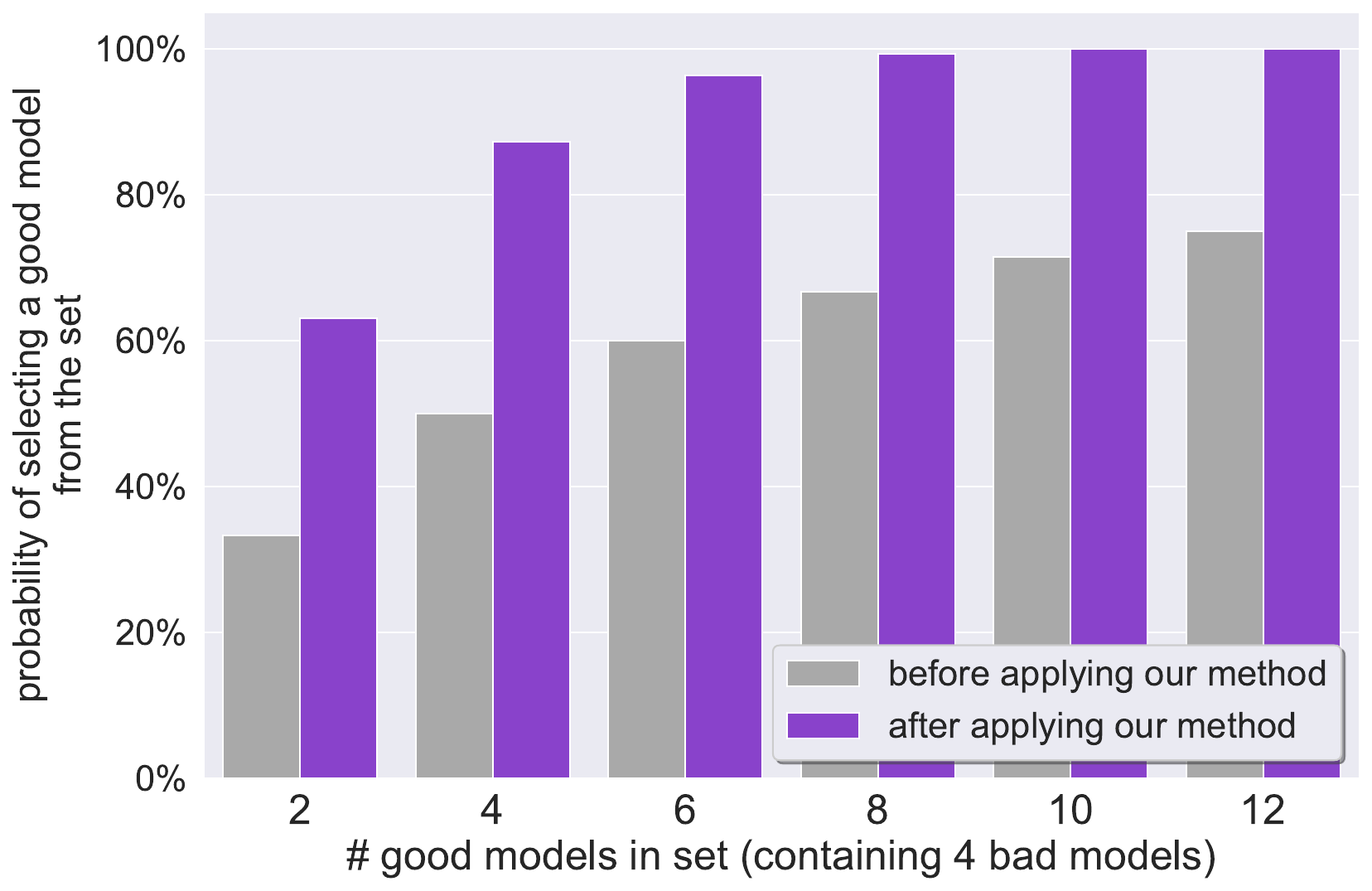}
		
		\caption{Mountain Car: our algorithm effectively increases the 
		probability 
			to choose a good model, due to effective filtering. The plot 
			corresponds 
			to 
			Table~\ref{table:results:MountainCarCombinatorialResultsTableMaxConditionMaxHyperparameter2}.}
		\label{fig:mountaincarProbabilitiesMaxConditionHyperparameter2}
	\end{figure}

	Due to the original bias of the initial set of candidates, in which $12$ 
	out of 
	the original $16$ models are good in the OOD setting, we set out to 
	validate 
	that the fact that our algorithm succeeded in returning solely good models 
	is 
	indeed due to its correctness, and not due to the inner bias among the set 
	of 
	models, to contain good models. In our experiments (summarized below) we 
	artificially generated new sets of models in which the ratio of good models 
	is 
	deliberately lower than in the original set. We then reran our algorithm on 
	all 
	possible combinations of the initial subsets, and calculated (for each 
	subset) 
	the probability of selecting a good model in this new setting, from the 
	models 
	surviving our filtering process. As we show, our method significantly 
	improves 
	the chances of selecting a good model \emph{even when these are a minority 
	in 
		the original set}. For example, the leftmost column of 
	Fig.~\ref{fig:mountaincarProbabilitiesMaxConditionHyperparameter2} shows 
	that 
	over sets consisting of $4$ bad models and only $2$ good ones, the 
	probability 
	of selecting a good model after running our algorithm is over $60\%$ (!) 
	--- 
	almost double the probability of randomly selecting a good model from the 
	original set before running our algorithm. These results were consistent 
	across 
	multiple subset sizes, and with various filtering criteria.
	
	\medskip
	\noindent
	\textbf{Note.} For the calculations demonstrating the chance to select a 
	good 
	model, we assume random selection from a subset of models: \emph{before} 
	applying our algorithm, the subset is the original set of models; and 
	\emph{after} our algorithm is applied --- the subset is updated based on 
	the 
	result of our filtering procedure. The probability is computed based on the 
	number of combinations of bad models surviving the filtering process, and 
	their 
	ratio relative to all the models returned in those cases (we assume uniform 
	probability, per subset).

	%
	

	
	
	\begin{table}[ht]
		\centering
		\captionsetup{justification=centering}     
		\begin{tabular}
			{ |P{0.07\linewidth}:P{0.07\linewidth}| 
				P{0.2\linewidth}|P{0.18\linewidth}:P{0.18\linewidth}|P{0.15\linewidth}:P{0.15\linewidth}|}
			\hline
			\multicolumn{2}{|P{0.14\linewidth}|}{\texttt{\textbf{COMPOSITION}}} 
			& 
			\multicolumn{1}{P{0.2\linewidth}|}{\texttt{\textbf{total $\#$ 
						experiments}}} & 
			\multicolumn{2}{ P{0.36\linewidth}|}{\texttt{\textbf{$\#$ 
			experiments 
						with}}} & 
			\multicolumn{2}{ P{0.3\linewidth}|}{\texttt{\textbf{probability to 
						choose 
						\textcolor{forestGreen}{good} models}}} \\ 
			\hline
			\textcolor{forestGreen}{good} & \textcolor{red}{bad} & 
			{\shortstack{subgroup \\combinations\\of 
					\textcolor{forestGreen}{good} models}} & 
			\shortstack{\textbf{all} \\surviving\\ models 
				\textcolor{forestGreen}{good}} & 
			\shortstack{\textbf{some}\\surviving\\ models \textcolor{red}{bad}} 
			& 
			naive 
			& our method \\ 
			\hline
			
			\textcolor{forestGreen}{12} & 
			\textcolor{red}{4} & 
			${12 \choose\textcolor{forestGreen}{12}} = 1$ & 
			1 & 
			0 & 
			75 \%
			&
			100\% \\
			
			\hline
			\textcolor{forestGreen}{10} &
			\textcolor{red}{4} & 
			${12 \choose \textcolor{forestGreen}{10}} = 66$ & 
			65 & 
			1 & 
			71.43 \%
			& 
			99.49 \% 
			\\
			
			\hline
			\textcolor{forestGreen}{8} & 
			\textcolor{red}{4} & 
			${12 \choose \textcolor{forestGreen}{8}} =  495$ & 
			423 & 
			72 & 
			66.67 \% &
			95.15 \%
			\\

			\hline
			\textcolor{forestGreen}{6} & 
			\textcolor{red}{4} & 
			${12 \choose \textcolor{forestGreen}{6}} = 924$ & 
			549 & 
			375 & 
			60 \% &
			86.26 \%  
			\\ 
			
			\hline
			\textcolor{forestGreen}{4} 
			& 
			\textcolor{red}{4} & 
			${12 \choose \textcolor{forestGreen}{4}} =  495$ & 
			123 & 372 & 
			50 \%
			& 
			74.01 \%
			\\

			\hline
			\textcolor{forestGreen}{2} & 
			\textcolor{red}{4} & 
			${12 \choose \textcolor{forestGreen}{2}} = 66$ & 
			66 & 
			0 & 
			33.33 \%
			& 
			54.04 \%   
			\\ 
			
			\hline
			
		\end{tabular}
		
		\vspace{3mm}
		\caption{Results summary of the combinatorial Mountain Car experiment, 
			using the \conditionPercentile filtering criterion.}
		\label{table:results:MountainCarCombinatorialResultsTablePercentileCondition}
	\end{table}
	\FloatBarrier

	
	\begin{table}[ht]
		\centering
		\captionsetup{justification=centering} 
		\begin{tabular}
			{ |P{0.07\linewidth}:P{0.07\linewidth}| 
				P{0.2\linewidth}|P{0.18\linewidth}:P{0.18\linewidth}|P{0.15\linewidth}:P{0.15\linewidth}|}
			\hline
			\multicolumn{2}{|P{0.14\linewidth}|}{\texttt{\textbf{COMPOSITION}}} 
			& 
			\multicolumn{1}{P{0.2\linewidth}|}{\texttt{\textbf{total $\#$ 
						experiments}}} & 
			\multicolumn{2}{ P{0.36\linewidth}|}{\texttt{\textbf{$\#$ 
			experiments 
						with}}} & 
			\multicolumn{2}{ P{0.3\linewidth}|}{\texttt{\textbf{probability to 
						choose 
						\textcolor{forestGreen}{good} models}}} \\ 
			\hline
			\textcolor{forestGreen}{good} & \textcolor{red}{bad} & 
			{\shortstack{subgroup \\combinations\\of 
					\textcolor{forestGreen}{good} models}} & 
			\shortstack{\textbf{all} \\surviving\\ models 
				\textcolor{forestGreen}{good}} & 
			\shortstack{\textbf{some}\\surviving\\ models \textcolor{red}{bad}} 
			& 
			naive 
			& our method \\ 
			\hline

			\textcolor{forestGreen}{12} & 
			\textcolor{red}{4} & 
			${12 \choose\textcolor{forestGreen}{12}} = 1$ & 
			1 & 
			0 & 
			75 \%
			&
			100\% \\
			
			\hline
			\textcolor{forestGreen}{10} &
			\textcolor{red}{4} & 
			${12 \choose \textcolor{forestGreen}{10}} = 66$ & 
			66 & 
			0 & 
			71.43 \%
			& 
			100 \% 
			
			\\
			
			\hline
			\textcolor{forestGreen}{8} & 
			\textcolor{red}{4} & 
			${12 \choose \textcolor{forestGreen}{8}} =  495$ & 
			486 & 
			9 & 
			66.67 \% &
			99.34 \%
			\\

			\hline
			\textcolor{forestGreen}{6} & 
			\textcolor{red}{4} & 
			${12 \choose \textcolor{forestGreen}{6}} = 924$ & 
			844 & 
			80 & 
			60 \% &
			96.33 \%  
			\\ 
			
			\hline
			\textcolor{forestGreen}{4} 
			& 
			\textcolor{red}{4} & 
			${12 \choose \textcolor{forestGreen}{4}} =  495$ & 
			375 & 120 & 
			50 \%
			& 
			87.32 \%
			\\

			\hline
			\textcolor{forestGreen}{2} & 
			\textcolor{red}{4} & 
			${12 \choose \textcolor{forestGreen}{2}} = 66$ & 
			31 & 
			35 & 
			33.33 \%
			& 
			63.03 \%   
			\\ 
			
			\hline
		\end{tabular}
		
		\vspace{3mm}
		\caption{Results summary of the combinatorial Mountain Car experiment, 
			using the \conditionMax filtering criterion.}
		\label{table:results:MountainCarCombinatorialResultsTableMaxConditionMaxHyperparameter2}
	\end{table}

	\begin{table}[ht]
		\centering
		\captionsetup[subfigure]{justification=centering}
		\captionsetup{justification=centering} 
		\begin{tabular}
			{ |P{0.07\linewidth}:P{0.07\linewidth}| 
				P{0.2\linewidth}|P{0.18\linewidth}:P{0.18\linewidth}|P{0.15\linewidth}:P{0.15\linewidth}|}
			\hline
			\multicolumn{2}{|P{0.14\linewidth}|}{\texttt{\textbf{COMPOSITION}}} 
			& 
			\multicolumn{1}{P{0.2\linewidth}|}{\texttt{\textbf{total $\#$ 
						experiments}}} & 
			\multicolumn{2}{ P{0.36\linewidth}|}{\texttt{\textbf{$\#$ 
			experiments 
						with}}} & 
			\multicolumn{2}{ P{0.3\linewidth}|}{\texttt{\textbf{probability to 
						choose 
						\textcolor{forestGreen}{good} models}}} \\ 
			\hline
			\textcolor{forestGreen}{good} & \textcolor{red}{bad} & 
			{\shortstack{subgroup \\combinations\\of 
					\textcolor{forestGreen}{good} models}} & 
			\shortstack{\textbf{all} \\surviving\\ models 
				\textcolor{forestGreen}{good}} & 
			\shortstack{\textbf{some}\\surviving\\ models \textcolor{red}{bad}} 
			& 
			naive 
			& our method \\ 
			\hline
			\textcolor{forestGreen}{12} & 
			\textcolor{red}{4} & 
			${12 \choose\textcolor{forestGreen}{12}} = 1$ & 
			1 & 
			0 & 
			75 \%
			&
			100\% \\
			
			\hline
			\textcolor{forestGreen}{10} &
			\textcolor{red}{4} & 
			${12 \choose \textcolor{forestGreen}{10}} = 66$ & 
			66 & 
			0 & 
			71.43 \%
			& 
			100 \% \\
			
			\hline
			\textcolor{forestGreen}{8} & 
			\textcolor{red}{4} & 
			${12 \choose \textcolor{forestGreen}{8}} =  495$ & 
			481 & 
			12 & 
			66.67 \% &
			99.26 \%
			\\

			\hline
			\textcolor{forestGreen}{6} & 
			\textcolor{red}{4} & 
			${12 \choose \textcolor{forestGreen}{6}} = 924$ & 
			842 & 
			82 & 
			60 \% &
			96.74 \%  
			\\ 
			
			\hline
			\textcolor{forestGreen}{4} 
			& 
			\textcolor{red}{4} & 
			${12 \choose \textcolor{forestGreen}{4}} =  495$ & 
			372 & 123 & 
			50 \%
			& 
			88.09 \%
			\\

			\hline
			\textcolor{forestGreen}{2} & 
			\textcolor{red}{4} & 
			${12 \choose \textcolor{forestGreen}{2}} = 66$ & 
			31 & 
			35 & 
			33.33 \%
			& 
			63.36 \%   
			\\ 
			
			\hline
		\end{tabular}
		
		\vspace{3mm}
		\caption{Results summary of the combinatorial Mountain Car experiment, 
			using the \conditionCombined filtering criterion.}
		\label{table:results:MountainCarCombinatorialResultsTableCombinedConditionMaxHyperparameter2}
	\end{table}
	\FloatBarrier

	\clearpage
	
	\section{Aurora: Supplementary Results}
	\label{sec:appendix:AuroraSupplementaryResults}
	\subsection{Additional Information}
	
	\begin{enumerate}
		\item A detailed explanation of Aurora's input statistics:
		\begin{inparaenum}[(i)]
			\item \textit{Latency Gradient}: a derivative of latency (packet 
			delays) 
			over the recent MI (``monitor interval'');
			\item \textit{Latency Ratio}: the ratio between the average latency 
			in the 
			current MI to the
			minimum latency previously observed; and
			\item \textit{Sending Ratio}: the ratio between the number of 
			packets sent 
			to the number of acknowledged packets over the recent MI.
		\end{inparaenum} As mentioned, these metrics indicate the link's 
		congestion 
		level.
		
		\item For all our experiments on this benchmark, we defined ``good'' 
		models 
		as models that achieved an average reward greater/equal to a threshold 
		of 
		\textbf{99}; ``bad'' models are models that achieved a reward lower 
		than 
		this threshold.
		
		\item \emph{In-distribution}, the average reward is not necessarily 
		correlated with the average reward \emph{OOD}. For example, in 
		Exp.~\ref{exp:auroraShort} with the short episodes during training (see 
		Fig.~\ref{fig:auroraRewards}):
		
		\begin{enumerate}
			\item In-distribution, model $\{4\}$ achieved a lower reward than 
			models $\{2\}$ and $\{5\}$, but a higher reward OOD.
			
			\item In-distribution, model $\{16\}$ achieved a lower reward than 
			model $\{15\}$, but a higher reward OOD.
		\end{enumerate}
	\end{enumerate}

	%
	

	
	\experiment{Aurora: Long Training Episodes}
	~\label{exp:auroraLong}
	
	Similar to Experiment~\ref{exp:auroraShort}, we trained a new set of $k=16$ 
	agents. In this experiment, we increased each training episode to consist 
	of 
	$400$ steps (instead of $50$, as in the ``short'' training). The remaining 
	parameters were identical to the previous setup in 
	Experiment~\ref{exp:auroraShort}. This time, $5$ models performed poorly in 
	the 
	OOD environment (i.e., did not reach our reward threshold of $99$), while 
	the 
	remaining $11$ models performed well both in-distribution and OOD.
	
	When running our method with the \conditionMax criterion, our algorithm 
	returned $4$ models, all being a subset of the group of $11$ models which 
	generalized successfully, and after fully filtering out all the 
	unsuccessful 
	models. Running the algorithm with the \conditionPercentile or the 
	\conditionCombined criteria also yielded a subset of this group, indicating 
	that the 
	filtering process was again successful (and robust to various algorithm 
	hyperparameters).
	
	\begin{figure}[ht]
		\centering
		\captionsetup{justification=centering}
		\subfloat[In-distribution 
		\label{subfig:auroraLongRewards:inDist}]{\includegraphics[width=0.49\textwidth]{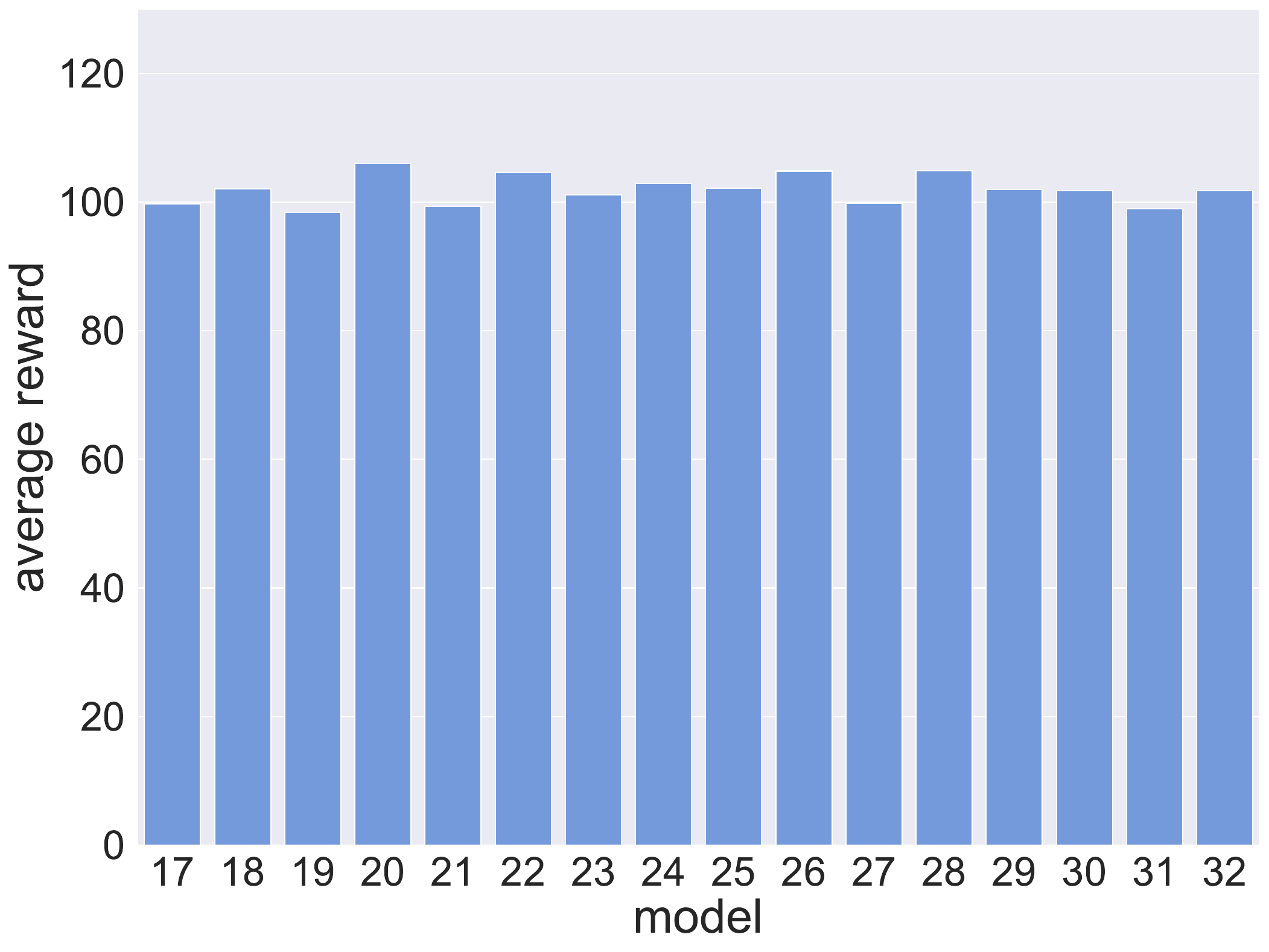}}
		\hfill
		\subfloat[OOD\label{subfig:auroraLongRewards:OOD}]{\includegraphics[width=0.49\textwidth]{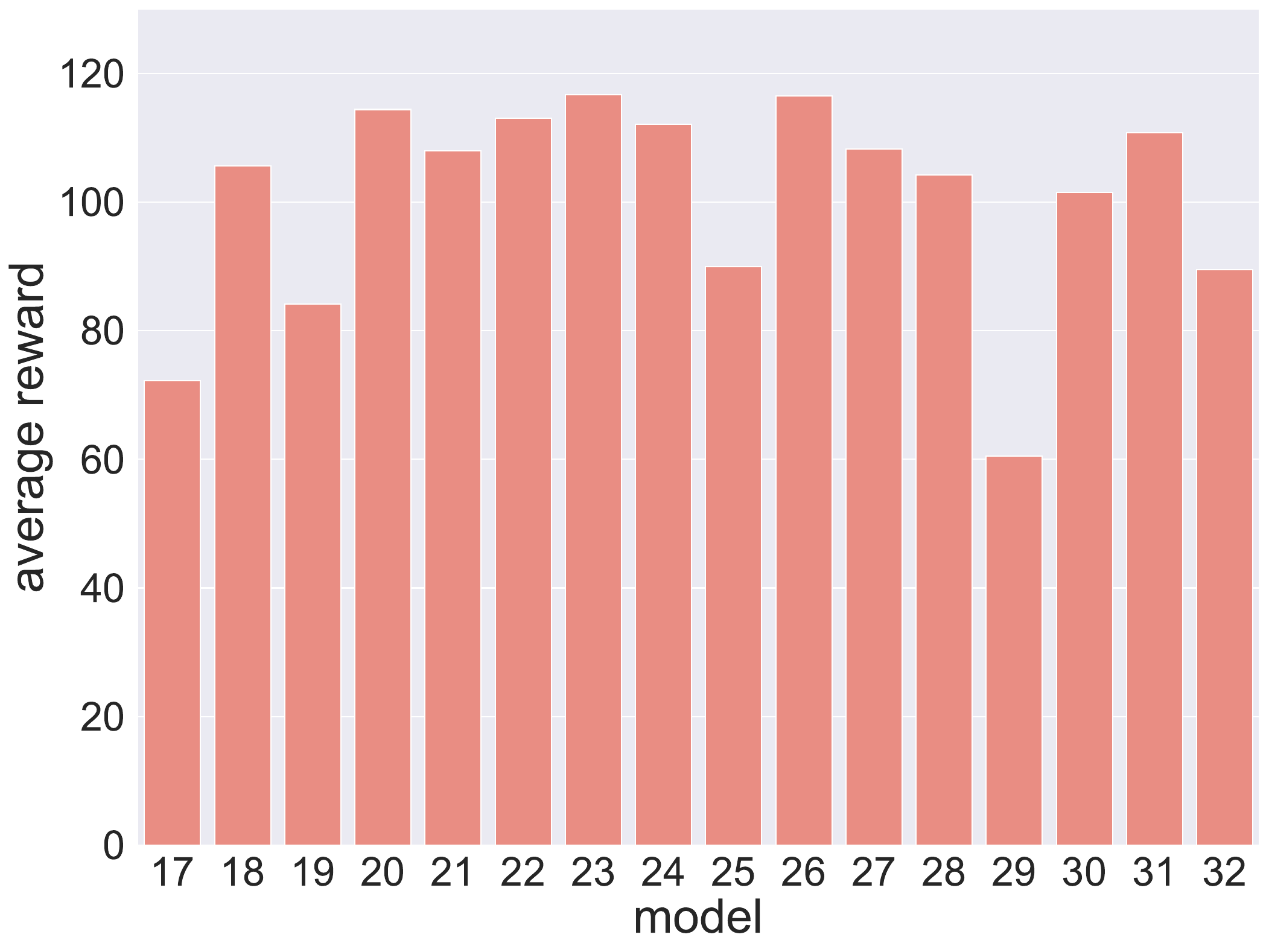}}\\
		
		\caption{Aurora Experiment~\ref{exp:auroraLong}: the models' average 
			rewards when simulated on different distributions.}
		\label{fig:auroraLongRewards}
	\end{figure}

	
	\begin{figure}[ht]
		\centering
		\captionsetup{justification=centering}
		\captionsetup[subfigure]{justification=centering}
		\captionsetup{justification=centering} 
		\begin{subfigure}[t]{0.49\linewidth}
			\centering
			\includegraphics[width=\textwidth, 
			height=0.67\textwidth]{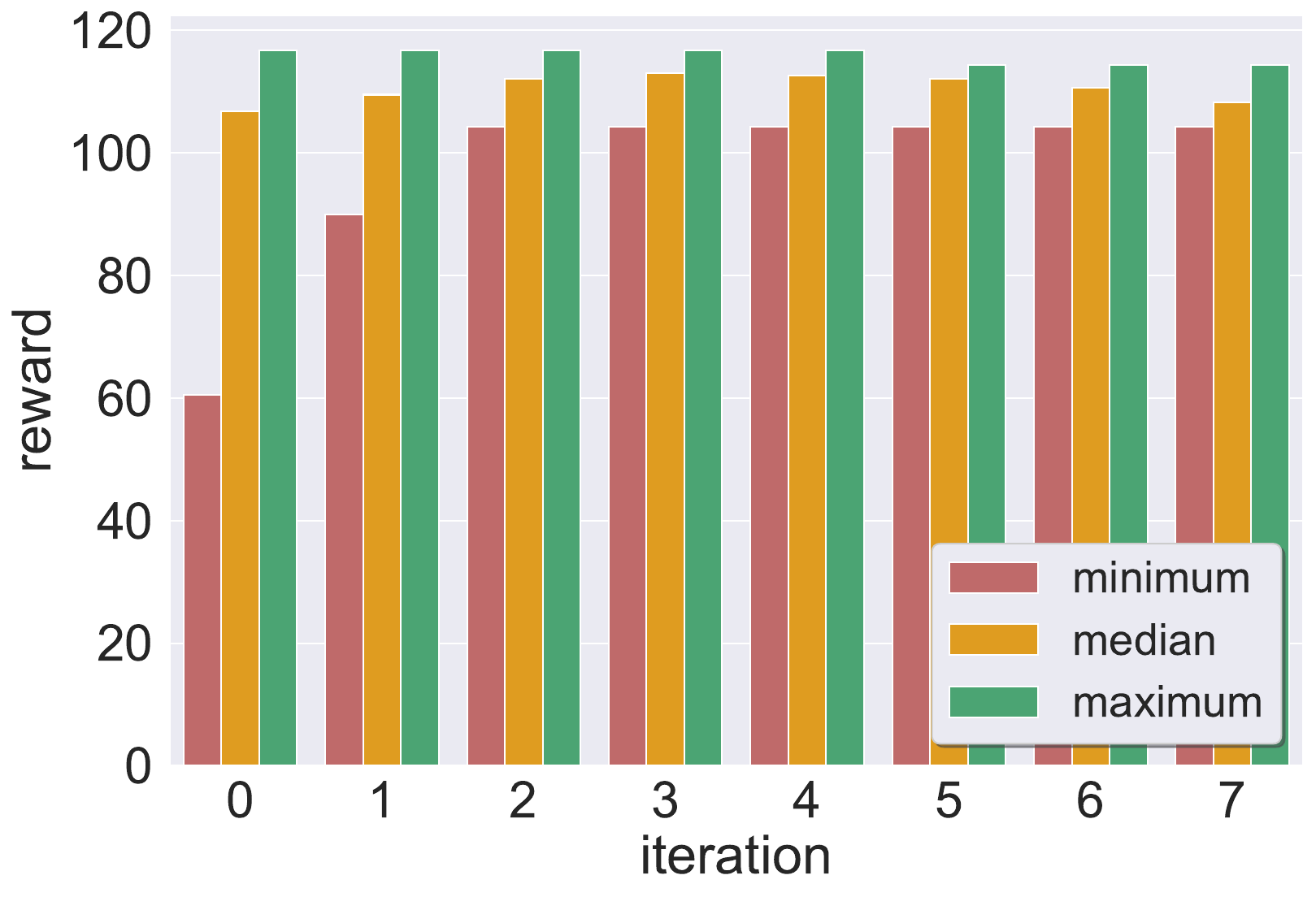}
			\caption{Reward statistics of remaining models}
			\label{}
		\end{subfigure}
		\hfill
		\begin{subfigure}[t]{0.49\linewidth}
			\includegraphics[width=\textwidth, 
			height=0.67\textwidth]{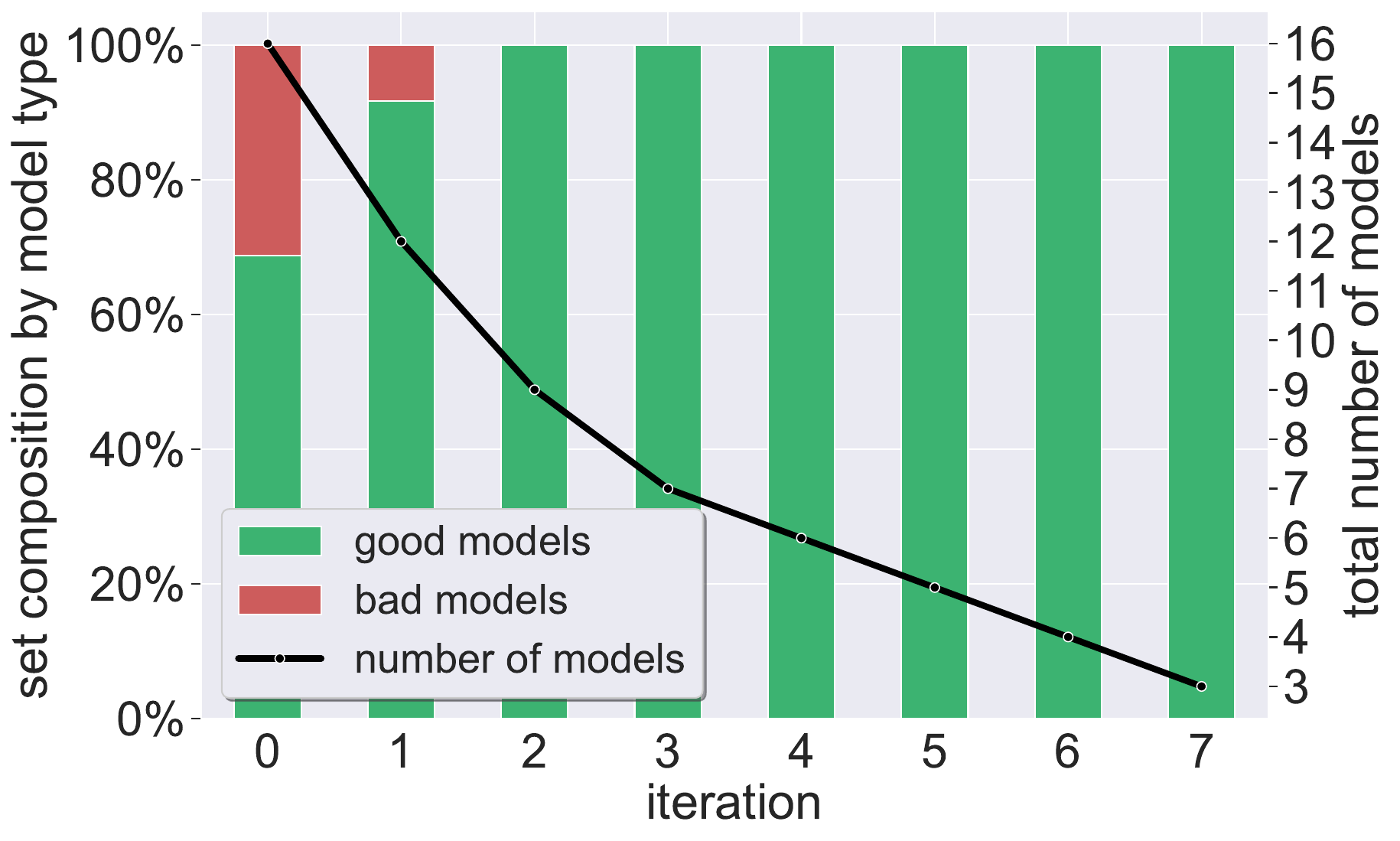}
			\caption{Ratio between good/bad models}
			\label{}
		\end{subfigure}

		\caption{Aurora Experiment~\ref{exp:auroraLong}: model selection 
		results.
			Our technique  selected models \{20, 27, 28\}.}
		\label{fig:auroraLongGoodBadModelsPercentages}
	\end{figure}

	\medskip
	\noindent
	\subsection{Additional Probability Density Functions}  
	
\begin{figure}
		\centering
		\captionsetup[subfigure]{justification=centering}
		\captionsetup{justification=centering}
		\begin{subfigure}[t]{0.32\linewidth}
			\includegraphics[width=\textwidth]{plots/aurora/models_rewards/short_episodes/rewards_In_distribution.pdf}
			
			\label{subfig:auroraShortRewards:inDist}
		\end{subfigure}
		\hfill
		\begin{subfigure}[t]{0.32\linewidth}
			\centering
			\includegraphics[width=\textwidth]{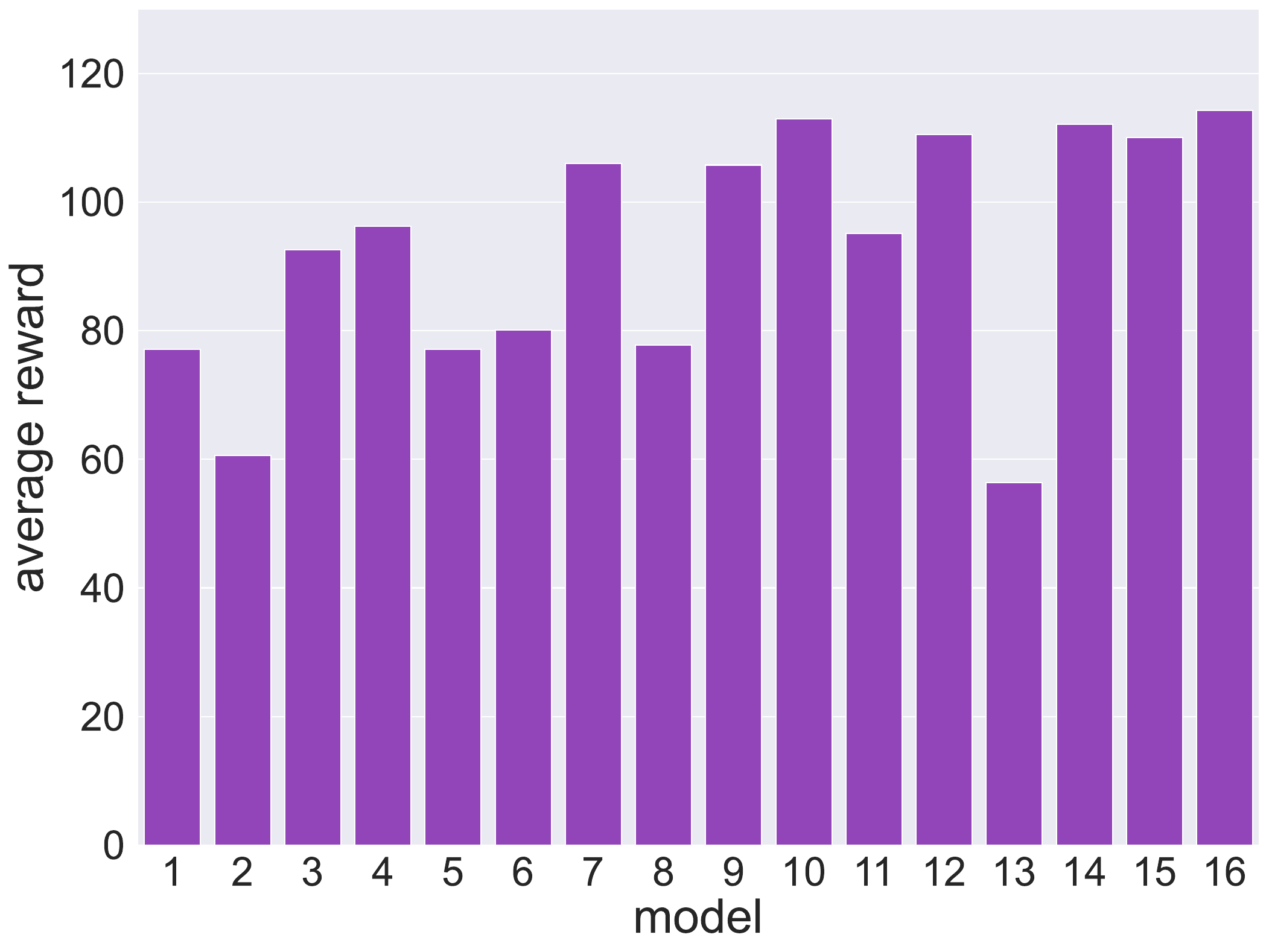}
			\label{subfig:auroraShortRewards:OODLowMean}
		\end{subfigure}
		\hfill
		\begin{subfigure}[t]{0.32\linewidth}
			\centering
			\includegraphics[width=\textwidth]{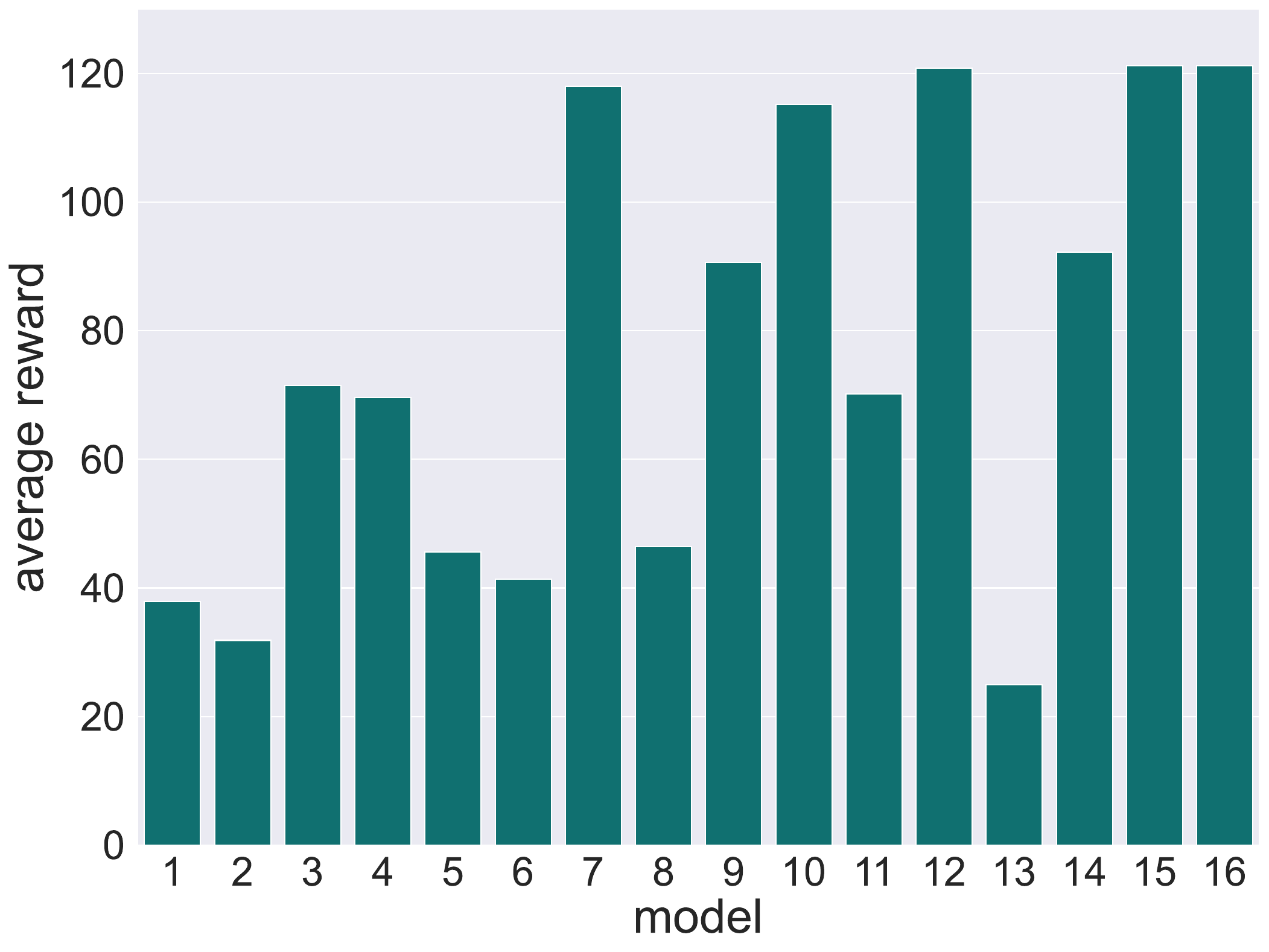}
			\label{subfig:auroraShortRewards:OODHighMean}
		\end{subfigure}     
		\hfill
		\begin{subfigure}[t]{0.32\linewidth}
			\includegraphics[width=\textwidth]{plots/aurora/models_rewards/long_episodes/rewards_In_distribution.pdf}
			
			\caption{In-distribution\\($\sim\mathcal{U}(low,high)$)}
			\label{subfig:auroraLongRewards:inDist}
		\end{subfigure}
		\hfill
		\begin{subfigure}[t]{0.32\linewidth}
			\centering
			\includegraphics[width=\textwidth]{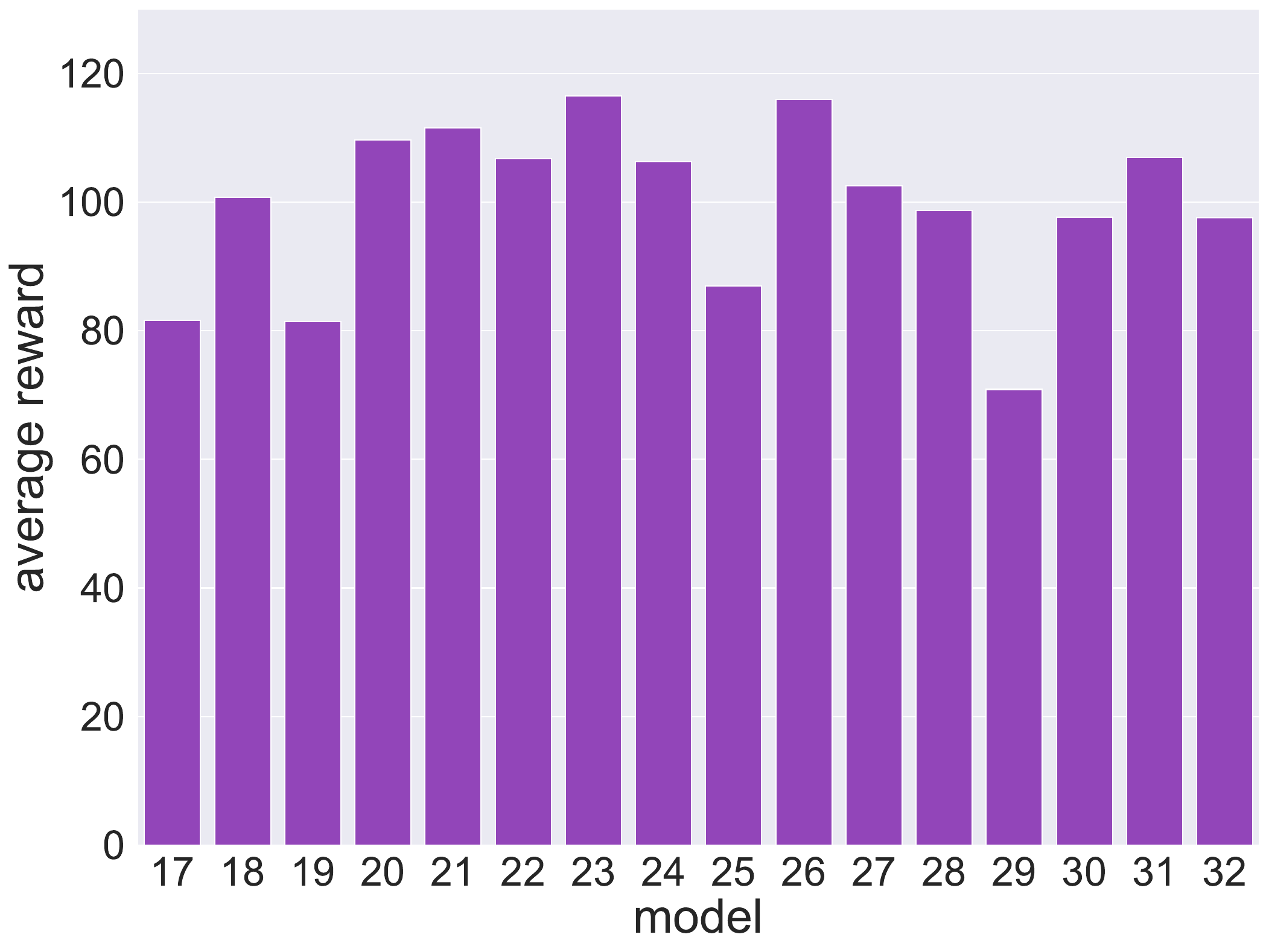}
			\caption{OOD ($\sim\mathcal{TN}(\mu_{low}, \sigma^{2}$))}
			\label{subfig:auroraLongRewards:OODLowMean}
		\end{subfigure}
		\hfill
		\begin{subfigure}[t]{0.32\linewidth}
			\centering
			\includegraphics[width=\textwidth]{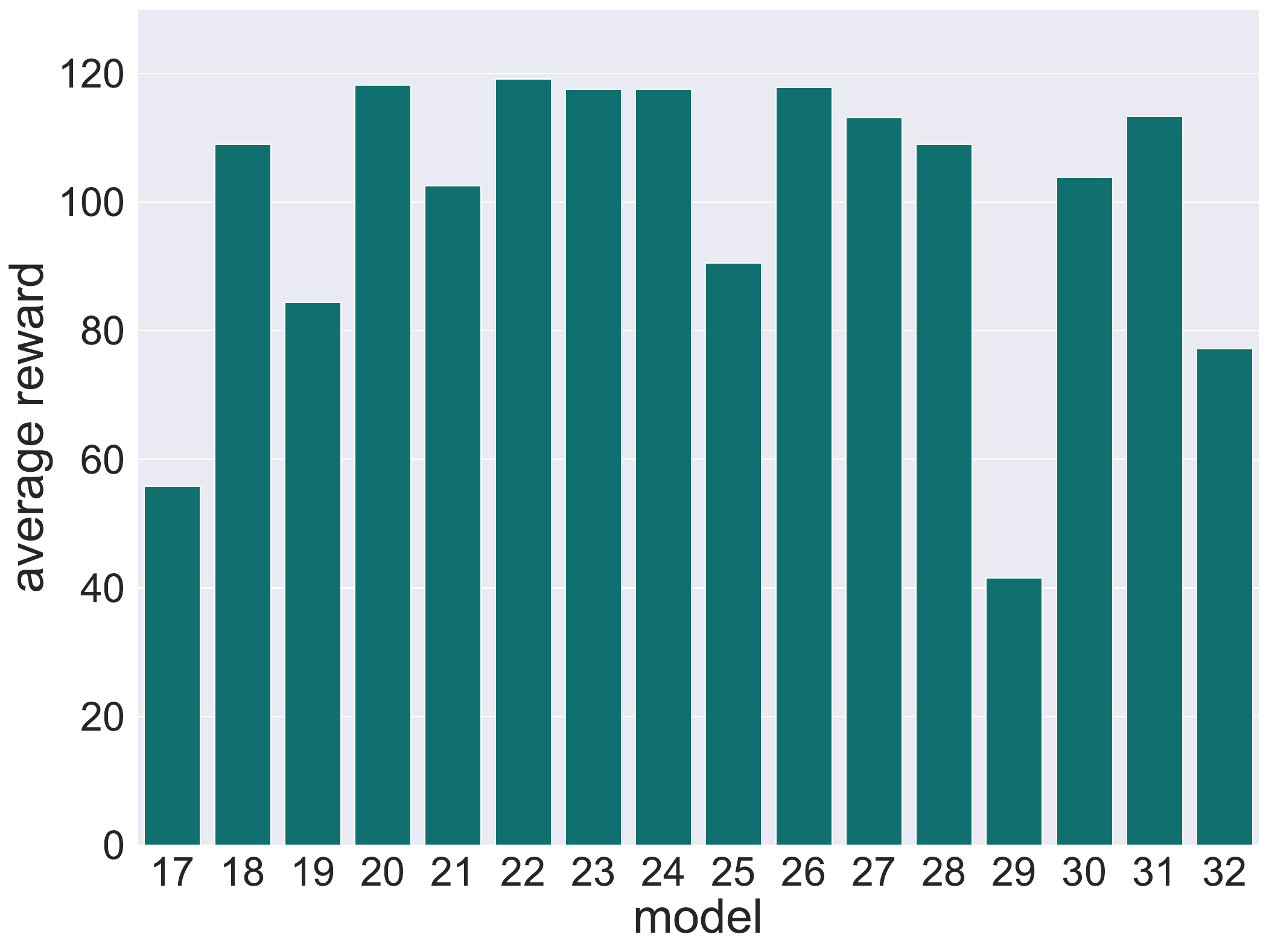}
			\caption{OOD ($\sim\mathcal{TN}(\mu_{high}, \sigma^{2}$))}
			\label{subfig:auroraLongRewards:OODHighMean}
		\end{subfigure}
		
		\caption{Aurora: the models' average rewards under different PDFs.
			\\
			Top row: results for the models used in 
			Experiment~\ref{exp:auroraShort}.
			\\
			Bottom row: results for the models used in 
			Experiment~\ref{exp:auroraLong}.}
		\label{fig:auroraShortDifferentPdsRewards}
	\end{figure}

	Following are the results discussed in subsection~\ref{subsec:aurora}. To 
	further demonstrate our method's robustness to different types of
	out-of-distribution inputs, we applied it not only to different
	\textit{values} (e.g., high \textit{Sending Rate} values) but also to
	various \textit{probability density functions} (PDFs) of values in the 
	(OOD) 
	input domain in question. More specifically, we repeated the OOD experiments
	(Experiment~\ref{exp:auroraShort} and Experiment~\ref{exp:auroraLong}) with 
	different
	PDFs. In their original settings, all of the environment's parameters
	(link's bandwidth, latency, etc.) are uniformly drawn from a range
	$[low, high]$. However, in this experiment, we generated two additional 
	PDFs:
	\emph{Truncated normal} (denoted as
	$\mathcal{TN}_{[low,high]}(\mu, \sigma^{2})$) distributions that are
	truncated within the range $[low, high]$. The first PDF was used with
	$\mu_{low}=0.3*high+(1-0.3)*low$, and the other with 
	$\mu_{high}=0.8*high+(1-0.8)*low$. For both PDFs, the variance 
	($\sigma^{2}$) was arbitrarily set to $\frac{high-low}{4}$. These new
	distributions are depicted in
	Fig.~\ref{fig:auroraShortDifferentPdsRewards} and were used to test
	the models from both batches of Aurora experiments 
	(Experiments~\ref{exp:auroraShort} and~\ref{exp:auroraLong}).
	
	\begin{figure}[ht]
		\centering
		\captionsetup[subfigure]{justification=centering}
		\captionsetup{justification=centering}
		\begin{subfigure}[t]{0.49\linewidth}
			\includegraphics[width=\textwidth]{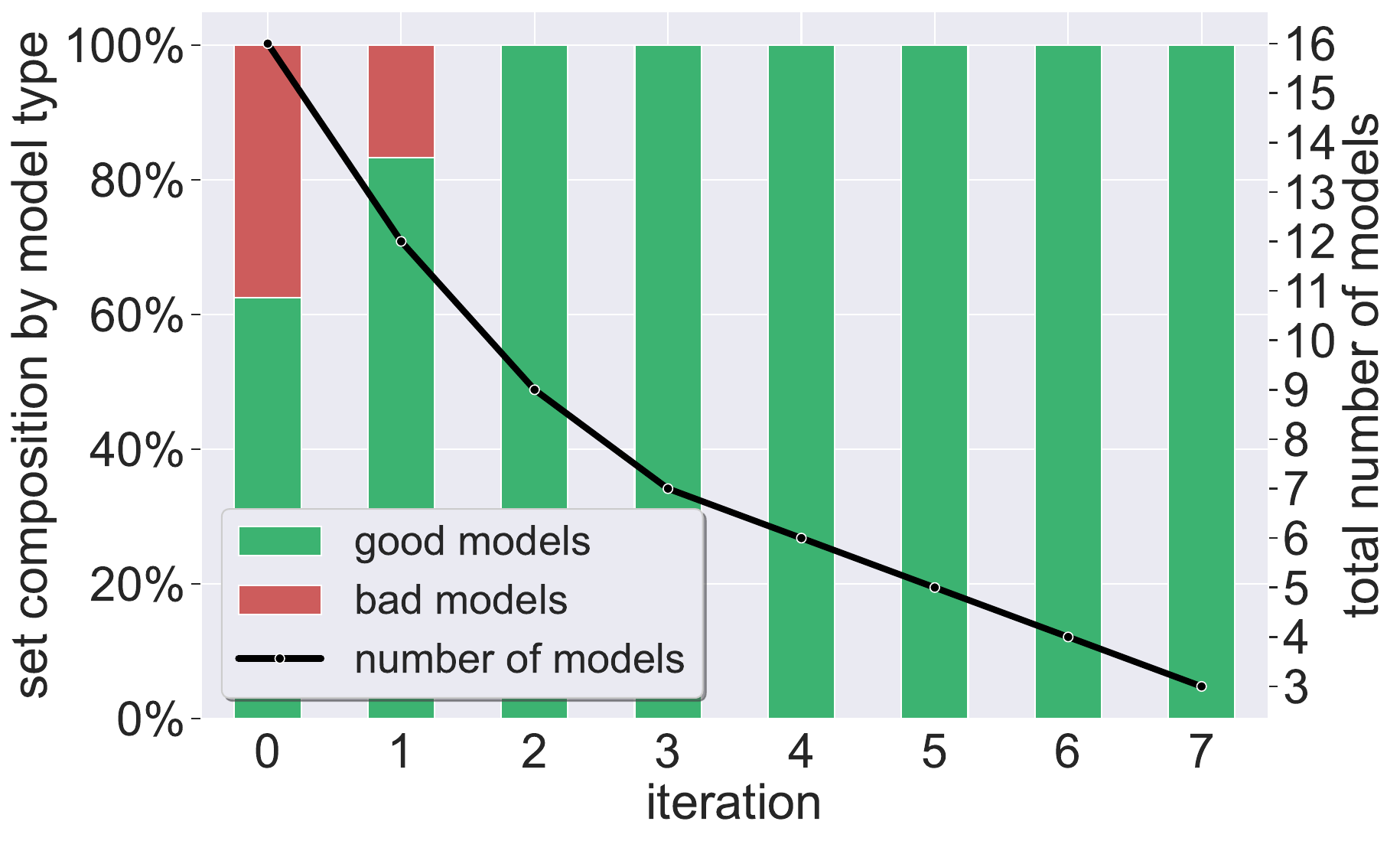}
			\caption{OOD $\sim\mathcal{TN}(\mu_{low}, \sigma^{2}$)}
			\label{subfig:auroraShortOodLowMeanPercentages}
		\end{subfigure}
		\hfill
		\begin{subfigure}[t]{0.49\linewidth}
			\centering
			\includegraphics[width=\textwidth]{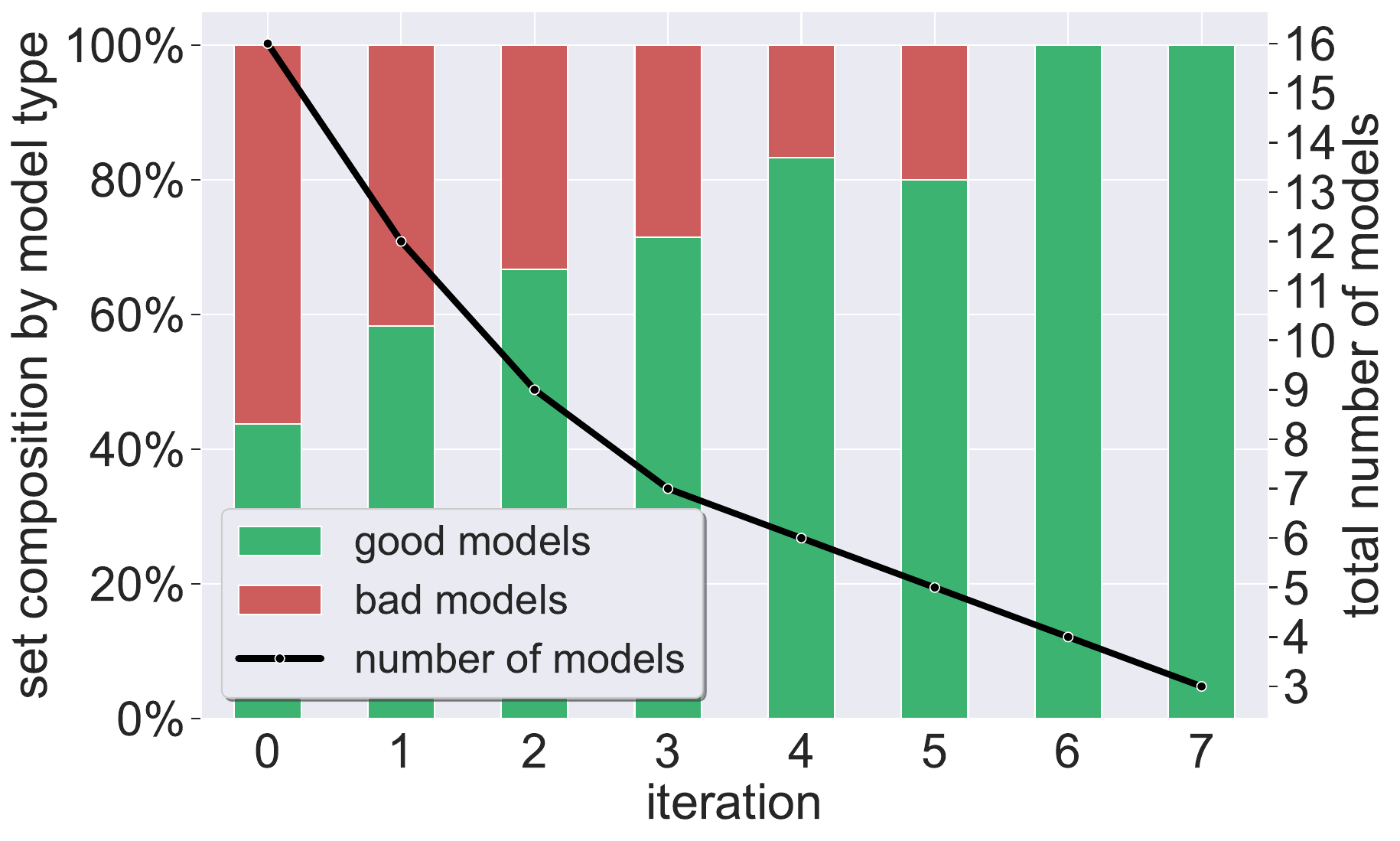}
			\caption{OOD $\sim\mathcal{TN}(\mu_{high}, \sigma^{2}$)}
			\label{subfig:auroraShortOodHighMeanPercentages}
		\end{subfigure}
		\caption{Aurora: Additional PDFs: model selection results for OOD 
		values; 
			the models are the same as in Experiment~\ref{exp:auroraShort}.}
		\label{fig:auroraShortDifferentPdfGoodBadModelsPercentages}
	\end{figure}

	\begin{figure}[ht]
		\centering
		\captionsetup[subfigure]{justification=centering}
		\captionsetup{justification=centering}
		\begin{subfigure}[t]{0.49\linewidth}
			\includegraphics[width=\textwidth]{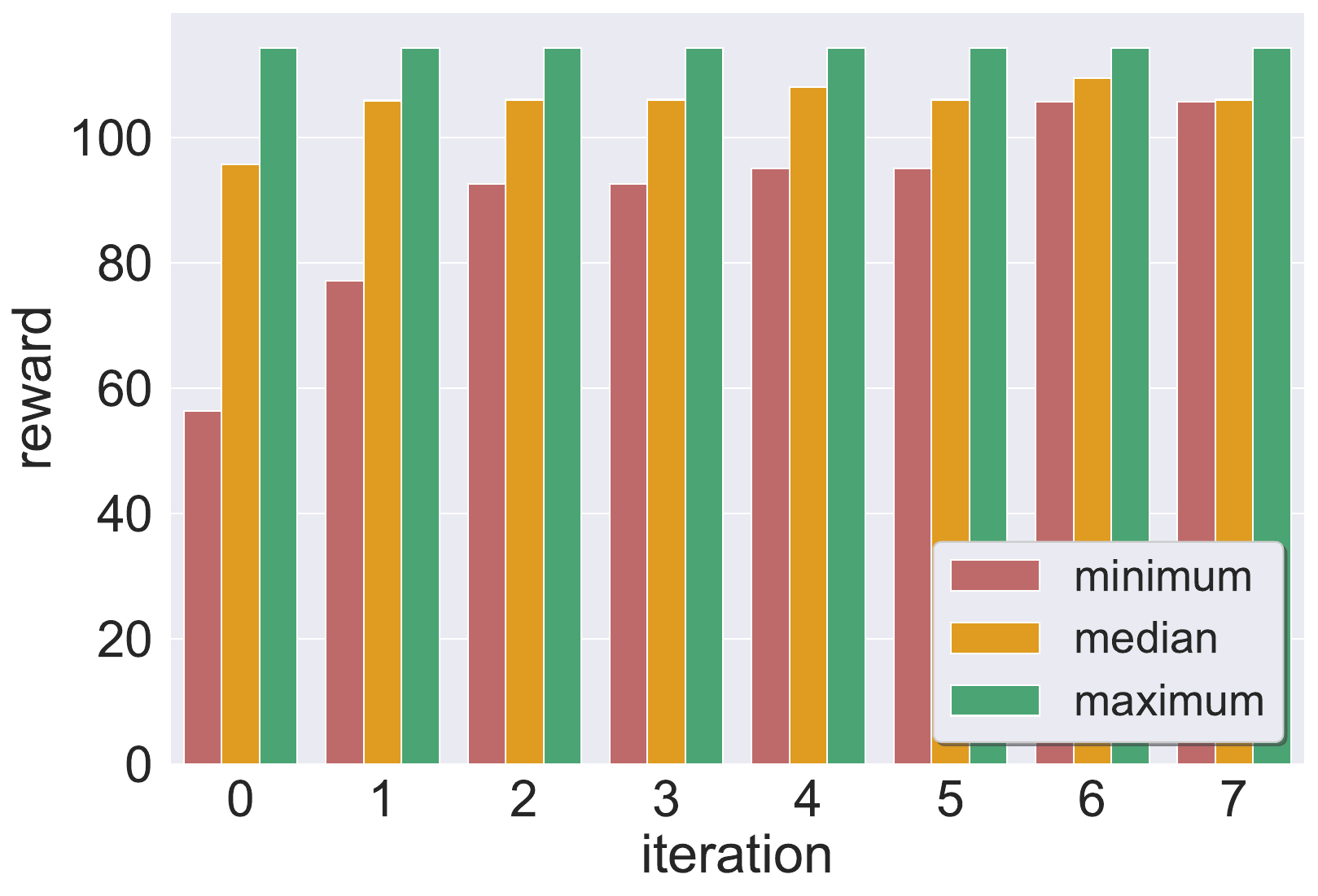}
			\label{subfig:auroraShortOodLowMeanRewardsStats}
			\caption{OOD $\sim\mathcal{TN}(\mu_{low
				}, \sigma^{2}$)}
		\end{subfigure}
		\hfill
		\begin{subfigure}[t]{0.49\linewidth}
			\centering
			\includegraphics[width=\textwidth]{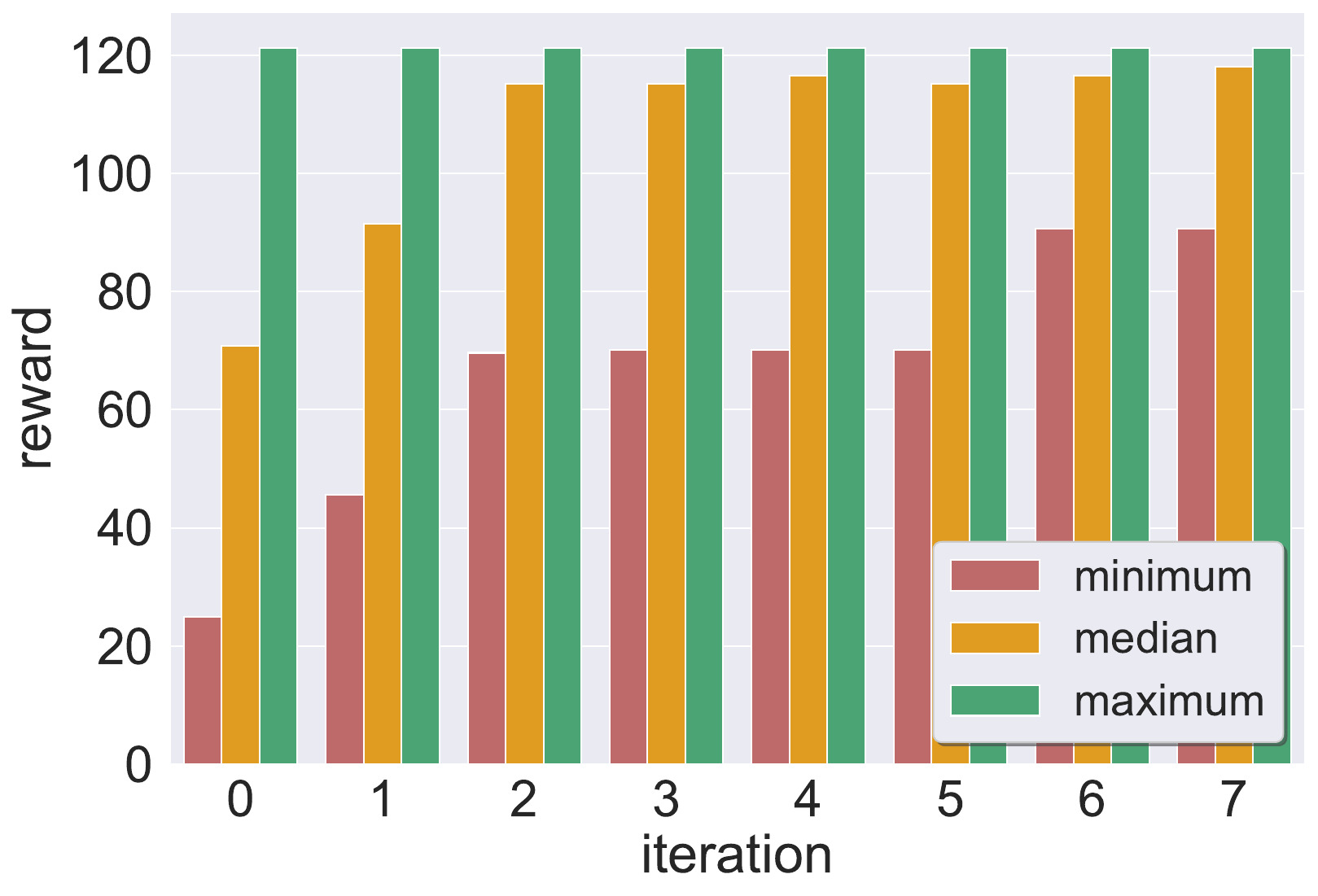}
			\label{subfig:auroraShortOodHighMeanRewardsStat}
			\caption{OOD $\sim\mathcal{TN}(\mu_{high}, \sigma^{2}$)}
		\end{subfigure}
		
		\caption{Aurora: Additional PDFs: model selection results: rewards 
			statistics per iteration;  the models are the same as in 
			Experiment~\ref{exp:auroraShort}.}
		\label{fig:auroraShortDifferentPdfGoodRewardsStats}
	\end{figure}
	
	\begin{figure}[ht]
		\centering
		\captionsetup[subfigure]{justification=centering}
		\captionsetup{justification=centering} 
		\begin{subfigure}[t]{0.49\linewidth}
			\includegraphics[width=\textwidth]{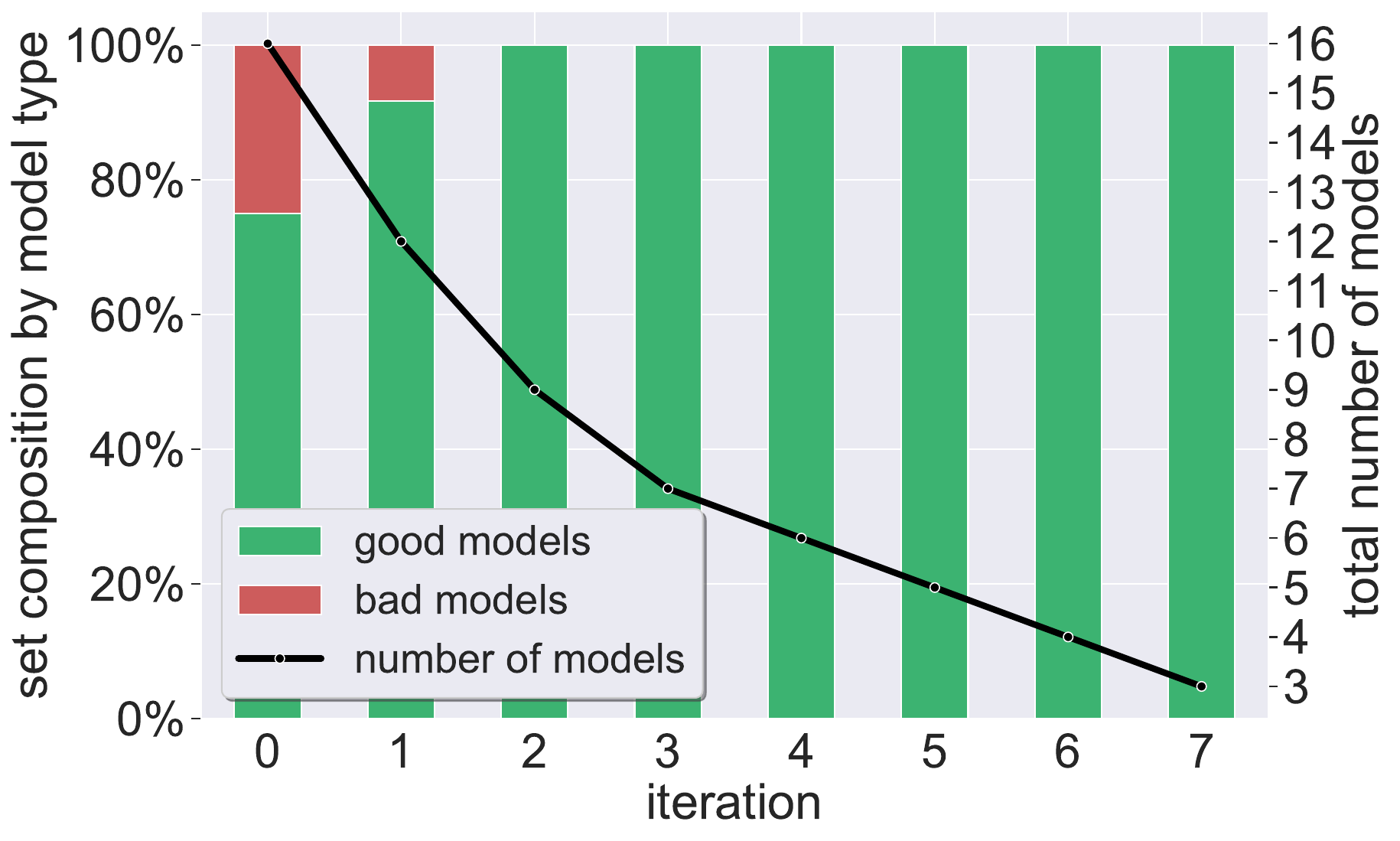}
			\caption{$\sim\mathcal{TN}(\mu_{low}, \sigma^{2}$)} 
			\label{subfig:auroraLongOodHighMeanPercentages}
		\end{subfigure}
		\hfill
		\begin{subfigure}[t]{0.49\linewidth}
			\centering
			\includegraphics[width=\textwidth]{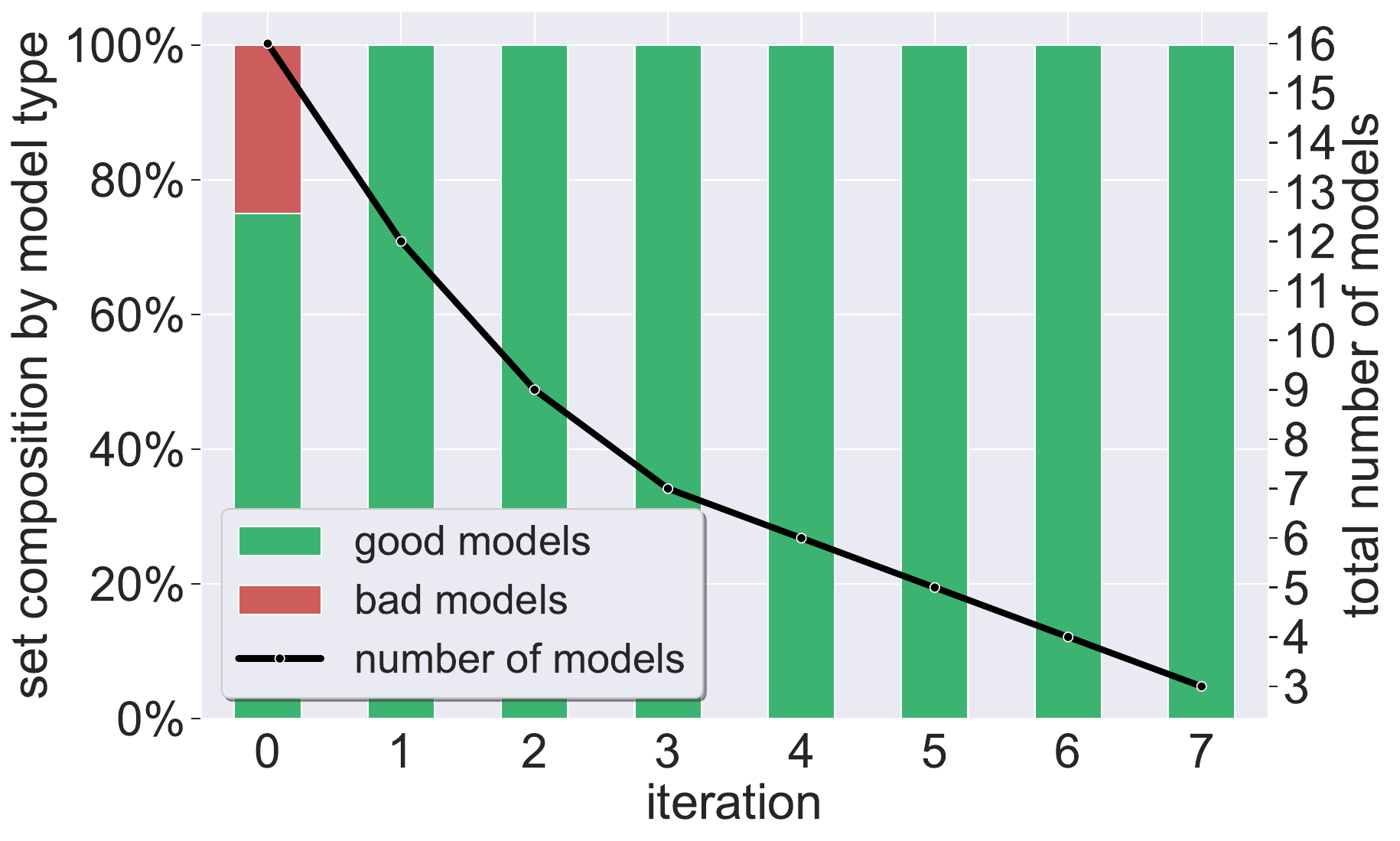}
			\caption{$\sim\mathcal{TN}(\mu_{high}, \sigma^{2}$)}
			\label{subfig:auroraLongOodHighMeanPercentages}
		\end{subfigure}
		\caption{Aurora: Additional PDFs: model selection results for OOD 
		values; 
			the models are the same as in Experiment~\ref{exp:auroraLong}.}
		\label{fig:auroraLongDifferentPdfGoodBadModelsPercentages}
	\end{figure}
	
	\begin{figure}[ht]
		\centering
		\captionsetup[subfigure]{justification=centering}
		\captionsetup{justification=centering} 
		\begin{subfigure}[t]{0.49\linewidth}
			\includegraphics[width=\textwidth]{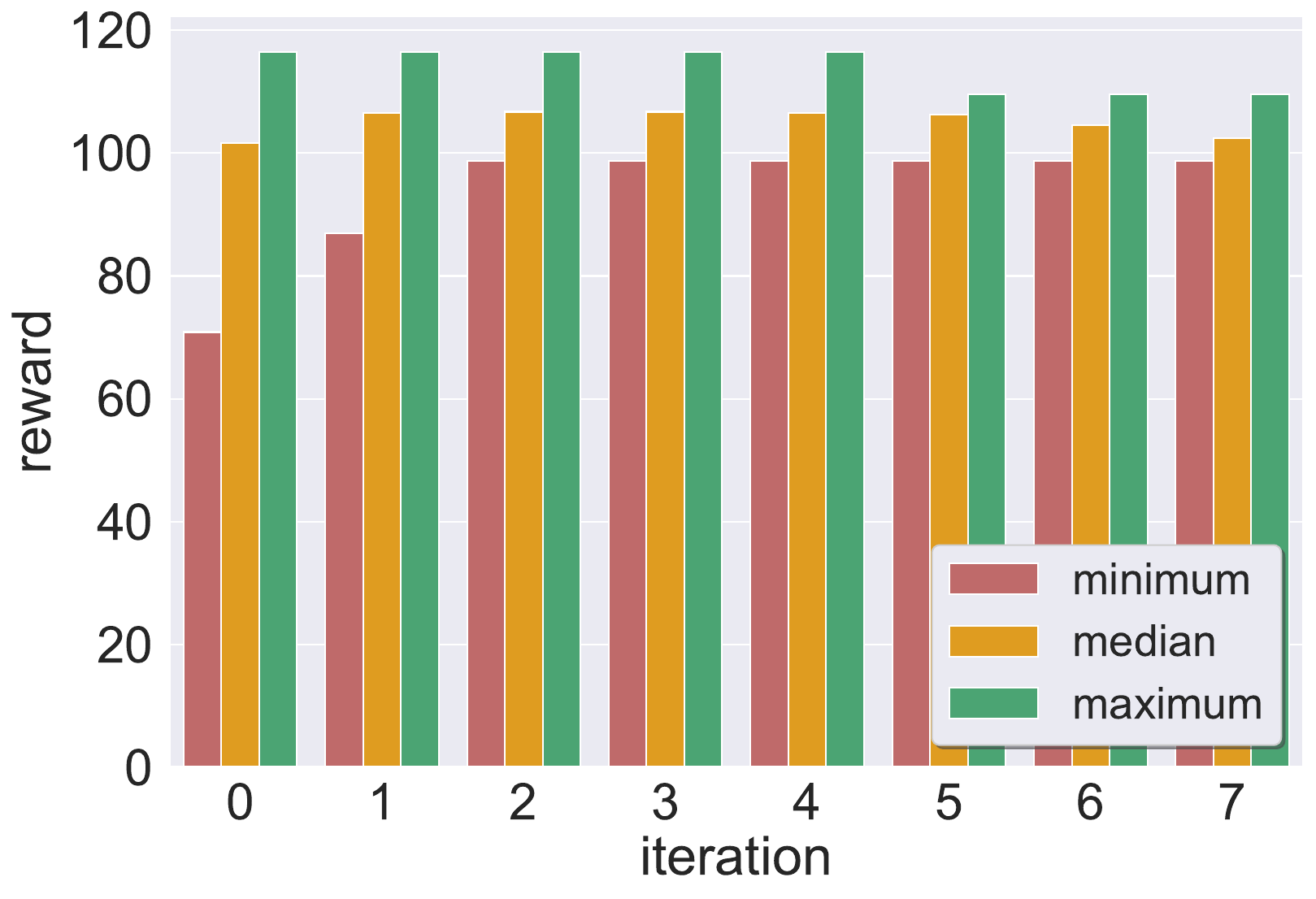}
			\caption{$\sim\mathcal{TN}(\mu_{low}, \sigma^{2}$)}
			\label{subfig:auroraLongOodLowMeanRewardsStats}
		\end{subfigure}
		\hfill
		\begin{subfigure}[t]{0.49\linewidth}
			\centering
			\includegraphics[width=\textwidth]{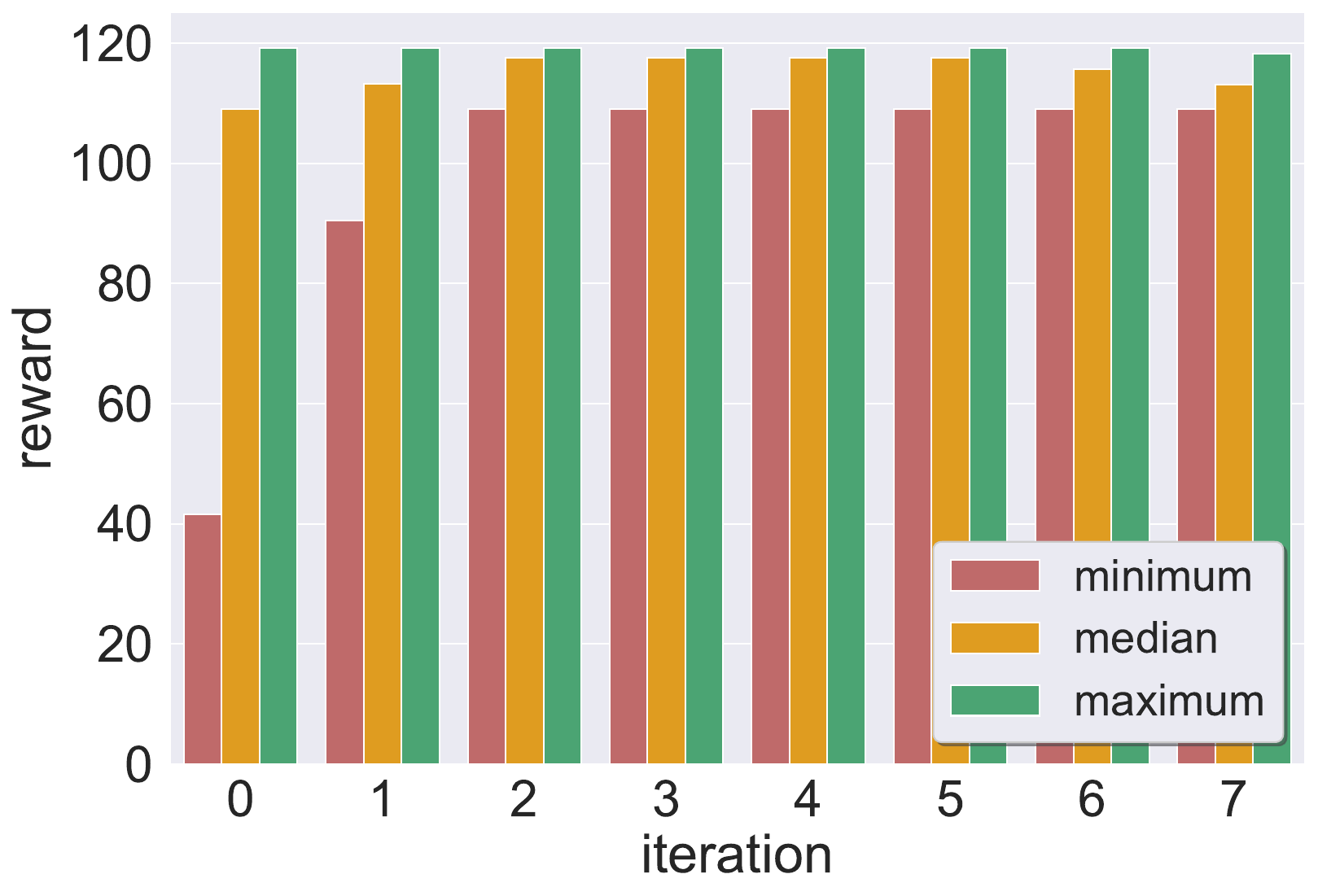}
			\caption{$\sim\mathcal{TN}(\mu_{high}, \sigma^{2}$)}
			\label{subfig:auroraLongOodLowMeanRewardsStats}
		\end{subfigure}
		\caption{Aurora: Additional PDFs: model selection results: rewards 
			statistics per iteration;  the models are the same as in 
			Experiment~\ref{exp:auroraLong}.}    
		\label{fig:auroraLongDifferentPdfGoodRewardsStats}
	\end{figure}
	
	\clearpage
	
	\subsection{Additional Filtering Criteria: Experiment~\ref{exp:auroraShort}}
	
	\begin{figure}[h]
		\centering
		\captionsetup[subfigure]{justification=centering}
		\captionsetup{justification=centering} 
		\begin{subfigure}[t]{0.49\linewidth}
			\centering
			\includegraphics[width=\textwidth, 
			height=0.67\textwidth]{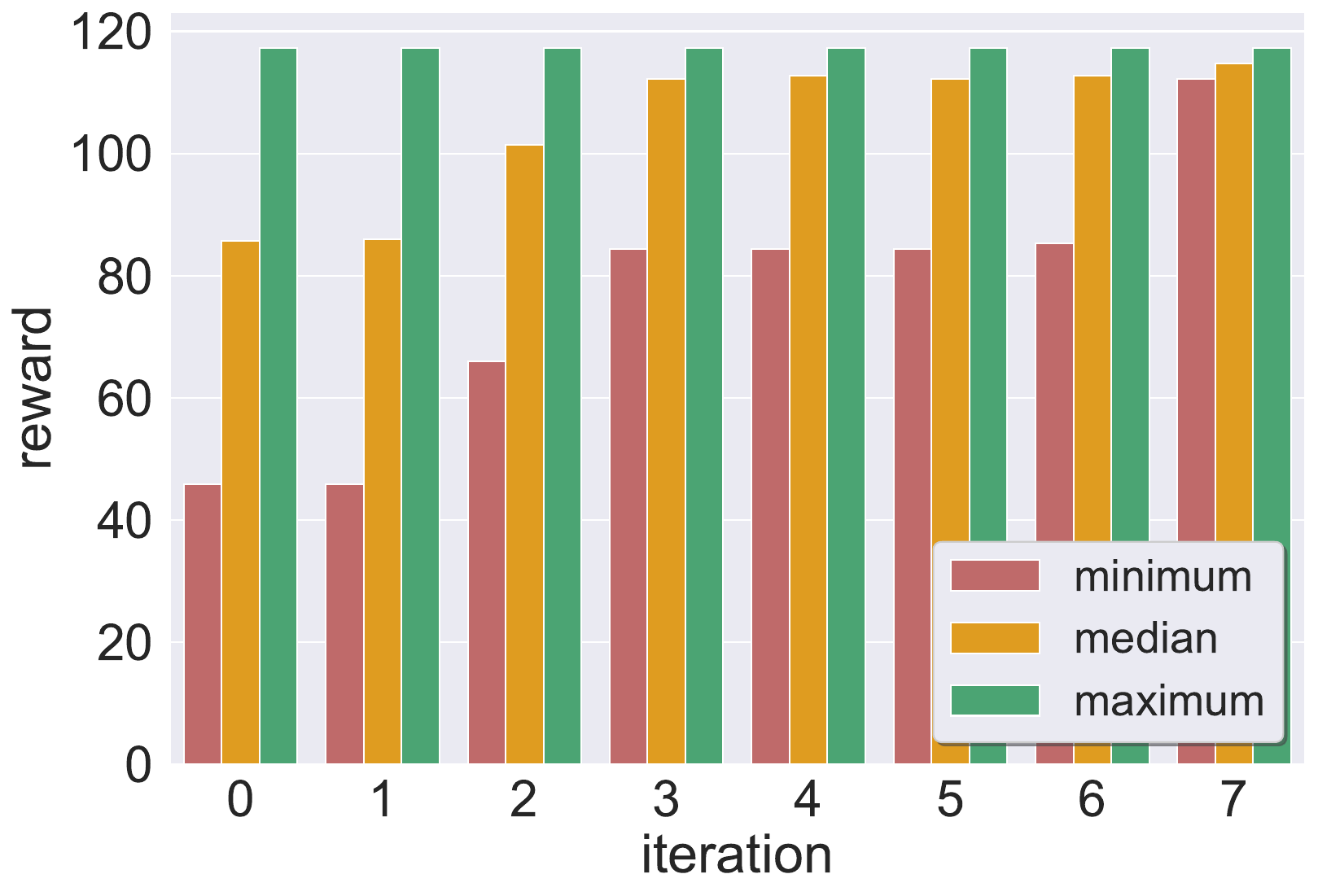}
			\caption{Reward statistics of remaining models}
			\label{}
		\end{subfigure}
		\hfill
		\begin{subfigure}[t]{0.49\linewidth}
			\includegraphics[width=\textwidth, 
			height=0.67\textwidth]{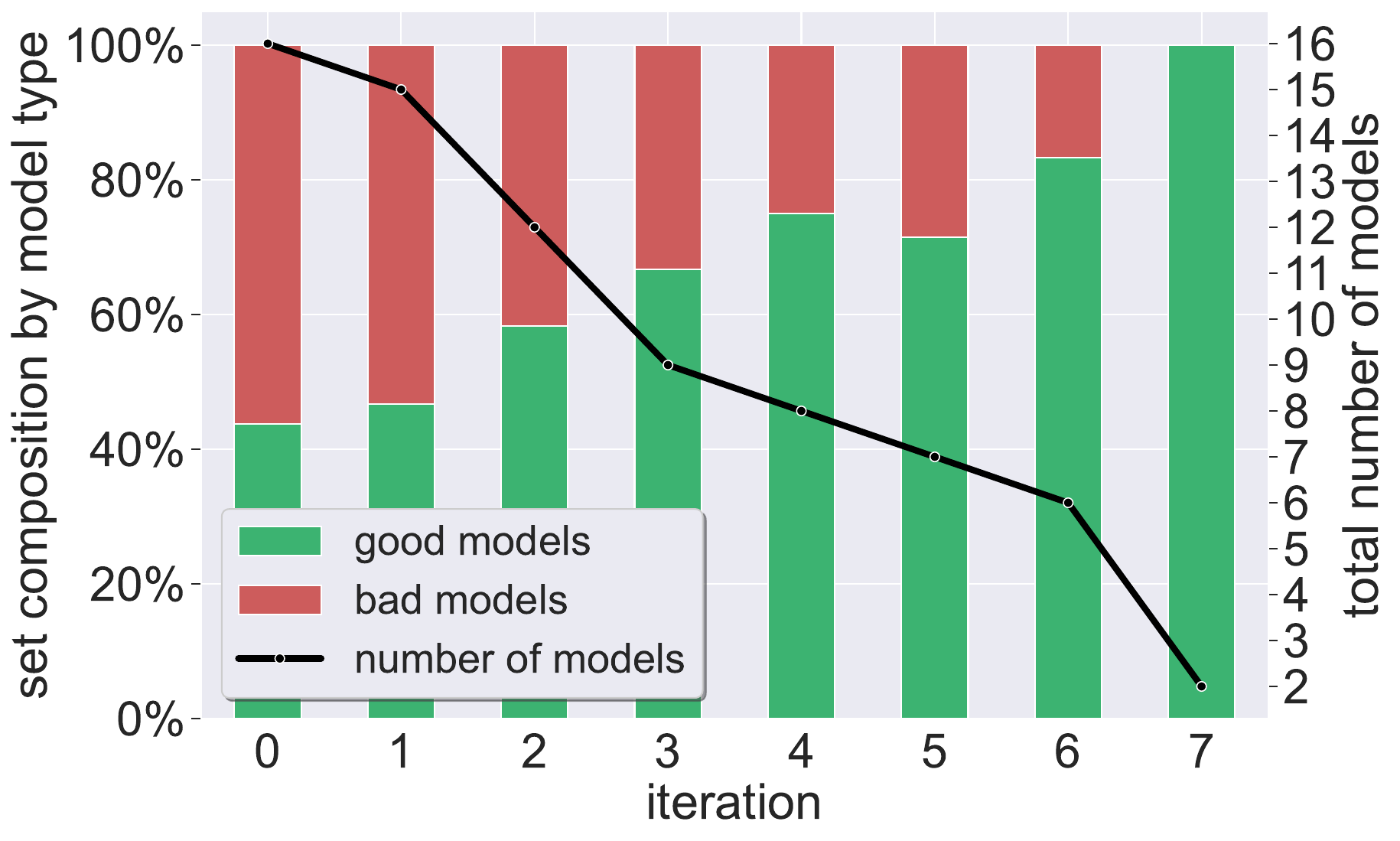}
			\caption{Ratio between good/bad models}
			\label{}
		\end{subfigure}
		\caption{Aurora Experiment~\ref{exp:auroraShort}: results using the 
		\maxAgg 
			filtering criterion.
			Our technique  selected models \{7,16\}.}
		\label{fig:auroraShortMaxFiltering}
	\end{figure}
	
	\begin{figure}[h]
		\centering
		\captionsetup[subfigure]{justification=centering}
		\captionsetup{justification=centering} 
		\begin{subfigure}[t]{0.49\linewidth}
			\centering
			\includegraphics[width=\textwidth, 
			height=0.67\textwidth]{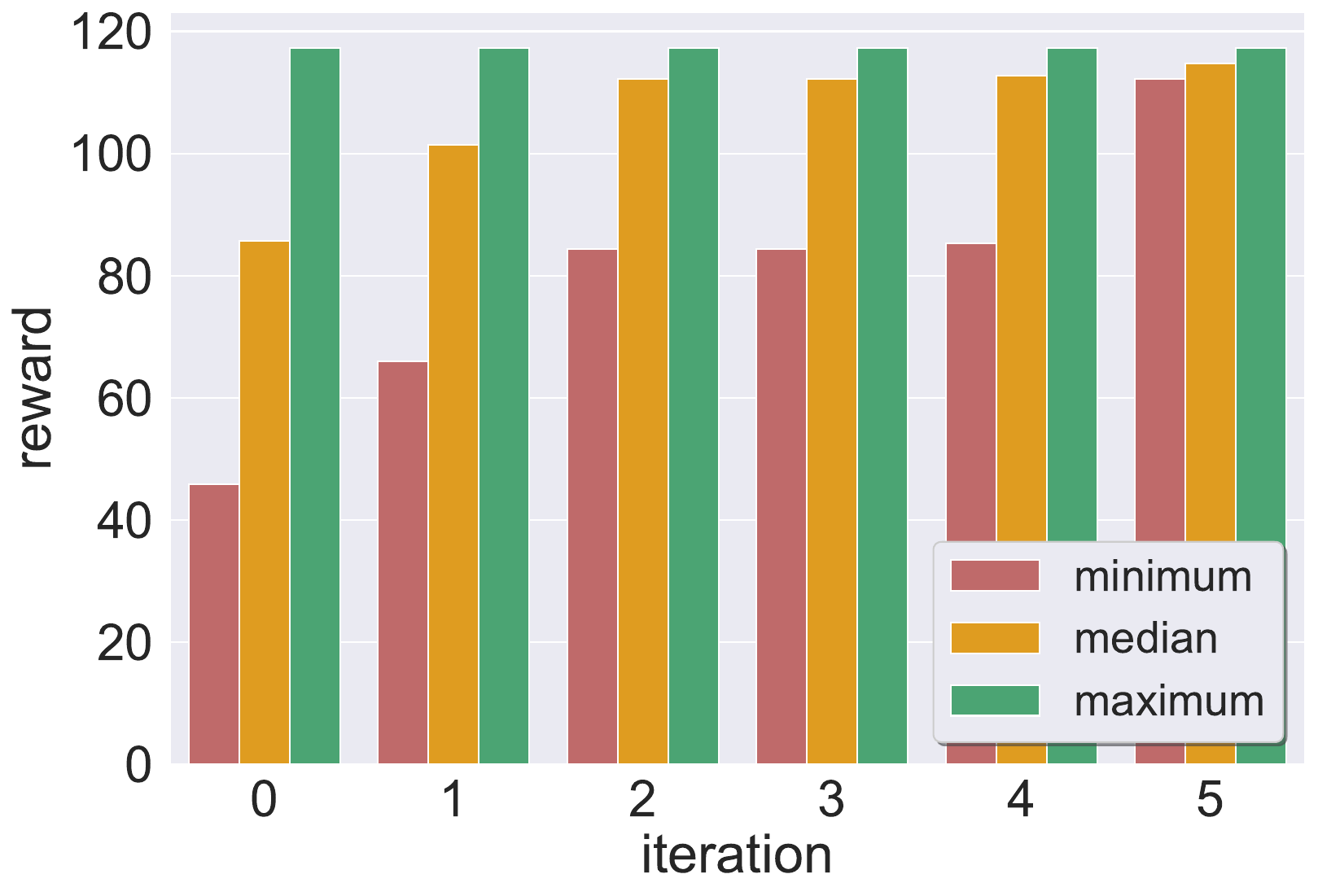}
			\caption{Reward statistics of remaining models}
			\label{}
		\end{subfigure}
		\hfill
		\begin{subfigure}[t]{0.49\linewidth}
			\includegraphics[width=\textwidth, 
			height=0.67\textwidth]{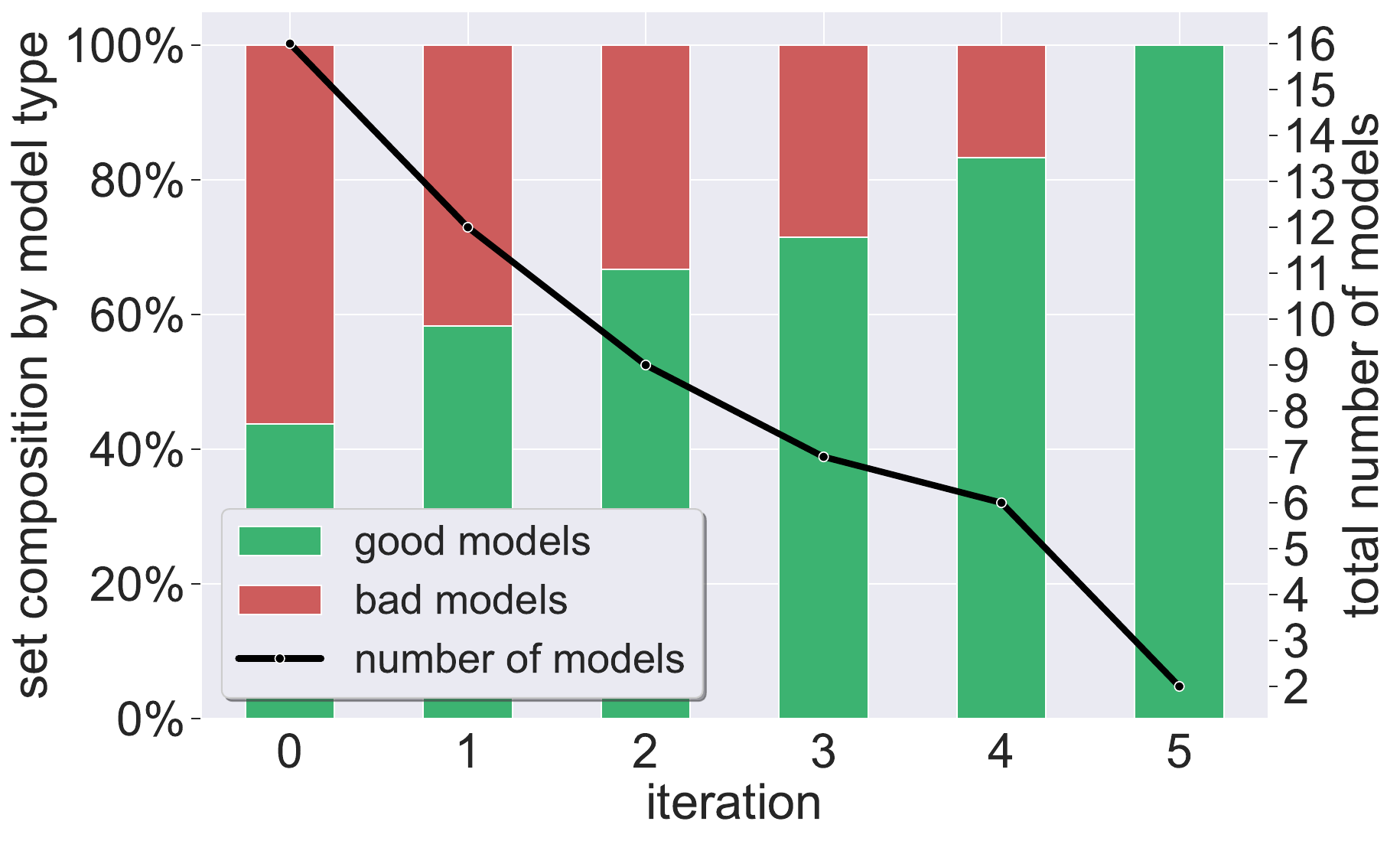}
			\caption{Ratio between good/bad models}
			\label{}
		\end{subfigure}
		\caption{Aurora Experiment~\ref{exp:auroraShort}: results using the 
			\conditionCombined filtering criterion.
			Our technique  selected models \{7,16\}.}
	\end{figure}
	\FloatBarrier
	
	\clearpage
	
	\subsection{Additional Filtering Criteria: Experiment~\ref{exp:auroraLong}}
	\begin{figure}[h]
		\centering
		\captionsetup[subfigure]{justification=centering}
		\captionsetup{justification=centering} 
		\begin{subfigure}[t]{0.49\linewidth}
			\centering
			\includegraphics[width=\textwidth, 
			height=0.67\textwidth]{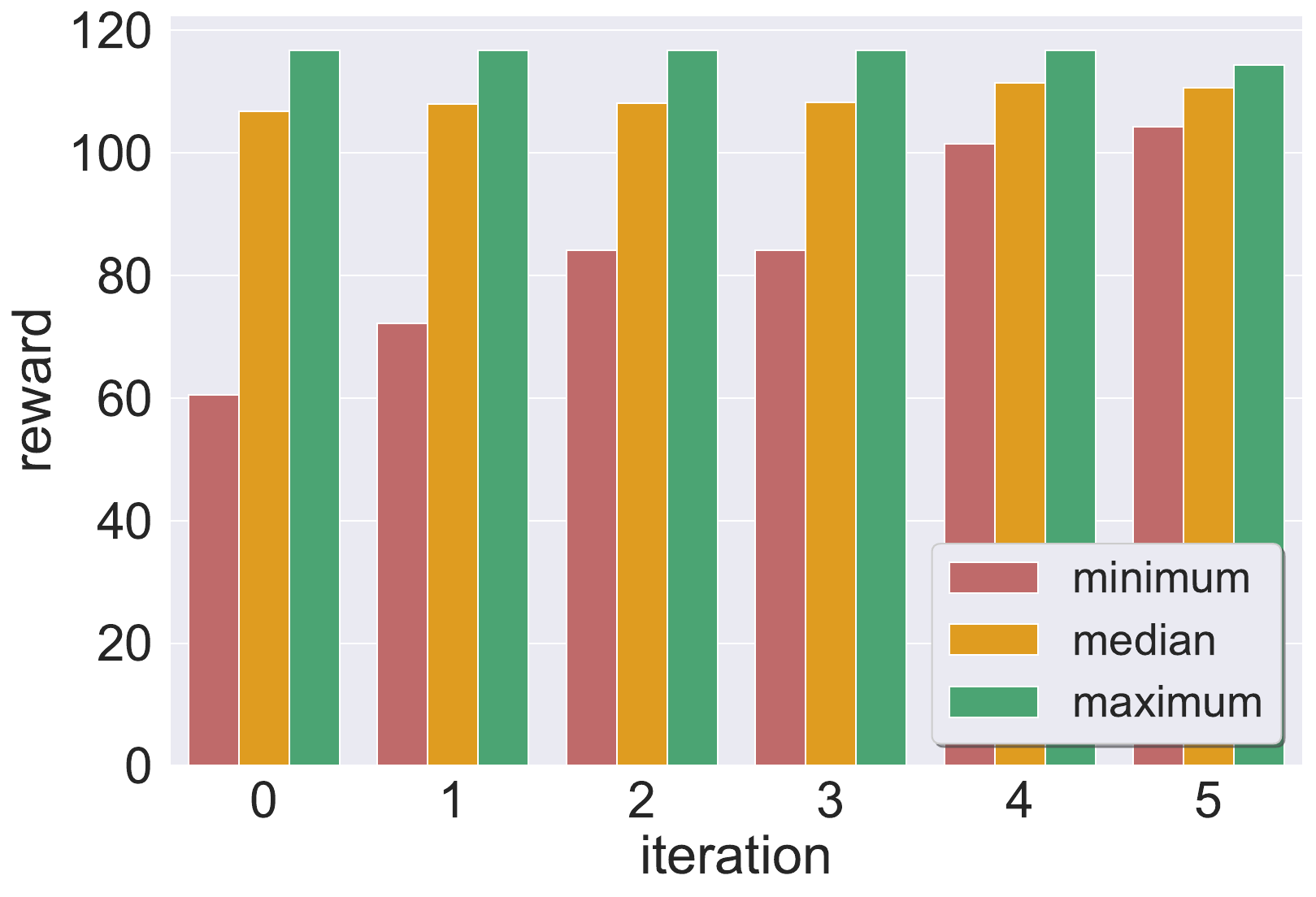}
			\caption{Reward statistics of remaining models}
			\label{}
		\end{subfigure}
		\hfill
		\begin{subfigure}[t]{0.49\linewidth}
			\includegraphics[width=\textwidth, 
			height=0.67\textwidth]{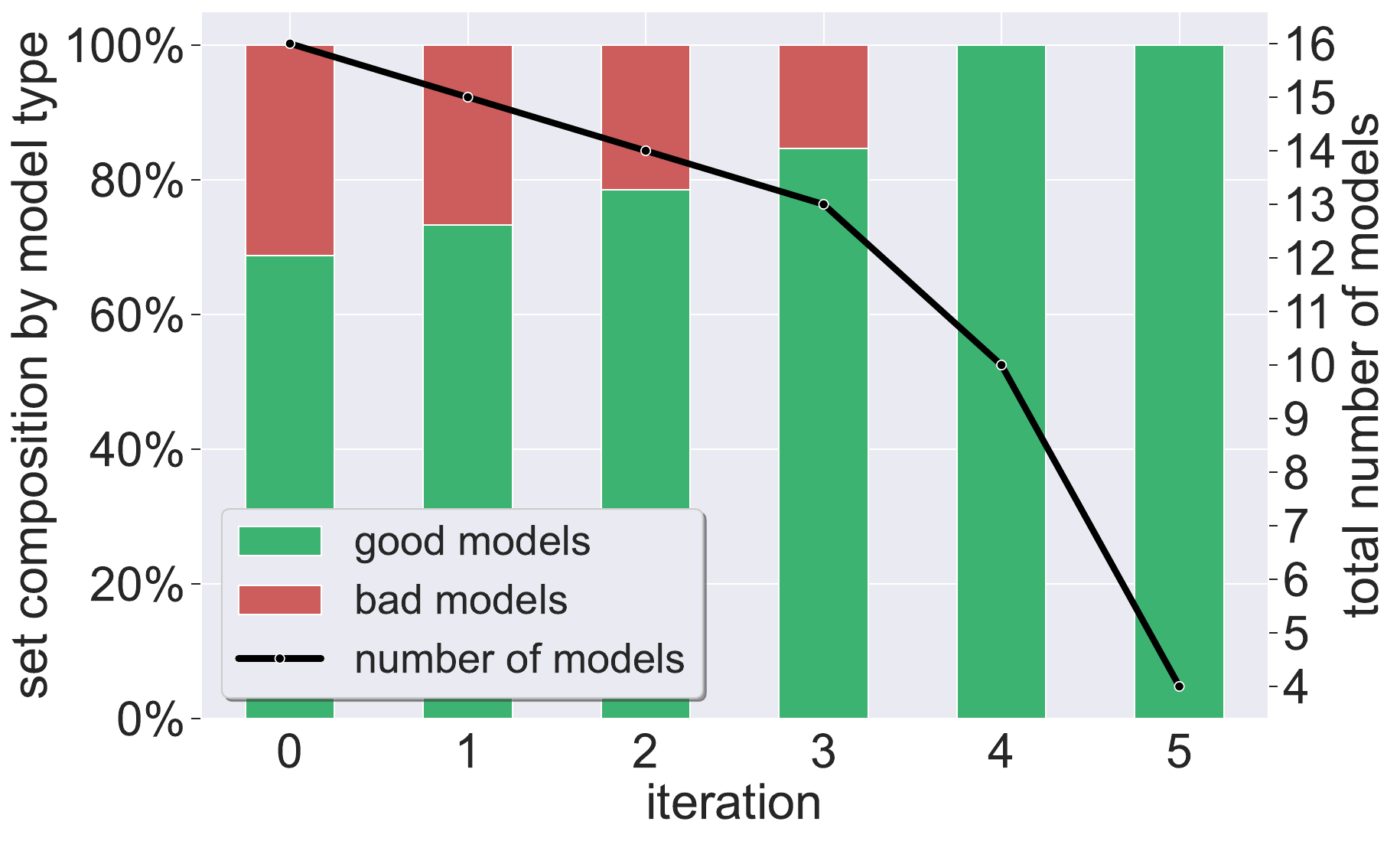}
			\caption{Ratio between good/bad models}
			\label{}
		\end{subfigure}
		\caption{Aurora Experiment~\ref{exp:auroraLong}: results using the 
		\maxAgg 
			filtering criterion.
			Our technique  selected models \{20, 22, 27, 28\}.}
	\end{figure}
	
	\begin{figure}[h]
		\centering
		\captionsetup[subfigure]{justification=centering}
		\captionsetup{justification=centering} 
		\begin{subfigure}[t]{0.49\linewidth}
			\centering
			\includegraphics[width=\textwidth, 
			height=0.67\textwidth]{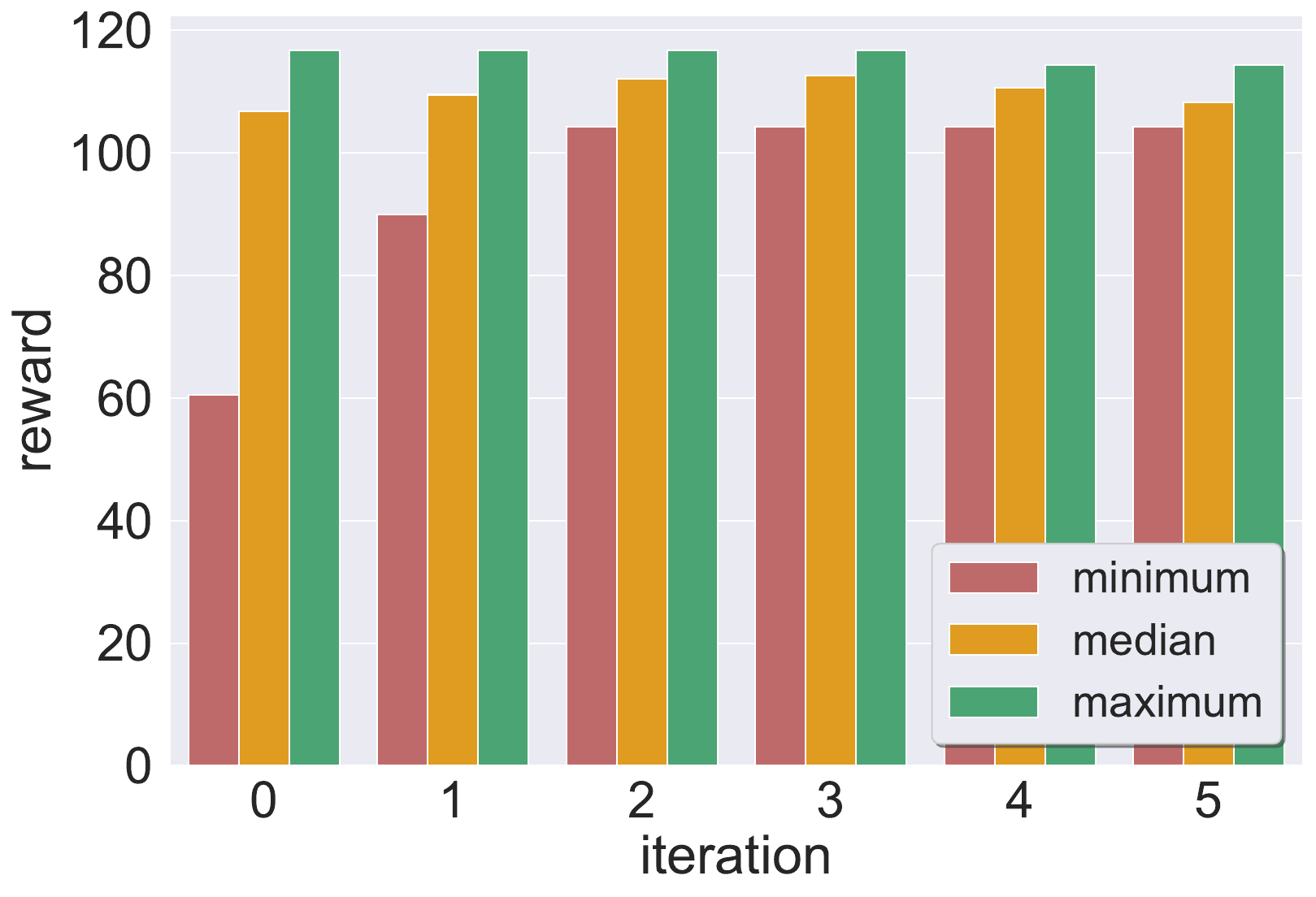}
			\caption{Reward statistics of remaining models}
			\label{}
		\end{subfigure}
		\hfill
		\begin{subfigure}[t]{0.49\linewidth}
			\includegraphics[width=\textwidth, 
			height=0.67\textwidth]{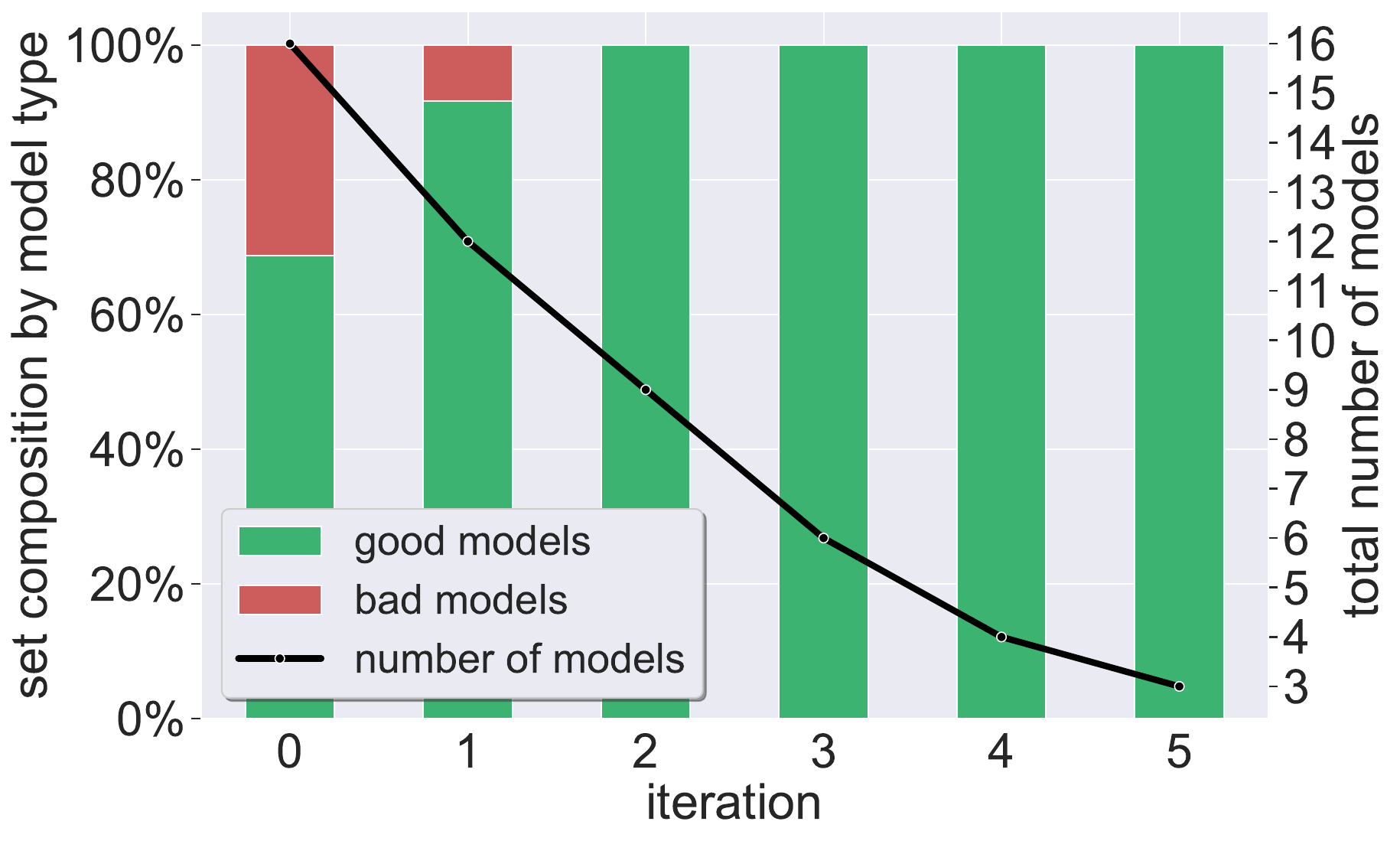}
			\caption{Ratio between good/bad models}
			\label{}
		\end{subfigure}
		\caption{Aurora Experiment~\ref{exp:auroraLong}: results using the 
			\conditionCombined filtering criterion.
			Our technique  selected models \{20, 27, 28\}.}
	\end{figure}
	\FloatBarrier
	
	\clearpage
	
	\subsection{Additional Filtering Criteria: Additional PDFs}
	
	\begin{figure}[h]
		\centering
		\captionsetup[subfigure]{justification=centering}
		\captionsetup{justification=centering} 
		\begin{subfigure}[t]{0.49\linewidth}
			\centering
			\includegraphics[width=\textwidth, 
			height=0.67\textwidth]{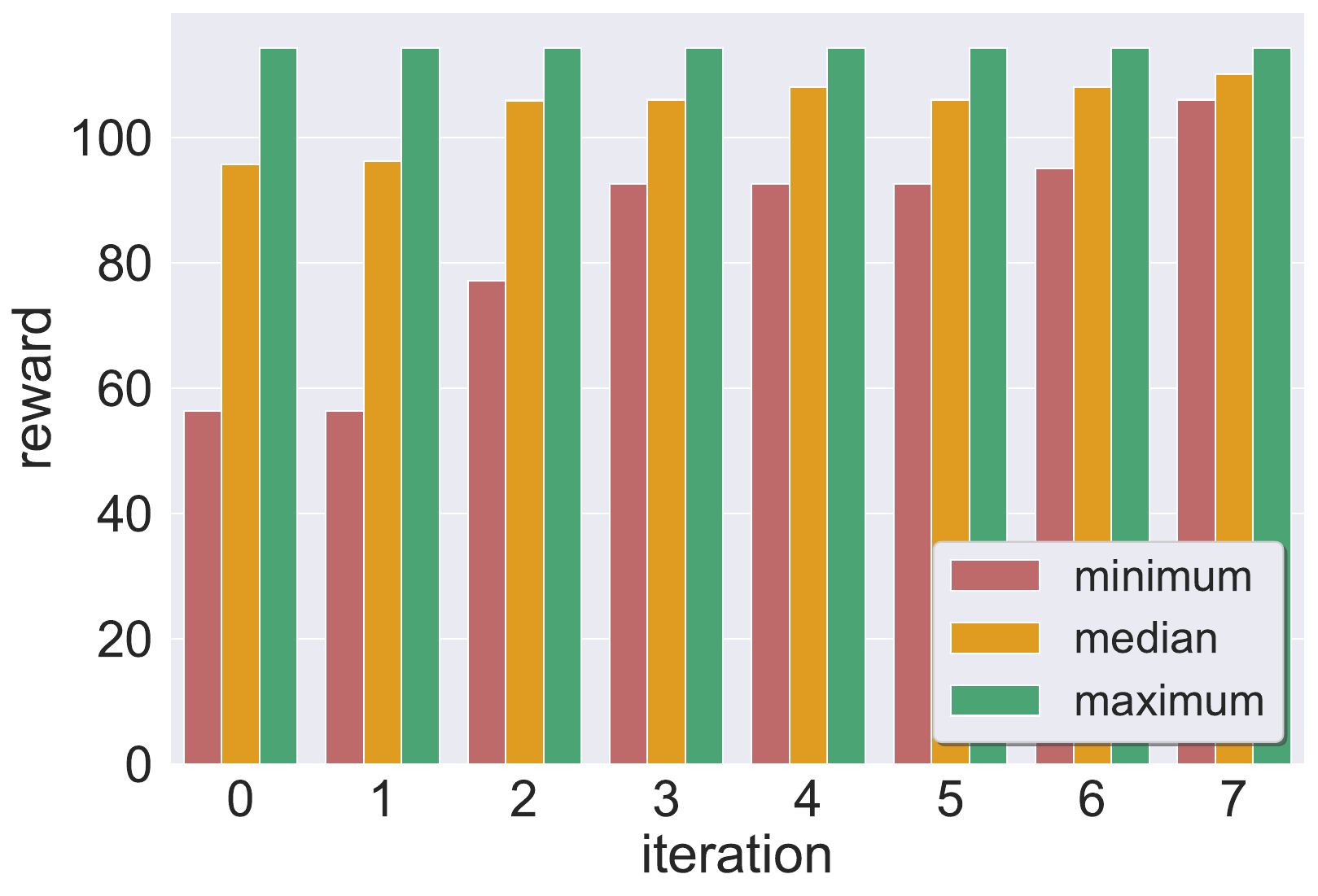}
			\caption{Reward statistics of remaining models}
			\label{}
		\end{subfigure}
		\hfill
		\begin{subfigure}[t]{0.49\linewidth}
			\includegraphics[width=\textwidth, 
			height=0.67\textwidth]{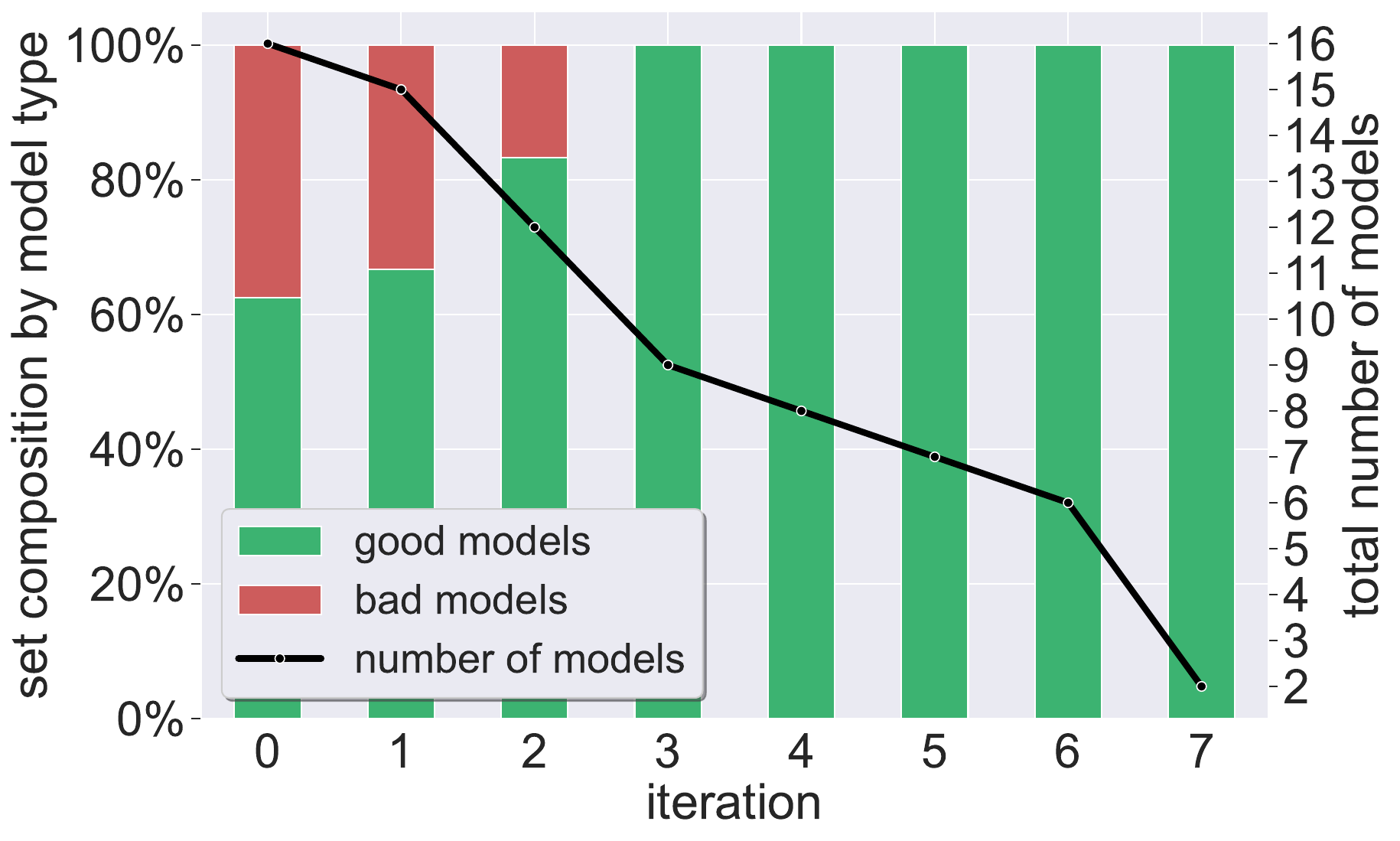}
			\caption{Ratio between good/bad models}
			\label{}
		\end{subfigure}
		\caption{Aurora Experiment~\ref{exp:auroraShort}: PDF 
			$\sim\mathcal{TN}(\mu_{low}, \sigma^{2}$): results using the 
			\maxAgg 
			filtering criterion.}
	\end{figure}
	
	\begin{figure}[h]
		\centering
		\captionsetup[subfigure]{justification=centering}
		\captionsetup{justification=centering} 
		\begin{subfigure}[t]{0.49\linewidth}
			\centering
			\includegraphics[width=\textwidth, 
			height=0.67\textwidth]{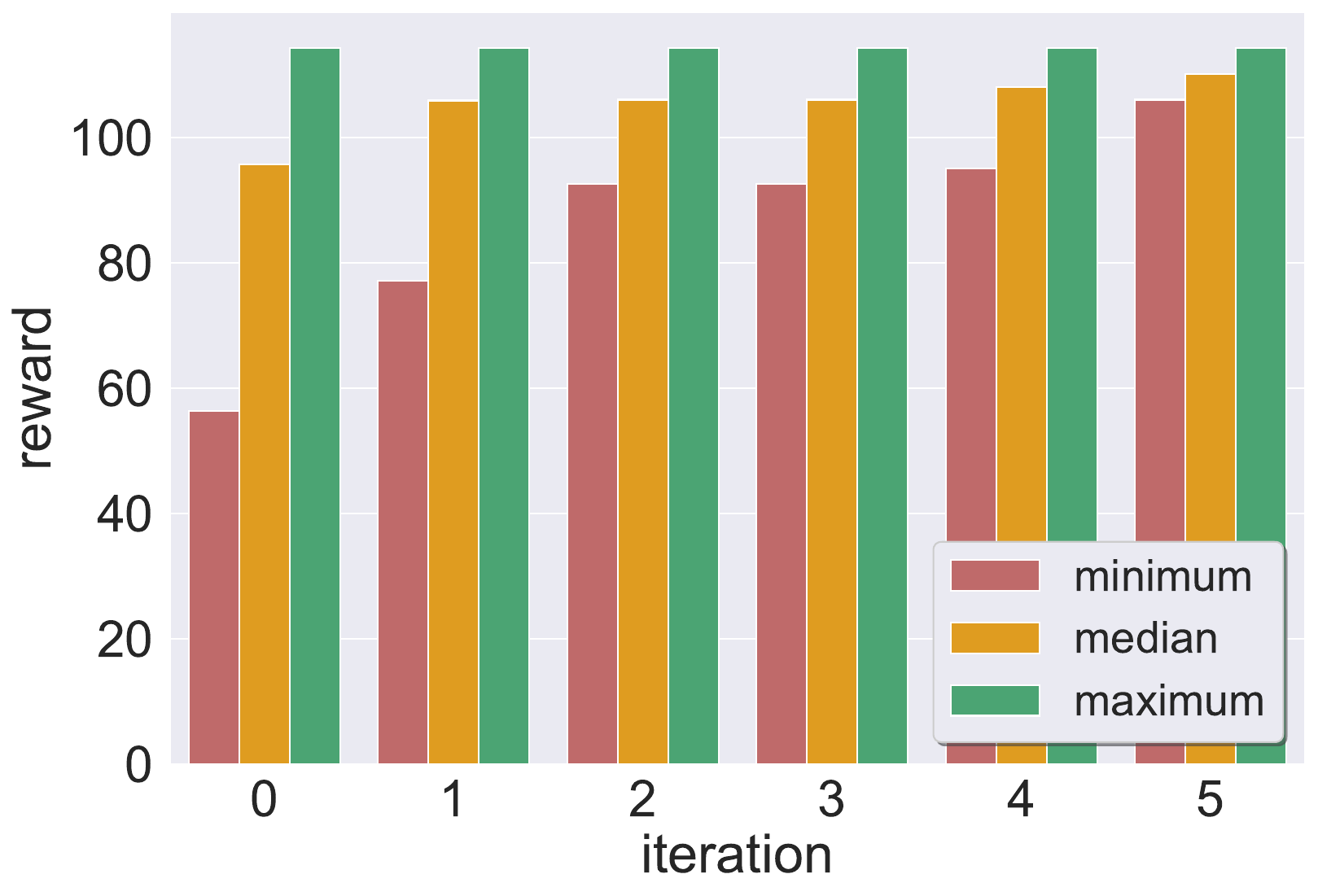}
			\caption{Reward statistics of remaining models}
			\label{}
		\end{subfigure}
		\hfill
		\begin{subfigure}[t]{0.49\linewidth}
			\includegraphics[width=\textwidth, 
			height=0.67\textwidth]{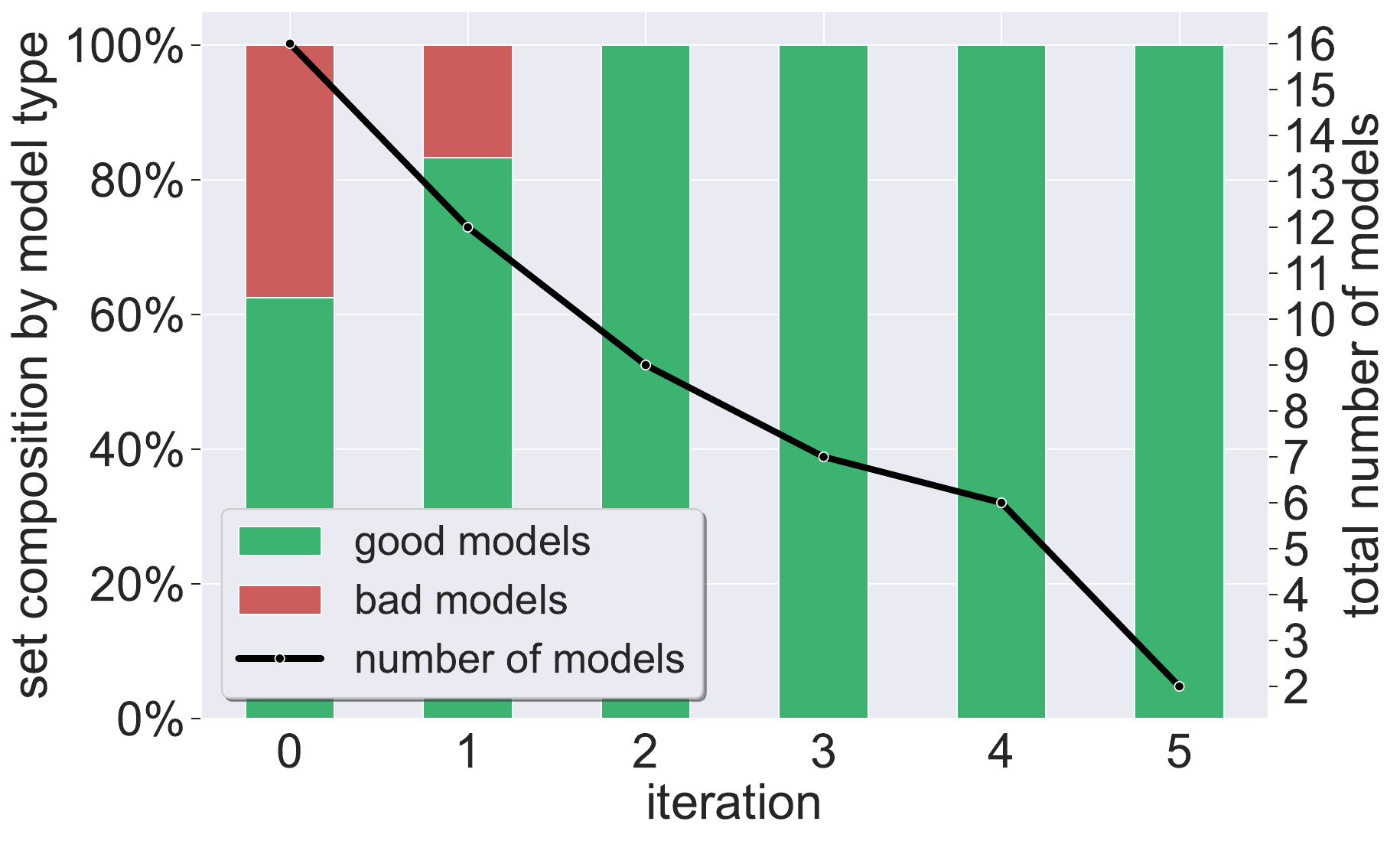}
			\caption{Ratio between good/bad models}
			\label{}
		\end{subfigure}
		\caption{Aurora Experiment~\ref{exp:auroraShort}: PDF 
			$\sim\mathcal{TN}(\mu_{low}, \sigma^{2}$): results using the 
			\conditionCombined filtering criterion.}
	\end{figure}
	
	\begin{figure}[h]
		\centering
		\captionsetup[subfigure]{justification=centering}
		\captionsetup{justification=centering} 
		\begin{subfigure}[t]{0.49\linewidth}
			\centering
			\includegraphics[width=\textwidth, 
			height=0.67\textwidth]{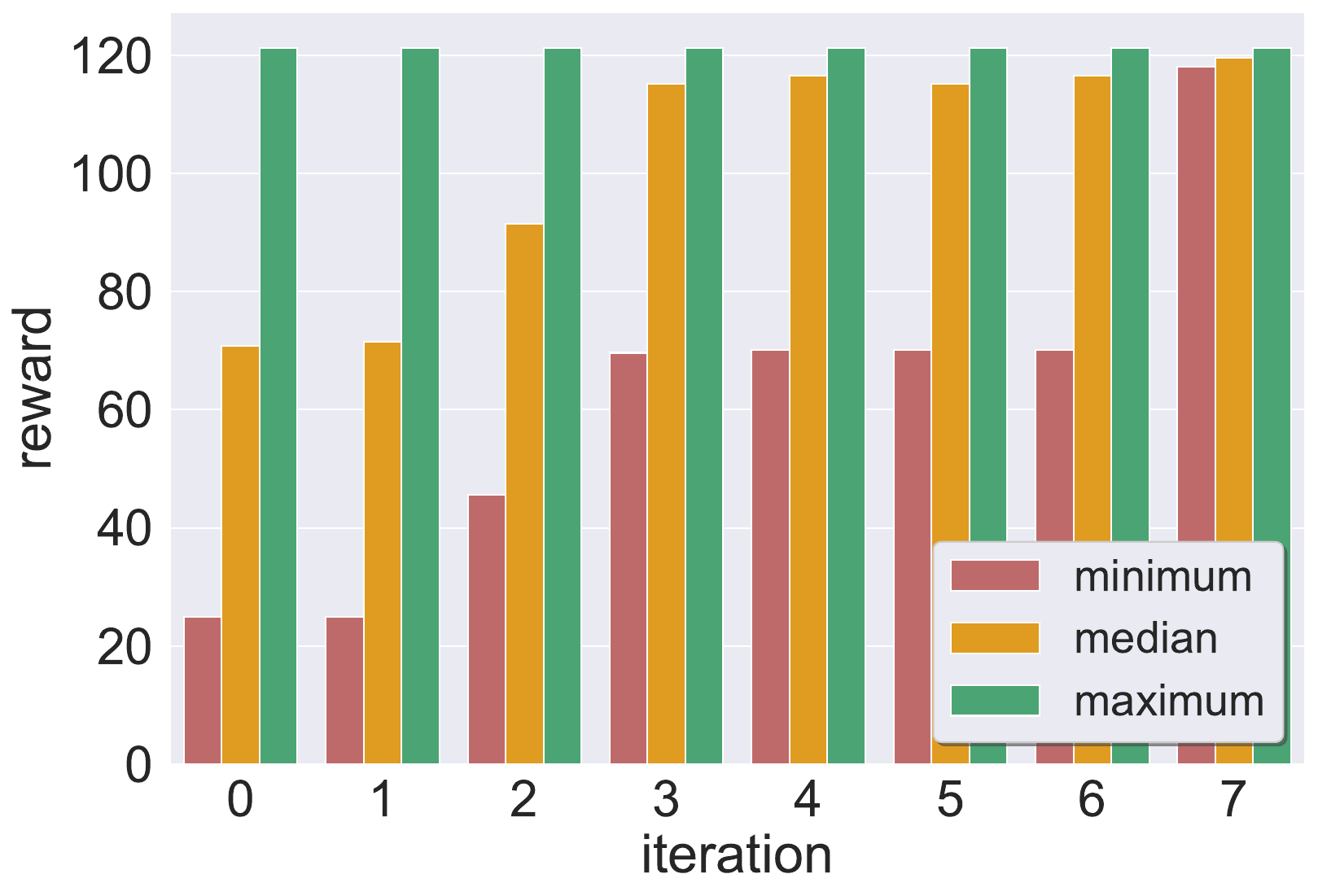}
			\caption{Reward statistics of remaining models}
			\label{}
		\end{subfigure}
		\hfill
		\begin{subfigure}[t]{0.49\linewidth}
			\includegraphics[width=\textwidth, 
			height=0.67\textwidth]{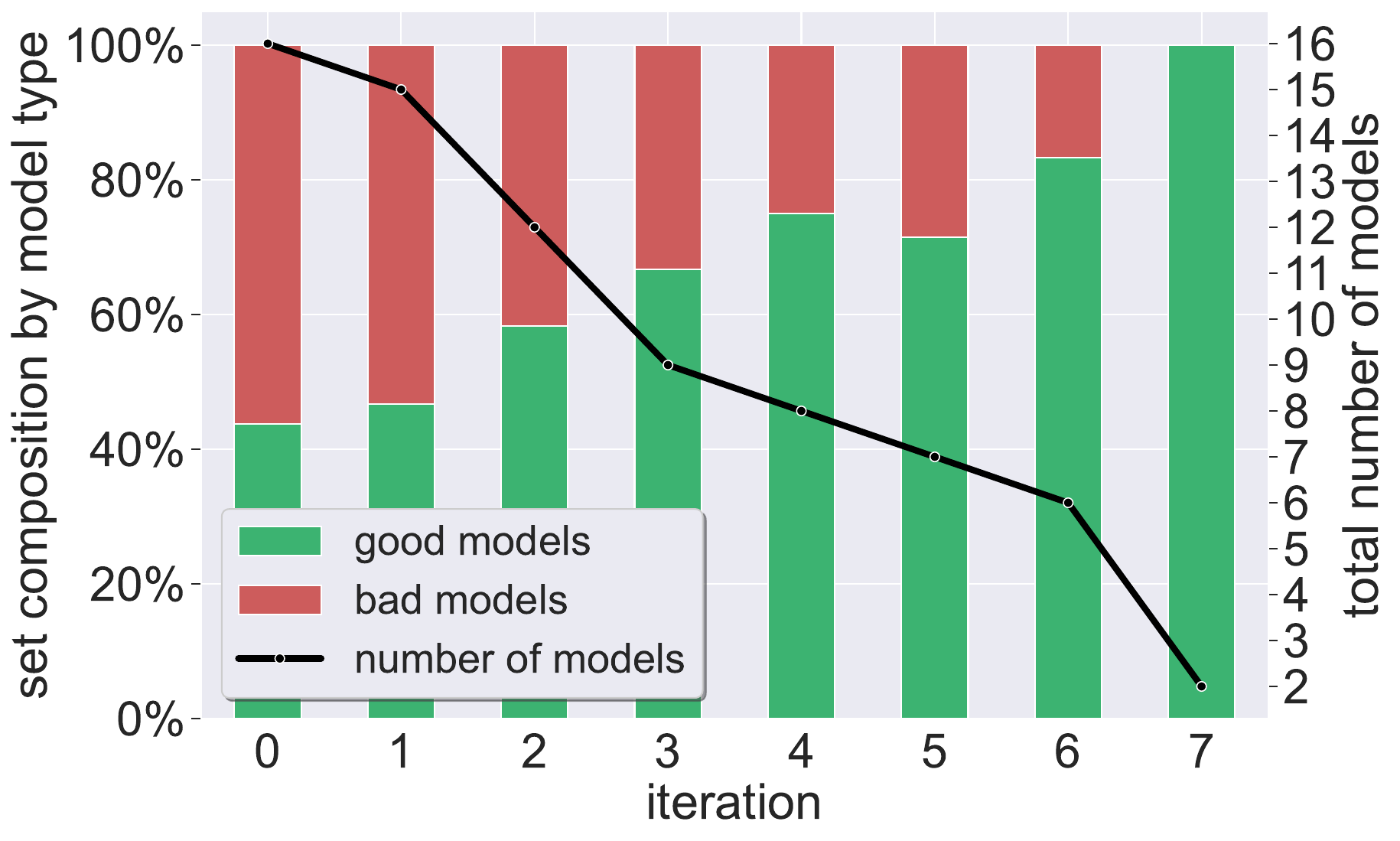}
			\caption{Ratio between good/bad models}
			\label{}
		\end{subfigure}
		\caption{Aurora Experiment~\ref{exp:auroraShort}: PDF 
			$\sim\mathcal{TN}(\mu_{high}, \sigma^{2}$): results using the 
			\maxAgg 
			filtering criterion.}
	\end{figure}
	
	\begin{figure}[h]
		\centering
		\captionsetup[subfigure]{justification=centering}
		\captionsetup{justification=centering} 
		\begin{subfigure}[t]{0.49\linewidth}
			\centering
			\includegraphics[width=\textwidth, 
			height=0.67\textwidth]{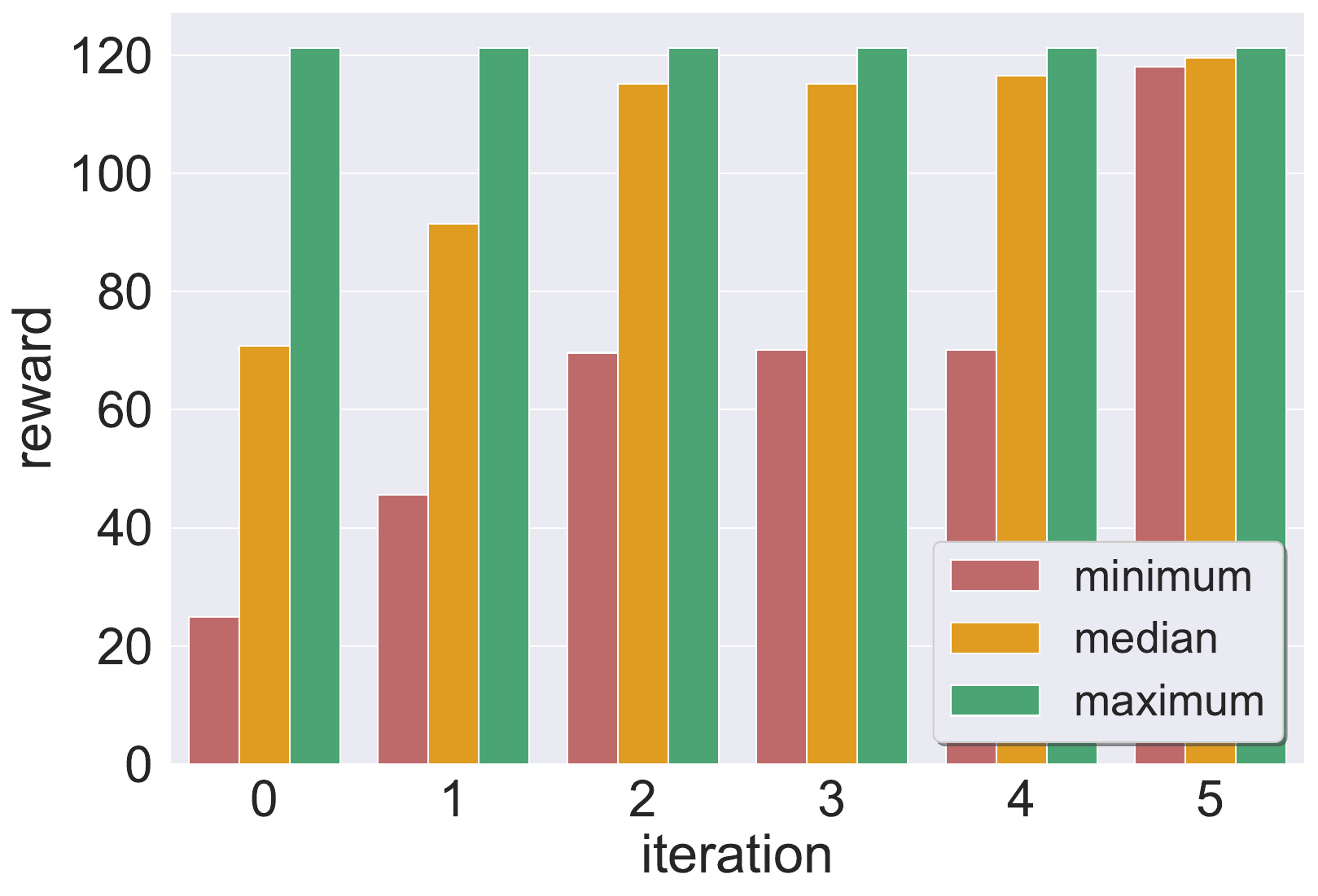}
			\caption{Reward statistics of remaining models}
			\label{}
		\end{subfigure}
		\hfill
		\begin{subfigure}[t]{0.49\linewidth}
			\includegraphics[width=\textwidth, 
			height=0.67\textwidth]{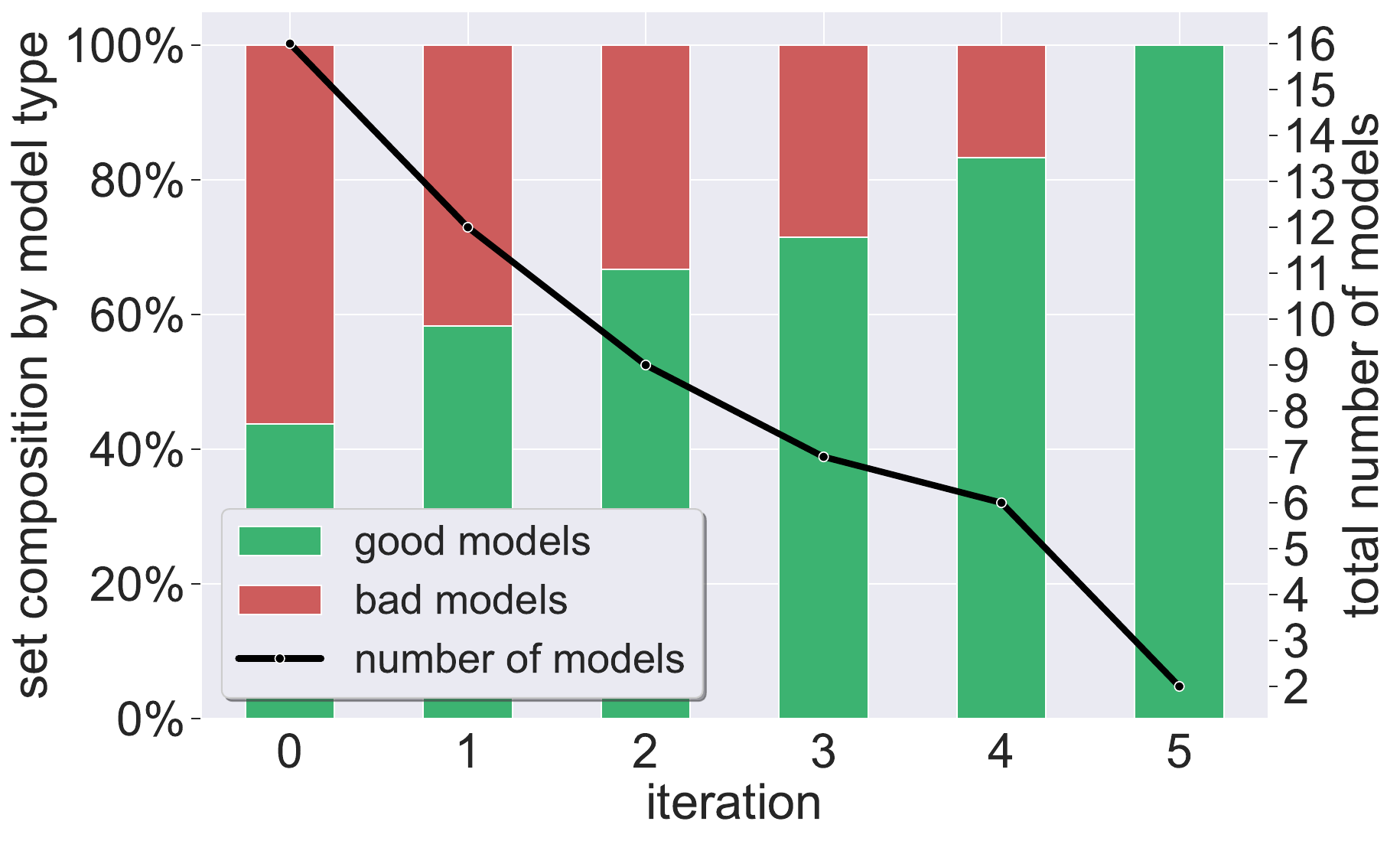}
			\caption{Ratio between good/bad models}
			\label{}
		\end{subfigure}
		\caption{Aurora Experiment~\ref{exp:auroraShort}: PDF 
			$\sim\mathcal{TN}(\mu_{high}, \sigma^{2}$): results using the 
			\conditionCombined filtering criterion.}
	\end{figure}
	
	\begin{figure}[h]
		\centering
		\captionsetup[subfigure]{justification=centering}
		\captionsetup{justification=centering} 
		\begin{subfigure}[t]{0.49\linewidth}
			\centering
			\includegraphics[width=\textwidth, 
			height=0.67\textwidth]{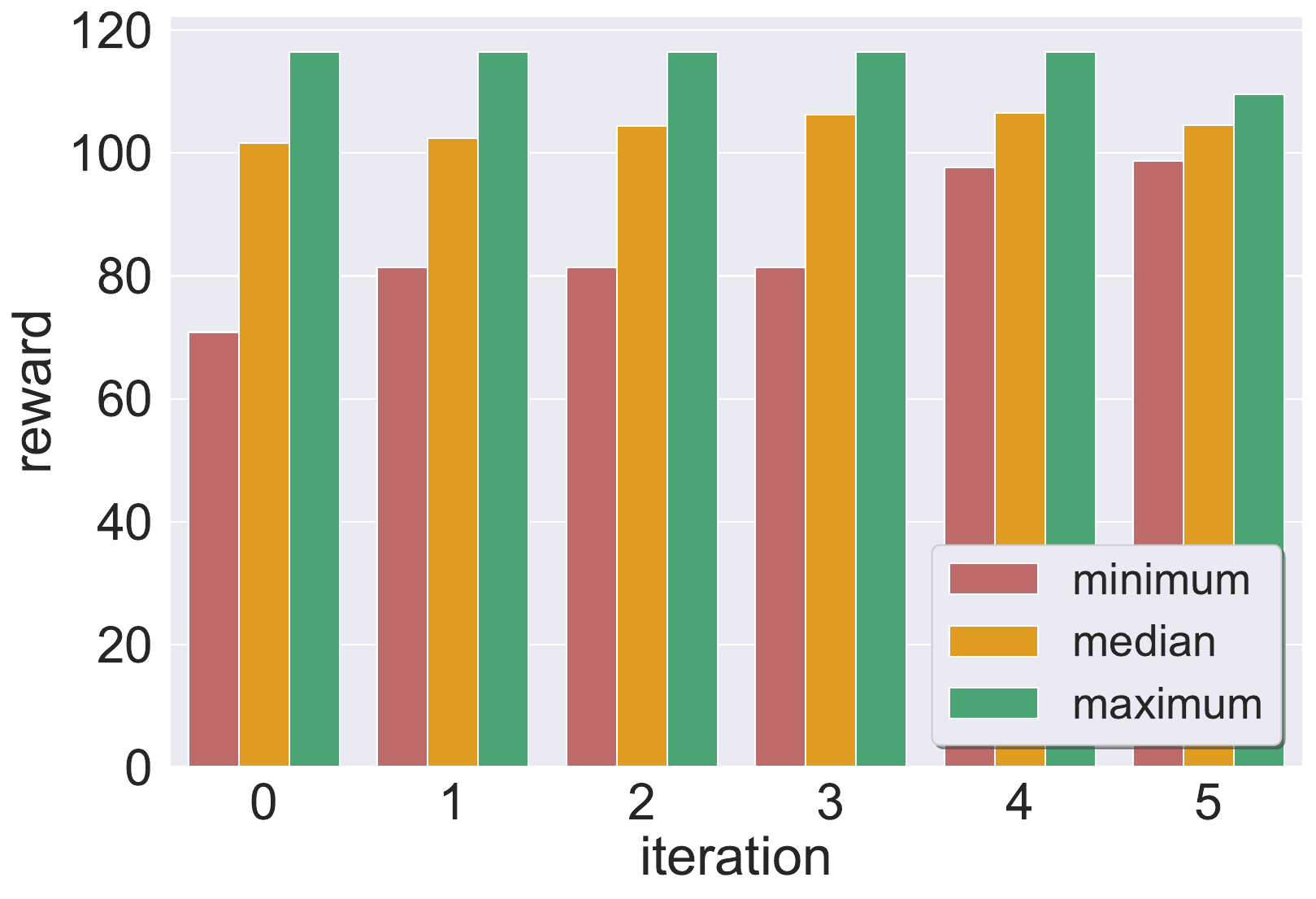}
			\caption{Reward statistics of remaining models}
			\label{}
		\end{subfigure}
		\hfill
		\begin{subfigure}[t]{0.49\linewidth}
			\includegraphics[width=\textwidth, 
			height=0.67\textwidth]{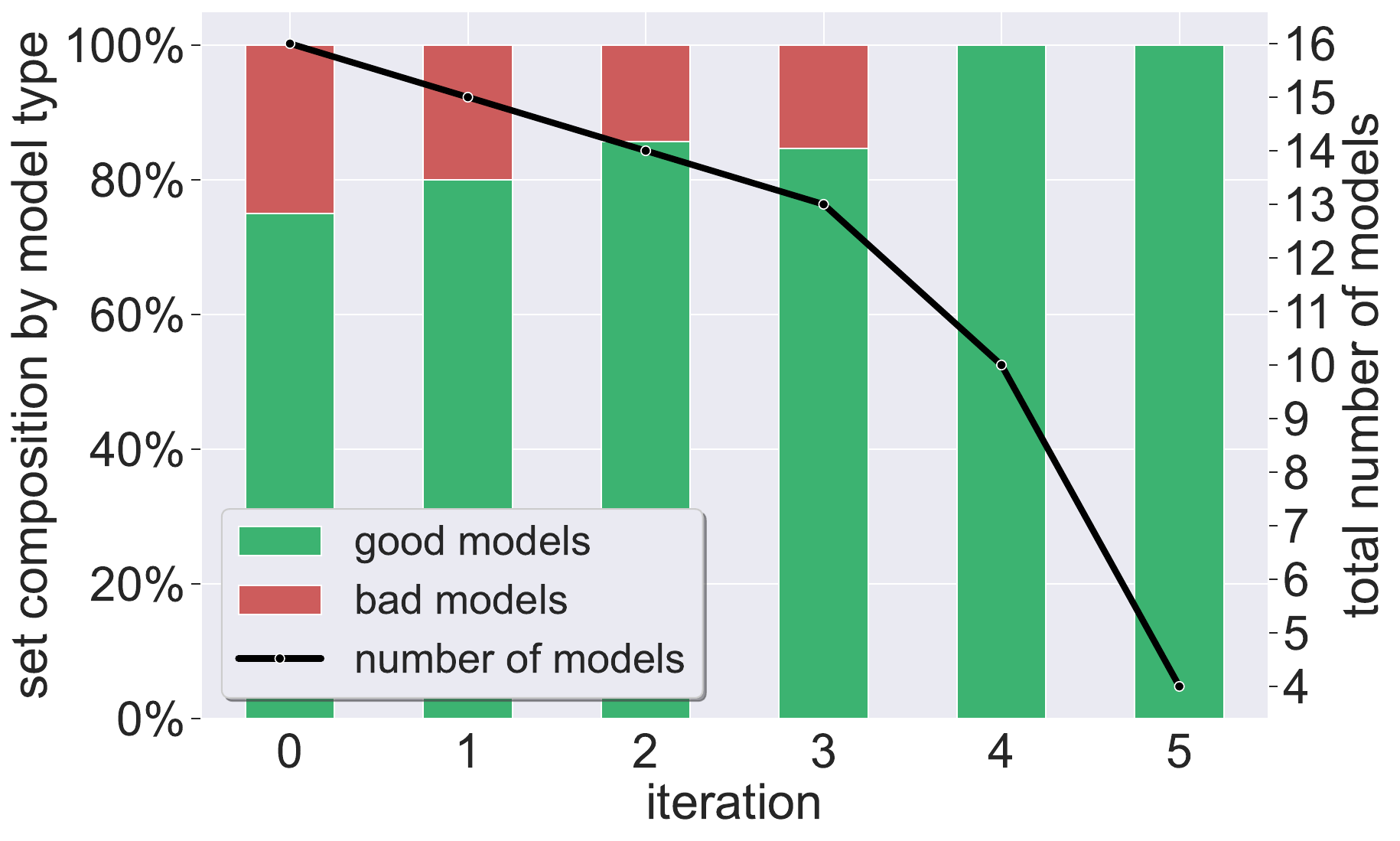}
			\caption{Ratio between good/bad models}
			\label{}
		\end{subfigure}
		\caption{Aurora Experiment~\ref{exp:auroraLong}:, PDF 
			$\sim\mathcal{TN}(\mu_{low}, \sigma^{2}$): results using the 
			\maxAgg 
			filtering criterion.}
	\end{figure}
	
	\begin{figure}[h]
		\centering
		\captionsetup[subfigure]{justification=centering}
		\captionsetup{justification=centering} 
		\begin{subfigure}[t]{0.49\linewidth}
			\centering
			\includegraphics[width=\textwidth, 
			height=0.67\textwidth]{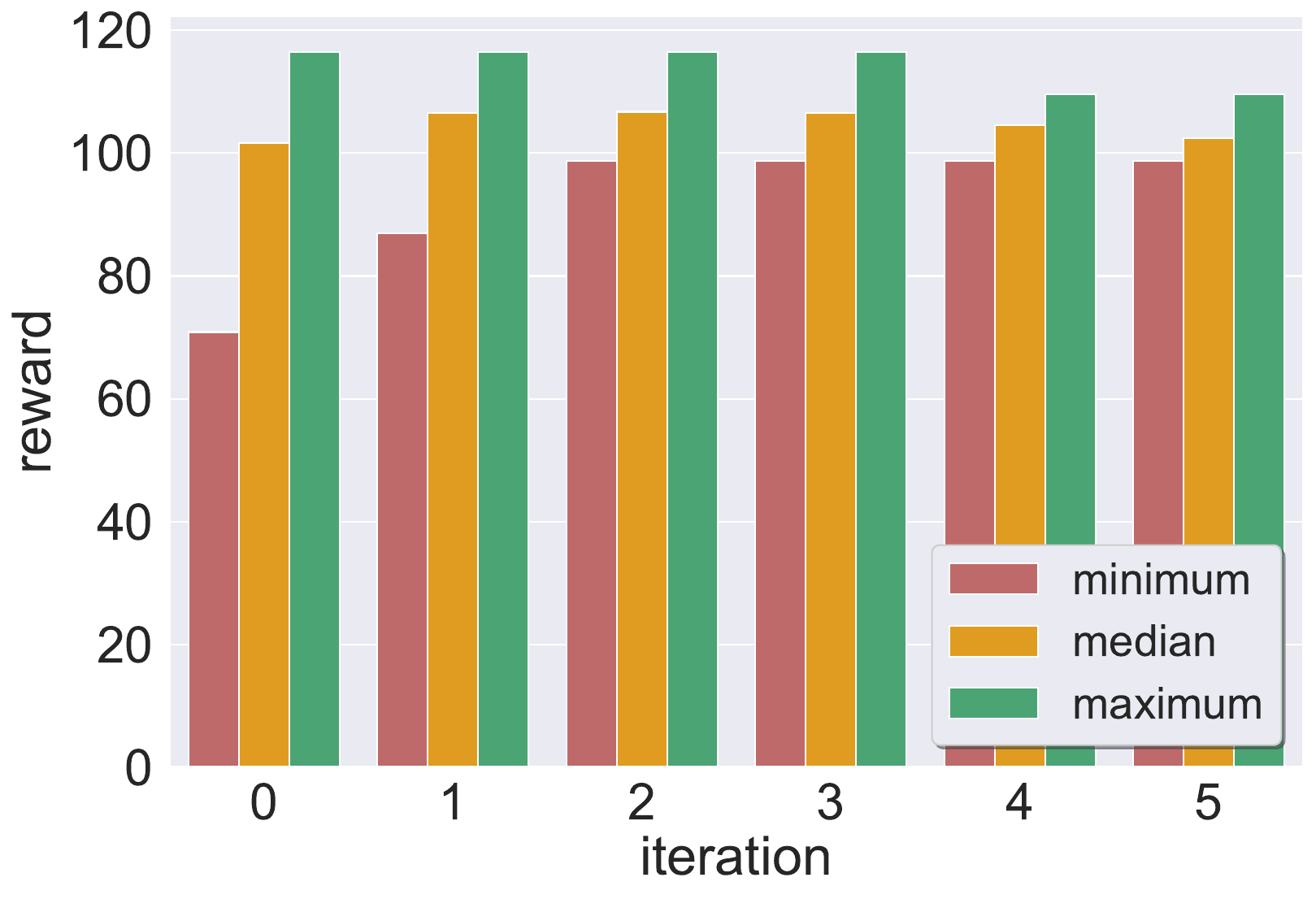}
			\caption{Reward statistics of remaining models}
			\label{}
		\end{subfigure}
		\hfill
		\begin{subfigure}[t]{0.49\linewidth}
			\includegraphics[width=\textwidth, 
			height=0.67\textwidth]{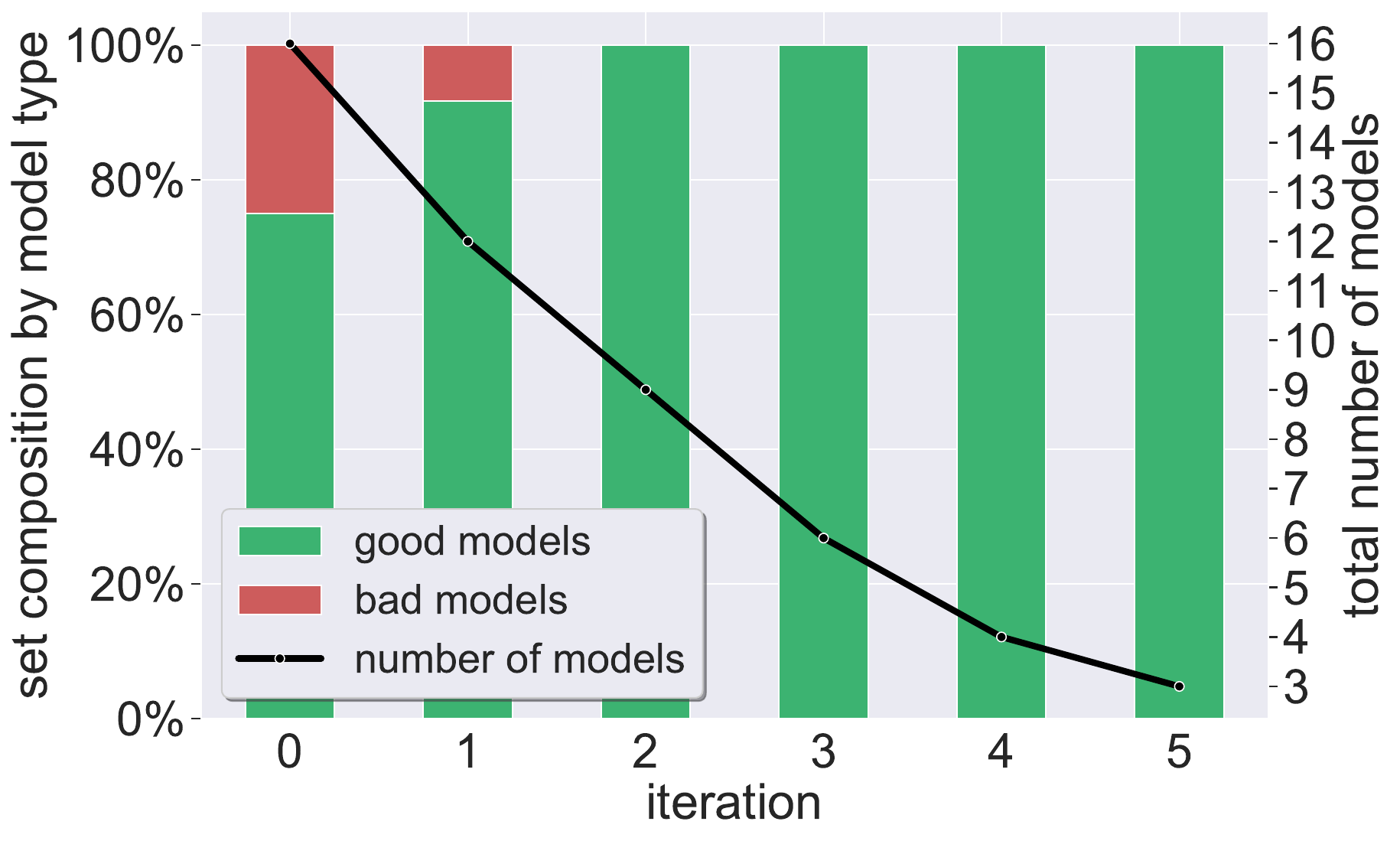}
			\caption{Ratio between good/bad models}
			\label{}
		\end{subfigure}
		\caption{Aurora Experiment~\ref{exp:auroraLong}: PDF 
			$\sim\mathcal{TN}(\mu_{low}, \sigma^{2}$): results using the 
			\conditionCombined filtering criterion.}
	\end{figure}
	
	\begin{figure}[h]
		\centering
		\captionsetup[subfigure]{justification=centering}
		\captionsetup{justification=centering} 
		\begin{subfigure}[t]{0.49\linewidth}
			\centering
			\includegraphics[width=\textwidth, 
			height=0.67\textwidth]{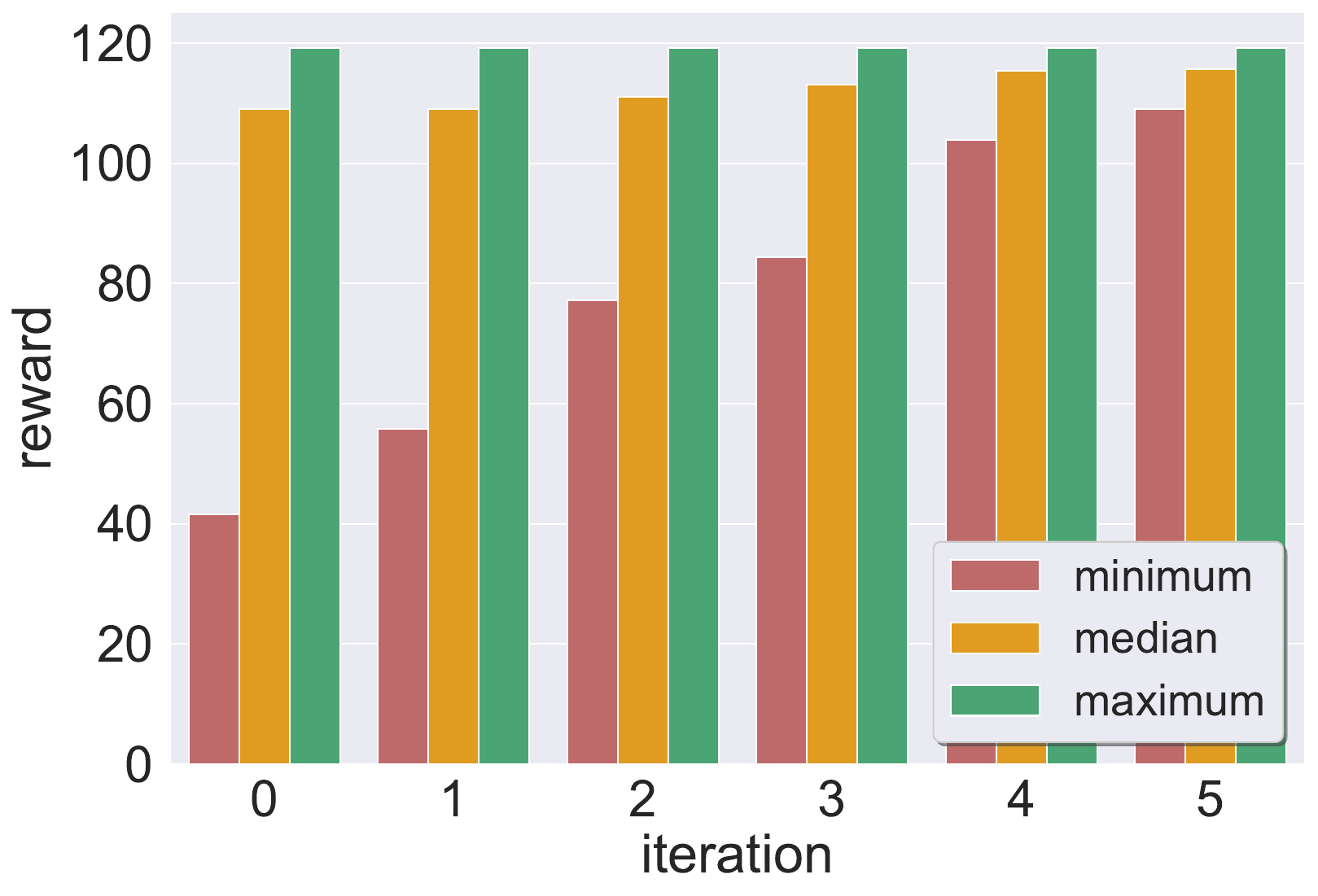}
			\caption{Reward statistics of remaining models}
			\label{}
		\end{subfigure}
		\hfill
		\begin{subfigure}[t]{0.49\linewidth}
			\includegraphics[width=\textwidth, 
			height=0.67\textwidth]{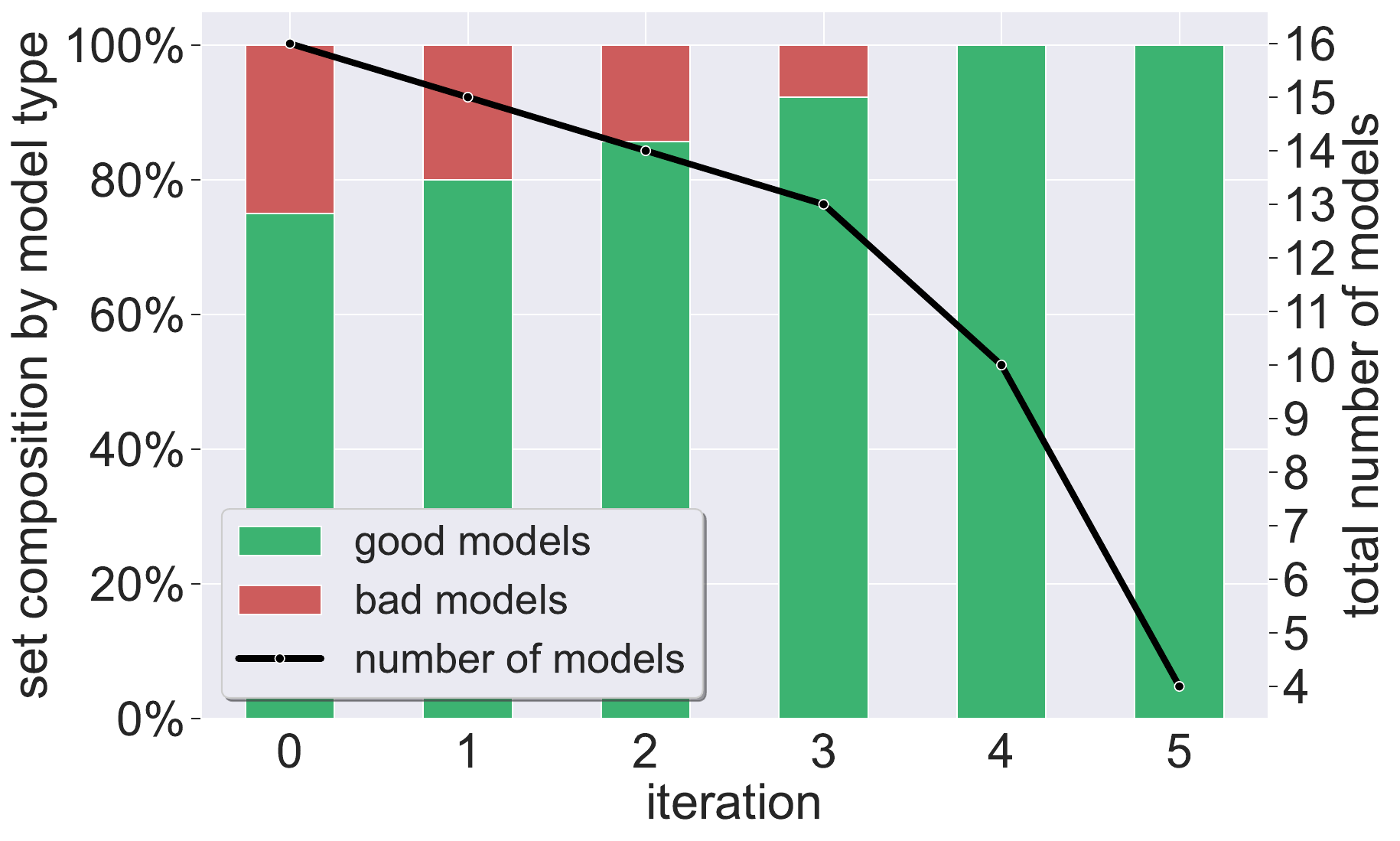}
			\caption{Ratio between good/bad models}
			\label{}
		\end{subfigure}
		\caption{Aurora Experiment~\ref{exp:auroraLong}: PDF 
			$\sim\mathcal{TN}(\mu_{high}, \sigma^{2}$): results using the 
			\maxAgg 
			filtering criterion.}
	\end{figure}
	
	\begin{figure}[h]
		\centering
		\captionsetup[subfigure]{justification=centering}
		\captionsetup{justification=centering} 
		\begin{subfigure}[t]{0.49\linewidth}
			\centering
			\includegraphics[width=\textwidth, 
			height=0.67\textwidth]{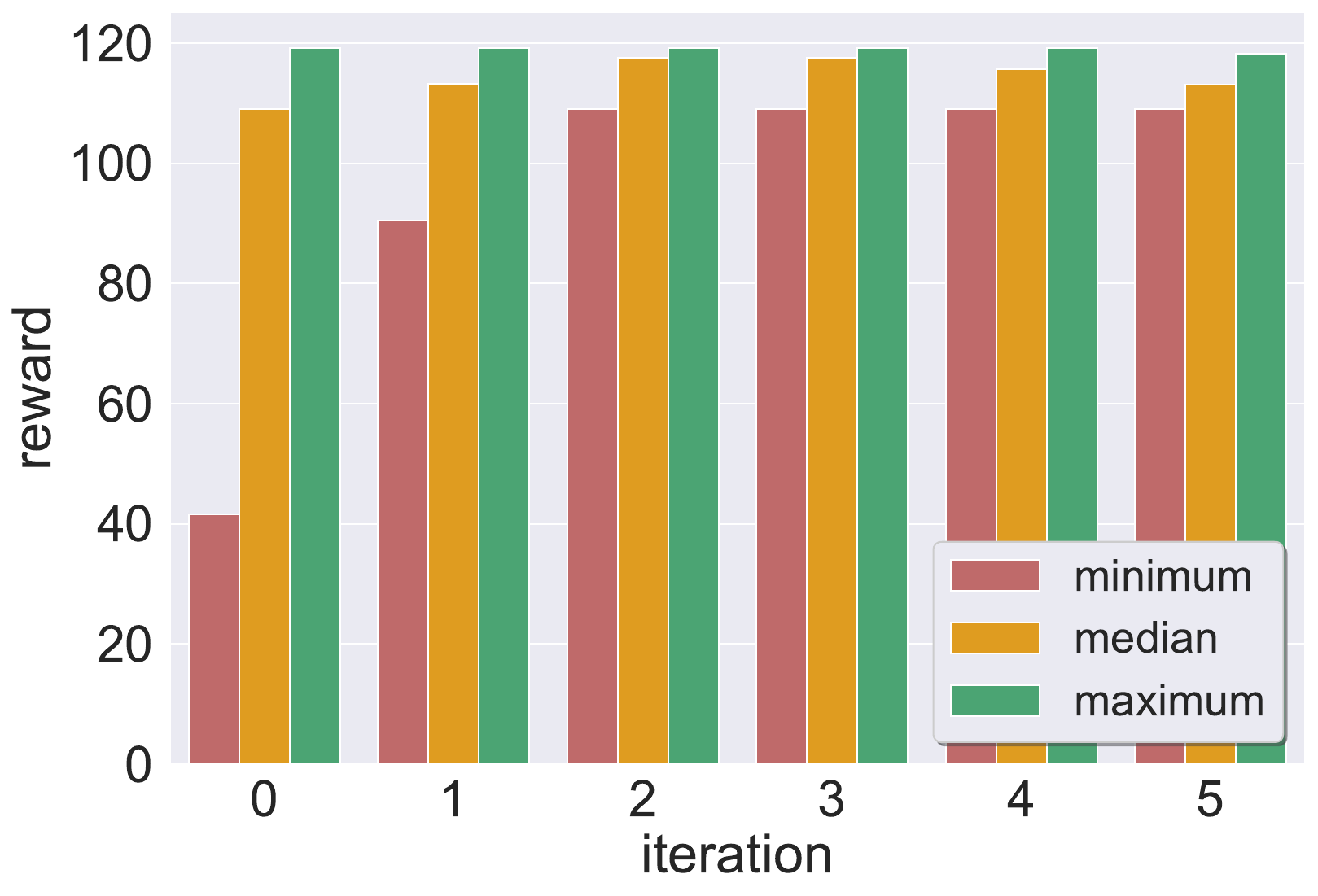}
			\caption{Reward statistics of remaining models}
			\label{}
		\end{subfigure}
		\hfill
		\begin{subfigure}[t]{0.49\linewidth}
			\includegraphics[width=\textwidth, 
			height=0.67\textwidth]{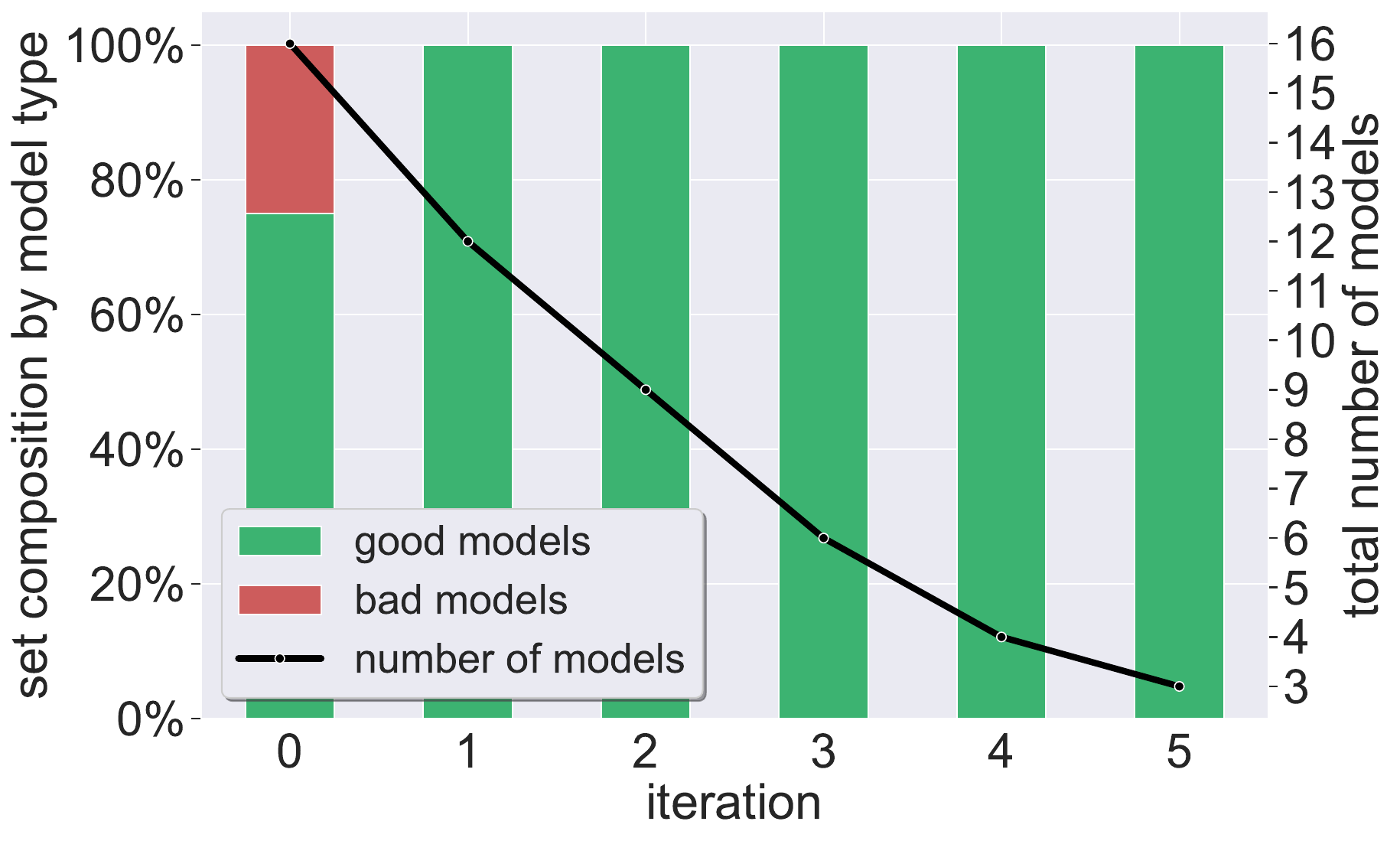}
			\caption{Ratio between good/bad models}
			\label{}
		\end{subfigure}
		\caption{Aurora Experiment~\ref{exp:auroraLong}: PDF 
			$\sim\mathcal{TN}(\mu_{high}, \sigma^{2}$): results using the 
			\conditionCombined filtering criterion.}
	\end{figure}
	\FloatBarrier
	
	\clearpage
	
	\section{Arithmetic DNNs: Supplementary Results}
	\label{sec:appendix:arithmeticDNNs}
	
	
	
	\begin{figure}[ht]
		\centering
		\captionsetup{justification=centering}
		\subfloat[In-distribution 
		\label{subfig:arithmeticDnns:inDist}]{\includegraphics[width=0.49\textwidth]{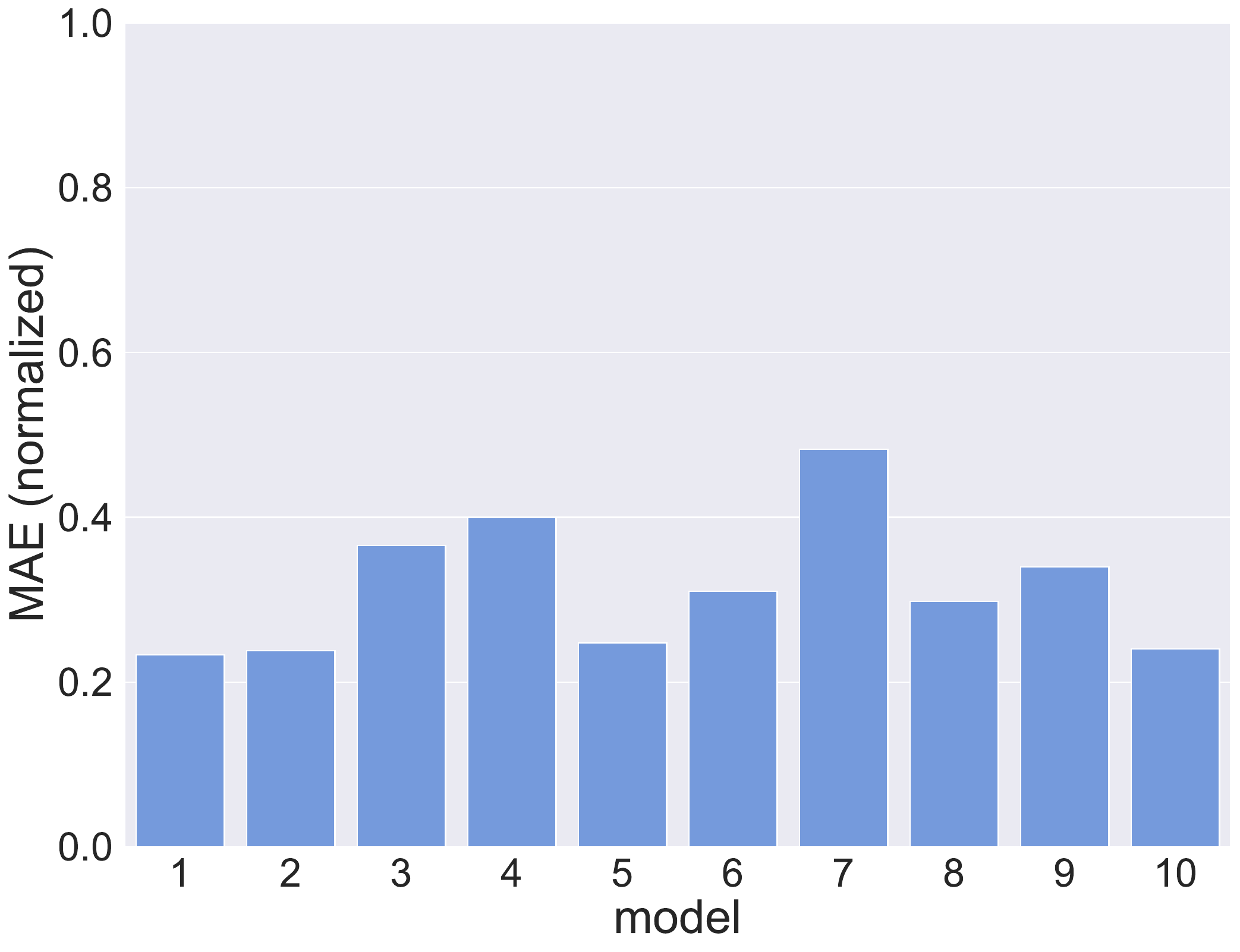}}
		\hfill
		\subfloat[OOD\label{subfig:arithmeticDnns:OOD}]{\includegraphics[width=0.49\textwidth]{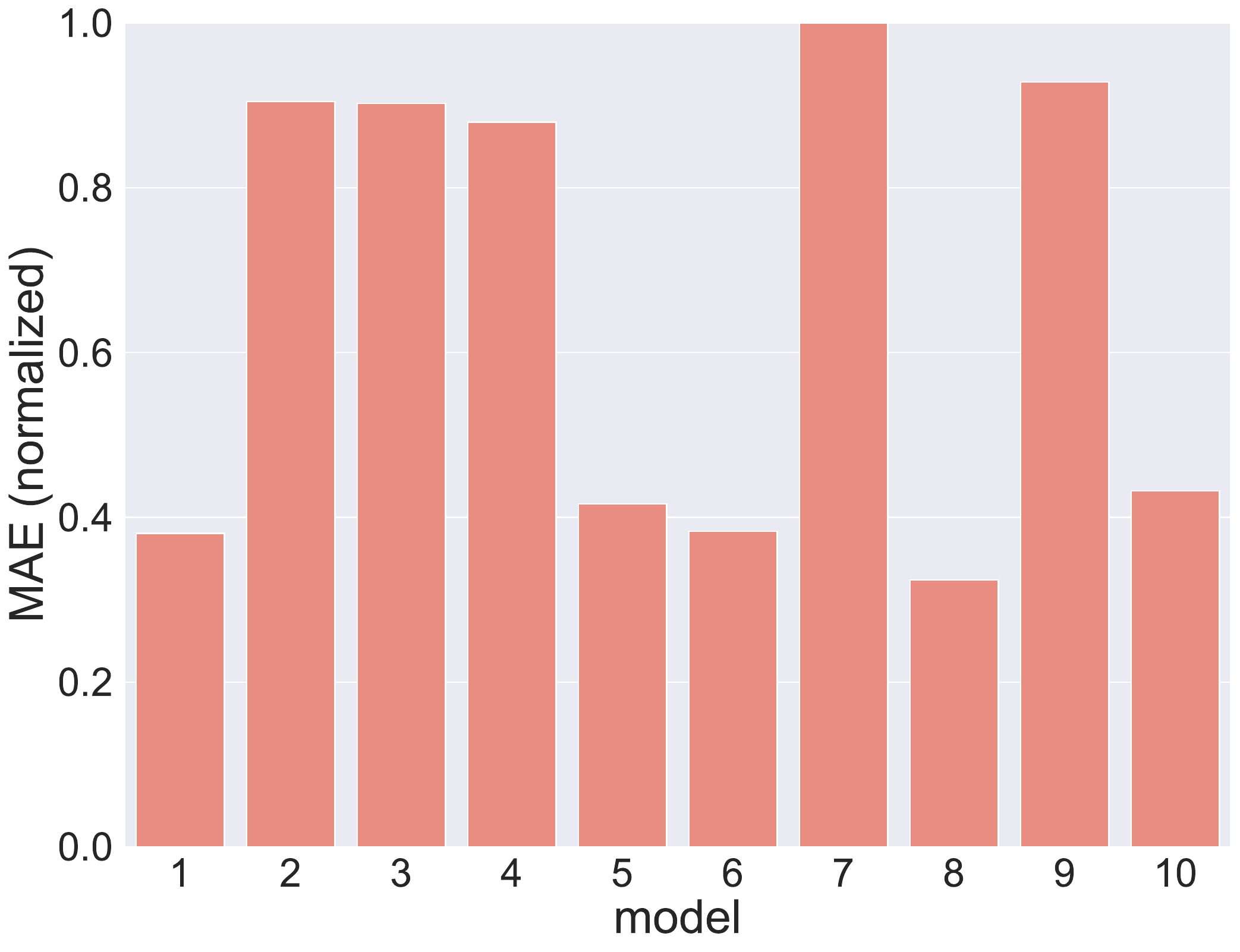}}\\
		
		\caption{Arithmetic DNNs: the models' maximal absolute error (MAE) when 
		simulated on different distributions. The in-distribution results are 
		normalized relative to the OOD range (i.e., multiplied by 100) and 
		divided by the maximal error in the OOD case. The OOD results are 
		normalized based on the maximal error in the OOD case (i.e., 210).}
		\label{fig:arithmeticDnns}
	\end{figure}

	\subsection{Additional Filtering Criteria}
	
	\begin{figure}[!h]
		\centering
		\captionsetup[subfigure]{justification=centering}
		\captionsetup{justification=centering} 
		\begin{subfigure}[t]{0.49\linewidth}
			\centering
			\includegraphics[width=\textwidth]{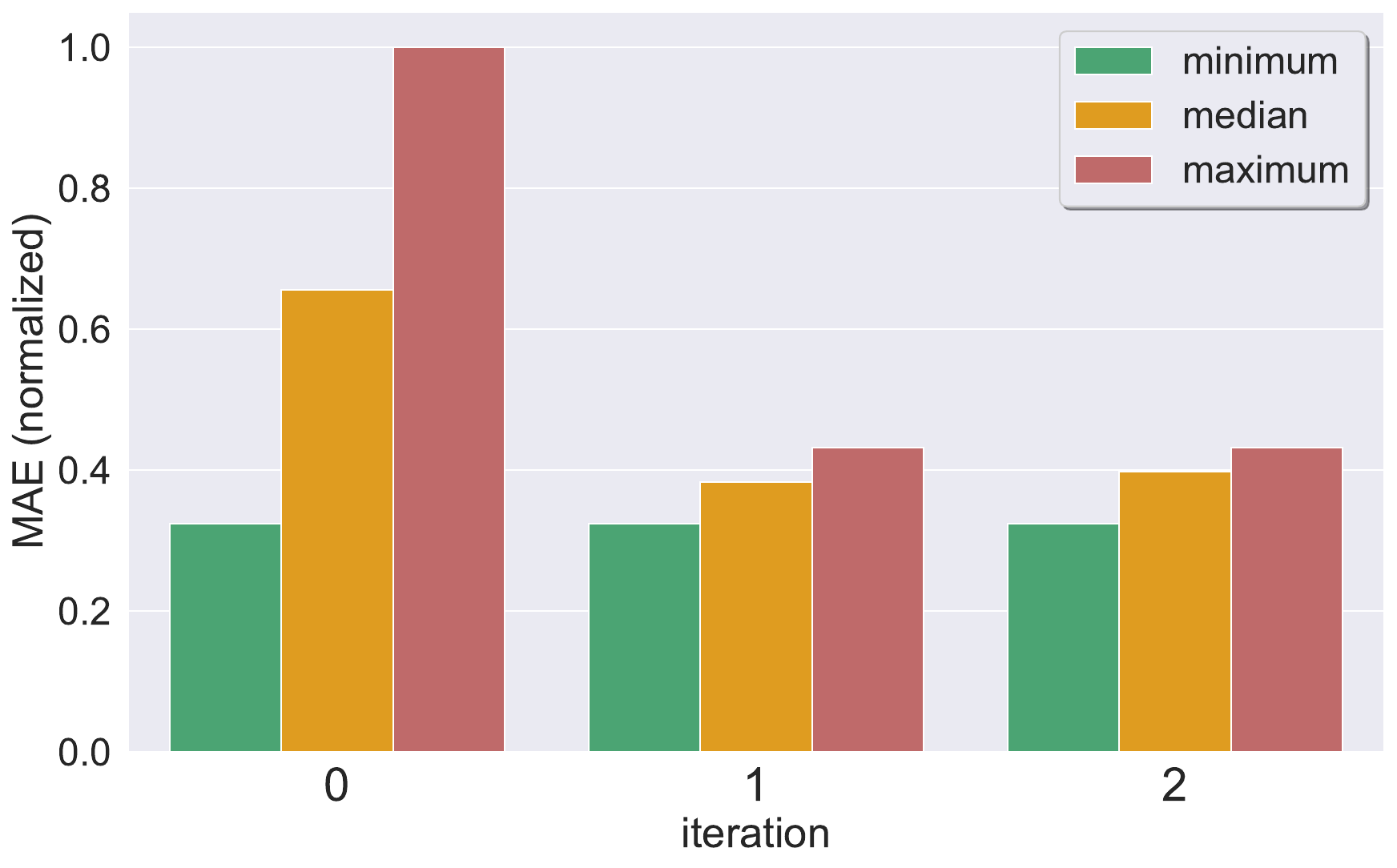}
			\caption{MAE statistics}
			\label{}
		\end{subfigure}
		\hfill
		\begin{subfigure}[t]{0.49\linewidth}
			\includegraphics[width=\textwidth]{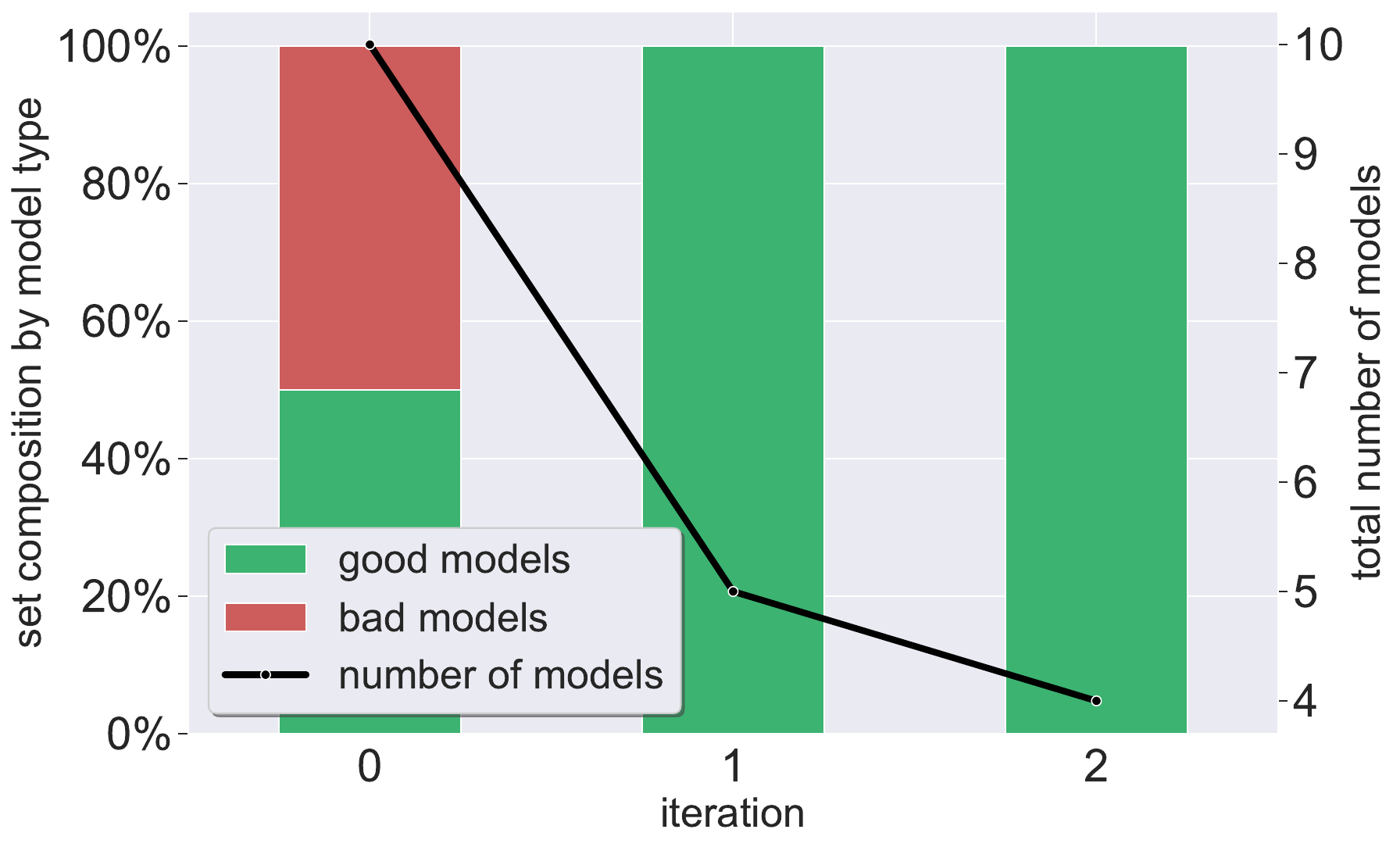}
			\caption{Ratio between good/bad models}
			\label{}
		\end{subfigure}
		\caption{Arithmetic DNNs: results using the \maxAgg filtering 
		criterion, and the \conditionCombined filtering criterion (identical 
		results).
			Our technique  selected models \{1,5,8,10\}.}
		\label{fig:arithmeticDnns:MaxCritResults}
	\end{figure}

	%
	%

	

\end{document}